\documentclass[examplefnt,biber,DBS,plain]{nowfnt} 

\usepackage[utf8]{inputenc}
\usepackage{tikz}
\usepackage{graphicx}
\usepackage{hyperref} 
\usepackage{multirow}
\usepackage[T2A,T1]{fontenc}
\usepackage[russian,english]{babel}
\usepackage{comment}
\usepackage{makeidx}
\usepackage{mdframed}
\usepackage{xcolor}
\usepackage{upquote}
\usepackage{listings} 
\usepackage[printwatermark]{xwatermark}
\newwatermark[allpages,color=black!80,angle=0,scale=0.4,xpos=0,ypos=134]{Submitted to Foundations and Trends in Databases}

\newcommand{\squishlist}{
 \begin{list}{$\bullet$}
  { \setlength{\itemsep}{0pt}
     \setlength{\parsep}{1pt}
     \setlength{\topsep}{1pt}
     \setlength{\partopsep}{0pt}
     \setlength{\leftmargin}{1.5em}
     \setlength{\labelwidth}{1em}
     \setlength{\labelsep}{0.5em} } }
 \newcommand{\squishend}{\end{list}}

\newcommand{\fms}[1]{{\color{orange}FMS: #1}}

\newcommand{\ent}[1]{{\small\tt #1}}
\newtheorem{defn}{Definition}

\newcommand{\ignore}[1]{}
\usepackage{amsmath}
\DeclareMathOperator*{\argmax}{argmax}

\title{Machine Knowledge: Creation and Curation\\ of Comprehensive Knowledge Bases}

\subtitle{Creation and Curation of Comprehensive Knowledge Bases}

\maintitleauthorlist{
Gerhard Weikum\\
Max Planck Institute for Informatics\\
weikum@mpi-inf.mpg.de
\and
Xin Luna Dong\\
Amazon\\
lunadong@amazon.com
\and
Simon Razniewski\\
Max Planck Institute for Informatics\\
srazniew@mpi-inf.mpg.de
\and
Fabian Suchanek\\
Telecom Paris University\\
suchanek@telecom-paris.fr
}

\issuesetup
{%
copyrightowner={G.~Weikum and L.~Dong and S.~Razniewski and F.~Suchanek},
 volume        = xx,
 issue         = xx,
 pubyear       = 2020,
 isbn          = xxx-x-xxxxx-xxx-x,
 eisbn         = xxx-x-xxxxx-xxx-x,
 doi           = 10.1561/XXXXXXXXX,
 firstpage     = 1, 
 lastpage      = 18
 }

\usepackage{mwe}

\author[1]{Weikum, Gerhard}
\author[2]{Dong, Xin Luna}
\author[3]{Razniewski, Simon}
\author[4]{Suchanek, Fabian}

\affil[1]{Max Planck Institute for Informatics;
weikum@mpi-inf.mpg.de}
\affil[2]{Amazon;
lunadong@amazon.com}
\affil[3]{Max Planck Institute for Informatics;
srazniew@mpi-inf.mpg.de}
\affil[4]{Telecom Paris University;
suchanek@telecom-paris.fr}

\begin{document}

\makeabstracttitle

\begin{abstract}
Equipping machines with comprehensive knowledge of the world's entities and their relationships has been a long-standing goal of AI. Over the last decade, large-scale knowledge bases, also known as knowledge graphs, have been automatically constructed from web contents and text sources, and have become a key asset for search engines. This machine knowledge can be harnessed to semantically interpret textual phrases in news, social media and web tables, and contributes to question answering, natural language processing and data analytics. 

This article surveys fundamental concepts and practical methods for creating and curating large knowledge bases. It covers models and methods for discovering and canonicalizing entities and their semantic types and organizing them into clean taxonomies. On top of this, the article discusses the automatic extraction of entity-centric properties. To support the long-term life-cycle and the quality assurance of machine knowledge, the article presents methods for constructing open schemas and for knowledge curation. Case studies on academic projects and industrial knowledge graphs complement the survey of concepts and methods.
\end{abstract}

%

\clearpage\newpage
\tableofcontents
\clearpage\newpage

\chapter{What Is This All About}

\section{Motivation}

Enhancing computers with
``machine knowledge''
that can power 
intelligent 
applications is a long-standing goal of computer science
(see, e.g., \citet{LenatFeigenbaum1991}). 

This formerly elusive vision has become practically viable today, made possible by major advances in {\em knowledge harvesting}. This comprises methods for turning noisy Internet content into crisp knowledge structures
on entities and relations.
Knowledge harvesting methods have enabled the
automatic construction of
{\em knowledge bases (KB)}:
collections of machine-readable facts about the real world.
Today, publicly available
KBs provide millions of
entities (such as people, organizations,  locations
and creative works like books, music etc.) 
and billions of statements about them (such as 
who founded which company when and where,
or which singer performed which song).
Proprietary KBs deployed
at major companies
comprise knowledge at an
even larger scale,
with one or two orders of magnitude
more entities.

A prominent use case where {knowledge bases} have become a key asset
is web search. 
When we send a query like 
``dylan protest songs'' to Baidu, Bing or Google,
we obtain a crisp list of songs such as
Blowin' in the Wind, Masters of War, or A Hard Rain's a-Gonna Fall.
So the search engine automatically detects that we are interested in
facts about an individual entity 
--- Bob Dylan in this case ---
and ask for specifically related entities of a certain type -- protest songs --
as answers.
This
is feasible because the search engine exploits a huge
knowledge base in its back-end data centers, aiding in the discovery of
entities in user requests (and their contexts) and in finding concise answers.

The KBs in this setting are centered on individual entities,
containing (at least) the following backbone information: 

\vspace*{0.05cm}
\squishlist
\item entities like people, places, organizations, products, events, such as\\
\ent{Bob Dylan}, the \ent{Stockholm City Hall}, 
the \ent{2016 Nobel Prize Award Ceremony}
\item the semantic classes to which entities belong, for example\\
\ent{$\langle$Bob Dylan, type, singer-songwriter$\rangle$}, \\
\ent{$\langle$Bob Dylan, type, poet$\rangle$}
\item relationships between entities, such as\\
\ent{$\langle$Bob Dylan, created, Blowin' in the Wind$\rangle$}, \\
\ent{$\langle$Bob Dylan, won, Nobel Prize in Literature$\rangle$}
\squishend
Some KBs also contain validity times 
such as 
\squishlist
\item \ent{$\langle$Bob Dylan, married to, Sara Lownds, [1965,1977]$\rangle$}
\squishend
This temporal scoping is optional, but very important
for the life-cycle management of a KB as
the real world evolves over time.
In the same vein of long-term quality assurance, KBs may also contain 
constraints and provenance information.

\section{History of Large-Scale Knowledge Bases}

The concept of a comprehensive KB goes back to pioneering work in Artificial Intelligence on universal knowledge bases in the 1980s and 1990s, most notably, the {\em Cyc} project 
(\citet{Lenat:CACM1995})
and the {\em WordNet} project 
(\citet{Fellbaum1998}).
\index{Cyc}
\index{WordNet}
However, these knowledge collections have been hand-crafted and curated manually. Thus, the knowledge acquisition was inherently limited in scope and scale.
With the Semantic Web vision in the early 2000s, domain-specific {\em ontologies} 
(\citet{StaabStuder2009})
and schema standards for Web data markup 
(\citet{Guha:CACM2016})
have been developed, but 
these were also manually created.
\index{Semantic Web}
\index{Ontology}
In the first decade of the 2000s,
{\em automatic knowledge harvesting}
from web and text sources 
became a major research avenue, and has made substantial practical impact.
Knowledge harvesting is the core methodology for the
automatic construction of large knowledge bases, going beyond manually
compiled knowledge collections like Cyc or WordNet.

These achievements are rooted in academic research and community projects.
Salient projects that started 
in the 2000s
are
DBpedia 
(\citet{Auer:ISWC2007}),
Freebase 
(\citet{Bollacker:Sigmod2008}),
KnowItAll 
(\citet{Etzioni:ArtInt2005}),
WebOfConcepts 
(\citet{Dalvi:PODS2009}),
WikiTaxonomy 
(\citet{DBLP:conf/aaai/PonzettoS07})
and 
YAGO 
(\citet{Suchanek:WWW2007}).
\index{DBpedia}
\index{Freebase}
\index{KnowItAll}
\index{WebOfConcepts}
\index{WikiTaxonomy}
\index{YAGO}
More recent projects with publicly available data include
BabelNet (\citet{Navigli:ArtInt2012}),
ConceptNet (\citet{Speer:LREC2012}),
DeepDive (\citet{Shin:VLDB2015}), 
EntityCube (aka. Renlifang) (\citet{Nie:IEEE2012}), 
KnowledgeVault (\citet{Dong:KDD2014}),
NELL (\citet{Carlson:AAAI2010}),
Probase (\citet{DBLP:conf/sigmod/WuLWZ12}),
WebIsALOD (\citet{DBLP:conf/semweb/HertlingP17}),
Wikidata (\citet{Vrandecic:CACM2014}), and
XLore (\citet{Wang:ISWC2013}). 
More on the history of KB technology can be found
in the overview article by
\citet{DBLP:journals/corr/abs-2003-02320}.

At the time of writing this survey, the largest 
general-purpose KBs 
with publicly accessible contents 
are 
Wikidata 
({\small\href{https://wikidata.org}{wikidata.org}}), 
BabelNet 
({\small\href{https://babelnet.org}{babelnet.org}}),
DBpedia 
({\small\href{https://dbpedia.org}{dbpedia.org}}), 
and YAGO 
({\small\href{https://yago-knowledge.org}{yago-knowledge.org}}).
They contain 
millions of entities, organized in
hundreds to hundred thousands of semantic classes, and hundred
millions to billions of relational 
statements on entities.  
These
and other knowledge resources are interlinked at the entity level,
forming the Web of Linked Open Data
(\citet{HeathBizer2011}, \citet{DBLP:books/sp/Hogan20}).

Over the 2010s, knowledge harvesting 
has been adopted at big industrial stakeholders
(see \citet{Noy:CACM2019} for an overview),
and large KBs
have become a key asset in a variety of commercial applications,
including semantic search 
(see 
surveys by 
\citet{Bast2016} and \citet{Reinanda2020}), 
analytics (e.g., aggregating by entities),
recommendations 
(see, e.g., \cite{DBLP:journals/corr/abs-2003-00911}),
and data integration (i.e., to combine heterogeneous datasets in and across enterprises).
Examples are the Google Knowledge Graph
\cite{Singhal2012},
KBs used in IBM Watson \cite{DBLP:journals/ibmrd/Ferrucci12},
the Amazon Product Graph \cite{Dong:ICDE2019,DBLP:conf/kdd/DongHKLLMXZZSDM20},
the Alibaba e-Commerce Graph 
\cite{DBLP:conf/sigmod/LuoLYBCWLYZ20},
the Baidu Knowledge Graph \cite{BaiduKG2020},
Microsoft Satori \cite{Qian2013},
Wolfram Alpha \cite{hoy2010wolfphram}
as  well as domain-specific knowledge bases in business, 
finance, life sciences, and more
(e.g., at Bloomberg \cite{Meij2019}).

Industrial KB construction also goes beyond search engines and other Internet giants, and the data may initially reside in structured databases. The obstacles faced in such applications are heterogeneous and ``dirty'' (i.e., partly erroneous) data.
Major advances on data integration methodology
(\citet{DoanHalevyIves2012} \cite{DBLP:journals/cacm/DoanKCGPCMC20}), 
most notably, 
entity matching and linking, have been key 
factors for these
settings.

In addition, KBs have found wide use as a source of distant supervision 
for a variety of tasks in natural language processing,
such as 
entity linking 
and question answering.

\section{Application Use Cases}
\label{sec:ApplicationUseCases}

Knowledge bases enable or enhance a wide variety
of applications.

\vspace{0.2cm}
\noindent{\bf Semantic Search and Question Answering:}

\noindent All major search engines have some form of KB as a background asset.
Whenever a user's information need centers around an entity
or a specific type of entities, such as singers, songs, 
tourist locations, companies, products, sports events etc.,
the KB can return a precise and concise list of entities
rather than merely giving ``ten blue links'' to web pages. 
The earlier example of asking for ``dylan protest songs''
is typical for this line of semantic search.
Even when the query is too complex or the KB is not complete
enough to enable entity answers, the KB information can
help to improve the ranking of web-page results by
considering the types and other properties of entities.
Similar use cases arise in enterprises as well,
for example, when searching for customers or products
with specific properties, or when forming a new team
with employees who have specific expertise and experience.

An additional step towards user-friendly interfaces 
is question answering (QA) where the user poses
a full-fledged question in natural language and
the system aims to return crisp entity-style answers
from the KB or from a text corpus or a
combination of both.
An example for KB-based QA is 
``Which songs written by Bob Dylan received Grammys?'';
answers include All Along the Watchtower, performed by
Jimi Hendrix, which received a Hall of Fame Grammy Award.
An ambitious example that probably requires tapping into
both KB and text would be
``Who filled in for Bob Dylan at the Nobel Prize ceremony
in Stockholm?''; the answer is Patti Smith.
Overviews on semantic search and question answering 
with KBs include \cite{Bast2016,Diefenbach2018,LeiOzcanEtal:DEbull2018,Reinanda2020,Unger2014}.

\vspace{0.2cm}
\noindent{\bf Language Understanding and Text Analytics:}

\noindent Both written and spoken language are full of ambiguities.
Knowledge is the key to mapping surface phrases to their
proper meanings, so that machines interpret language
as fluently as humans. AI-style use cases include 
machine translation, and conversational assistants like chatbots. 
Prominent examples include Amazon's Alexa,
Apple's Siri, and
Google's
Assistant and new chatbot initiatives
\cite{DBLP:journals/corr/abs-2001-09977},
and Microsoft's Cortana.

In these applications, world knowledge plays a crucial role. Consider, for example,
sentences like ``Jordan holds the record of 30 points per match'' or ``The forecast for Jordan is a record
high of 110 degrees''.
The meaning of the word ``Jordan'' can be inferred by having
world knowledge about the basketball champion Michael Jordan
and the middle-eastern country Jordan.

Understanding entities (and their attributes and associated relations) in text is also key to large-scale analytics over
news articles, scientific publications,
review forums, or social media discussions. 
For example, we can identify mentions of products
(and associated consumer opinions), link them
to a KB, and then perform comparative and
aggregated studies. We can even incorporate
filters and groupings on product categories,
geographic regions etc., by combining the
textual information with structured data
from the KB or from product and customer databases.
All this can be enabled by the KB as a clean and
comprehensive repository of entities
(see 
\citet{Shen:TKDE2015} 
for a survey
on the core task of entity linking).

A trending example of semantic text analytics
is detecting gender bias in news and other online
content (see, e.g., \cite{SuchanekPreda:VLDB2014}). 
By identifying people, and the KB
knowing their gender, we can compute statistics
over male vs. female people in political offices
or on company boards.
If we also extract earnings from movies and
ask the KB to give us actors and actresses,
we can shed light into potential unfairness
in the movie industry.

\vspace{0.2cm}
\noindent{\bf Visual Understanding:}

\noindent For detecting objects and concepts in images (and videos),
computer vision has made great advances using machine learning.
The training data for these tasks are collections of 
semantically annotated images, which can be viewed as visual
knowledge bases. The most well-known example is
ImageNet (\cite{Deng:CVPR2009}) which has populated a subset
of WordNet concepts with a large number of example images.
A more recent and more advanced endeavor along these lines is
VisualGenome (\cite{Krishna:IJCV2017}).
By themselves, these assets already go a long way, but
their value can be further boosted by combining them
with additional world knowledge.

For example, knowing that lions and tigers are both predators
from the big cat family and usually prey on deer or antelopes, 
can help to automatically label scenes
as ``cats attack prey''.
Likewise, recognizing landmark sites such as the Brandenburg gate
and having background knowledge about them (e.g., other sites
of interest in their vicinity) helps to understand 
details and implications of an image.
In such computer vision and further AI applications, 
a KB often serves as an informed prior for machine learning models,
or as a reference for consistency (or plausibility) checks.

\vspace{0.2cm}
\noindent{\bf Data Cleaning:}

\noindent Coping with incomplete and erroneous records in large
heterogeneous data is a classical topic in database research
(see, e.g., \cite{DBLP:journals/debu/RahmD00}).
The problem has become more timely and pressing than ever.
Data scientists and business analysts want to rapidly tap into diverse datasets,
for comparison, aggregation and joint analysis.
So different kinds of data need to be combined and fused, more or less on the fly and thus largely depending on automated tools. This trend amplifies the crucial role of 
identifying and repairing missing and incorrect values.

In many cases, the key to spotting and repairing errors or to
infer missing values is consistency across a set of records.
For example, suppose that a database about music has a new tuple stating that Jeff Bezos won
the Grammy Award. A background knowledge base would tell that
Bezos is an instance of types like
businesspeople, billionaires, company founders etc.,
but there is no type related to music.
As the Grammy is given only for songs, albums
and musicians, the tuple about Bezos is likely a
data-entry error.
In fact, the requirement that Grammy winners,
if they are of type person, have to be
musicians, can be encoded into a logical consistency constraint for automated inference.
Several KBs contain such consistency constraints. They typically include:
\squishlist
\item type constraints, e.g.:
a Grammy winner who belongs to the type
\ent{person} must also be an instance
of type \ent{musician} (or a sub-type),
\item functional dependencies, e.g.: 
for each year and each award category, there is exactly one Grammy winner,
\item inclusion dependencies, e.g.: 
composers are also musicians and thus can win a Grammy, and all Grammy winners must have at least
one song to which they contributed (i.e., the set
of Grammy winners is a subset of the set of
people with at least one song),
\item disjointness constraints, e.g.: %
songs and albums are disjoint, so no piece of music can simultaneously win both of these award categories for the Grammy,
\item temporal constraints, e.g.: 
the Grammy award is given only to living people, so that someone who died in a certain year cannot win it in any later year.
\squishend

\noindent Data cleaning as a key stage in data integration
is surveyed by
\citet{DBLP:journals/ftdb/IlyasC15,DBLP:books/acm/IlyasC19}
and the articles in \cite{DBLP:journals/debu/000118}.

\clearpage\newpage
\section{Scope and Tasks}

\subsection{What is a Knowledge Base}

This article covers methods for automatically constructing and
curating large knowledge bases from online content,
with emphasis on semi-structured 
web pages with lists, tables etc.,
and unstructured text sources.

Throughout this survey, we use the following
pragmatic definition of a KB and its salient properties. 

\begin{mdframed}[backgroundcolor=blue!5,linewidth=0pt]
\squishlist
\item[ ] A {\bf knowledge base (KB)} is a 
collection of structured data about {\em entities} and 
{\em relations} with the following characteristics:
\squishlist
\item {\bf Content:} 
The data contains entities and their semantic types for a given domain of interest.
Additionally, attributes of entities (including numeric and string literals) and relationships between entities are captured.
The domain of interest can be of broad enyclopedic nature,
with emphasis on notable entities (like those with Wikipedia articles), or can have specific themes such as indie music
or medical and nutritional health products, with a long tail of relevant entities. 
\item {\bf Quality:} We expect the KB content to be
of {\em near-human quality}, with the rate of invalid statements
below the error rate that a collection with
expert-level curation would achieve.
The KB content should be continuously updated for freshness,
and maintained in a consistent way (e.g., no
contradictory statements).
\item {\bf Schema and Scale:} 
Unlike a conventional database, there is
often no pre-determined relational schema where all
knowledge has to fit into a 
static 
set of relations. 
Even when KB construction starts with a specified schema, its
longitudinal evolution must allow ad-hoc additions where the set of types and relations may grow
to ten or hundred thousands. At this scale, 
a database-style rigorous schema is 
not
manageable. 
Therefore, KBs adopt the
{\em dataspace} ``pay-as-you-go'' principle
\cite{DBLP:conf/pods/HalevyFM06}:
the content is augmented and refined by adding
new types, attributes and relations, as the KB grows.
\item {\bf Open Coverage:}
An ideal KB would contain {\em all} entities and their properties that are of interest for the domain or enterprise.
However, this complete coverage is often a 
hopeless if not ill-defined target.
New entities and 
facts
emerge and get covered in new web sources at high rate. %
Therefore, we have to view KB construction
and maintenance as a ``never-ending'' task,
following an open world assumption and
acknowledging the high pace of real-world changes.
\squishend
\squishend
\end{mdframed}

These characteristics are most strongly expressed in thematically broad KBs of encyclopedic nature.
Therefore, we will 
largely 
focus on this line of
exemplary cases. Nevertheless, domain-specific KBs, say on the health domain, involve 
the same
tasks and challenges,
and many enterprise projects have similar
characteristics.
For example, retail companies such as Amazon or Alibaba Taobao have a huge and ever-changing set of 
entities, with much of the data coming from
external partners like independent merchants.
In other cases, building an enterprise-wide reference catalog of entities and their properties is
closer to traditional database integration \cite{DoanHalevyIves2012}, such as matching and merging data on customers and a fixed set of products. These 
contexts are sometimes referred to as 
Enterprise Data Warehouses, Master Data Management or 360 Databases. 
The material in this article is potentially relevant
for these settings as well, 
but specifically covering them
is out of scope.

\subsection{Tasks for KB Creation and Curation}

{\bf Creating a KB} involves populating it with
entities, types, attributes of entities and
relationships between entities. 
This is a major endeavor on its own, but it is
only the first big step towards deployment and
providing value to applications.
A KB should be maintained, grown and enhanced
over an extended life-time. 
This necessitates
a longitudinal form of {\bf KB curation}
for continuously augmenting and cleaning the KB content and for quality assurance.

\begin{figure} [bt!]
  \centering
   \includegraphics[width=0.9\textwidth]{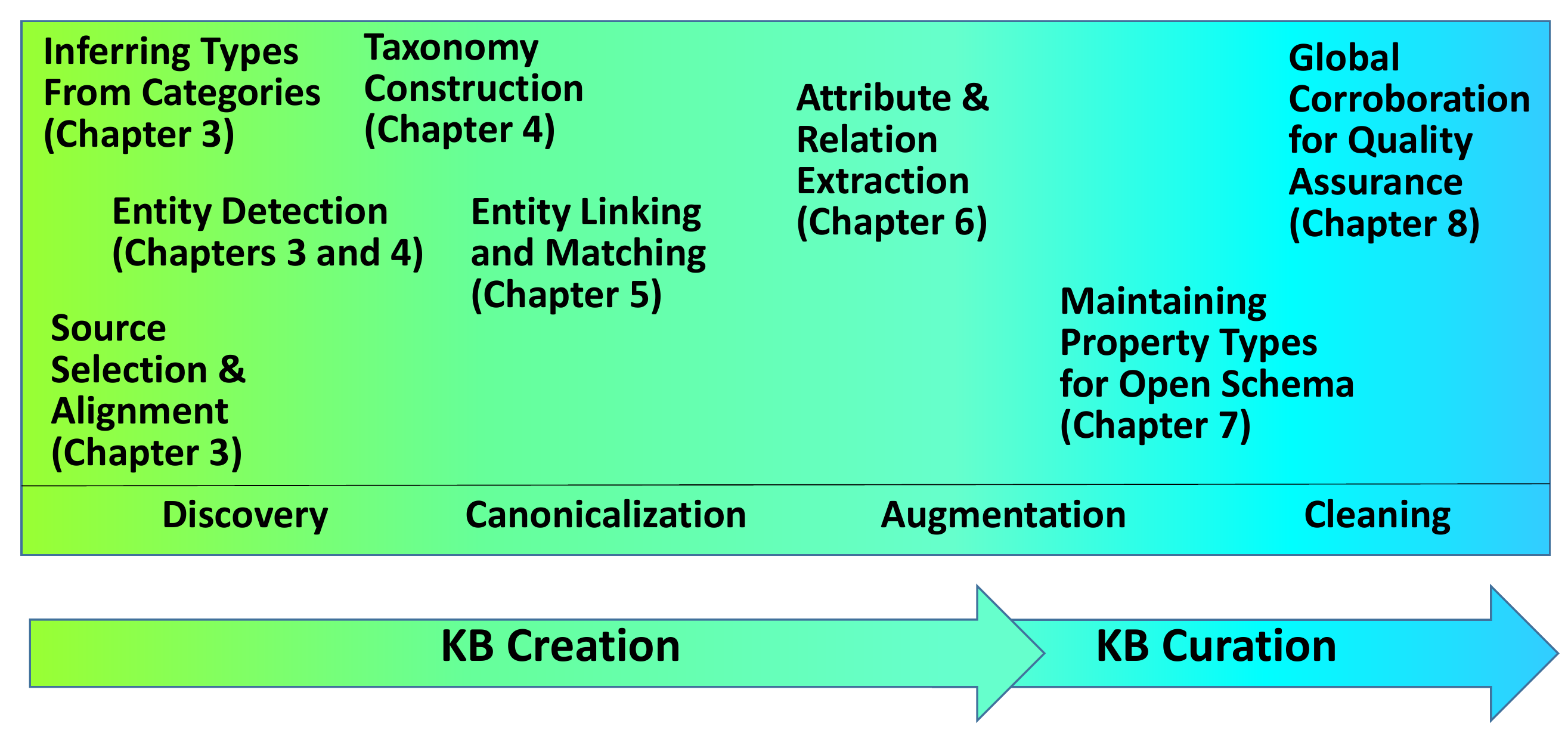}
      \caption{Major Tasks for KB Construction} 
      \label{ch1-fig-knowledgeharvesting-tasks}
\end{figure}

Creating a knowledge base and curating it throughout its life-cycle involves a number of tasks and their interplay. 
We outline the most important of these tasks as major building blocks that this article is 
focused on. Figure \ref{ch1-fig-knowledgeharvesting-tasks} gives a pictorial overview.
From left to right it depicts major phases in
the 
value-adding pipeline;
for each phase (or pair of overlapping phases), 
it shows the technical topics that we discuss in
the respective chapters. 

\vspace*{0.2cm}
\noindent {\bf Discovery:} 
KB construction needs to start
with
deciding on the intended scope of entities and types that should be covered,
and selecting promising sources for these %
entities. 
Thus, {\em source selection} and {\em content discovery} are early steps.
When 
tapping into several structured or semi-structured sources simultaneously,
it can be beneficial to first compute
{\em alignments} between sources, at the entity or type level.
In some cases, semantic types can be derived from category labels or
tags provided by the sources. 
Chapter 3 discusses these first-stage tasks.

A next step in the discovery phase is {\em entity detection},
which spots entity mentions in tables, lists or text. 
Analogously, types are discovered in or derived from input sources, 
and organized
into a hierarchical taxonomy, as a backbone for further populating
and augmenting the KB at later stages.
Chapter 4 presents methods for entity detection and taxonomy construction.

\vspace*{0.2cm}
\noindent {\bf Canonicalization:}
The entities discovered in the input sources are first captured 
in the form of {\em mentions}, where different names for the same entity
and ambiguous names that can denote different entities
are not yet resolved. 
Identifying synonyms (aka duplicates) and disambiguating 
mention strings
is a key step for a high-quality KB where all entities should be
properly canonicalized to be uniquely identifiable.
When the input is semi-structured web content or unstructured text
and there is already an initial reference repository of entities,
this task is called {\em entity linking}. When the reconciliation is
over two or more structured datasets, it is often referred to as
{\em entity matching}. 
Chapter 5 is devoted to this %
task.

\vspace*{0.2cm}
\noindent {\bf Augmentation:}
Once we have a core KB with canonicalized entities and a clean type system,
the next task is to populate properties with instance-level statements:
attribute values and entity pairs that stand in a certain relation.
This phase augments the KB content and value for downstream applications.
At the heart of this task is the algorithmic and learning-based {\em extraction
of properties} 
from text and semi-structured web contents (i.e., tables, lists etc.).
Chapter 6 presents a suite of methods for this purpose.
At some point, when the original KB schema is not sufficient anymore to satisfy
the application needs, we need to address the evolving requirements for
{\em open coverage}. 
Dynamically {\em augmenting the schema} is the subject of
Chapter 7.

\vspace*{0.2cm}
\noindent {\bf Cleaning:}
There is no recipe for building a perfect KB, without any errors or gaps.
Therefore, there is always a need for {\em curating} a freshly created KB.
This involves corroborating its statements, pruning out invalid ones, 
and comprehensive KB cleaning for quality assurance.
Moreover, gradually growing the KB, to keep up with real-world changes
and evolving requirements, necessitates continuously scrutinizing the KB
for correctness and coverage. 
Chapter 8 discusses this important, but often neglected,
aspect of KB quality and longitudinal maintenance.

\section{Audience}

We hope that 
this article
will be useful for doctoral students 
(and potentially also advanced undergraduates) and faculty 
interested in a wide spectrum of topics -- from machine knowledge and
data quality to machine learning and data science
as well as applications in web content mining and natural language understanding. 
In addition, the article aims to be useful also for industrial researchers and practitioners working on semantic technologies for web, social media, or enterprise contents, including all kinds of applications where sense-making from text or semi-structured data is an issue. 
Prior knowledge on natural language processing or statistical learning is not required; we will introduce relevant methods as they are needed (or at least give specific pointers to literature).

KB technology has originated from and is pursued in different areas of computer science.
We hope that our audience cuts across a variety of relevant communities, from databases and web research to machine learning and general AI.
This diversity entails 
serving
readers with very different backgrounds.
We aim to minimize assumptions on prior knowledge
by introducing
basics as we go along.
As perfection exists only in songs and movies,
we apologize for whatever gaps are left
and for shortcuts taken.

\noindent{\bf Semantic Web:} 
This community pretty much led the advent of
large, automatically constructed KBs, like
DBpedia, Freebase and YAGO, in the late 2000s.
Knowledge extraction and KB curation are still
hot topics, and so are use cases from
semantic search to reasoning and fact checking.
A substantial amount of research is devoted to
integrating KBs with Semantic Web standards,
like RDF, OWL etc., grounded in logics.
This article refers to these standards, but
does not provide deeper coverage.
For in-depth treatment, we refer to the survey by
\citet{DBLP:journals/corr/abs-2003-02320}
with focus on modeling, representation and
logical foundations.

\noindent{\bf Database Systems (DB):}
KB technology has become increasingly important for DB tasks, most notably, data cleaning where reference repositories of entities are vital.
Highly heterogeneous ``data lakes'' 
(i.e., many independently designed DBs) 
have become the
norm rather than the exception, in both enterprises
and Open Data such as public government content or
data science resources.
Explorative querying of these data sources and 
discovering relevant pieces of knowledge and insight
pose major challenges. The necessary stage of data preparation and data integration is highly related
to methods for KB construction and curation.
Although this article places emphasis on unstructured and semi-structured web and text sources, 
many design considerations and building blocks
potentially carry over to structured data with high
heterogeneity and noisy content.

\noindent {\bf Data Mining (DM):} The data mining community pursues the strategic goal of knowledge discovery from data.
Thus, building KBs from raw data is of natural interest here.
DM research has richly contributed to methods for
knowledge extraction, from foundational algorithms to
tool suites.

\noindent {\bf Information Retrieval (IR):}
With search engines being the premier use case of
knowledge bases, the IR community has shown 
great interest in KB methodology, especially
knowledge extraction from text which is 
a major focus of this article.
For querying, answer ranking and related IR topics,
we refer the reader to the survey by
\citet{Reinanda2020} 
with focus on the IR dimension.

\noindent{\bf Machine Learning (ML):} The construction of a KB often involves ML methodology. This article presents a fair amount of these methods. Readers with good ML background may thus skip some parts, but
hopefully appreciate the discussion of how ML contributes to knowledge extraction and curation at the bigger system level.
Conversely, many ML applications leverage KBs
as a source of distant supervision.
Thus, machine learning and machine knowledge 
are complementary and synergistic.
The more a computer knows, the better it can learn;
and better learning enables acquiring more and
deeper knowledge.

\noindent{\bf Natural Language Processing (NLP):}
As KB construction often taps into textual sources,
NLP methods are a very useful asset.
This spans classical techniques like word-sense lexicons and word tagging all the way to 
recent methods on context-sensitive language models
using neural embeddings.
KBs have thus received much attention in NLP research.
Conversely, KBs are often a valuable source
of distant supervision for all kinds of NLP tasks,
including sentiment and argumentation mining,
discourse analysis, summarization and question answering.

\noindent{\bf AI and Web:}
Machine knowledge is a long-standing theme in general AI. As modern AI puts emphasis on machine learning and language understanding, the above arguments
on high relevance apply here as well.
Classical AI has intensively investigated
formalisms for knowledge representation, but hardly
embarked on large-scale KB construction.
The advent of the web, with its rich online content, has enabled the population of comprehensive KBs.
The web community is a very productive and fertile research ground on the theme of 
structured knowledge.

\clearpage\newpage
\section{Outline}

\vspace*{0.2cm}
\noindent{\bf Structure:}\\
The article is organized into ten chapters.
Chapter 2 gives foundational basics on knowledge representation
and discusses the design space for building a KB.
Chapters 3, 4 and 5 cover the methodology for constructing the core of a KB that comprises entities and their types (aka classes).
Chapter 3 discusses tapping premium sources with rich and clean semi-structured contents, and Chapter 4 addresses knowledge harvesting from textual contents.
Chapter 5 specifically focuses on the important issue
of canonicalizing entities into unique representations (aka entity linking).
Chapters 6 and 7 extend the scope of the KB by methods for discovering and extracting attributes of entities and
relations between entities.
Chapter 6 focuses on the case where a schema is designed upfront for the properties of interest.
Chapter 7 discusses the case of discovering
new
property types for attributes and relations that are not
(yet) specified in the KB schema.
Chapter 8 discusses the issue of quality assurance for KB curation and the long-term
maintenance of KBs.
Chapter 9 presents several case studies on specific KBs including
industrial knowledge graphs (KGs).
We conclude in Chapter 10 with key lessons and an
outlook on where the 
theme of machine knowledge may be heading.

\vspace*{0.2cm}
\noindent{\bf Reading Flow:}\\
Depending on prior background or specific interests,
the reader may deviate from the linear flow of the ten chapters. Readers who are familiar with
general KB concepts may skip Chapter 2.
Readers who are interested in particular 
tasks and methods for knowledge extraction
could visit each of the Chapters 3 through 6
separately.
Readers with good background in machine learning
may skip over some subsections where we introduce
basic concepts (CRF, LSTM etc.).
Advanced readers who want to know more about
open challenges could be most interested in
Chapters 7 and 8, which emphasize the need for
long-term KB maintenance and life-cycle
management.
Chapter 9 on use cases is self-contained,
but refers back to methodological constituents
in Chapters 3 to 8 -- which can be looked up
on a need-to-know basis.
Finally, the outlook on open issues in 
Chapter 10 should appeal to everybody who
is actively pursuing research on machine knowledge.

\clearpage\newpage

\chapter{Foundations and Architecture}
\label{ch:foundations}

\section{Knowledge Representation}
\label{ch2-sec:knowledgerepresentation}

Knowledge bases, KBs for short, 
comprise salient information about 
entities, semantic classes to which entities belong, attributes of entities,
and relationships between entities.
When the focus is on classes and their logical connections such as subsumption and disjointness,
knowledge repositories are often referred to as {\bf ontologies}. In database terminology, this is
referred to as the {\bf schema}.
The class hierarchy alone is often called a {\bf taxonomy}.
The notion of KBs in this article covers all these aspects of knowledge, %
including ontologies and taxonomies.

This chapter presents foundations for casting knowledge into
formal representations.
Knowledge representation has a long history, spanning decades of AI research,
from the classical model of frames
to recent variants of description logics. 
Overviews on this spectrum are given by
{
\citet{RussellNorvig2003} and \citet{StaabStuder2009}. 
}%
In this article, we restrict ourselves to the knowledge representation that has emerged as a pragmatic consensus 
for entity-centric knowledge bases
(see~\cite{DBLP:conf/rweb/SuchanekLBW19} for an extended discussion).
More on the wide spectrum of knowledge modeling
can be found in the survey 
{
by
\citet{DBLP:journals/corr/abs-2003-02320}.
}%

\subsection{Entities}

The most basic element of a KB is an \emph{entity}. 

\begin{mdframed}[backgroundcolor=blue!5,linewidth=0pt]
\squishlist
\item[ ] An {\bf entity} is any abstract or concrete object of fiction or reality. 
\squishend
\end{mdframed}

This definition 
includes people, places, products and also events and creative works (books, poems, songs etc.),
real people (living or dead) as well as fictional people (e.g., Harry Potter),
and also general concepts such as empathy
and Buddhism. 
KBs take a pragmatic approach: they model only entities
that match their scope and purpose.
A KB on writers and their biographies would include
Shakespeare and his drama Macbeth, but it may not
include the characters of the drama such as King Duncan or
Lady Macbeth. However, a KB for literature scholars --
who want to analyze character relationships 
in literature content --
should 
include all characters from 
Shakespeare's works.
A KB for a library would even include individual book copies,
readers' comments and ratings, and further information
which is irrelevant for the KB for literature scholars.
Likewise, a health KB on diseases would 
include different types of diabetes, but not
individual case histories of patients,
and a literature KB would include different editions of all
Shakespeare works but not individual copies
owned by readers (postulating that each copy has a unique
identifier, as e-books would have).
On the other hand, a KB for personal medicine
should include individual patients and clinical
trials with diabetes diagnoses and treatments.

\vspace{0.2cm}
\noindent{\bf Individual Entities (aka Named Entities):}
We often narrow down the set of entities of interest 
by emphasizing uniquely identifiable entities
and distinguishing them from general concepts.

\begin{mdframed}[backgroundcolor=blue!5,linewidth=0pt]
\squishlist
\item[ ] An {\bf individual entity} is an entity that can 
be uniquely identified against all other entities.
\squishend\end{mdframed}

To uniquely identify a location, we can use
its geo-coordinates: longitude and latitude with
a sufficiently precise resolution. To identify a person,
we would -- in the extreme case -- have to use the
person's DNA sequence, but for all realistic purpose
a combination of full name, birthplace and birthdate
are sufficient. In practice, we are typically even
coarser and just use location or person names when
there is near-universal social consensus about what or who
is denoted by the name. 
Wikipedia article names usually follow this 
principle of using names that are sufficiently unique.
For these arguments, individual entities are
also referred to as {\bf named entities}.

\vspace{0.2cm}
\noindent{\bf Identifiers and Labels:}
To denote an entity unambiguously, we need a name that can refer to only a single entity.
Such an \emph{identifier} can be a unique name, but can also
be specifically 
introduced keys such as URLs for web sites, ISBNs for books (which can even distinguish different editions
of the same book), DOIs for publications, Google Scholar URLs or ORCID IDs for authors, etc.
In the data model of the Semantic Web, {\bf RDF} (for {\bf Resource Description Framework}) \cite{W3C:RDF},
identifiers always take the form of URIs (Unique Resource Identifiers, a generalization of URLs).

\begin{mdframed}[backgroundcolor=blue!5,linewidth=0pt]
\squishlist
\item[ ] {An {\bf identifier} for an entity is a string of characters that uniquely denotes the entity.}
\squishend
\end{mdframed}

As identifiers are not necessarily of a form that is nicely readable and directly interpretable by a human,
we often want to have human-readable {\bf labels} or {\bf names} in addition.
For example, the protagonist in Shakespeare's Hamlet, Prince Hamlet of Denmark, may have an unwieldy identifier like
\url{https://www.wikidata.org/wiki/Q2447542} (in the Wikidata KB), but we want to refer to him also
by labels like ``Prince Hamlet'', ``Prince Hamlet of Denmark'' or simply ``Hamlet''.
Labels may be ambiguous, however: ``Hamlet'' alone could also refer to his father, the King of Denmark,
to the play itself (titled ``Hamlet''), 
or to the Hamlet castle (which exists in reality).
When an entity has several labels, they are called {\bf synonyms}
 or {\bf alias names} (different names, same meaning).
When different entities share a label such as ``Hamlet'', this label is a {\bf homonym} (same name, different meanings).
When an entity label appears in a text, we call that appearance a {\bf mention} of that entity. The label that appears in the mention is called the {\bf surface form} of the entity (two different mentions can use the same surface form).

We will later embark on the task of
detecting entities in input sources, from
lists and tables in web sites to natural language text. 
This entails spotting cues about the appearance
of entities.

\begin{mdframed}[backgroundcolor=blue!5,linewidth=0pt]
\squishlist
\item[ ] {A {\bf mention} of an entity in a data source (including text) is a string (including acronyms and abbreviations) intended to denote an entity.}
\squishend
\end{mdframed}
Ideally, a mention is one of the labels that the entity has associated with it. 
As such alias names may be ambiguous, 
a subsequent disambiguation stage for {\bf entity canonicalization} 
is often needed (see Chapter \ref{ch3-sec-EntityDisambiguation}).
In more difficult cases, acronyms or abbrevations may be used that are not (yet) in the list of known labels, and for the case of text as input,
mentions in the form of
pronouns like ``she, he, his, her'' etc. 
or indefinite noun phrases such as ``the singer''
need to be handled.
This case of entity mentions is known as
{\em co-references}, when a more explicit mention is usually nearby (e.g., in the preceding sentence).

\subsection{Classes (aka Types)}

Entities of interest often come in groups where all elements have a shared characteristic.
For example, \ent{Bob Dylan}, \ent{Elvis Presley} and \ent{Lisa Gerrard} are all musicians 
and singers. We capture this knowledge by organizing entities into {\bf classes}
or, synonomously, {\bf types}.

\begin{mdframed}[backgroundcolor=blue!5,linewidth=0pt]
\squishlist
\item[ ] {A {\bf class}, or interchangeably {\bf type}, 
is a named set of entities that share a common trait. An element of that set is called an 
{\bf instance} of the class.}
\squishend
\end{mdframed}

For Dylan, Presley and Gerrard, classes of interest include:
\ent{musicians} and \ent{singers} with all three as instances,
\ent{guitarists} with Dylan and Presley as instances,
\ent{men} with these two, \ent{women} containing only Gerrard, and so on.

Note that an entity can belong to multiple classes, and classes can relate to each other
in terms of their members as being disjoint (e.g. \ent{men} and \ent{women}), 
overlapping, or one subsuming the other (e.g., \ent{musicians} and \ent{guitarists}).
Classes can be quite specific; for example \ent{left-handed electric guitar players}
could be a class in the KB containing Jimi Hendrix, Paul McCartney and others.

It is not always obvious whether something should be modeled as an entity or as a class. 
We could construct, for every entity, a singleton class that contains just this entity.
Classes of interest typically have multiple instances, though.
By this token, we do not consider general concepts such as love, Buddhism or pancreatic cancer as classes, unless we were interested
in specific instances (e.g., the individual cancer of one particular patient).

There are borderline cases, depending on the scope and purpose of the KB.
For example,
when a health KB is about biomedical knowledge without any interest in individual patients,
there is little point in viewing \ent{pancreatic cancer} as a (singleton) class without instances; it is better modeled
as an \ent{entity}, belonging to classes like %
\ent{pancreas disorders} and \ent{diseases}. 
A pragmatic guideline could be: if we are interested in the plural form of a label (expressed by a word or phrase,
such as ``disorders''),
then it should be considered as a class. 
Ultimately, though, the modeling choice of 
class vs. entity is up to the KB architect 
and depends on the intended use case(s).

\vspace{0.2cm}
\noindent{\bf Taxonomies (aka Class Hierarchies):}

By relating the instance sets of two classes, we can specify
invariants that must hold between the classes, most notably,
subsumption, also known as the {\bf subclass}/{\bf superclass} relation.
By combining these pairwise invariants across all classes,
we can thus construct a {\bf class hierarchy}.
We refer to this aspect of the KB as a {\bf taxonomy}.

\begin{mdframed}[backgroundcolor=blue!5,linewidth=0pt]
\squishlist
\item[ ] Class $A$ is a {\bf subclass} of (is subsumed by) class $B$ if 
all instances of 
$A$ 
must also be instances 
of $B$.
\squishend
\end{mdframed}

For example, the classes \ent{singers} and \ent{guitarists}
are subclasses of \ent{musicians} because every singer and every guitarist is a musician. 
We say that class X is a {\em direct} subclass of Y if there is no other class that
subsumes X and is subsumed by Y. %
Classes can have multiple superclasses, but there should not be any cycles in
the subsumption relation. 
For example, \ent{left-handed electric guitar players} are a subclass of
both \ent{left-handed people} and \ent{guitarists}.

\begin{mdframed}[backgroundcolor=blue!5,linewidth=0pt]
\squishlist
A {\bf taxonomy} is a directed acyclic graph, where the nodes are classes and there is an edge from class X to class Y if X is a 
direct 
subclass of Y.
\squishend
\end{mdframed}

Note that the subsumptions in the taxonomy do not just {\em describe} the current instances of 
class pairs, but actually {\em prescribe} that the invariants hold for all
possible instance sets. So the taxonomy acts like a database {\em schema}, and is instrumental
for keeping the KB consistent.
In database terminology, a subclass/superclass pair is also called an {\bf inclusion dependency}.
In  the Semantic Web, the {\bf RDFS} extension of the RDF model (RDFS = RDF Schema)
allows specifying such constraints \cite{W3C:RDFS}.
Other Semantic Web models, notably, OWL, support also disjointness constraints (aka mutual exclusion),
for example, specifying that \ent{men} and \ent{women} are disjoint.

\begin{figure} [tb]
  \centering
   \includegraphics[width=1.0\textwidth]{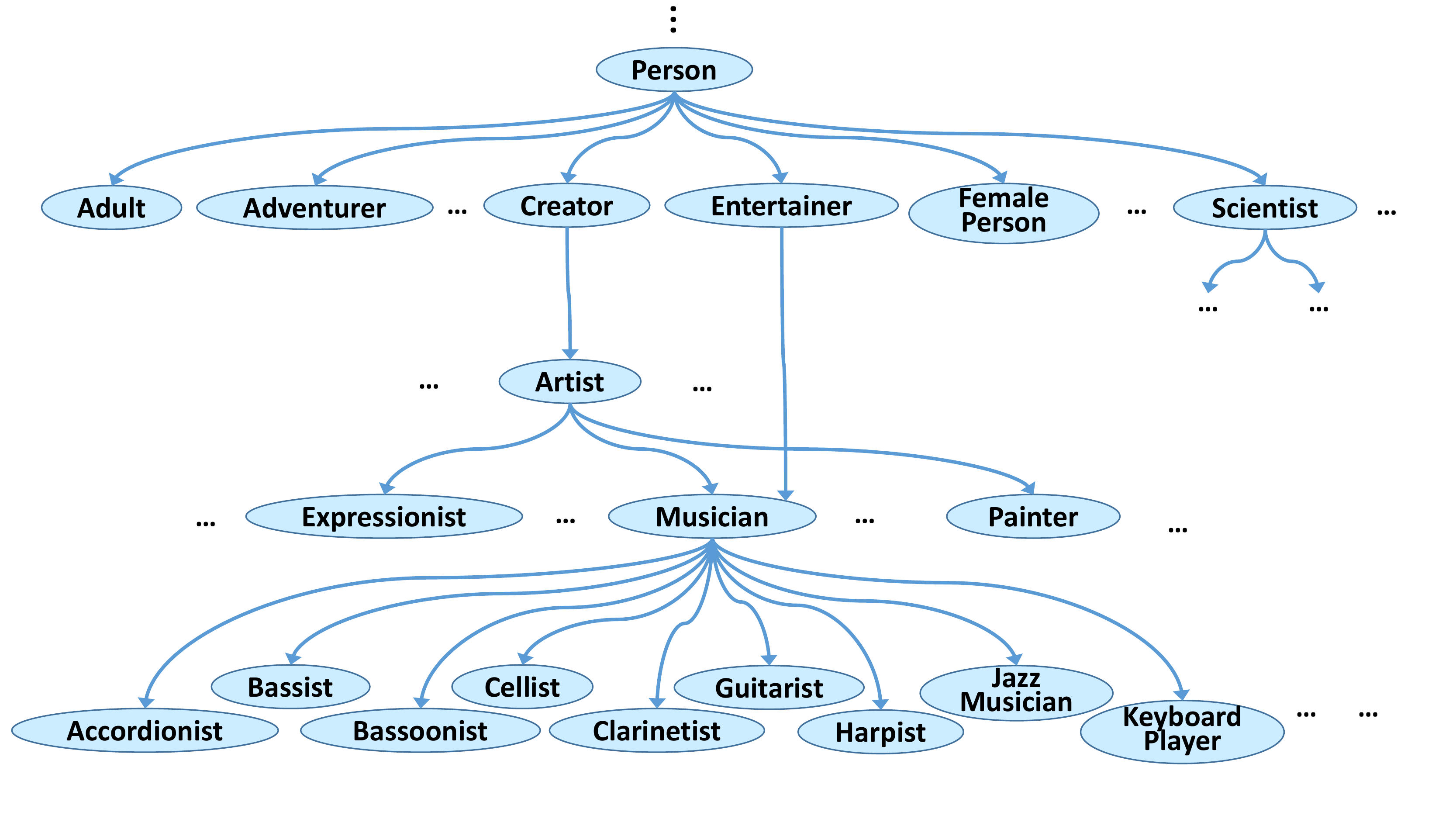}
   \vspace*{-0.4cm}
      \caption{Excerpt from the WordNet taxonomy}
      \label{ch2-fig-WordNet-excerpt}
\end{figure}

One of the largest taxonomic repositories 
is the {\bf WordNet} lexicon
(\citet{Fellbaum1998}),
comprising more than 100,000 classes.
Figure \ref{ch2-fig-WordNet-excerpt} visualizes an excerpt of the WordNet taxonomy.
The nodes are classes, called {\em word senses} in WordNet,
the edges indicate subsumption. 
In linguistic terminology, this lexical
relation is called {\bf hypernymy}:
an edge connects a more special class,
called {\em hyponym}, 
with a generalized class, called {\em hypernym}.
\cite{DBLP:series/synthesis/2016Gurevych} gives
an overview of this kind of lexical resources.
Further examples of (potentially unclean) taxonomies include the Wikipedia category system or 
product catalogs.

\vspace{0.2cm}
\noindent{\bf Subsumption vs. Part-Whole Relation:}\\
Class subsumption 
should not be confused or conflated
with the relationship between parts and wholes. 
For example, a \ent{soprano saxophone} is {\em part of} a \ent{jazz band}. 
This does not mean, however, that every soprano saxophone {\em is a} jazz band. 
Likewise,
New York is part of the USA, but \ent{New York} is not a subclass of the \ent{USA}.

\vspace{0.2cm}
\noindent{\bf Instance-of vs. Subclass-of:}\\
Some KBs do not make a clear distinction between classes and instances, and they collapse the
{\em instance-of} and {\em subclass-of} relations into a single {\bf is-a hierarchy}.
Instead of stating that \ent{Bob Dylan} is an instance of the class \ent{singers} and that
\ent{singers} are a subclass of \ent{musicians}, they would view all three as general entities
and connect them in a generalization graph.

\subsection{Properties of Entities: Attributes and Relations}
\label{ch2-subsec:properties}

Entities have {\bf properties} such as birthdate, birthplace and height of a person,
prizes won, 
books, songs or software
written, and so on.
KBs capture these in the form of mathematical relations:

\begin{mdframed}[backgroundcolor=blue!5,linewidth=0pt]
\squishlist
\item[ ] A {\bf relation} or {\bf relationship} for the instances of classes\\ 
\mbox{$C_1, ..., C_n$}
is a subset of the Cartesian product $C_1 \times ... \times C_n$, along
with an identifier (i.e., unique name) for the relation.
\squishend
\end{mdframed}

For example, we can state the birthdate and birthplace of Bob Dylan in the relational form:\\
\hspace*{1cm}  $\langle$\ent{Bob Dylan}, \ent{24-5-1941}, \ent{Duluth (Minnesota)}$\rangle$ $~\in$ \ent{birth}\\
where \ent{birth} is the identifier of the relation.
This instance of the \ent{birth} relation is a ternary tuple, that is, it has three arguments:
the person entity, the birthdate, and the birthplace. The underlying Cartesian product of the relation
is \ent{persons} $\times$ \ent{dates} $\times$ \ent{cities}.

In logical notation, we also write $R(x_1,...,x_n)$ instead of $\langle x_1,...,x_n\rangle \in R$,
and we refer to $R$ as a {\bf predicate}.
The number of arguments, $n$, is called the \emph{arity} of $R$.
The domain $C_1 \times ... \times C_n$ is also called the relation's
{\bf type signature}.

\begin{mdframed}[backgroundcolor=blue!5,linewidth=0pt]
\squishlist
\item[ ] The {\bf schema} of a KB consists of 
\squishlist
\item a 
set of types, organized into a taxonomy, and
\item a set of property types -- attributes and relations -- with type signatures.
\squishend
\squishend
\end{mdframed}

As most KBs are of encyclopedic nature, the instances of a relation are 
often referred to as
{\em facts}. We do not want to exclude knowledge that is not fact-centric
(e.g., commonsense knowledge with a socio-cultural dimension);
so we call relational instances more generally
{\bf statements}. The literature also speaks of {\bf facts}, and sometimes uses the
terminology {\em assertion} as well. For this article, the three terms {\em statement}, {\em fact}
and {\em assertion} are more or less interchangeable.

In logical terms, statements are grounded expressions of
first-order predicate logic (where ``grounded'' means that the expression has no variables).
In the KB literature, the term ``relation'' is sometimes used to denote both
the relation identifier $R$ and an instance $\langle x_1,...,x_n\rangle$.
We avoid this ambiguity, and more precisely speak of the relation and its (relational) tuples.

\subsubsection{Attributes of Entities}

In the above example about the \ent{birth} relation, we made use of the class \ent{dates}.
By stating this, we consider individual dates, such as \ent{24-5-1941}, entities. %
It is a design choice whether we regard numerical expressions like dates, heights or monetary
amounts as entities or not. Often, we want to treat them simply as {\bf{values}} for which we do not have any additional
properties.
In the RDF data model, such values are called {\bf literals}.
Strings such as nicknames of people (e.g., 
``Air Jordan'')
are another
frequent type of literals.

We introduce a special class of relations with two arguments
where the first argument is an entity of interest, such as \ent{Bob Dylan} or
\ent{Michael Jordan} (the basketball player), and the second argument is a value
of interest, such as their heights ``171 cm'' and ``198 cm'', respectively. 

The case for binary relations with values as second argument largely corresponds
to the modeling of entity {\bf attributes} in database terminology. 
Such relations are restricted to be {\em functions}: for
each entity as first argument there is only one value for the second argument.
We denote attributes in the same style as other relational properties, but we use
numeric or string notation to distinguish the literals from entities:
\begin{center}
\begin{tabular}{| l l |}\hline
$\langle$\ent{Michael Jordan}, 198cm$\rangle$ $\in$ \ent{height}
& $~~~$ or  \\
\ent{height} (\ent{Michael Jordan}, 198cm) & \\ \hline
$\langle$\ent{Michael Jordan}, ``Air Jordan''$\rangle$ $\in$ \ent{nickname}
& $~~~$ or  \\
\ent{nickname} (\ent{Michael Jordan}, ``Air Jordan'') & \\ \hline
\end{tabular}
\end{center}

\subsubsection{Relations between Entities}

In addition to their attributes, entities are characterized by their {\bf relationships} with
other entities, for example, the birthplaces of people, prizes won, songs written or
performed, and so on.
Mathematical relations over classes, as introduced above, are the proper formalism
for representing this kind of knowledge.
The frequent case of {\bf binary relations} captures the relationship between exactly two entities.

Some KBs focus exclusively on binary
relations, and the Semantic Web data model RDF has specific terminology and
formal notation for this case of so-called 
{\bf subject-predicate-object triples}, or {\bf SPO triples},  or merely {\bf triples} for short.

\begin{samepage}
\begin{mdframed}[backgroundcolor=blue!5,linewidth=0pt]
\squishlist
\item[ ] The RDF model restricts the three roles in 
a {\bf subject-predicate-object (SPO) triple} as follows: 
\squishlist
\item S must be a URI identifying an entity,
\item P must be a URI identifying a relation, and 
\item O must be a URI identifying an entity
for a relationship between entities,
or a literal denoting the value of an attribute.
\squishend
\squishend
\end{mdframed}
\end{samepage}

As binary relations can be easily cast into a labeled graph 
 -- with node labels for S and O and edge labels for P --
knowledge bases that focus on SPO triples are widely referred to
as {\bf knowledge graphs}.
SPO triples are often written in the form $\langle$ \ent{S, P, O} $\rangle$ 
or as \ent{S P O} with
the relation between subject and object.
Examples of SPO triples are:

{\small\tt
\begin{center}
\begin{tabular}{|lll|}\hline
Bob Dylan $~$  & married to $~$ & Sara Lownds \\ \hline
Bob Dylan $~$ & composed $~$ & Blowin' in the Wind \\ \hline
Blowin' in the Wind $~$ & composed by $~$ & Bob Dylan \\ \hline
Bob Dylan $~$ & has won $~$ & Nobel Prize in Literature \\ \hline
Bob Dylan $~$ & type $~$ & Nobel Laureate \\ \hline
\end{tabular}
\end{center}
}

The examples also illustrate the notion of {\bf inverse relations}:
\ent{composed by} 
is inverse to
\ent{composed}, and can be written also as \ent{composed}$^{-1}$:
$$\langle S,O \rangle \in P ~ \Leftrightarrow ~ \langle O,S  \rangle \in P^{-1}.$$
The last example in the above table
shows that an entity belonging to a certain class
can also be written as a binary relation, with
{\bf type} as the predicate, following the RDF standard.
It also shows that knowledge can sometimes be expressed either
by class membership or by a binary-relation property.
In this case, the latter adds information (Nobel Prize
in {\em Literature}) and the former is convenient for
querying (about all Nobel Laureates).
Moreover, having a class \ent{Nobel Laureate}
allows us to define further relations and attributes
with this class as domain.
To get the benefits of all this, we may want to have 
both of the example triples in the KB.

An advantage of binary relations is that they can express facts in a self-contained manner,
even if some of the arguments for a higher-arity relation is missing or the instances
of the relations are only partly known.
For example, if we know only Dylan's birthplace but not his birthdate (or vice versa),
capturing this in the ternary relation \ent{birth} is a bit awkward as the unknown argument
would have to be represented as a {\em null value} (i.e., a placeholder for an unknown
or undefined value). In database systems, null values are standard practice, but they
often make things complicated. In KBs, the common practice is to avoid null values
and prefer binary relations where we can simply have a triple for the known argument
(birthplace) and nothing else.

\subsubsection{Higher-Arity Relations}
\label{ch2-subsubsec:higherarity-relations}

Some KBs emphasize binary relations only,
leading to the notion of {\bf knowledge graphs} (KGs).
However, ternary and higher-arity relations
can play a big role, and these cannot
be directly captured by a graph.

At first glance, it may seem that we
can always decompose a higher-arity
relation into multiple binary relations.
For example, instead of introducing
the ternary relation
$birth: person \times date \times city$,
we can alternatively use two binary relations:
$birthdate: person \times date$ and
$birthplace: person \times city$.
In this case, no information is lost
by using simpler binary relations.
Another case where such decomposition
works well is the relation that contains
all tuples of parents, sons and daughters:
$children: person \times boys \times girls$.
This could be equally represented by two 
separate relations $sons: person \times boys$
and $daughters: person \times girls$.
In fact, database design theory tells
us that this decomposition is a better
representation, based on the notion
of multi-valued dependencies \cite{DBLP:books/daglib/0011318}.

However, not every higher-arity relation
is decomposable without losing information.
Consider a quarternary relation
$won: person \times award \times year \times field$
capturing who won which prize in which
year for which scientific field. 
Instances would include \\
\hspace*{1cm}{\small\tt
$~~~~~\langle$ \ent{Marie Curie}, \ent{Nobel Prize}, \ent{1903}, \ent{physics}  $\rangle$\\
}
and\\
\hspace*{1cm}{\small\tt
$~~~~~\langle$ \ent{Marie Curie}, \ent{Nobel Prize}, \ent{1911}, \ent{chemistry}  $\rangle$.\\
}
If we simply split these 4-tuples into a set
of binary-relation tuples (i.e., SPO triples), we
would end up with:\\
{\small\tt
$~~\langle \ent{Marie Curie}, \ent{Nobel Prize} \rangle$,
$\langle \ent{Marie Curie}, \ent{1903} \rangle$,
$\langle \ent{Marie Curie}, \ent{Physics} \rangle$,\\
$~~\langle \ent{Marie Curie}, \ent{Nobel Prize} \rangle$,
$\langle \ent{Marie Curie}, \ent{1911} \rangle$,
$\langle \ent{Marie Curie}, \ent{Chemistry} \rangle$.\\
}
Leaving the 
technicality
of two identical tuples
aside, the crux here is
that we can no longer reconstruct in which 
year Marie Curie won which of the two prizes.
Joining the binary tuples using database
operations would produce spurious tuples,
namely, all four combinations of
1903 and 1911 with physics and chemistry.

The Semantic Web data model RDF and its associated W3C standards (including the SPARQL
query language) support only binary relations. They therefore exploit clever ways of
encoding higher-arity relations into a binary representation, based on techniques related to
{\bf reificiation} \cite{W3C:RDFprimer}. Essentially, each instance of the higher-arity relation
is given an identifier of type \ent{statement} and that identifier is combined with the original relation's
arguments into a set of binary tuples.

\begin{mdframed}[backgroundcolor=blue!5,linewidth=0pt]
\squishlist
\item[ ] For the $n$-ary relation instance $R (X_1, X_2 \dots X_n)$
the {\bf reified representation} consists of the set of binary instances\\
\hspace*{0.5cm} $type (id,statement)$, 
$arg_1(id,X_1), arg_2(id,X_2) \dots arg_n(id,X_n)$ \\
where $id$ is an identifier.
\squishend
\end{mdframed}

With this technique,
the triple 
$\langle$\ent{id type statement}$\rangle$ asserts the existence of the
higher-arity tuple, and the additional triples fill in the
arguments. 
In some KBs, the technique is referred to as
{\bf compound objects}, as the $\langle$\ent{id type statement}$\rangle$
is expanded into a set of
{\em facets}, often called {\bf qualifiers}, with the number of facets
even being variable 
(see, e.g., \cite{DBLP:conf/semweb/HernandezHK15}).
Reification can be applied to binary relations as well
(if desired): the
representation of $\langle$\ent{S P O}$\rangle$ then becomes
$\langle$\ent{id type statement}$\rangle$, 
$\langle$\ent{id hasSubject S}$\rangle$, 
$\langle$\ent{id hasPredicate P}$\rangle$, 
$\langle$\ent{id hasObject O}$\rangle$.
Our use case for reification is higher-arity relations, though,
most importantly, to capture events and their different aspects.
Attaching provenance or belief information to statements is another case for reification.

The identifier \ent{id} could  be a number or a URI (as required by RDF).
The names of the facets of arguments $arg_i ~ (i=1..n)$ can be arbitrarily chosen, but
often capture certain properties that can be aptly reflected in their names.
For example, a tuple for the higher-arity relation
\ent{wonAward} 
may result in the following triples: 

{\small\tt
\begin{center}
\begin{tabular}{|lll|}\hline
id $~$  & type $~$ & statement \\ \hline
id $~$  & hasPredicate $~$ & wonAward \\ \hline
id $~$  & winner $~$ & Marie Curie \\ \hline
id $~$  & award $~$ & Nobel Prize \\ \hline
id $~$  & year $~$ & 1903 \\ \hline
id $~$  & field $~$ & physics \\ \hline
\end{tabular}
\end{center}
}

The example additionally includes a triple that encodes the name of the property \ent{wonAward} for which the n-tuple holds.
Strictly speaking, we could then drop the \ent{id type statement} triple without losing information, and the remaining triples
are a typical knowledge representation for n-ary predicates
(e.g., for events):
one triple for the predicate itself and one for each of the n-ary predicate arguments.
If we want to emphasize two of the arguments as
major subject and object, we could also use
a hybrid form with triples like
$\langle$\ent{id hasSubject Marie Curie}$\rangle$,
$\langle$\ent{id hasObject Nobel Prize}$\rangle$,
$\langle$\ent{id year 1903}$\rangle$,
$\langle$\ent{id field physics}$\rangle$.

The advantage of the triples 
representation is that it stays in the world
of binary relations, and the notion of a {\em knowledge graph}
still applies. In a graph model, the SPO triple that encodes the
existence of the original n-ary $R$ instance is often called a
{\em compound node} (e.g., in Freebase and in Wikidata), and serves as 
the ``gateway'' to the {\em qualifier triples}.

The downside of reification and related techniques 
for casting n-ary relations into RDF
is that they make
querying more difficult if not to say tedious.
It requires more joins, and considering paths and not just
single edges when dealing with compound nodes in the graph model.
For this reason, some KBs have also pursued hybrid representations
where for each higher-arity relation, the most salient pair of
arguments are represented as a standard binary relation and 
reification is used only for the other arguments.

\subsection{Canonicalization}
\label{sec:ch2-canonicalization}

An important objective for a clean knowledge base 
is the uniqueness of the subjects, predicates and objects in SPO triples
and other relational statements.
We want to capture every entity and every fact about it exactly once,
just like an enterprise company should contain every customer and 
her orders and account balance once and only once.
As soon as redundancy creeps in, this opens the door for 
variations of the same information and hence potential inconsistency.
For example, if we include two entities in our KB, \ent{Bob Dylan} and
\ent{Robert Zimmerman} (Dylan's real name) without knowing that they
are the same, we could attach different facts to them that may
eventually contradict each other. 
Furthermore, we would distort the results of counting queries (counting two people instead of one person).
This motivates the following {\bf canonicalization} principle:

\begin{mdframed}[backgroundcolor=blue!5,linewidth=0pt]
\squishlist
\item[ ] Each entity, class and property in a KB is {\bf canonicalized}
by having a unique identifier and being included exactly once. 
\squishend
\end{mdframed}

For entities this implies the need for 
{\bf named entity disambiguation}, also known as {\bf entity linking}
(\cite{Shen:TKDE2015}). For example, we need to infer that
Bob Dylan and Robert Zimmerman are the same person and should
have him as one entity with two different labels rather than
two entities with different identifiers. 
The same principle should hold for classes, for example, avoiding
that we have both \ent{guitarists} and \ent{guitar players},
and for properties as well.

Entity linking, as part of the KB construction process, has different flavors depending on the input setting. In this article, we will mostly discuss
the case where the input is web pages or text documents together with an initial reference repository of target entities, like an early-stage core KB. Other settings include the case where the inputs is two or more %
datasets, such as independently developed databases, without any pre-existing reference point. In this case, the goal is to
identify equivalence classes of entities across all data sources. This task is known as 
{\bf entity matching},
{\bf record linkage} or
{\bf deduplication}
(and sometimes {\em entity resolution}) 
\cite{DBLP:journals/corr/abs-1905-06397}.
Methods for tackling it largely overlap with those
for entity linking. We will come back to this in
Chapter \ref{ch3-sec-EntityDisambiguation}.

We strive to avoid redundancy and the  resulting ambiguities and
potential inconsistencies.
However, this goal is not always perfectly achievable in the entire life-cycle
of KB creation, growth and curation.
Some KBs settle for softer standards and allow diverse representations for
the same facts to co-exist, effectively demoting entities and relations into
literal values. Here is an example of such a softer 
(and hence less desirable, but still useful) representation:

{\small\tt
\begin{center}
\begin{tabular}{|lll|}\hline
Bob Dylan  & has won  & ``Nobel Prize in Literature'' \\ \hline
Bob Dylan  & has won  & ``Literature Nobel Prize'' \\ \hline
Bob Dylan  & has won award  & ``Nobel'' \\ \hline
\end{tabular}
\end{center}
}

\subsection{Logical Invariants}

In addition to the {\em grounded statements} about entities, classes
and relational properties, KBs can also contain {\em intensional knowledge}
in the form of logical constraints and rules.
This theme dates back to the pioneering work
on hand-crafted KBs like Cyc 
(\citet{DBLP:journals/aai/GuhaL91})
and SUMO
(\citet{DBLP:conf/fois/NilesP01}), 
and is
of great importance for automated knowledge harvesting as well.
The purpose of {\bf constraints}
is to enforce the consistency of the KB:
grounded statements that violate a constraint cannot be entered. For example, we do not allow
a second birthdate for a person, as the 
\ent{birthdate} property is a function, and
we require creators of songs to be musicians
(including composers and bands).
The former is an example of a {\em functional dependency},
and the latter is an example of a {\em type constraint}.
We discuss consistency constraints and their
crucial role for KB curation in 
Chapter \ref{chapter:KB-curation} (particularly, Section~\ref{sec:constraints}).
The purpose of {\bf rules} is to derive additional
knowledge that logically follows from what the KB already contains. For example,
if Bob is married to Sara, then Sara is married to Bob,
by
the symmetry of the \ent{spouse} relation). We discuss rules in Section~\ref{sec:rules}.

\section{Design Space for Knowledge Base Construction}

Machines cannot create any knowledge on their own; all knowledge about our
world is created by humans (and their instruments)
and documented in the form of encyclopedia, 
scientific publications, books, daily news, all the way to contents in
online discussion forums and other social media.
What machines actually do to construct a knowledge base is to tap
into these sources, {harvest} their nuggets of knowledge,
and refine and semantically organize them into a formal knowledge representation.
This process of distilling knowledge from online contents is best characterized as
{\bf knowledge harvesting} \cite{DBLP:conf/pods/WeikumT10,DBLP:journals/debu/WeikumHS16,DBLP:series/lncs/WeikumHS19}.

This big picture unfolds 
into a spectrum
of specific goals, and the design space for
knowledge harvesting architectures can be organized
along different dimensions.
In Subsection \ref{ch2-subsec:requirements}, we discuss
requirements for an ideal KB and trade-offs that must
be considered in design choices and priorities.
Subsections \ref{ch2-subsec:inputs} and
\ref{ch2-subsec:outputs} discuss options for
selecting inputs and targeting outputs of the
KB creation process.
Subsection \ref{ch2-subsec:methods} outlines
design alternatives for the
methodologies to extract and infer the
desired outputs from
the chosen input sources.

\subsection{Requirements}
\label{ch2-subsec:requirements}

An ideal KB would have no errors, comprise all entities and types of interest,
contain expressive statements that cover all aspects of interest,
and maintain itself automatically by adding new entities, types and statements
and removing obsolete data
throughout its life-cycle.
All this should be fully automated, without manual supervision and human intervention.
Unfortunately, satisfying all of these desiderata together is wishful thinking.
KB projects have to understand trade-offs, and differ in how they prioritize
requirements. The following list gives the most important dimensions in this
space of desiderata.

\squishlist
\item {\bf Correctness:}
Entities in the KB should indeed exist in the real-world domain of interest,
and each entity should be captured exactly once (i.e., with proper
canonicalization, after de-duplication). 
Likewise, all associated properties (incl. types) should correspond to
valid real-world facts.
\item {\bf Coverage:}
This goal requires capturing {\em all} entities of interest and 
{\em all} their properties that are relevant for the use case(s). 
Regarding properties, this applies to all attributes of entities and
relations between entities that are foreseen in the {\em KB schema}.
The schema is another design point for the KB architect, where
coverage is a key issue. A KB can be complete on a small set of properties
with every entity being covered, but may have a fairly limited schema.
For people, for example, sometimes basic properties are sufficient,
like birthdate, birthplace,
spouses, children, profession, awards, and a generic \ent{knownFor} property.
When the KB requires more comprehensive knowledge, then high coverage
additionally 
means to capture a {\em wider set of property types} (e.g., about education,
career, important events, notable particularities, and more).
\item {\bf Expressiveness:}
Entities organized into classes is the backbone of any KB and more or less mandatory. 
By adding statements about attributes and relations, we can go deeper and make the KB more expressive. 
With more property types, the depth of the KB increases further. Some KBs may consider only truly binary relations for simplicity, whereas others aim to capture also higher-arity relations (such as events and their associated roles). This is regardless of whether the representation is in triple form with reification or qualifiers (see Section \ref{ch2-subsubsec:higherarity-relations}) or as unrestricted relations.
Finally, advanced KBs would require also statements about the temporal scope of when triples or tuples
are valid,
often in combination with versioning. 
This kind of ``higher-order knowledge'' could be
further extended by adding scopes and degrees of belief, for example: {\em many} Americans {\em believe} that Norway is a member of the European Union, and {\em some} Chinese {\em believe}
that Munich is the capital of Switzerland.
However, this article focuses on factual knowledge, and disregards the direction of 
capturing claims and beliefs.
\item {\bf Agility:}
As the world keeps changing, a KB needs to be continuously updated
and maintained over a long time horizon. By KB agility we refer to the
ability to keep up with evolving domains of interest in a timely manner, while
preserving the required levels of correctness, coverage and expressiveness.
The {\em life-cycle management} needs to support adding statements, removing
obsolete ones, and also changing the KB schema to revise and extend the scope of
property types. Furthermore, the underlying algorithmic methods must be
extensible as well, for example, to integrate new kinds of sources and
new extractors.
Ideally, all this is implemented in an {\em incremental manner},
without having to re-build the entire KB.
\item {\bf Efficiency:}
Building a KB at scale, with many millions of entities and billions of statements,
is a challenging effort. This requires both {\em time} efficiency and {\em cost} efficiency.
Simple methods may take weeks or months to complete the KB creation, which would
be unacceptable in some settings. Even if the underlying methods allow
scaling out the computation across many computers, there is a monetary cost to
be paid (e.g., for cloud server rental) 
which could be prohibitive. 
It may be crucial to employ highly efficient methods for knowledge extraction
and KB creation, and judiciously selecting only the highest-yield input sources,
for acceptable run-time at affordable cost.
This holds for both building the entire KB and incremental maintenance -- either in periodic batches or with continuous updates. 
\item {\bf Scrutability:}
For each statement in the KB, we need to be able to track down the evidence
from where it was extracted or inferred. 
This kind of {\em provenance tracking} is crucial for several reasons.
First, it helps to {\em explain} the context and validity of a statement to KB curators
or crowdsourcing workers who contribute to scrutinizing the KB and 
quality assurance.
As some statements may actually be invalid, this process should ideally include
multiple sources of evidence or 
counter-evidence, even when the
statement itself is extracted solely from one source.
Second, provenance is essential for {\em maintaining} an evolving KB where 
newly added statements may conflict with existing ones and require 
conflict resolution, for example, by invalidating (and discarding) prior
statements. Keeping (timestamped) {\em versions} of statements is one way to go about this issue,
but there is still a need for identifying the currently valid master version of the KB.
\item {\bf Automation vs. Human in the Loop:}
It is
desirable to automate the process of KB creation and curation as much as possible. Ideally, no human effort would be required at all.
However, this is unrealistic if we also strive
for the other requirements in this list of desiderata. 
Human inputs may range from manually specifying the schema for types and properties of interest, along with consistency constraints, to judiciously selecting input sources, all the way to providing the system with labeled training samples for supervised knowledge acquisition.
We can obtain this suite of inputs from
highly qualified knowledge engineers and domain experts, or from crowdsourcing workers at larger scale. 
All this is about the trade-off between achieving high correctness, coverage and expressiveness, and
limiting or minimizing the monetary cost of
building and maintaining the KB.
\squishend

The most crucial trade-off among these requirements is between
correctness and coverage. As no method for knowledge harvesting
can be perfect, striving for high degrees of correctness implies
having to sacrifice on the coverage dimension: capturing fewer statements,
for the most salient property types, or even capturing fewer entities.
Conversely, striving for complete coverage implies ingesting also 
some fraction of spurious statements. 
This trade-off, usually referred to as {\em precision-recall trade-off},
is prevalent and inevitable in all kinds of classification tasks,
search engines (e.g., answer ranking and query suggestions), and recommender systems.
For KB quality, we define precision and recall as follows.

\begin{mdframed}[backgroundcolor=blue!5,linewidth=0pt]
\squishlist
\item[ ] The {\bf precision} of a KB of statements is the ratio
$$\frac{\#~correct~statements~in~KB}{\#~statements~in~KB}.$$
The {\bf recall} of a KB of statements is the ratio
$$\frac{\#~correct~statements~in~KB}{\#~correct~statements~in~real~world}.$$
\squishend
\end{mdframed}

Precision can be evaluated by inspecting a KB alone, but recall can only
be estimated as the complete real-world knowledge is not known to us.
However, we can sample on a per-entity basis and compare what a KB
knows about an entity (in the form of relational tuples) against
what a human reader can obtain from
high-coverage online sources about the entity.

\vspace*{0.2cm}
\noindent{\bf Coping with Trade-offs:}\\
Often, the prioritization of requirements and corresponding design choices
depend on the intended downstream applications. 
For example, when the use case is a search engine for the Internet or
broad usage in an enterprise intranet, the most important point is to
identify entities and classes (or categories and labeled lists) in the users' keywords, telegraphic phrases
or 
full
questions, matching them against entities in documents and the KB 
(see, e.g., \cite{DBLP:conf/www/LinPGKF12,DBLP:journals/corr/abs-2001-04828}).
Statements on relationships between entities would be needed only in 
complex queries which are infrequent on the Internet. Hence, KBs for
search engines may well restrict their properties to basic ones of wider interest,
like spouses and awards (relevant for search about entertainment celebrities), 
or songs and albums for musicians (but not composer, instrumentation, lyrics etc.)

Answering Internet queries may even tolerate some
errors in the entity identification and in the KB itself, 
as long as there is a %
good match between user input and search-engine results.
For example, users searching for ``Oscar winners''
and users looking for ``Academy Award winners'' should
ideally be provided with the same answers, if the KB can
tell that ``Oscar'' and ``Academy Award'' denote the same entity.
But if the KB lacks this knowledge and assumes two different entities, all users would still be satisfied with ample of good results regardless of their query formulations,  
because of the topic's large coverage on the Internet.
Along similar lines, if the KB is leveraged by a
machine learning system to generate
recommendations for new movies, the recommendations 
would still be very good despite the KB's limitations or
even a small amount of incorrect entries.
Of course, this is not to say that search engines deliberately accept
errors.
The big stakeholders
put enormous effort into KB quality assurance,
and this is indeed crucial for some use cases.
Nevertheless, search engines 
have a regime of popular queries where 
KB limitations do not affect the end-user experience too badly.

The situation is different, though, for 
advanced
usage, for example, by
patent researchers, financial enterprises, or medical professionals, and possibly even in online shopping. 
Such power users often need near-complete answers (and not just 
diversified top-10), to form a bigger picture
and compare results for groups of entities. For example, it is crucial 
to recognize that Xarelto and Rivaroxaban are different names for the
same medication (brand name and biochemical agent)
and that it belongs to the class of oral anticoagulants (as opposed to other anticoagulants like Warfarin and Heparin).
So the KB must be of very high quality in terms of avoiding entity duplicates
(under different names), the proper typing of entities, and the associated facts
such as treatments and known risks and side effects. When in doubt, coverage must be
sacrificed to avoid incorrect data in the KB.
The same rationale and prioritization of correctness applies to all use cases
of analytic nature, where entity-associated results are grouped, aggregated
and compared.

\begin{figure} [tb!]
  \centering
   \includegraphics[width=1.0\textwidth]{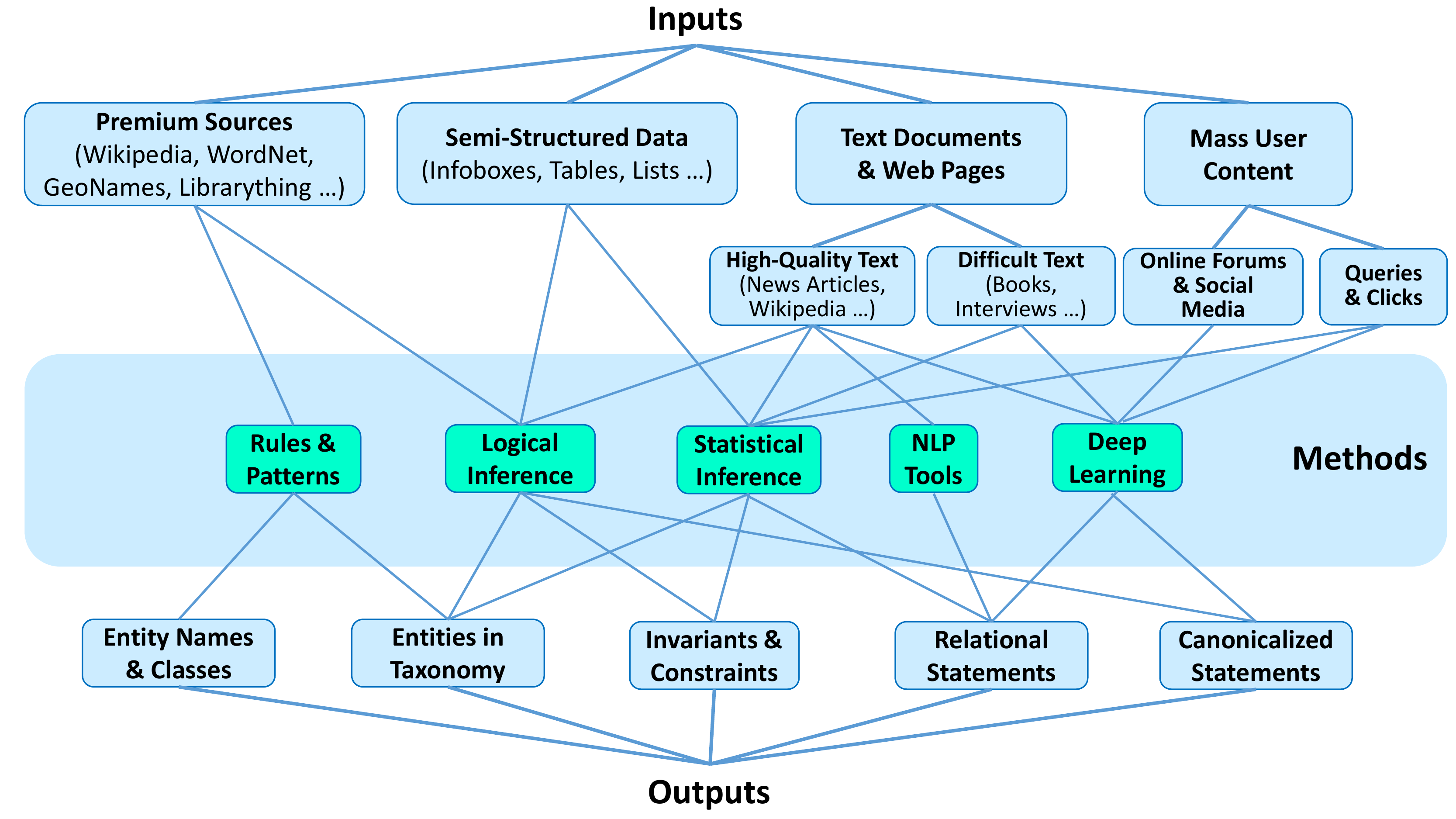}
      \caption{Design space for knowledge harvesting}      %
      \label{ch2-fig-knowledgeharvesting-designspace}
\end{figure}

Obviously, we cannot provide recipes for KB building across 
a wide spectrum of different settings;
there is no universally viable 
solution.
Instead, this article presents {\em building blocks} that KB architects can consider for their specific endeavors. 
Nevertheless, some of the introduced methods are more versatile than others, and thus can be seen at work
in many KB projects across a wider spectrum.
In the following sections, we outline a 
selected set of 
design points regarding the 
choice of input sources, the output qualities, and the algorithmic methods
for extracting and organizing knowledge. 
Figure \ref{ch2-fig-knowledgeharvesting-designspace}
gives a pictorial overview, but is not meant to
be (near-) exhaustive.

\subsection{Input Sources}
\label{ch2-subsec:inputs}

There is a wide spectrum of input sources to be considered.
The top part of Figure \ref{ch2-fig-knowledgeharvesting-designspace}
shows some notable design points, with difficulty increasing from left to right.
The difficulties arise from the decreasing ratio of valuable knowledge to noise in the
respective sources.

To build a high-quality KB, we  advocate starting with the cleanest sources,
called {\bf premium sources} in the figure.
These include well-organized and curated encyclopedic content like Wikipedia.
For example, Wikipedia's set of article names is a great source for KB construction, as these
names constitute the world's notable entities in a reasonably standardized form:
millions of entities with human-readable unique 
labels. 
Harvesting these entities first (and cues about their classes, e.g., via Wikipedia
categories)  forms a strong backbone for subsequent extensions and refinements.
This design choice can be seen as an instantiation of the 
folk wisdom to 
``pick low-hanging fruit'' first, widely applied in systems engineering.
Beyond Wikipedia as a general-purpose source, knowledge harvesting should generally
start with the most authoritative high-quality sources for the domains of interest.
For example, for a KB about movies, it would be a mistake to disregard
IMDB 
as it is the
world's largest and cleanest (manually constructed) repository of
movies, characters, actors, 
keywords and phrases about movie plots, etc. 
Likewise, we must not overlook sources like 
GeoNames 
and
OpenStreetMap 
for
geographic entities, 
GoodReads and Librarything for books,
MusicBrainz 
(or even Spotify's catalog)
for music, DrugBanks 
for medical drugs, and so on.
Note that some sources may be proprietary and require licensing.

The next step is to tap
into {\bf semi-structured elements} in online data, like infoboxes (in Wikipedia
and other wikis), tables, lists, headings, category systems, etc.
This is almost always easier than comprehending textual contents.
However, if we aim at rich coverage of facts about entities, we eventually
have to extract knowledge from {\bf natural-language text} as well -- ranging from high-quality
sources like Wikipedia articles all the way to user contents in online forums
and other social media. Finally, mass-user data about online behavior -- like queries and clicks --
is yet another source, which comes with a large amount of noise and potential bias.

\subsection{Output Scope and Quality}
\label{ch2-subsec:outputs}

Depending on our goals on precision and recall of a KB and our choice
on dealing with the trade-off, we can expect different kinds of outputs
from a knowledge-harvesting machinery.
Figure \ref{ch2-fig-knowledgeharvesting-designspace} lists major options
in the bottom part.

The minimum that every KB-building endeavor should have is that all entities
in the KB are semantically typed in a system of classes.
Without any classes, the KB would merely be a flat collection of entities,
and a lot of the mileage that search applications get from
KBs is through the class system. 
For example, queries or questions of the form
``songs/movies/phones by/with \dots'' are frequent, and
recognizing the expected answer types is a key issue.
Advanced queries or questions of the form
``singers who are also poets'' can even
be answered by intersecting the entities
of two classes (returning, e.g., Bob Dylan, Leonard Cohen, Serge Gainsbourg). 

As for the entities themselves, some KBs do {\em not} normalize,
lacking unique
identifiers and including real-world duplicates. Such a KB of names, containing, 
for example, both Bob Dylan and Robert Zimmerman as if they were two entities
(see Section \ref{sec:ch2-canonicalization}), can still be useful for many applications.
However, a KB with disambiguated names and canonicalized representation of
entities clearly offers more value for high-quality use cases such as
data analytics.

Larger coverage and application value come from capturing also properties of entities,
in the form of relational statements. Often, but not necessarily, this goes hand
in hand with logical invariants about properties, which could be acquired
by rule mining or by hand-crafted modeling using expert or crowdsourced inputs.
Analogously to surface-form versus canonicalized entities, 
relational statements also come in these two forms: with a diversity of names
for the same logical relation, or with unique names and no redundancy
(see Section \ref{sec:ch2-canonicalization} for examples).

\subsection{Methodological Repertoire}
\label{ch2-subsec:methods}

To go from input to output, knowledge harvesting has a variety of
options for its algorithmic methods and tools.
The following is a list of most notable options;
there are further choices, and practical techniques
often combine several of the listed options.

\begin{itemize}
    \item {\bf Rules and Patterns:} When inputs have rigorous structure and
    the desired output quality mandates conservative techniques, rule-based
    extraction can achieve best results. The System T project at IBM 
    \cite{DBLP:conf/naacl/ChiticariuDLRZ18}
    is a prime
    example of rule-based knowledge extraction for industrial-strength applications.
    \item {\bf Logical Inference:} Using consistency constraints can often
    eliminate spurious candidates for KB statements, and deduction rules can
    generate additional statements of interest. Both cases require reasoning
    with computational logics. This is usually combined with other paradigms
    such as extraction rules or statistical inference.
    \item {\bf Statistical Inference:} Distilling crisp knowledge from
    vague and ambiguous text content or semi-structured tables and lists
    often builds on the observation that there is redundancy in content sources:
    the same KB statement can be spotted in (many) different places.
    Thus we can leverage statistics and corresponding inference methods.
    In the simplest case, it boils down to frequency arguments, but it can
    be much more elaborated by considering different statistical measures
    and joint reasoning. In particular, statistical inference can be
    combined with logical invariants, for example, by 
    probabilistic graphical models 
    such as Markov Logic 
    \cite{DBLP:journals/cacm/DomingosL19}
    \item {\bf NLP Tools:} Modern tools for natural language processing
    (see, e.g., \cite{Eisenstein2019,JurafskyMartin2019})
    encompass a variety of methods, from rule-based to deep learning.
    They reveal structure and text parts of interest, such as dependency-parse trees
    for syntactic analysis, pronoun resolution, 
    identification of entity names,
    sentiment-bearing phrases, and much more. However, as language becomes
    more informal with incomplete sentences and colloquial expressions
    (incl. social-media talks such as ``LOL''), mainstream NLP does not always
    work well.
    \item {\bf Deep Learning:} The most recent addition to the methodogical
    repertoire is deep neural networks, trained in a supervised or distantly
    supervised manner (see, e.g., \cite{Burkov2019hundred,Goldberg2017}).
    The sweet spot here is when there is a large amount
    of ``gold-standard'' labeled training data, and often in combination with learning
    so-called {\em embeddings} from large text corpora.
    Thus, deep learning is most naturally used for increasing the coverage of a KB
    after initial population, such that the initial KB can serve as a source
    of distant supervision. 
\end{itemize}

In Figure \ref{ch2-fig-knowledgeharvesting-designspace},
the edges between input and methods and between outputs and methods indicate choices for 
methods being applied to different kinds of inputs and outputs.
Note that this is not meant to exclude 
further choices and additional combinations.

The outlined design space and the highlighted options are by no means complete,
but merely reflect some of the prevalent choices as of today. 
We will largely use this big picture as a ``roadmap'' for organizing material
in the following chapters.
However, there are further options and plenty of underexplored (if not unexplored) opportunities
for advancing the state of the art in knowledge harvesting.

\clearpage\newpage

\chapter{Knowledge Integration from Premium Sources}
\label{ch2:knowledge-integration}

This chapter presents a powerful method for populating a 
knowledge base with entities and classes,
and for organizing these into a systematic taxonomy.
This is the backbone that any high-quality KB
-- broadly encyclopedic or focused on a vertical domain --
must have.
Following the rationale of our design-space
discussion, we 
focus here on
knowledge harvesting
from premium sources such as Wikipedia
or domain-specific repositories such as 
GeoNames for spatial entities or 
GoodReads and Librarything for the domain of books.
This emphasizes the philosophy of ``picking low-hanging fruit first''
for the best benefit/cost ratio.

\section{The Case for Premium Sources}
\label{sec:wikipedia}

We recommend to start every KB construction project by tapping one or a few premium sources first.
Such sources should have the following characteristics:
\squishlist
\item authoritative high-quality content about entities of interest,
\item high coverage of many entities, and
\item clean and uniform representation of content, like having
clean HTML markup or even wiki markup,
unified headings and structure,
well-organized lists,
informative categories, and more.
\squishend

\noindent Distilling some of the contents from such sources into machine-readable knowledge can create a strong {\em core KB} 
with a good yield-to-effort ratio.
In particular, relatively simple extraction and cleaning
methods can go a long way. 
The core KB can then be further expanded
from other sources -- with more advanced methods, as presented in subsequent chapters. 

\vspace*{0.2cm}
\noindent{\bf Wikipedia:}\\
For a general-purpose encyclopedic knowledge base,
{\bf Wikipedia} is presumably the most suitable starting point, 
with its huge number of entities, highly informative descriptions
and annotations, and quality assurance via curation
by a large community and a sophisticated system of moderators.
The English edition of Wikipedia (\url{https://en.wikipedia.org}) 
contains more than 6 million articles 
with 500 words of text on average
(as of July 1, 2020), all
with unique names most of which
feature individual entities. 
These include
more than 
1.5 million notable people,
more than 750,000 locations of interest,
more than 250,000 organizations, and instances of further
major classes including events (i.e., sports tournaments, natural disasters, battles and wars, etc.)
and creative works (i.e., books, movies, musical pieces, etc.).

Wikipedia serves as an archetype of
knowledge-sharing communities, which
can be seen as ``proto-KBs'': the right
contents for a KB, but not yet in the
proper representation.
Another case in point would be the 
Chinese encyclopedia
Baidu Baike with almost
20 million articles 
({\small\url{https://baike.baidu.com}}).
In this chapter, we focus on the English Wikipedia as an exemplary case.
We will see that Wikipedia alone does not
lend itself to building a clean KB as
easily as one would hope.
Therefore, we combine input from Wikipedia
with another premium source: 
the {\bf WordNet} lexicon \cite{Fellbaum1998}
as a key asset for the KB taxonomy.

\vspace*{0.2cm}
\noindent{\bf Geographic Knowledge:}\\
For building a KB about geographic and geo-political entities, like countries, cities, rivers, mountains,
natural and cultural landmarks, Wikipedia itself is a good starting point, but
there are very good alternatives as well. Wikivoyage ({\small\url{https://www.wikivoyage.org}}) is a 
travel-guide wiki with specialized articles about travel destinations. 
GeoNames ({\small\url{https://www.geonames.org}}) is a huge repository
of geographic entities, from mountains and volcanos
to churches and city parks, more than 10 million in total.
If city streets, highways, shops, buildings and hiking trails are of interest, too,
then OpenStreetMaps ({\small\url{https://www.openstreetmap.org/}})
is another premium source to consider
(or alternatively commercial maps if you can afford to license them). 
Even commercial review forums such as
TripAdvisor 
({\small\url{https://www.tripadvisor.com/}})
could be of interest, to include hotels, restaurants and tourist services.

These sources complement each other, but they also overlap in entities.
Therefore, simply taking their union as an entity repository is not a viable solution.
Instead, we need to carefully integrate the sources, using techniques for
{\em entity matching} to avoid duplicates and to combine their different pieces
of knowledge for each entity (see Chapter \ref{ch3-sec-EntityDisambiguation}, especially Section \ref{sec:entity-matching}).

To obtain an expressive and clean taxonomy of classes, we could tap each of the
sources separately, for example, by interpreting categories as semantic types.
But again, simply taking a union of several category systems does not make sense.
Instead, we need to find ways of aligning equivalent (and possibly also subsumption) 
pairs of categories, as a basis for
constructing a {\em unified type hierarchy}.
For example, can and should we map \ent{craters} from one source to \ent{volcanoes}
in a second source, and how are both related to \ent{volcanic national parks}?
This alignment and integration is not an easy task, but it is still 
much simpler than extracting all volcanoes and craters
from textual contents in a wide variety of diverse web pages.

\begin{samepage}
\vspace*{0.2cm}
\noindent{\bf Health Knowledge:}

\noindent Another vertical domain of great importance
for society is health: building a KB with entity
instances of diseases, symptoms, drugs, therapies etc.
\end{samepage}
There is no direct counterpart to Wikipedia for this case, but there are
large and widely used tagging catalogs and %
terminology lexicons
like
MeSH ({\small\url{https://www.nlm.nih.gov/mesh/meshhome.html}}) and 
UMLS ({\small\url{https://www.nlm.nih.gov/research/umls/}}
including the SNOMED clinical terminology for healthcare), and these can be
treated as analogs to Wikipedia: rich categorization but not always
semantically clean. The next step would 
then be to clean these raw assets,
using methods like the presented ones,
and populate the resulting classes with entities. 

For the latter, additional premium sources could be considered:
either Wikipedia articles about biomedical entities, or curated 
sources such as DrugBank ({\small\url{https://www.drugbank.ca/}}) or
Disease Ontology ({\small\url{http://disease-ontology.org/}}),
and also human-oriented Web portals like the one by the Mayo Clinic
({\small\url{https://www.mayoclinic.org/patient-care-and-health-information}}).

Research projects along the lines of this knowledge integration and taxonomy construction for health
include KnowLife/DeepLife  \cite{Ernst:BMCbioinformatics2015,Ernst:ACL2016},
Life-INet \cite{Ren:ACL2017} and
Hetionet \cite{himmelstein2017systematic};
see also \cite{DBLP:conf/ecir/JimmyZK18}
for general discussion of health knowledge.

\section{Category Cleaning}
\label{sec:ch3-category-cleaning}

Many premium sources come
 with a rich {\bf category system}:
 assigning pages to
 relevant categories that can be
 viewed as proto-classes but are 
 too noisy to be considered as
 a semantic type system.
Wikipedia, as our canonical example, 
organizes its articles in 
a hierarchy of more than 1.9 million categories (as of July 1, 2020).
For example, \ent{Bob Dylan} (the entity corresponding to
article {\small\url{en.wikipedia.org/wiki/Bob_Dylan}}) is placed in categories such as\\
\hspace*{1cm}\ent{American male guitarists},\\ 
\hspace*{1cm}\ent{Pulitzer Prize winners},\\ \hspace*{1cm}\ent{Songwriters from Minnesota} etc.,\\
and \ent{Blowin' in the Wind} (corresponding to
{\small\url{en.wikipedia.org/wiki/Blowin'_in_the_Wind}}) is in categories such as\\
\hspace*{1cm}\ent{Songs written by Bob Dylan}, \\
\hspace*{1cm}\ent{Elvis Presley songs},\\
\hspace*{1cm}\ent{Songs about freedom} and\\ 
\hspace*{1cm}\ent{Grammy Hall of Fame recipients},
among others.\\
Using these categories as classes with their respective entities, it seems we could
effortlessly construct an initial KB.
So are we done already?

\begin{figure} [tb]
  \centering
   \includegraphics[width=1.0\textwidth]{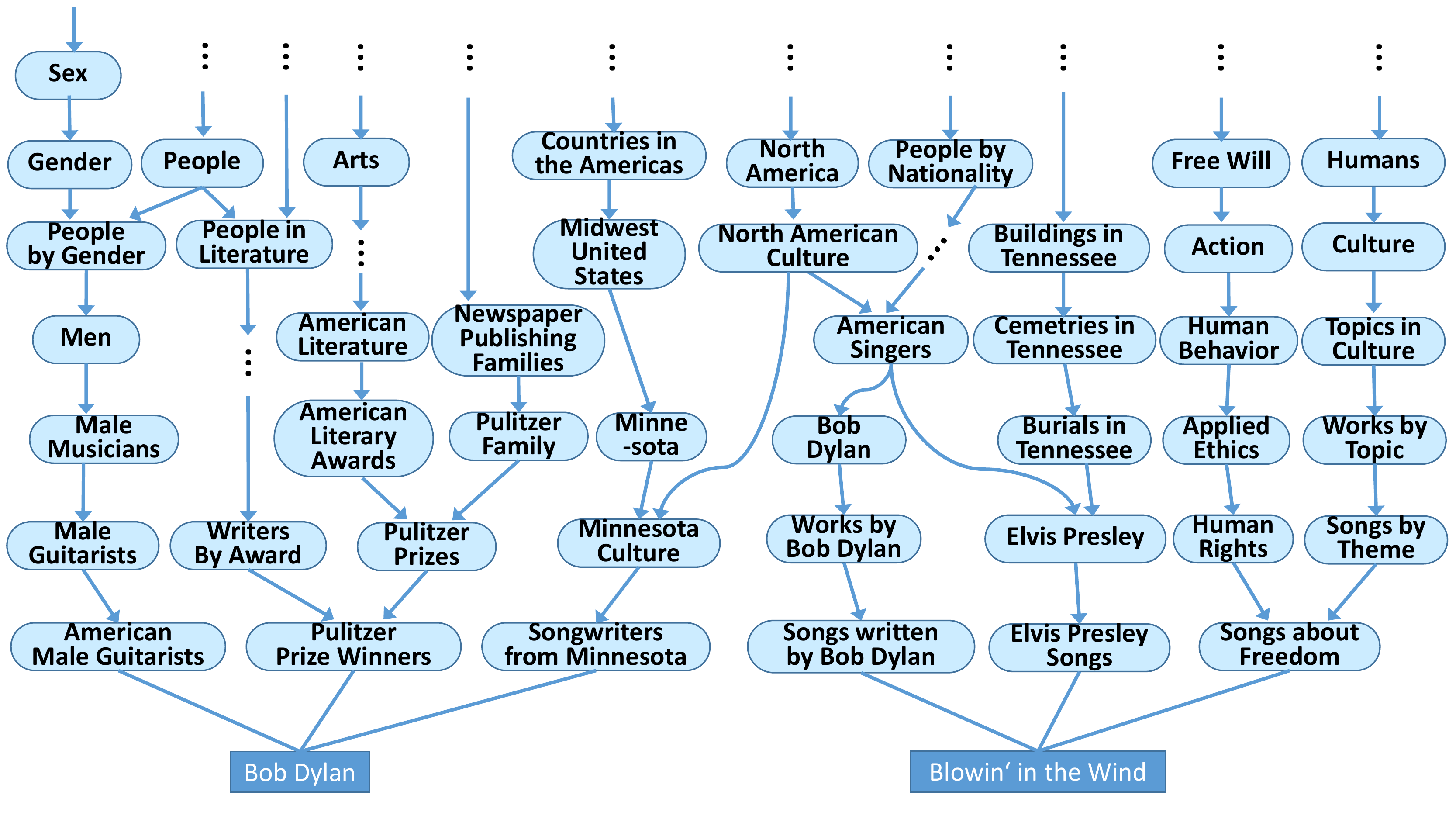}
      \caption{Example categories and super-categories from Wikipedia}
      \label{ch3-fig-wikipedia-categories}
\end{figure}

Unfortunately, the Wikipedia category system is almost a class taxonomy,
but only almost. 
Figure~\ref{ch3-fig-wikipedia-categories} shows excerpts of the category hierarchy for the entities
\ent{Bob Dylan} and \ent{Blowin' in the Wind}.
We 
face the following difficulties:

\squishlist
\item {\bf High Specificity of Categories:}
The direct categories of entities (i.e., leaves in the category hierarchy)
tend to be highly specific and often combine multiple classes into one
multi-word phrase. Examples are 
\ent{American male singer-songwriters}, 
\ent{20th-Century American gui\-tarists}
or \ent{Nobel laureates absent at the ceremony}.
For humans, it is obvious that this implies membership in classes
\ent{singers}, \ent{men}, \ent{guitar  players}, \ent{Nobel} \ent{laureates} etc.,
but for a computer, the categories are 
initially just noun phrases.
\item {\bf Drifting Super-Categories:}
By considering also super-categories (i.e., non-leaf nodes in the hierarchy)
and the paths in the category system, we could possibly generalize 
the leaf categories and derive
broader classes of interest, such as \ent{men}, \ent{American people}, \ent{musicians}, etc.
However, the Wikipedia category system exhibits conceptual drifts where super-categories
imply classes that are incompatible with those of the corresponding leaves and the entity
itself. 
Consider the category-to-category linkage in Figure
\ref{ch3-fig-wikipedia-categories}.
By transitivity, the super-categories would imply that Bob Dylan 
is a location, a piece of art, a family and a kind of reproduction process (generalizing
the ``sex'' category). 
For the example song, the category system alone would likewise lead
to flawed or meaningless classes:
locations, buildings, singers, actions, etc.
\item {\bf Entity Types vs. Associative Categories:}
Some of the super-categories are general concepts, for example,
\ent{Applied Ethics} and
\ent{Free Will} in Figure \ref{ch3-fig-wikipedia-categories}.
Some of the edges between a category and its immediate super-category
are conceptual leaps, for example, moving from songs and works to the
respective singers in Figure \ref{ch3-fig-wikipedia-categories}.
\squishend

All this makes sense when the category hierarchy is viewed
as a means to support user browsing in an associative way,
but it is not acceptable for the taxonomic backbone of a clean KB.
For example, queries about buildings on North America may return
\ent{Blowin' in the Wind} as an answer, and statistical analytics on
prizes, say by geo-region or gender, would confuse the awards and
the awardees.

In the following we present a 
simple but powerful
methodology to leverage
the Wikipedia categories as raw input with thorough cleansing
of their noun-phrase names and integration with upper-level taxonomies
for clean KB construction based on the works of 
\citet{DBLP:conf/aaai/PonzettoS07}
\cite{DBLP:journals/ai/PonzettoS11}
and
\citet{Suchanek:WWW2007} (see also \cite{DBLP:journals/ai/HoffartSBW13}),
applied in the WikiTaxonomy and YAGO projects.

\vspace*{0.2cm}
\noindent{\bf Head Words of Category Names:}\\
As virtually all category names are noun phrases, a basic building block
for uncovering their semantics is to parse these multi-word phrases into
their syntactic constituents. This task is known as {\em noun-phrase parsing}
in NLP (see, e.g., \cite{JurafskyMartin2019} and \cite{Eisenstein2019}).
In general, noun phrases consist of nouns, adjectives, determiners (like ``the'' or ``a''),
pronouns, coordinating conjunctions (``and'' etc.), prepositions (``by'', ``from'' etc.),
and possibly even further word variants.

A typical first step is to perform {\bf part-of-speech tagging}, or {\bf POS tagging}
for short. This tags each word with its syntactic sort: noun, adjective etc.
Nouns are further classified into common nouns, which can have an article (e.g.,
``{\em the} guitarist'' or ``{\em a} guitarist''),
and proper nouns which denote names (e.g., ``Bob Dylan'' or ``Minnesota'') that are not prefixed by an article (in English and many other languages).
POS tagging usually works by dynamic programming over a pre-trained statistical
model of word-variant sequences in large corpora.
The subsequent {\bf noun-phrase parsing} computes a syntactic tree structure
for the POS-tagged word sequence, inferring which word modifies or refines which
other word. This is usually based on stochastic context-free grammars, again using
some form of dynamic programming. 
Later chapters in this article will
make intensive use of such NLP methods, too.

The root of the resulting tree is called the {\bf head word}, and this is what
we are mostly after. For example, for ``American male guitarists'' the head word
is ``guitarists'' and for ``songwriters from Minneota'' it is ``songwriters''.
The head word is preceded by so-called {\em pre-modifiers}, and followed by {\em post-modifiers}.
Sometimes, words from these modifiers can be combined with the head word to form
a semantically meaningful class as well (e.g., ``female guitarists'').

Equipped with this building block, we can now tackle the desired category cleaning.
Our goal is to distinguish 
taxonomic
categories (such as ``guitarists'') from
associative categories (such as ``music'').
The key idea is
the heuristics that plural-form common
nouns are likely to denote classes, whereas single-form nouns tend to correspond
to general concepts (e.g., ``free will''). 
The reason for this is that classes regroup several instances, and a plural form is thus a strong indicator for a class.
We can possibly relax this heuristics to consider also singular-form nouns 
where the corresponding plural form is frequently occurring in corpora
such as news. For example, if a Wikipedia category were named 
``jazz band'' rather
than ``jazz bands'' 
we should accept it as a class, while still
disregarding categories such as ``free will'' or ``ethics'' (where ``wills'' is very rare,
and ``ethics'' is a singular word despite ending with ``s'').
These ideas can be cast into the following algorithm.

\begin{samepage}
\begin{mdframed}[backgroundcolor=blue!5,linewidth=0pt]
\squishlist
\item[ ] {\bf Algorithm for Category Cleaning}\\
Input: Wikipedia category name $c$: leaf node or non-leaf node\\
Output: string for semantic class label or \ent{null}
\squishlist
\item[1.] Run noun-phrase parsing to identify headword $h$
and modifier structure: $c = pre_1 ~ .. ~ pre_k ~ h ~ post_1 ~ .. ~ post_l$.
\item[2.] Test if $h$ is in plural form or has a frequently occurring plural form.\\
If not, {\em return} \ent{null}.\\
Optionally, consider also $pre_i ~ .. ~ pre_k ~ h$ as class candidates,
with increasing $i$ from 0 to $k-1$.
\item[3.] For leaf category $c$, {\em return} $h$ (and optionally additional
class labels $pre_i ~ .. ~ pre_k ~ h$).
\item[4.] For non-leaf category $c$, test if the class candidate
$h$ is a synonym or hypernym (i.e., generalization) of an
already accepted class (including $h$).\\
If so, {\em return} $c$; otherwise return \ent{null}.
\squishend
\squishend
\end{mdframed}
\end{samepage}

The rationale for the additional test in Step 4 is that
non-leaf categories in Wikipedia are often merely associative,
as opposed to denoting semantically proper super-classes (see discussion above).
So we impose this additional scrutinizing, while still being able to harvest
the cases when head words of super-categories are meaningful outputs (e.g., \ent{Musicians} and \ent{Men}
in Figure~\ref{ch3-fig-wikipedia-categories}).
The test itself can be implemented by looking up head words in existing
dictionaries like {\bf WordNet} \cite{Fellbaum1998} 
or {\bf Wiktionary} ({\small\url{https://www.wiktionary.org/}}),
which list synonyms and hypernyms for many words. 
This is a prime case of harnessing a second premium source.

\vspace{0.2cm}
\noindent{\bf Class Candidates from Wikipedia Articles:}\\
We have so far focused on Wikipedia {\em categories} as a source
of semantic class candidates. However, normal Wikipedia {\em articles}
may be of interest as well. Most articles represent individual entities,
but some feature concepts, among which some may qualify as classes.
For example, the articles {\small\url{https://en.wikipedia.org/wiki/Cover_version}}
and {\small\url{https://en.wikipedia.org/wiki/Aboriginal_Australians}}
correspond to classes as they have instances,
whereas {\small\url{https://en.wikipedia.org/wiki/Empathy}} is a singleton concept as it has no individual 
entities as instances of interest.

Simple but effective heuristics to capture these cases have been studied by
\cite{Gupta:COLING2016,Pasca:WWW2018}. The key idea is that an article qualifies
as a class if its textual body mentions the article's title in both singular and plural forms.
For example, ``cover version'' and ``cover versions'' are both present in the article
about cover versions (of songs), but the article on empathy does not refer to ``empathys''.
Obviously, this heuristic technique is just another building block that should be combined with
other heuristics and statistical inference.

It works well for English and many other
Indo-European languages with
morphological cues about singular vs. plural.
For other languages, specific models would be needed,
possibly learned via language embeddings with neural networks.

\section{Alignment between Knowledge Sources}
\label{ch3-sec:alignment}

By applying the category cleaning algorithm to all Wikipedia categories,
we can obtain a rich set of class labels for each entity. 
However, as the Wikipedia community does not enforce strict
naming standards, we could arrive at duplicates for the same class,
for example, accepting both \ent{guitarist} and \ent{guitar player}.
Moreover, as we are mostly harvesting the leaf-node categories
and expect to prune many of the more associative super-categories, our KB
taxonomy may end up with many disconnected classes.

To fix these issues, we resort to pre-existing high-quality taxonomies
like {\bf WordNet} 
(\citet{Fellbaum1998}). 
This lexicon already covers 
more than hundred thousand concepts and classes -- called {\em word senses}
or {\em synsets} for sets of synonyms --
along with clean structure for hypernymy.
Alternatively, we could consider
{\bf Wiktionary} ({\small\url{https://www.wiktionary.org/}}).
Both of these lexical resources also
have multilingual extensions, covering
a good fraction of mankind's languages;
see 
\citet{DBLP:series/synthesis/2016Gurevych} 
for
an overview of this kind of lexical resources.

A major caveat, however, is that 
WordNet has hardly any entity-level instances for its classes;
you can think of it as an un-populated {\em upper-level taxonomy}.
The same holds for Wiktionary.
The goal now is to align the 
selected categories (i.e., class {\em candidates})
harvested from
Wikipedia with the classes in 
WordNet.

\vspace*{0.2cm}
\noindent{\bf Similarity between Categories and Classes:}\\
The key idea is to perform a {similarity test}
between a 
category
from Wikipedia and potentially
corresponding classes in WordNet.
In the simplest case, this is just a surface-form {\bf string similarity}.
For example, the Wikipedia-derived candidate ``guitar player''
has high similarity with the WordNet entry ''guitarist''.
There are two problems to address, though.
First, we could still observe low string similarity for two
matching classes, for example, 
``award'' from Wikipedia against ``prize'' in WordNet.
Second, we can find multiple matches with high similarity,
for example ``building'' from Wikipedia matching two
different senses in WordNet, namely, building in the sense
of a man-made structure (e.g., houses, towers etc.)
and building in the sense of a construction process
(e.g., building a KB).
We have to make the right choice among such alternatives
for 
ambiguous words.

The solution for the first problem -- similarity-based matching -- is to consider contexts as well.
WordNet, and Wiktionary alike, provide synonym sets as well as short descriptions (so-called glosses),
for their entries, and the Wikipedia categories can be contextualized by
related words occurring in super-categories or (articles for) their instances.
This way, we are able to map 
``award'' to ``prize''
because
WordNet has the entry 
``prize, award (something given for victory or superiority in a contest or competition or for winning a lottery)'', stating that ``prize'' and ``award'' are synonyms
(for this specific word sense).
More generally, we could consider also entire neighborhoods of WordNet entries
defined by hypernyms, hyponyms, derivationally related terms, and more.
Such contextualized {\bf lexical similarity} measures have been investigated in research
on {\bf word sense disambiguation (WSD)}, see 
\cite{Navigli:ComputingSurvey2009,DBLP:series/synthesis/2016Gurevych} for overviews.

Another approach to strengthen the similarity comparisons is
to incorporate {\bf word embeddings} such as Word2Vec \cite{Mikolov:NIPS2013}
or Glove \cite{DBLP:conf/emnlp/PenningtonSM14} (or even deep neural networks along these lines, such as
 BERT \cite{DBLP:conf/naacl/DevlinCLT19}). We will not go into this topic now, but will come back
 to it in Section \ref{subsec:word-entity-embeddings}.

For the second problem -- ambiguity -- 
we could apply state-of-the-art WSD methods, but it turns out
that there is a very simple heuristic that works so well that it is hardly
outperformed by any advanced WSD method. It is known as the 
{\bf most frequent sense (MFS)} heuristic:
whenever there is a choice among different word senses, pick the one that
is more frequently used in large corpora such as news or literature.
Conveniently, the WordNet team has already manually annotated large corpora with WordNet senses, and has recorded the frequency of each word sense.  It is thus easy to identify, for each given word, its most frequent meaning.
For example, the MFS for ``building'' is indeed the man-made structure.
There are exceptions to the MFS heuristics,
but they can be handled in other ways.

\vspace*{0.2cm}
\noindent{\bf Putting Everything Together:}\\
Putting these considerations together, we arrive at the following 
heuristic algorithm
for aligning Wikipedia categories and WordNet senses.

\begin{mdframed}[backgroundcolor=blue!5,linewidth=0pt]
\squishlist
\item[ ] {\bf Algorithm for Alignment with WordNet}\\
Input: Class name $c$ derived from Wikipedia category\\
Output: synonym or hypernym in WordNet, or \ent{null}
\squishlist
\item[1.] Compute string or lexical similarity of $c$
to WordNet entries $s$ (for candidates $s$ with certain overlap
of character-level substrings). Then pick the $s$ with
highest similarity if this is above a given threshold;
otherwise return \ent{null}.
\item[2.] If the highest-similarity entry $s$ is unambiguous
(i.e., the same word has only this sense), then return the 
WordNet sense for $s$.
If $s$ is an ambiguous word, then return the MFS for $s$
(or use another WSD method for $c$ and the $s$ candidates).
\squishend
\squishend
\end{mdframed}

Recall that WordNet has extremely low coverage of
individual entities for its classes.
Therefore, signals from entity-set overlap are
completely disregarded, as they would not lead
to a viable approach.

Once we have mapped the accepted Wikipedia-derived classes
onto WordNet, we have a complete taxonomy, with the upper-level
part coming from the clean WordNet hierarchy of hypernyms.
The last thing left to decide for the alignment task
is whether the category-based class
$c$ is synonymous to the identified WordNet sense $s$
or whether $s$ is a hypernym of $c$.
The latter occurs when $c$ does not have a direct counterpart
in WordNet at all. For example, we could keep the category $c =$
``singer-songwriter'', but WordNet does not have this class
at all. Instead we should align $c$ to \ent{singer} or
\ent{songwriter} or both. If WordNet did not have an entry
for songwriters, we should map $c$ to the next hypernym, which
is \ent{composer}.

This final issue can be settled heuristically, for example, 
by assuming a synonymy match if the similarity score is very high
and assuming a hypernym otherwise, or we could resort to
leveraging information-theoretic measures over additional text corpora
(e.g., the full text of all Wikipedia articles).
Specifically, for a pair $c$ and $s$ (e.g., \ent{songwriter} and 
\ent{composer} -- a case of hypernymy), a symmetry-breaking measure
is the conditional probability $P[s|c]$ estimated from co-occurrence frequencies of words. 
If $P[s|c] \gg P[c|s]$, that is, $c$ sort of implies $s$ but not vice versa,
then $s$ is likely a hypernym of $c$, not a synonym.
Various measures along these lines are
investigated in \cite{DBLP:conf/coling/WeedsWM04,DBLP:conf/acl/GeffetD05}.

The presented methodology can also be adapted to other cases of taxonomy alignment, 
for example, to GeoNames, Wikivoyage and OpenStreetMap (and Wikipedia or Wikidata) 
about geographic categories and classes (see Section \ref{sec:wikipedia}).

\section{Graph-based Alignment}

\begin{figure} [tb]
  \centering
   \includegraphics[width=1.0\textwidth]{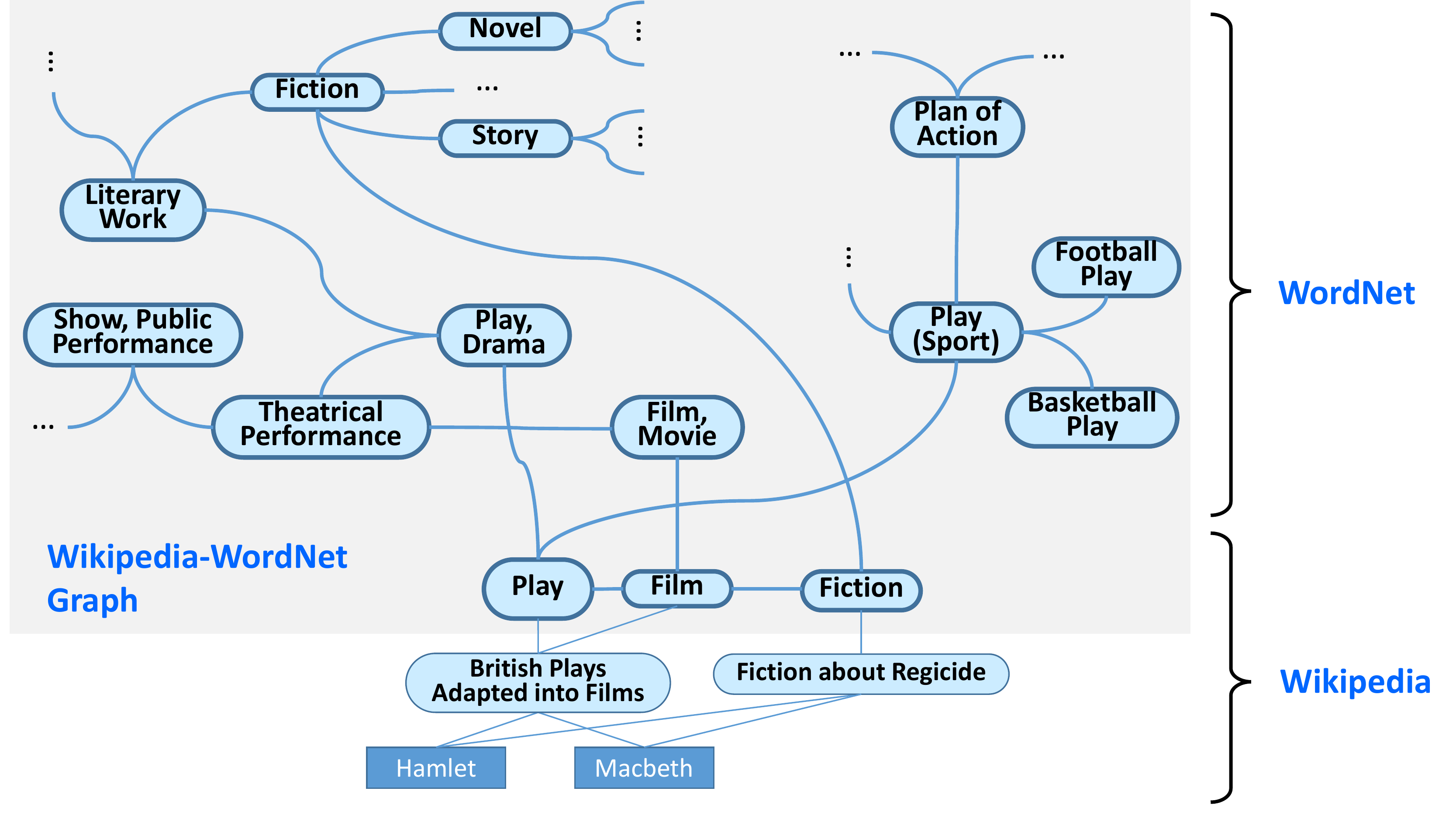}
      \caption{Example for graph-based alignment}
      \label{ch3-fig-mapping-wk-wn}
\end{figure}

A different paradigm for aligning Wikipedia categories with WordNet classes,
as prime examples of premium sources,
has been developed by
\citet{Navigli:ArtInt2012} 
for constructing the
{\em BabelNet} knowledge base. 
It is based on a candidate graph derived from the Wikipedia category
at hand and potential matches in WordNet, and then uses graph metrics
such as shortest paths or graph algorithms like random walks to
infer the best alignment. Figure \ref{ch3-fig-mapping-wk-wn} illustrates
this approach.
The graph construction extracts the head word of interest and salient
context words from Wikipedia -- ``play'' as well as ``film'' and ``fiction''
in the example.
Then all approximate matches are identified in WordNet, and their
respective local neighborhoods are added to the graph casting WordNet's
lexical relations like hypernymy/hyponymy, holoynymy/meronymy (whole-part) etc.
into edges. Edges could even be weighted based on similarity or salience metrics.

In the example, we have two main candidates to which ``play'' could refer:
``play, drama'' or ``play (sports)'' (WordNet contains even more).
To rank  these candidates, the simplest method is to look at how close
they are to the Wikipedia-based start nodes ``play'', ``film'' and ``fiction'',
for example, by aggregating the shortest paths from start nodes to
candidate-class nodes. 
Alternatively, and typically better performing, we can use methods based on
{\bf random walks} over the graph, analogously to how Google's
original PageRank and PersonalizedPageRank measures were computed
(\cite{BrinPage:WWW1998,JehWidom:WWW2003}).

The walk starts at the Wikipedia-derived head word ``play'' and randomly
traverses edges to visit other nodes -- where the probabilities for picking
edges should be proportional to edge weights (i.e., uniform over all
outgoing edges if the edges are unweighted). 
Occasionally, the walk could jump back to the start node ``play'',
as decided by a probabilistic coin toss.
By repeating this random procedure sufficiently often, we obtain
statistics about how often each node is visited. This statistics
converges to well-defined {\em stationary visiting probabilities}
as the walk length (or repetitions after jumping back to the start node)
approaches infinity.
The candidate class with the highest visiting probability is the winner:
``play, drama'' in the example.
Such random-walk methods are amazingly powerful, easy to implement and
widely applicable.

\section{Discussion}

There are various extensions of the presented method.
Wikipedia category names do not just indicate class memberships,
but often reflect other relations as well. For example, 
for Bob Dylan being in the category 
\ent{Nobel Laureates in Literature}
we can infer a relational triple $\langle$ \ent{Bob Dylan},
\ent{has won}, \ent{Literature Nobel Prize} $\rangle$.
Such extensions have been developed by
\citet{DBLP:journals/ai/NastaseS13};
we will revisit these techniques in Chapter \ref{ch4:properties}.
Another line of generalization is to leverage
the taxonomies from the presented methods as training data
to learn generalized patterns in category names and other
Wikipedia structures (infobox templates, list pages etc.).
This way, the taxonomy can be further grown and refined.
Such methods have been developed in the
Kylin/KOG project by
\citet{WuWeld:CIKM2007}
\cite{WuWeld:WWW2008}.

\vspace{0.2cm}
\noindent{\bf Relationship with Ontology Alignment:}\\
The alignment between Wikipedia categories and WordNet classes
can be seen as a special case of {\bf ontology alignment}
(see, e.g., 
\citet{StaabStuder2009} for an overview,
\cite{COMA++2005,RiMOM2009} for representative state-of-the-art methods,
and {\small\url{http://oaei.ontologymatching.org/}} for a prominent benchmark series). 
Here, the task is to match classes and properties of one ontology
with those of a second ontology, where the two ontologies are given
in crisp logical forms like RDFS schemas or, even better, in the OWL
description-logic language \cite{W3C:OWL2}.
Ontology matching
is in turn highly related to
the classical task of {\bf database schema matching} \cite{DoanHalevyIves2012}. %

The case of matching Wikipedia categories and WordNet classes is a special case, though, for two reasons. First, WordNet has hardly
any instances of its classes, and we chose
to ignore the few existing ones.
Second, the upper part of the Wikipedia category hierarchy is more associative than taxonomic so that it had to be cleaned first
(as discussed in Section \ref{sec:ch3-category-cleaning}). For these reasons, the case of Wikipedia and WordNet
benefits from tailored alignment methods,
and similar situations are likely to arise
also for domain-specific premium sources.

\vspace*{0.2cm}
\noindent{\bf Beyond Wikipedia:}\\
We used Wikipedia and WordNet as
exemplary cases of premium sources,
and pointed out a few vertical domains
for wider applicability of the presented methods.
Aligning and enriching pre-existing knowledge sources
is also a key pillar for industrial-strength KBs
about retail products (see, e.g., \cite{Deshpande2013kosmix,DBLP:conf/kdd/DongHKLLMXZZSDM20}).
More discussion on this use case is offered
in Section \ref{ch9-sec:industrialKG}.

Apart from this mainstream, similar cases for knowledge integration
can be made for less obvious domains, too,
examples being food 
\cite{DBLP:conf/smap/Arens-VollandGN18,DBLP:conf/semweb/HaussmannSCNCCM19} or
fashion \cite{Kari:DINAcon2019,DBLP:conf/wsdm/Grauman2020}.
Food KBs have integrated sources like
the FoodOn ontology \cite{dooley2018foodon},
the nutrients catalog {\small\url{https://catalog.data.gov/dataset/food-and-nutrient-database-for-dietary-studies-fndds}}
and a large recipe collection
\cite{DBLP:journals/corr/abs-1810-06553},
and fashion KBs could make use of contents from catalogs
such as {\small\url{https://ssense.com}}.
More
exotic verticals to which the Wikipedia-inspired methodology
has been carried over are 
{\em fictional universes} such as 
Game of Thrones, the Simpsons, etc., with input from rich category systems
by fan-community wikis ({\small\url{https://www.fandom.com/}}).
Recent research on this topic includes 
\cite{HertlingPaulheim:ICBK2018} and \cite{DBLP:conf/www/ChuRW19}.
Finally, another non-standard theme for
KB construction is {\em how-to knowledge}:
organizing human tasks and procedures for solving them in a principled taxonomy.
Research on this 
direction includes \cite{DBLP:conf/sigir/YangN15,DBLP:conf/www/ChuTW17}.

\section{Take-Home Lessons}

We summarize this chapter by the following
take-home lessons.
\squishlist
\item For building a 
core KB,
with individual entities organized into
a clean taxonomy of semantic types,
it is often wise to start with one
or a few {\em premium sources}.
Examples are Wikipedia for general-purpose
encyclopedic knowledge, GeoNames and
WikiVoyage for geo-locations, or IMDB
for movies.
\item A key asset are
categories 
by which entities are annotated
in these sources. 
As categories are often merely associative,
designed 
for
manual browsing, this typically involves
a {\em category cleaning} step
to identify taxonomically clean
classes.
\item To construct an expressive and clean taxonomy, while harvesting two or more premium sources, it is often necessary
to {integrate different
type systems}. This can be achieved
by {\em alignment heuristics} based
on NLP techniques (such as noun phrase parsing), or by {\em random walks}
over candidate graphs.
\squishend

\clearpage\newpage

\chapter{KB Construction: Entity Discovery and Typing}
\label{ch3:entities}

This chapter presents advanced methods for populating a 
knowledge base with entities and classes (aka types),
by tapping into textual and semi-structured sources.
Building on the previous chapter's insights on harvesting
premium sources first,
this chapter
extends the regime for
inputs to discovering entities
and type information in Web pages and
text documents.
This will often yield noisier output, in the
form of entity duplicates.
The following chapter,
Chapter 
\ref{ch3-sec-EntityDisambiguation}, will address this issue
by presenting methods for
canonicalizing entity mentions into uniquely
identified
subjects in the KB.

\section{Problem and Design Space}
\label{ch3-sec-EntityDiscovery}

Harvesting entities and their classes from premium sources
goes a long way, but it is bound to be incomplete when
the goal is to fully cover a certain domain such as 
music or health, and to associate all entities with
their relevant classes.
Premium sources alone are typically insufficient to
capture long-tail entities, such as less prominent
musicians, songs and concerts, as well as long-tail classes
such as left-handed cello players or cover songs in a language different from the original.
In this section, we present a suite of methods for automatically 
extracting such additional entities and classes from sources
like web pages and other text documents.

In addressing this task, we typically leverage that
premium sources already give us a taxonomic backbone
populated with prominent entities.
This leads to various discovery tasks:
\squishlist
\item[1.] Given a class $T$ containing a set of entities $E=\{e_1 \dots e_n\}$,
find more entities for $T$ that are not yet in $E$.
\item[2.] Given an entity $e$ and its associated classes $T=\{t_1 \dots t_k\}$,
find more classes for $e$ not yet captured in $T$.
\item[3.] Given an entity $e$ with known names, find additional (alternative) names
for $e$, such as acronyms or nicknames. This is often referred to as {\em alias name discovery}.
\item[4.] Given a class $t$ with known names, find additional names for $t$.
This is sometimes referred to as {\em paraphrase discovery}.
\squishend
In the following, we organize different approaches by methodology rather than by
these four tasks, as most methods apply to several tasks.
Recall that we use the terms {\em class} and
{\em type} interchangeably.

In principle, there is also a case where the initial repository of entities
and classes is empty -- that is, when there is no premium source to be harvested first.
This is the case for {\em ab-initio taxonomy induction} (or ``{\em zero-shot} learning'')
from noisy observations,
which will be discussed in
Section \ref{ch3-sec-TaxonomyConstruction}.

\section{Dictionary-based Entity Spotting}
\label{subsec:dictionary-based-entity-spotting}

An important building block for all of the outlined discovery tasks
is to detect {\bf mentions} {of already known entities} in web pages and text documents.
These mentions do not necessarily take the form of an entity's full name
as derived from premium sources.
For example, we want to be able to spot occurrences of
\ent{Steve Jobs} and \ent{Apple Inc.} in a sentence such as
``Apple co-founder Jobs gave an impressive demo of the new iPhone.''
Essentially, this is a string matching task where we compare known names of
entities in the existing KB against textual inputs.
The key asset here is to have a rich {\bf dictionary of alias names} from the KB.
Early works on information extraction
from text made extensive use of name
dictionaries, called {\em gazetteers},
in combination with NLP techniques
(POS tagging etc.) and string patterns.
Two seminal projects of this sort are
the {\em GATE} toolkit
by
\citet{DBLP:journals/lre/Cunningham02} \cite{DBLP:conf/acl/CunninghamMBT02}
and
the {\em UIMA} framework by \citet{DBLP:journals/nle/FerrucciL04}.

A good dictionary should include 
\squishlist
\item abbreviations (e.g., ``Apple'' instead of ``Apple Inc.''),
\item acronyms (e.g., ``MS'' instead of ``Microsoft''),
\item nicknames and stage names (e.g., ``Bob Dylan'' vs. his real name ``Robert Zimmerman'',
or ``The King'' for ``Elvis Presley''), 
\item titles and roles (e.g., ``CEO Jobs'' or
``President Obama''), and possibly even 
\item rules for deriving 
short-hand names (e.g., ``Mrs. Y'' 
for female people with 
last name ``Y'').
\squishend

Where do we get such dictionaries from?
This is by itself a research issue, tackled, for example, by \cite{DBLP:journals/debu/ChakrabartiCCGH16}.
The easiest approach is to exploit redirects in premium sources
and hyperlink anchors.
Whenever a page with name $X$ is redirected to an official page with name $Y$
and whenever a hyperlink with anchor text $X$ points to a page $Y$,
we can consider $X$ an alias name for entity $Y$.
In Wikipedia, for example, 
a page with title ``Elvis'' ({\small\url{https://en.wikipedia.org/w/index.php?title=Elvis}})
redirects to the proper article 
({\small\url{https://en.wikipedia.org/wiki/Elvis_Presley}}),
and the page about the Safari browser contains a hyperlink with 
anchor ``Apple'' that points to the article
{\small\url{https://en.wikipedia.org/wiki/Apple_Inc.}}.
This approach extends to links from Wikipedia disambiguation pages:
for example, the page {\small\url{https://en.wikipedia.org/wiki/Robert_Zimmerman}}
lists 11 people with this name, including a link to 
{\small\url{https://en.wikipedia.org/wiki/Bob_Dylan}}.
Of course, this does not resolve the ambiguity of the name,
but it gives us one additional alias name for \ent{Bob Dylan}.
Hyperlink anchor texts in Wikipedia were first exploited for entity alias names by
\cite{DBLP:conf/eacl/BunescuP06}, and this simple technique was
extended to hyperlinks in arbitrary web pages by
\cite{DBLP:conf/lrec/SpitkovskyC12}.
A special case of interest is to harvest multilingual names from
interwiki links in Wikipedia (connecting different language editions)
or non-English web pages linking to English Wikipedia
articles. For example, French pages with anchor text ``Londres'' link to
the article {\small\url{https://en.wikipedia.org/wiki/London}}. 
All this is low-hanging fruit from an engineering perspective, and
gives great mileage towards rich dictionaries for entity names.

An analogous issue arises for class names as well.
Here, redirect and anchor texts are useful, too, but are fairly sparse.
The WordNet thesaurus and the Wiktionary lexicon contain synonyms for many word 
 senses 
(e.g., ``vocalists'' for \ent{singers}, and vice versa)
and can serve to populate 
a
dictionary of class paraphrases.

The ideas that underlie the above heuristics can be
cast into a more general principle of {\bf strong co-occurrence}:

\begin{mdframed}[backgroundcolor=blue!5,linewidth=0pt]
\squishlist
\item[ ] {\bf Strong Co-Occurrence Principle:}
\item[ ] If an entity or class name $X$ co-occurs with name $Y$
in a context with cue $Z$ (defined below), then $Y$ is (likely) an alias name for $X$.
\vspace*{0.1cm}
\item[ ] This principle can be instantiated in various ways, depending on what we
consider as context cue $Z$:
\squishlist
\item The cue $Z$ is a hyperlink where $X$ is the link target and
$Y$ is the anchor text.
\item The cue $Z$ is a specific wording in a sentence, like
``also known as'', ``aka.'', ``born as'', ``abbreviated as'', ``for short'' etc.
\item The context cue $Z$ is a query-click pair (observed by a search engine),
where $X$ is the query and $Y$ is the title of a clicked result
(with many clicks by different users).
\item The context cue $Z$ is the frequent occurrence of $X$ in documents
about $Y$ (e.g., Wikipedia articles, biographies, product reviews etc.).
\squishend
\squishend
\end{mdframed}

For example, when many users who query for ``Apple'' (or ``MS'') 
subsequently click on the
Wikipedia article or homepage of \ent{Apple Inc.} (or ``Microsoft''), we learn that
``Apple'' is a short-hand name for the company.
In the same vein, queries about ``Internet companies'' with subsequent clicks on pages of
Amazon or Google provide hints that these
are instances of a class \ent{Internet companies}.
This co-clicking technique has been studied by
\cite{DBLP:conf/www/TanevaCCH13}.
Extending this to textual co-occurrence (i.e., the last of the above itemized cases)
comes with a higher risk of false positives, but could still be worthwhile for cases like short names or acronyms for products.
The technique can be tuned towards either precision or recall by
thresholding on the observation frequencies and by making the context cue
more or less restrictive.
More advanced techniques for learning co-occurrence cues about alias names
-- so-called synonym discovery,
have been investigated by \cite{DBLP:conf/kdd/QuR017}, among others.

\section{Pattern-based Methods}
\label{ch3-subsect-patterns}

\subsection{Patterns for Entity-Type Pairs}

\vspace{0.2cm}
\noindent{\bf Hearst Patterns:}\\
To discover more entities of a given class or more classes of a given entity,
a powerful approach is to consider specific patterns that co-occur with
input (class or entity) and desired output (entity and class).
For example, a text snippet like ``singers such as Bob Dylan, Elvis Presley and
Frank Sinatra'' suggests that Dylan, Presley and Sinatra belong to the class of \ent{singers}.
Such patterns have been identified in the seminal work of 
\citet{DBLP:conf/coling/Hearst92}, 
and are thus known as {\bf Hearst patterns}.
They are a special case of the strong co-occurrence principle where
the Hearst patterns serve as context cues.

In addition to the ``such as'' pattern, the most important Hearst patterns
are:
``$X$ like $Y$'' (with class $X$ in plural form and entity $Y$),
``$X$ and other $Y$'' (with entity $X$ and class $Y$ in plural form),
and
``$X$ including $Y$'' (with class $X$ in plural form and entity $Y$).

Some of the Hearst patterns also apply to discovering subclass relations between classes.
In the pattern ``$X$ including $Y$'', $X$ and $Y$ could both be classes,
for example, ``singers including rappers''.
In fact, the patterns alone cannot distinguish between observations of
entity-class (type) relations (entity $\in$ class) versus subclass relations
(class1 $\subseteq$ class2).
Additional techniques can be applied to identify which surface strings
denote entities and which ones refer to classes. Simple heuristics can already go a long way: in English, for example, words that start with an uppercase letter are often entities whereas common nouns in plural form are more likely class names. Full-fledged approaches 
make use of dictionaries
as discussed above or more advanced methods for entity recognition, discussed 
further below. 

Hand-crafted patterns are useful also for
discovering entities in semi-structured web contents
like lists and tables. 
For example, if a list heading or a column header
denotes a type, then the list or column entries
could be considered as entities of that type.
The pattern for this purpose would refer to HTML tags
that mark headers and entries 
\cite{Etzioni:ArtInt2005,DBLP:journals/pvldb/LimayeSC10}.
Of course, this is just a 
heuristics
that has a non-negligible risk of failing.
We will discuss more advanced methods that handle
this case more robustly.
Particularly, Sections \ref{ch6:subsubsec:patterns-text-lists-trees}
and \ref{ch6-sec:properties-from-semistructured}
go into depth on extraction from semi-structured contents
for the more general scope of entity properties.

\vspace{0.2cm}
\noindent{\bf Multi-anchored Patterns:}\\
Hearst patterns may pick up spurious observations.
For example, the sentence ``protest songs against war like 
Universal Soldier'' could erroneously yield that
\ent{Universal Solider} is an instance of the class \ent{wars}.
One way of making the approach more robust is to include part-of-speech tags
or even dependency-parsing trees
(see \cite{Eisenstein2019,JurafskyMartin2019} for these NLP basics) in the specification of patterns.
Another approach is to extend the context cue and strengthen its role.
In addition to the pattern itself, we can demand that the context contains
at least one additional entity which is already known to belong to the observed class.
For example, the text ``singers such as Elvis Presley'' alone may be considered insufficient evidence
to accept Elvis as a singer, but the text ``singers such as 
Elvis Presley and Frank Sinatra'' would have a stronger cue if \ent{Frank Sinatra} is a known singer
already.
This idea has been referred to as {\em doubly-anchored patterns} in the literature
\cite{DBLP:conf/acl/KozarevaH10} for the case of observing two entities of the same class.
With a strong cue like
``singers such as'', insisting on a known witness
may be an overkill, but the principle
equally applies to weaker cues, for
example, ``voices such as ...'' for
the target class \ent{singers}.

Multi-anchored patterns are particularly useful when going beyond text-based
Hearst patterns by considering strong co-occurrence in enumerations, lists and tables.
For simplicity, consider only the case of tables in web pages -- 
as opposed to relational tables in databases.
The co-occurring entities are typically the names in the cells of the same column,
and the class is the name in the column header. 
Due the ambiguity of words and the ad-hoc nature of web tables,
the spotted entities in the same column may be very heterogeneous,
mixing up apples and oranges.
For example, a table column on Oscar winners could have both
actors and movies as rows. Thus, we may incorrectly learn that
\ent{Godfather} is an actor and \ent{Bob Dylan} is a movie.
To overcome these difficulties, we can require that a new entity name
is accepted only when co-occurring with a certain number of known entities
that belong to the proper class (\cite{DBLP:conf/wsdm/DalviCC12}),
for example, at least 10 actors in the same column for a table of 15 rows.
Needless to say, all these are still heuristics that may occasionally fail,
but they are easy to implement, powerful and have practical value.

\subsection{Pattern Learning}
\label{ch4:subsec:PatternLearning}

Pre-specified patterns are inherently limited in their coverage.
This motivates approaches for automatically learning patterns,
using initial entity-type pairs and/or initial patterns for {\em distant supervision}.
For example, when frequently observing a phrase like ``great voice in'' for
entities of type \ent{singers}, this phrase could be added to a set of
indicative patterns for discovering more singers.
This idea is captured in the following principle of
{\em statement-pattern duality}, first formulated by 
\citet{DBLP:conf/webdb/Brin98}
(see also \citet{DBLP:conf/acl/RavichandranH02}
in the context of question answering):

\begin{mdframed}[backgroundcolor=blue!5,linewidth=0pt]
{\bf Principle of Statement-Pattern Duality}
\squishlist
\item[ ] When correct statements about entities $x$ (e.g., $x$ belonging to class $y$)
frequently co-occur with textual pattern $p$, then $p$ is likely a good pattern
to derive statements of this kind.\\
Conversely, when statements about entities $x$ frequently co-occur with a good
pattern $p$, then these statements are likely correct.\\
Thus, observations of good statements and good patterns 
reinforce
each other; hence the name {statement-pattern duality}.
\squishend
\end{mdframed}

This insightful paradigm gives rise to a straightforward algorithm
where we start with statements in the KB as {\em seeds}
(and possibly also with pre-specified patterns), and then iterate
between deriving patterns from statements and deriving statements 
from patterns
(\citet{DBLP:conf/dl/AgichteinG00},
\citet{Etzioni:ArtInt2005},
\citet{DBLP:conf/icdm/WangC08}).

\begin{samepage}
\begin{mdframed}[backgroundcolor=blue!5,linewidth=0pt]
\squishlist
\item[ ] {\bf Algorithm for Seed-based Pattern Learning}\\
Input: Seed statements in the form of known entities for a class\\
Output: Patterns for this class, and new entities of the class
\squishlist
\item[ ] Initialize:\\
 $~~~ S$ $\leftarrow$ seed statements\\
 $~~~ P$ $\leftarrow$ $\emptyset$ (or pre-specified patterns like Hearst patterns)
\item[ ] Repeat
\squishlist
\item[1.] Pattern discovery:
\item[-] search for mentions of entity $x \in S$ in web corpus, and identify co-occurring phrases;
\item[-] generalize phrases into patterns by substituting $x$ with a placeholder $\$X$;
\item[-] analyze frequencies of patterns (and other statistics);
\item[-] $P$ $\leftarrow$ $P ~ \cup ~$ frequent patterns;
\item[2.] Statement expansion:
\item[-] search for occurrences of patterns $p \in P$ in web corpus, and identify co-occurring entities;
\item[-] analyze frequencies of entities co-occurring with multiple patterns (and other statistics);
\item[-] $S$ $\leftarrow$ $S ~ \cup ~$ frequent entities;
\squishend
\squishend
\squishend
\end{mdframed}
\end{samepage}

A toy example for running this algorithm is shown in Table \ref{table:ExampleSeedBasedPatternLearning}.
\begin{table}
\begin{center}
{\small
\begin{tabular}{|l l l|}\hline
0    & sentence: & Singers like Elvis Presley were kings of rock'n'roll.\\ \hline
1.1       & new pattern: & {\em singers like $\$X$} \\ \hline
1.2   & sentence: & Singers like %
Nina Simone fused jazz, soul and gospel.\\
       & new entity: & \ent{Nina Simone} \\\hline
2.1  & sentence: & Nina Simone's vocal performance was amazing.\\
       & new pattern: & {\em $\$X$'s vocal performance}\\\hline
2.2 & sentence: &  Amy Winehouse's vocal performance won a Grammy. \\
      & new entity: & \ent{Amy Winehouse} \\
 & sentence: & Queen's vocal performance led by Freddie Mercury \dots\\  
 & new entity: & \ent{Queen} \\   \hline
3.1 & sentence: & The voice of Amy Winehouse reflected her tragic life. \\
 & sentence: & The voice of Elvis Presley is an incredible baritone. \\
 & new pattern: & {\em voice of $\$X$} \\ \hline
3.2 & sentence: & The melancholic voice of Francoise Hardy \dots \\
 & new entity: & \ent{Francoise Hardy} \\
 & sentence: & The great voice of Donald Trump got loud and angry.\\
 & new entity: & \ent{Donald Trump} \\ \hline
  ... & ... & ...    \\ \hline
\end{tabular}
}%
\caption{Toy example for Seed-based Pattern Learning, with seed ``Elvis Presley''}
\label{table:ExampleSeedBasedPatternLearning}
\end{center}
\end{table}

\ignore{
\begin{table}
\begin{center}
\begin{tabular}{|l|l|l|l|}\hline
Round & Seeds & Patterns & New Entities \\ \hline\hline
 1  & Bob Dylan & singers like $\$X$ & Frank Sinatra \\
                & Elvis Presley & &  Mick Jagger \\
                &  & &  Nina Simone \\ \hline
2  &   & $\$X$'s vocal performance & Lisa Gerrard \\
                 &   & Grammy winner $\$X$ & Amy Winehouse \\
                 &   &   & Rolling Stones \\ \hline
3  &   & $\$X$ and her chansons &  Francoise Hardy \\
                 &   &  $\$X$'s great voice & Donald Trump \\\hline
 ... & ... & ...\\\hline
\end{tabular}
\caption{Toy example for the Seed-based Pattern Learning algorithm.
\fms{May I suggest to replace this Table by Table~\ref{table:ExampleSeedBasedPatternLearning}? Feel free to further modify that table.}}
\label{table:ExampleSeedBasedPatternLearning2}
\end{center}
\end{table}
}%

In the example, phrases such as 
``$\$X$'s vocal performance'' 
(with $\$X$ as a placeholder for entities) 
are 
generalized patterns. They co-occur with at least one but
typically multiple of
the entities in $S$ known so far, and their strength is the
cumulative frequency of these occurrences.
Newly discovered patterns also co-occur with seed entities, and
this would further strengthen the patterns' usefulness. 
The NELL project \cite{nell}
has run a variant of this algorithm at large scale, and has found 
patterns for musicians, 
such as ``original song by $X$'', ``ballads reminiscent of $X$'', ``bluesmen , including $X$'', ``was later covered by  $X$'', ``also performed with $X$'', ``$X$ 's backing bands'', and hundreds more (see {\small\url{http://rtw.ml.cmu.edu/rtw/kbbrowser/predmeta:musician}}).

Despite its elegance, the algorithm, in this basic form, has severe limitations:
\squishlist
\item[1.] Over-specific patterns:\\
Some patterns are overly specific. For example, a possible pattern ``$\$X$
and her chansons'' would apply only to female singers. This can be overcome
by generalizing patterns into {\em regular expressions} over words and part-of-speech tags \cite{Etzioni:ArtInt2005}:
``$\$X$ $\ast$ and $PRP$ chansons'' in this example, where $\ast$ is a wildcard
for any word sequence and $PRP$ requires a personal pronoun.
Likewise, the pattern ``$\$X$'s great voice'' could be generalized into
``$\$X$'s $JJ$ voice'' to allow for other adjectives (with part-of-speech tag $JJ$)
such as ``haunting voice'' or ``angry voice''.
Moreover, instead of considering these as sequences over the surface text,
the patterns could also be derived from paths in dependency-parsing trees
(see, e.g., 
\cite{DBLP:conf/naacl/BunescuM05,DBLP:conf/conll/MoschittiPB06,DBLP:conf/acl/SuchanekIW06}).
\item[2.] Open-ended iterations:\\
In principle, the loop could be run forever. Recall will continue to increase,
but precision will degrade with more iterations, as the acquired patterns get diluted.
So we need to define a meaningful stopping criterion.
A simple heuristic could be to consider the fraction of seed occurrences 
observed in the output at the end of iteration $i$, as the new patterns should still co-occur
with known entities. So a sudden drop in the observations of seeds could possibly indicate
a notable loss of quality. 
\item[3.] False positives and pattern dilution:\\
Even after one or two iterations, some of the newly observed statements
are false positives: 
Queen is a band, not a singer,
and Donald Trump does not have a lot of musical talent.
This is caused by picking up overly broad or ambiguous patterns.
For example,
``Grammy winner'' applies to bands as well, and 
``$\$X$'s great voice'' may be observed in sarcastic news about politics,
besides music.
\squishend

As already stated for points 1 and 2, these weaknesses can be ameliorated by
extending the method.
The hardest issue is point 3. 
To mitigate the potential dilution of patterns, a number of techniques
have been explored. One is to impose additional constraints for
pruning out misleading patterns and spurious statements; we will
discuss these in Chapter \ref{ch4:properties} for the more general task of acquiring relational statements.
A second major technique is to compute statistical measures of
pattern and statement quality after each iteration, and use these to
drop doubtful candidates
\cite{DBLP:conf/dl/AgichteinG00}.
In the following, we list some of the salient measures that
can be leveraged for this purpose.
Recall that statements are usually 
subject-predicate-object triples, here of the form
{\em $\langle$ entity \ent{type} class $\rangle$},
and patterns often refer to sentences or other text passages, but would also cover contextual cues from lists or tables.

\begin{mdframed}[backgroundcolor=blue!5,linewidth=0pt]
\squishlist
\item[ ] 
The {\bf support} of pattern $p$ for seed statements $S_0$, 
$supp(p,S_0)$, is
the 
ratio of the frequency of
joint occurrences of $p$ with any of the entities $x \in S_0$
to the 
co-occurrence frequency of any pattern with any $x \in S_0$:
$$\textit{supp}(p,S_0) = 
\frac{\sum_{x \in S_0} \textit{freq}(p,x)}{\sum_q \sum_{x \in S_0} \textit{freq}(q,x)}$$
where $\sum_q$ ranges over all observed patterns $q$ 
(possibly with a lower bound on absolute frequency) and
$\textit{freq}(~)$ is the total number of observations of a pattern or 
pattern-entity pair.
\squishend
\end{mdframed}

\begin{mdframed}[backgroundcolor=blue!5,linewidth=0pt]
\squishlist
\item[ ] The {\bf confidence} of pattern $p$ with regard to seed statements $S_0$
is the ratio of the frequency of $p$ jointly with seed entities $x \in S_0$
to the frequency of $p$ with any entities:
$$\textit{conf}(p,S_0) = \frac{\textit{freq}(p,S_0)}{\textit{freq}(p)}$$
\squishend
\end{mdframed}

\begin{mdframed}[backgroundcolor=blue!5,linewidth=0pt]
\squishlist
\item[ ] The {\bf diversity} of pattern $p$ with regard to seed statements $S_0$,
$\textit{div}(p,S_0)$, is the number of distinct entities $x \in S_0$ that co-occur with 
pattern $p$:
$$\textit{div}(p,S_0) = |\{x \in S_0 ~|~ \textit{freq}(p,x) > 0\}|$$
\squishend
\end{mdframed}

We can also contrast the positive occurrences of a pattern $p$, that is,
co-occurrences with correct statements from $S_0$, against the negative
occurrences with statements known to be incorrect.
To this end, we have to additionally compile a set of incorrect statements
as {\bf negative seeds}, for example, specifying that the Beatles and
Barack Obama are not singers to prevent noisy patterns that led to
acquiring 
Queen and Donald Trump as new singers.
Let us denote these negative seeds as $\overline{S}_0$.
This allows us to revise the definition of confidence:

\begin{mdframed}[backgroundcolor=blue!5,linewidth=0pt]
\squishlist
\item[ ] Given positive seeds $S_0$ and negative seeds $\overline{S}_0$,
the {\bf confidence} of pattern $p$,
$\textit{conf}(p)$, is 
the ratio of positive occurrences to occurrences with either positive or negative seeds:
$$\textit{conf}(p) = 
\frac
{\sum_{x \in S_0} \textit{freq}(p,x)}
{\sum_{x \in S_0} \textit{freq}(p,x) ~+~ \sum_{x \in \overline{S}_0} \textit{freq}(p,x)}$$
\squishend
\end{mdframed}

These quality measures can be used to restrict the acquired patterns
to those for which support, confidence or diversity -- or any combination
of these -- are above a given threshold.
Moreover, the measures can also be carried over to the observed
statements. This is again based on the principle of statement-pattern duality.
To this end, we now identify a subset $P$ of {\em good patterns}
using the statistical quality measures.

\begin{samepage}
\begin{mdframed}[backgroundcolor=blue!5,linewidth=0pt]
\squishlist
\item[ ] 
The {\bf confidence of statement $x$} (i.e., that an entity $x$ belongs to
the class of interest) is
the normalized aggregate frequency of co-occurring with good patterns,
weighted by the confidence of these patterns:
$$\textit{conf}(x) = \frac{\sum_{p \in P} \textit{freq}(x,p) \cdot \textit{conf}(p)}{\sum_{q} \textit{freq}(x,q)}$$
where $\sum_q$ ranges over all observed patterns.
That is, we achieve perfect confidence in statement $x$ if it is
observed only in conjunction with good patterns and all these patterns have
perfect confidence 1.0.
\squishend
\end{mdframed}
\end{samepage}

\begin{mdframed}[backgroundcolor=blue!5,linewidth=0pt]
\squishlist
\item[ ] The {\bf diversity of statement $x$} is the number of
distinct patterns $p \in P$ that  $x$ co-occurs with:
$$\textit{div}(x) = |\{p \in P ~|~ \textit{freq}(p,x) > 0\}|$$
\squishend
\end{mdframed}

For both of these measures, variations are possible as well as
combined measures.
Diversity is a useful signal to avoid that a single pattern
drives the acquisition of statements, which would incur
a high risk of error propagation.
Rather than relying directly on these measures,
it is also possible to use probabilistic models
or random-walk techniques for scoring and ranking
newly acquired statements.
In particular, algorithms for {\bf set expansion}
(aka. concept expansion)
can be considered to this end 
(\citet{DBLP:conf/icdm/WangC08}, see also \cite{DBLP:conf/www/HeX11,DBLP:conf/www/WangCHGCB15,DBLP:conf/wsdm/ChenCJ16}).

We can now leverage these considerations to extend the
seed-based pattern learning algorithm.
The key idea is to prune, in each round of the iterative method,
both patterns and statements that do not exceed certain thresholds
for support, confidence and/or diversity.
Conversely, we can {\em promote}, in each round, the best statements
to the status of seeds, to incorporate them into the calculation
of the quality statistics for the next round.
For example, when observing \ent{Amy Winehouse} as a high-confidence
statement in some round, we can add her to the seed set, this
way enhancing the informativeness of the statistics in the next round.
We sketch this extended algorithm, whose key ideas have been
developed by \cite{DBLP:conf/dl/AgichteinG00} and \cite{Etzioni:ArtInt2005}.

\begin{samepage}
\begin{mdframed}[backgroundcolor=blue!5,linewidth=0pt]
\squishlist
\item[ ] {\bf Extended algorithm for Seed-based Pattern Learning}\\
Input: Seed statements in the form of known entities for a class\\
Output: Patterns for this class, and new entities of the class
\squishlist
\item[ ] Initialize:\\
 $~~~ S$ $\leftarrow$ seed statements\\
 $~~~ S^+$ $\leftarrow$ $S$ //acquired statements\\
 $~~~ P$ $\leftarrow$ $\emptyset$ (or pre-specified patterns like Hearst patterns)
\item[ ] Repeat
\squishlist
\item[1.] Pattern discovery:
\item[-] same steps as in base algorithm
\item[-] $P$ $\leftarrow$ $P~ \setminus$ $\{$patterns below quality thresholds$\}$
\item[2.] Statement expansion:
\item[-] same steps as in base algorithm
\item[-] $S^+$ $\leftarrow$ $S^+ ~ \cup \{$newly acquired statements$\}$
\item[-] $S$ $\leftarrow$ $S ~ \cup$ $\{$new statements above quality thresholds$\}$
\squishend
\squishend
\squishend
\end{mdframed}
\end{samepage}

All of the above assumes that patterns are either matched or not.
However, it is often the case that a pattern is almost but not exactly
matched, with small variations in the wording or using synonymous words.
For example, the pattern ``Nobel prize winner $\$X$'' (for \ent{scientists} for
a change) could be considered
as {\em approximately matched} by phrases such as
``Nobel winning $\$X$ or
``Nobel laureate $\$X$''.
If these phrases are frequent and we want to consider them
as evidence, we can add a {\em similarity kernel} $sim(p,q)$ to the 
observation statistics, based on {\em edit distance} or {\em n-gram overlap}
where n-grams are sub-sequences of $n$ consecutive characters.
This simply extends the frequency of pattern $p$ into a 
weighted count $\textit{freq}(p) = \sum_q \textit{sim}(p,q)$ if $\textit{sim}(p,q) > \theta$
where $\sum_q$ ranges over all approximate matches of $p$ in the corpus
and $\theta$ is a pruning threshold to eliminate 
weakly matching phrases.

The presented techniques are also applicable to acquiring
subclass/superclass pairs (for the \ent{subclass-of} relation,
as opposed to \ent{instance-of}). 
For example, we can detect from patterns in web pages that
\ent{rappers} and \ent{crooners} are subclasses of \ent{singers}.
This has been further elaborated in work on
ontology/taxonomy learning
\cite{DBLP:conf/ecai/MaedcheS00,Etzioni:ArtInt2005}.

\section{Machine Learning for Sequence Labeling}
\label{ch3-subsect-machinelearning}

A major alternative to learning patterns for entity discovery
is to devise {\bf end-to-end machine learning}
models. In contrast to the paradigm of seed-based distant supervision
of Subsection \ref{ch3-subsect-patterns}, 
we now consider {\em fully supervised} methods that
require {\bf labeled training data} in the form of
{\em annotated sentences} (or other text snippets).
Typically, these methods work well only if the training
data has a substantial size, in the order of ten thousand
or higher. 
By exploiting large corpora with appropriate markup,
or annotations from crowdsourcing workers, 
or high-quality outputs of employing methods like those
for premium sources,
such large-scale training data is indeed available today.
On the first direction, Wikipedia is again a first-choice asset,
as it has many sentences where a named entity appears and is
marked up as a hyperlink to the entity's Wikipedia article.

In the following, we present two major families of
end-to-end learning methods. The first one is 
{\bf probabilistic graphical models} where a sequence
of words, or, more generally, {\em tokens}, is mapped into
the joint state of a graph of random variables, with
states denoting tags for the input words.
The second approach is {\bf deep neural networks}
for classifying the individual words of an input
sequence onto a set of tags.
Thus, both of these methods are geared for the task
of {\bf sequence labeling}, also known as {\bf sequence tagging}.

\subsection{Probabilistic Graphical Models}
\label{ch4-subsec:CRF}

This family of models considers a set of coupled random variables
that take a finite set of tags as values.
In our application, the tags are primarily used to
demarcate entity names in a token sequence, like
an input sentence or other snippet.
Each random variable corresponds to one token in the
input sequence, and the coupling reflects short-distance dependencies (e.g., between the tags for adjacent or nearby tokens).
In the simplest and most widely used case, the coupling is pair-wise
such that a random variable for an input token depends only on the
random variable for the immediately preceding token.
This setup amounts to learning conditional probabilities for
subsequent pairs of tags as a function of the input tokens.
As the input sequence is completely known upfront, we can generalize
this to each random variable being a function of all tokens or
all kinds of {\em feature functions} over the entire token sequence.

\vspace*{0.2cm}
\noindent {\bf Conditional Random Fields (CRF):}\\
The most successful method from this family of probabilistic graphical models
is known as {\bf Conditional Random Fields}, or {\bf CRF}s for short,
originally developed by \citet{DBLP:conf/icml/LaffertyMP01}.
More recent tutorials are by \citet{sutton2012introduction} 
on foundations and algorithms, and
\citet{sarawagi2008information} on applying CRFs for information extraction. 
CRFs are in turn a generalization of the prior notion of
{\em Hidden Markov Models (HMMs)}, the difference lying in the 
incorporation of feature functions on the entire input sequence
versus only considering two successive tokens.

\begin{mdframed}[backgroundcolor=blue!5,linewidth=0pt]
\squishlist
\item[ ] A {\bf Conditional Random Field (CRF)}, operating
over input sequence $X = x_1 \dots x_n$ is an undirected graph,
with a set of finite-state random variables 
$Y = \{Y_1 \dots Y_m\}$ as nodes 
and pair-wise coupling of variables as edges.\\
An edge between variables $Y_i$ and $Y_j$ denotes
that their value distributions are coupled.
Conversely, it denotes that, in the absence of any
other edges, two variables are conditionally independent,
given their neighbors.\\
More precisely, the following {\em Markov condition}
is postulated for all variables $Y_i$ and all possible values $t$:
$$ P[Y_i=t ~|~ x_1 \dots x_n, ~ Y_1 \dots Y_{i-1}, Y_{i+1} \dots Y_m]
=$$
$$ P[Y_i=t ~|~ x_1 \dots x_n, ~ \text{all}~ Y_j \text{ with an edge ~} (Y_i, Y_j)] $$

\squishend
\end{mdframed}

In general, the coupling of random variables may go
beyond pairs by introducing {\em factor nodes} (or {\em factors} for short)
for dependencies between two or more variables.
In this section, we restrict ourselves to the basic case of
pair-wise coupling. Moreover, we assume that the graph forms
a linear chain: a so-called {\bf linear-chain CRF}.
Often the variables correspond one-to-one to the input tokens:
so each token $x_i$ is associated with variable $Y_i$ (and we have $n=m$
for the number of tokens and variables).

\vspace*{0.2cm}
\noindent{\bf CRF Training:}\\
The {\em training} of a CRF from labeled sequences, in the form of
$(X,Y)$ value pairs, involves
the posterior likelihoods
$$P[Y_i|X] = P[Y_i=t_i ~|~ x_1 \dots x_n, ~ \text{all  neighbors } Y_j \text{~of~} Y_j]$$
With feature functions $f_k$ over input $X$ and 
subsets $Y_c \subset Y$ of coupled random variables
(with known values in the training data),
this can be shown to be
equivalent to 
$$P[Y|X] \sim \frac{1}{Z} \prod_{c} \exp(~\sum_{k} w_k \cdot f_k(X, ~ Y_c )]$$
with an input-independent normalization constant $Z$,
$k$ ranging over all feature functions,
and $c$ ranging over all coupled subsets of variables, the so-called
{\em factors}. 
For a linear-chain CRF, the factors are all pairs of adjacent variables:
$$P[Y|X] \sim \frac{1}{Z} \prod_{i} \exp(~\sum_{k} w_k \cdot f_k(X, Y_{i-1}, Y_i)]$$
The parameters of the model are the feature-function weights $w_k$;
these are the output of the training procedure.
The training objective is to choose weights $w_k$ to minimize the error between 
the model's maximum-posterior $Y$ values
and the ground-truth values,
aggregated over all training samples.
The error, or {\em loss function}, can take different forms,
for example, the negative log-likelihood that the trained model
generates the ground-truth tags.

As with all machine-learning models, the objective function
is typically combined with a {\em regularizer} to counter the
risk of overfitting to the training data.
The training procedure is usually implemented as a form of
{\em (stochastic) gradient descent} 
(see, e.g.,
\cite{Burkov2019hundred} and further references given there). 
This guarantees convergence
to a local optimum and empirically approximates the
global optimum of the objective function fairly well (see
\cite{sutton2012introduction}). 
For the case of linear-chain CRFs, the optimization is convex;
so we can always approximately achieve the global optimum.

\vspace*{0.2cm}
\noindent{\bf CRF Inference:}\\
When a {trained CRF} is presented with a previously unseen
sentence, the {\em inference} stage computes the posterior values of all
random variables, that is, the tag sequence for the entire input,
that has the {\em maximum likelihood} given the input and the trained model
with weights $w_k$:
$$Y^* = \argmax_Y P[Y ~|~ X, ~ \text{all} ~ w_k]$$
For linear-chain CRFs, this can be computed using
{\em dynamic programming}, 
namely, variants of the Viterbi algorithm (for HMMs). 
For general CRFs, other -- more expensive -- 
techniques like
{\em Monte Carlo sampling} or variational inference are needed.
Alternatively, the CRF inference can also be cast into
{\em an Integer Linear Program (ILP)} and solved
by optimizers like the Gurobi software ({\small\url{https://www.gurobi.com/}}),
following \cite{DBLP:conf/icml/RothY05}.
We will go into more depth on ILPs as a modeling and inference tool
in Chapter \ref{chapter:KB-curation},
especially Section \ref{ch8-subsec:ILP}.

\begin{figure} [h!]
  \centering
   \includegraphics[width=0.8\textwidth]{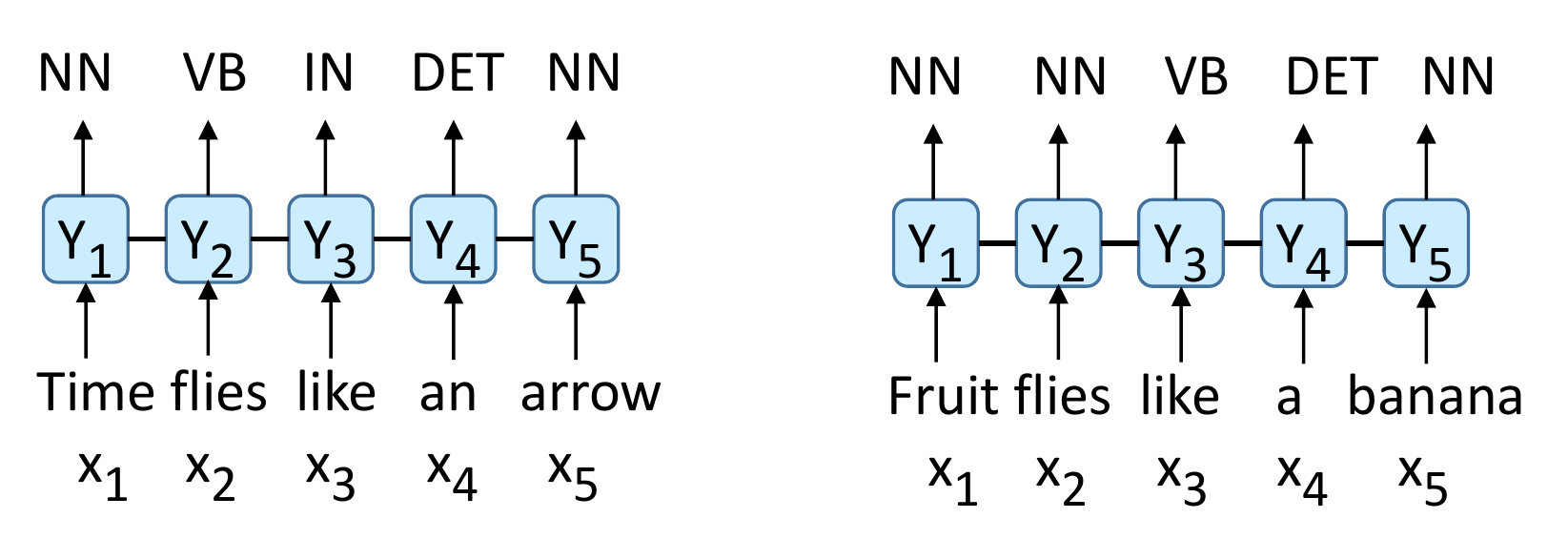}
      \caption{Examples for CRF-based Part-of-Speech Tagging}
      \label{ch3-fig-crf4pos}
\end{figure}

\vspace*{0.2cm}
\noindent{\bf CRF for Part-of-Speech Tagging:}\\
A classical application of CRF-based learning is
part-of-speech tagging: labeling each word in an input sentence
with its word category, like noun (NN), verb (VB), preposition (IN), article (DET) etc.
Figure \ref{ch3-fig-crf4pos} shows two examples for this task,
with their correct output tags.

The intuition why this works so well is that word sequence frequencies
as well as tag sequence frequencies from large corpora can inform
the learner, in combination with other feature functions, to
derive very good weights.
For example, nouns are frequently followed by verbs, verbs are 
frequently followed by prepositions, and some word pairs are
composite noun phrases (such as ``fruit flies'').
Large corpus statistics also help to cope with exotic or
even non-sensical inputs. For example, the sentence
``Pizza flies like an eagle'' would be properly tagged
as {\em NN VB IN DET NN} because, unlike fruit flies,
there is virtually no mention of pizza flies in any corpus.

\vspace{0.2cm}
\noindent{\bf CRF for Named Entity Recognition (NER) and Typing:}\\
For the task at hand, entity discovery, the CRF tags of interest
are primarily $NE$ for Named Entity and $O$ for Others.
For example, the sentence \\
\hspace*{0.5cm} ``Dylan composed the song Sad-eyed Lady of the Lowlands,\\
\hspace*{0.5cm} about his wife Sara Lownds,
while staying at the Chelsea hotel'' \\
should yield the tag sequence\\
\hspace*{0.5cm} {\em NE O O O NE NE NE NE NE O O O NE NE O O O O NE NE}.\\
Many entity mentions correspond to the part-of-speech tag
NNP, for proper noun, that is, nouns that should not be prefixed with
an article in any sentence, such as names of people.
However, this is not sufficient, as it would accept false positives
like abstractions (e.g., ``love'' or ``peace'') and would miss out
on multi-word names that include non-nouns, such as song or book titles
(e.g. ``Sad-eyed Lady of the Lowlands''),
and names that come with an article (e.g., ``the Chelsea hotel'').
For these reasons,  CRFs for {\em Named Entity Recognition}, or {\em NER}
for short, have been specifically developed, with training over
annotated corpora. The seminal work on this
is \citet{DBLP:conf/acl/FinkelGM05};
advanced extensions are implemented in the 
Stanford CoreNLP software suite (\cite{DBLP:conf/acl/ManningSBFBM14}).

As the training of a CRF involves annotated corpora,
instead of merely distinguishing entity mentions versus other words,
we can piggyback on the annotation effort and incorporate
more expressive tags for {\em different types of entities},
such as people, places, products (incl. songs and books).
This idea has indeed been pursued already in the work of
\cite{DBLP:conf/acl/FinkelGM05}, integrating
{\bf coarse-grained entity typing} into the CRF for NER.
The simple change is to move from output variables
with tag set $\{NE, O\}$ to a larger tag set like
$\{PERS, LOC, ORG, MISC, O\}$
denoting persons (PERS), locations (LOC), organizations (ORG),
entities of miscellanous types (MISC) such as products or events, 
and non-entity words (O).
For the above example about Bob Dylan, we should then obtain the 
tag sequence\\
\hspace*{0.5cm}{\em PERS O O O MISC MISC MISC MISC MISC}\\
\hspace*{0.5cm}{\em O O O PERS PERS O O O O LOC LOC}.

The way the CRF is trained and used for inference stays the same,
but the training data requires annotations for the entity types.
Later work has even devised (non-CRF) classifiers for more {\bf fine-grained entity typing},
with tags for hundreds of types, such as politicians, scientists, artists, musicians,
singers, guitarists, etc.
\cite{DBLP:conf/coling/FleischmanH02,DBLP:conf/aaai/LingW12,DBLP:conf/acl/NakasholeTW13,DBLP:conf/acl/LevyZCC18}.
An easy way of obtaining training data for this task is
consider hyperlink anchor texts in Wikipedia as entity names
and derive their types from Wikipedia categories, 
or directly from a core KB constructed
by methods from Chapter \ref{ch2:knowledge-integration}. %
An empirical comparison of various NER methods with fine-grained typing
is given by
\cite{DBLP:conf/coling/MaiPNDBSS18}.

Widely used feature functions for CRF-based NER tagging, or features for other
kinds of NER/type classifiers, include the following:
\squishlist
\item part-of-speech tags of words and their co-occurring words in left-hand and right-hand proximity,
\item uppercase versus lowercase spelling,
\item word occurrence statistics in type-specific dictionaries, such as
dictionaries of people names, organization names, 
or location names, along with short descriptions
(from yellow pages and so-called gazetteers),
\item co-occurrence frequencies of word-tag pairs in the training data,
\item further statistics for word n-grams and their co-occurrences with tags.
\squishend

There is also substantial work on {\bf domain-specific NER}, especially for the
biomedical domain (see, e.g., \cite{DBLP:journals/bmcbi/FunkBGRBCHV14} and references given there),
and also for chemistry, restaurant names and menu items, and titles of
entertainment products.
In these settings, domain-specific dictionaries play a strong role
as input for feature functions
\cite{DBLP:conf/conll/RatinovR09,DBLP:conf/emnlp/ShangLGRR018}.

\subsection{Deep Neural Networks}
\label{subsec:deep-neural-networks}

In recent years, neural networks have become the
most powerful approach for supervised machine learning
when sufficient training data are available.
This holds for a variety of NLP tasks \cite{Goldberg2017},
potentially including Named Entity Recognition.

The most primitive neural network is a single {\em perceptron}
which takes as input a set of real numbers, aggregates them
by weighted summation, and applies a 
{\em non-linear activation function} (e.g., logistic function or
hyperbolic tangent) to the sum 
for producing its output.
Such building blocks can be connected to construct
entire networks, typically organized into layers.
Networks with many layers are called {\em deep networks}.
As a loose metaphor, one may think of the nodes as neurons
and the interconnecting edges as synapses.

The weights of incoming edges (for the weighted summation)
are the parameters of such neural models, to be learned
from labeled training data.
The inputs to each node are usually entire vectors, not just
single numbers, and the top-layer's output are real values for
regression models or, after applying a softmax function,
scores for classification labels.
The loss function for the training objective can take various
forms of error measures, possibly combined with regularizers
or constraint-based penalty terms.
For neural learning, it is crucial that the loss function is
differentiable in the model parameters (i.e., the weights),
and that this can be backpropagated through the entire network.
Under this condition, training is effectively performed by
methods for {\em stochastic gradient descent}
(see, e.g., \cite{Burkov2019hundred}), and modern software libraries 
(e.g., TensorFlow) support scaling out these computations across many processors.
For inference, with new inputs outside the training data, 
input vectors are simply fed forward through the network
by performing matrix and tensor operations at each layer.

\vspace*{0.2cm}
\noindent{\bf LSTM Models:}\\
There are various families of neural networks, with different
topologies for interconnecting layers. 
For NLP where the input to the entire network is a 
text sequence, so-called {\bf LSTM networks} 
(for ``Long Short Term Memory'') have become prevalent
(see \cite{DBLP:journals/nn/Schmidhuber15,Goldberg2017}
and references there).
They belong to the broader family of {\em recurrent neural networks}
with feedback connections between nodes of the same layer.
This allows these nodes to aggregate latent state computed
from seeing an input token and the latent state derived from
the preceding tokens.
To counter potential bias from processing the input sequence
in forward direction alone, {\bf Bi-directional LSTM}s (or {\bf Bi-LSTMs} for short)
connect nodes in both directions. 

We can think of LSTMs as the neural counterpart of CRFs.
A key difference, however, is that neural networks do not
require the explicit modeling of feature functions.
Instead, they take the raw data (in vectorized form) as inputs
and automatically learn latent representations that implicitly capture 
features and their cross-talk.

\begin{figure} [h!]
  \centering
   \includegraphics[width=1.0\textwidth]{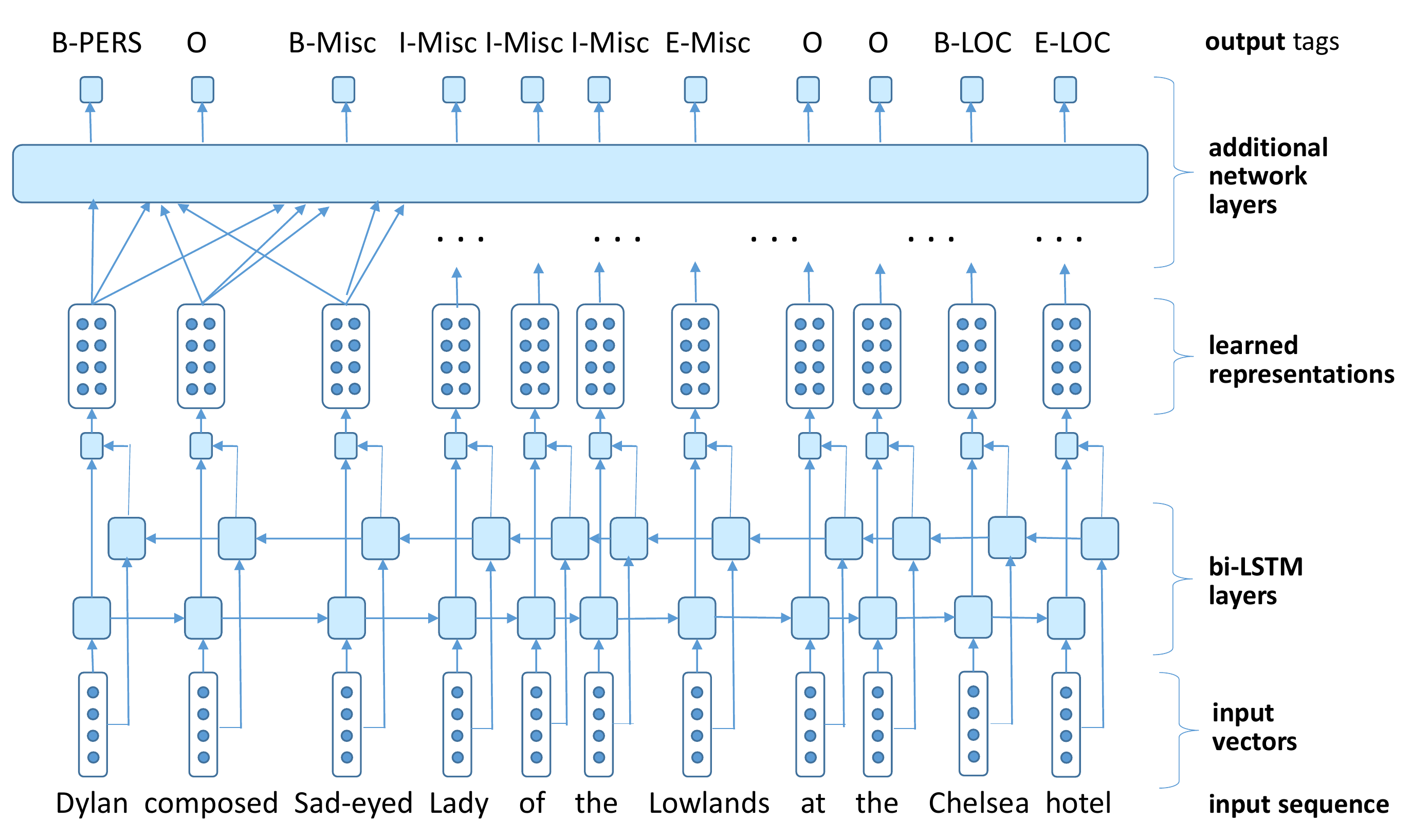}
      \caption{Illustration of LSTM network for NER}
      \label{ch3-fig-lstm4ner}
\end{figure}

Figure \ref{ch3-fig-lstm4ner} gives a pictorial illustration of an
LSTM-based neural network for NER, applied to a variant of our
Bob Dylan example sentence.
The outputs of the forward LSTM and the backward LSTM are
combined into a latent data representation, for example,
by concatenating vectors.
On top of the bi-LSTM, further network layers (learn to) compute the
scores for each tag, typically followed by a softmax function
to choose the best label.
The output sequence of tags is slightly varied here, by prefixing
each tag with its role in a subsequence of identical tags:
B for Begin, I for In, and E for End.
This serves to distinguish the case of a single multi-word mention
from the case of different mentions without any interleaving ``Other'' words.
The E tags are not really needed as a following B tag indicates the next mention
anyway, hence E is usually omitted.
This simple extension of the tag set is also adopted by CRFs and other 
sequence labeling learners; we disregarded this earlier for simplicity.

LSTM-based neural networks can be combined with a CRF on top of the neural layers,
this way combining the strengths of the two paradigms
(\citet{DBLP:journals/corr/HuangXY15,DBLP:conf/acl/MaH16,DBLP:conf/naacl/LampleBSKD16}).
The strengths of LSTM and CRF models complement each other.
The neural LSTM is a highly effective 
{\em encoder} for representation learning,
capturing non-linear patterns 
without requiring any hand-crafted features.
The CRF, on the other hand, is better geared for considering
dependencies between the output labels of different input tokens.
It operates as a {\em decoder} on top of the neurally learned
latent representation.

Other enhancements (see \citet{DBLP:journals/corr/abs-1812-09449} for a survey of
neural NER)
include additional bi-LSTM layers for learning
character-level representations, capturing character n-grams in a latent manner.
This can leverage large unlabeled corpora, analogously to the role of
dictionaries in feature-based taggers. 
This line of methods has also extended the scope of {\em fine-grained entity typing},
yielding labels for thousands of types
(\citet{DBLP:conf/acl/LevyZCC18}).

Overall, deep neural networks, in combination with CRFs, tend to outperform
other methods for NER whenever a large amount of training data is at hand.
When training data is not abundant, for example, in specific domains such as
health, pattern-based methods and feature-driven
graphical models (incl. CRFs) are still a good choice.

\section{Word and Entity Embeddings}
\label{subsec:word-entity-embeddings}

Machine learning methods, and especially neural networks,
do not operate directly on text, as they require numeric
vectors as inputs. A popular way of casting text into
this suitable form is by means of {\bf embeddings}.

\begin{samepage}
\begin{mdframed}[backgroundcolor=blue!5,linewidth=0pt]
\squishlist
\item[ ] {\bf Word Embedding:}\\
Embeddings of words (or multi-word phrases) are 
real-valued vectors of fixed dimensionality, such that
the 
distance between two vectors (e.g., by their cosine)
reflects the {\em relatedness}
(sometimes called ``semantic similarity'') of the two words,
based on the respective contexts in which the words typically occur.
\squishend
\end{mdframed}
\end{samepage}

Word embeddings are computed (or ``learned'') from
co-occurrences and neighborhoods of words in large corpora.
The rationale is that the meaning of a word is
captured by the 
contexts in which it is often used,
and that two words are highly related if they
are used in similar contexts.
This is often referred to as ``distributional semantics''
or ``distributed semantics'' of words \cite{DBLP:journals/tacl/LevyGD15}.

This hypothesis should not be confused with two words directly
co-occurring together. 
Instead, we are interested in indirect co-occurrences
where the contexts of two input words share many
words.
For example, the words ``car'' and
``automobile'' have the same semantics, but rarely 
co-occur directly in the same text span. 
The point rather is that
both often co-occur with third words such as ``road'',
``traffic'', ``highway'' and names of car models.

Technically, this becomes an optimization problem.
Given a large set of text windows 
$C$ of length $k+1$ with word sequences
$w_0 \dots w_k$,
we aim to compute a fixed-length vector $\vec{w}$
for each word $w$ such that the error for predicting
the word's surrounding window 
$C(w_t) = w_{t-k/2} \dots w_t \dots w_{t+k/2}$,
from all word-wise vectors alone,
is minimized.
This consideration
leads to a non-convex continuous optimization 
with the objective function \cite{Mikolov:NIPS2013}:
$$maximize ~~ \sum_C 
\sum_{j \in C(w_t), j \ne t}
\log \frac{exp(\vec{w_j}^T \cdot \vec{w_t})}
{\sum_v exp(\vec{v~}^T \cdot \vec{w_t})}$$
\noindent where the outermost sum ranges over all possible
text windows (with overlapping windows).
The dot product between the output vectors $\vec{w_j}$
and $\vec{w_t}$ reflects 
overlapping-context
likelihoods
of word pairs,
and the softmax function normalizes these scores
by considering all possible words $v$.
Intuitively, the objective is maximized if the
resulting word vectors can predict the surrounding
window from a given word with high accuracy.
This specific objective is known as the {\em skip-gram model};
there are also other variations, with a similar flavor.
Computing solutions for the non-convex optimization is
typically done via gradient descent methods.

Embedding vectors can be readily plugged into
machine-learning models, and they are a major asset
for the power of neural networks for NLP tasks.
A degree of freedom is the choice for the dimensionality
of the vectors. 
For most use cases, this is set to a few hundred,
say 300, for robust behavior.

The most popular models and tools for this
line of text embeddings are word2vec,
by \citet{Mikolov:NIPS2013},
and GloVE, by \citet{DBLP:conf/emnlp/PenningtonSM14}.
Both come with pre-packaged embeddings derived from
news and other collections, but applications can also 
compute new embeddings from customized corpora.
The word2vec approach has been further extended 
to compute embeddings
for short paragraphs and entire documents (called doc2vec).
Important earlier work on latent text representations, with similar but less expressive models, includes
Latent Semantic Indexing (LSI) and
Latent Dirichlet Allocation (LDA)
(\cite{DBLP:journals/jasis/DeerwesterDLFH90,DBLP:journals/ml/Hofmann01,DBLP:journals/jmlr/BleiNJ03}).

A recent, even more advanced way of computing and
representing embeddings is by training deep neural networks
for word-level or sentence-level prediction tasks, and
then keeping the learned model as a building block for
training an encompassing network for downstream tasks
(e.g., question answering or conversational chatbots).

The pre-training utilizes large corpora like the full text
of all Wikipedia articles or the Google Books collection.
A typical objective function is to minimize the error 
in predicting a masked-out word given a text window of
successive words.
The embeddings for all words are jointly given by
the learned weights of the network's synapses 
(i.e., connections
between neurons),
with 100 millions of real numbers or even more.
Popular instantiations of this approach are
ElMo
(\citet{DBLP:conf/naacl/PetersNIGCLZ18}), 
BERT (\citet{DBLP:conf/naacl/DevlinCLT19}) and
RoBERTa (\citet{DBLP:journals/corr/abs-1907-11692}).
\cite{DBLP:journals/tacl/RogersKR20} is a survey
on the underlying issues for network architecture and
training methodology.

\vspace*{0.3cm}
\noindent{\bf Embeddings for Word and Entity Pair Relatedness:}\\
\noindent Embedding vectors are not directly interpretable;
they are just vectors of numbers.
However, we can apply linear algebra operators to them
to obtain further results.
Embeddings are additive (hence also subtractive), which allows
forming analogies of the form:

\begin{center}
\begin{tabular}{rcl}
$\overrightarrow{man} + \overrightarrow{king}$ & = & $\overrightarrow{woman} + \overrightarrow{queen}$\\
$\overrightarrow{France} + \overrightarrow{Paris}$ & = & $\overrightarrow{Germany} + \overrightarrow{Berlin}$\\
$\overrightarrow{Einstein} + \overrightarrow{scientist}$ & = & $\overrightarrow{Messi} + \overrightarrow{footballer}$
\end{tabular}
\end{center}

\vspace*{-0.3cm}
\noindent So we can solve an equation like\\
\vspace*{0.1cm}
\hspace*{1cm}$\overrightarrow{Rock~and~Roll} + \overrightarrow{Elvis~Presley} = \overrightarrow{Folk~Rock} + \overrightarrow{X}$\\
\noindent yielding\\
\vspace*{0.1cm}
\hspace*{1cm}$\overrightarrow{X} = \overrightarrow{Rock~and~Roll} + \overrightarrow{Elvis~Presley} - \overrightarrow{Folk~Rock} ~~~ \approx \overrightarrow{Bob~Dylan}$\\

Most importantly for practical purposes, we can compare
the embeddings of two words (or phrases) by computing
a distance measure between their respective vectors,
typically, the cosine or the scalar product.
This gives us a measure of how strongly the two words
are related to each other, where (near-) synonyms would often have
the highest relatedness.

\begin{samepage}
\begin{mdframed}[backgroundcolor=blue!5,linewidth=0pt]
\squishlist
\item[ ] {\bf Embedding-based Relatedness:}\\
For two words $v$ and $w$, their relatedness can be computed
as $cos(\vec{v},\vec{w})$ from their embedding vectors
$\vec{v}$ and $\vec{w}$.
\squishend
\end{mdframed}
\end{samepage}

The absolute values of the relatedness scores are not crucial,
but we can now easily order related words by descending scores.
For example, for the word ``rock'', the most related words
and short phrases
are ``rock n roll'', ``band'', ``indie rock'' etc.,
and for ``knowledge'' we obtain the most salient words
``expertise'', ``understanding'', ``knowhow'', ``wisdom'' etc.
We will later see that such relatedness measures are very
useful for many sub-tasks in knowledge base construction.

The embedding model can capture not just words or other
text spans, but we can also apply it to compute
{\bf distributional representations of entities}.
This is achieved by associating each entity in the
knowledge base with a textual description of the entity,
typically the Wikipedia article about the entity
(but possibly also the external references given there,
homepages of people and organizations, etc.).

Once we have per-entity vectors, we can again compute
cosine or scalar-product distances for entity pairs.
This results in measures for
{\bf entity-entity relatedness}.
Moreover, by coupling the computations of per-word and
per-entity embeddings, 
we also obtain scores for {\bf entity-word relatedness},
which is often useful when we need salient keywords
or keyphrases for an entity
(most notably, for the task of named entity disambiguation, see Chapter \ref{ch3-sec-EntityDisambiguation}).
For example, the embedding for Elvis Presley should be
close to the embeddings for ``king'', ``rock n roll'', etc.
Technical details for these models can be found in
\cite{DBLP:conf/emnlp/WangZFC14,DBLP:conf/sigir/ZwicklbauerSG16,yamada2020wikipedia2vec}; the latter includes data and code for
the {\em wikipedia2vec} tool.

Such embeddings have also been computed from domain-specific
data sources, most notably, for biomedical entities and
terminology, with consideration of the standard MeSH
vocabulary. Resources of this kind include
BioWordVec (\cite{Zang2019:BioWordVec}) and BioBERT (\cite{10.1093/bioinformatics/btz682}).

An important predecessor to all these works is 
the semantic relatedness model of
\cite{DBLP:conf/ijcai/GabrilovichM07}, which was the
first to harness Wikipedia articles for this purpose.

A related, recent direction is {\em knowledge graph (KG) embeddings} (see
\citet{DBLP:journals/tkde/WangMWG17}
for a survey).
These kinds of embeddings capture the neighborhood
of entities in an existing graph-structured KB.
They do not use textual inputs, however,
and serve different purposes.
We will discuss KG embeddings in Chapter \ref{chapter:KB-curation},
specifically Section \ref{ch8-sec:kgembeddings}.

\section{Ab-Initio Taxonomy Construction}
\label{ch3-sec-TaxonomyConstruction}

Assuming that we can extract a large pool of types, and optionally also
entities for them, the task discussed here is to construct
a taxonomic tree or DAG (directed acyclic graph) for these types
-- without assuming any prior structure such as WordNet.
In the literature, the problem is also referred to as {\bf taxonomy induction}
(\citet{DBLP:conf/acl/SnowJN06,DBLP:journals/ai/PonzettoS11}),
as its output is a generalization of bottom-up observations.
The input can take different forms, for example, starting from the noisy set
of Wikipedia categories (but ignoring the
graph structure), or from noisy and sparse
pairs of hyponym-hypernym candidates
derived by applying patters to large text and web corpora.
A good example for the latter is the {\em WebIsALOD} project
({\small\url{http://webisa.webdatacommons.org/}},
which used more than 50 patterns to extract candidate pairs and a supervised classifier
to prune out the most noisy ones \cite{DBLP:conf/semweb/HertlingP17}.
This collection, and others of similar flavor, does not strictly focus
on hypernymy but also captures
meronymy/holonymy (part-of, hasA) and, to some extent, instance-type pairs.
Hence the broader term {\em IsA} in the project name.

\vspace*{0.2cm}
\noindent{\bf Methods for Wikipedia Categories:}\\
Seminal work that considered all Wikipedia
categories as noisy type candidates and the
subcategory-supercategory pairs as hypernymy candidates
was the {\em WikiTaxonomy} project
by
\citet{DBLP:conf/aaai/PonzettoS07}
\cite{DBLP:journals/ai/PonzettoS11}. 
Its approach can be characterized by three steps:

\begin{samepage}
\begin{mdframed}[backgroundcolor=blue!5,linewidth=0pt]
\squishlist
\item[ ] {\bf Wikipedia-based Taxonomy Induction:}\\
\squishlist
\item {\em Category Cleaning:} 
eliminating noisy categories that do not really denote types.
\item {\em Category-Pair Classification:}
using a rule-based classifier
to eliminate pairs that do not denote hypernymy.
\item {\em Taxonomy Graph Construction:} 
building a tree or DAG from the remaining types and hypernymy pairs.
\squishend
\squishend
\end{mdframed}
\end{samepage}

The first step is very similar to the techniques
presented in Section \ref{sec:ch3-category-cleaning}.
State-of-the-art techniques for this purpose
are discussed by
\citet{Pasca:WWW2018}.
The second step is based on heuristic but powerful
rules that compare 
stems or lemmas of head words in multi-word noun phrases.
The following shows two examples for rules:
\squishlist
\item For sub-category $S$ and direct super-category $C$: if $head(S)$ is the same as $head(C)$, then this is likely a hyponym-hypernym pair (e.g., $S$ = ``American baritones'',
$C$ = ``baritones by nationality'').
\item For $C$ and $S$: if $head(C)$ appears in $S$,
but $head(S)$ is different from $head(C)$, then this is likely not a good pair (e.g., $S$ = ``American baritones'',
$C$ = ``baritone saxophone players''),
\squishend
Additional rules are used to refine the first case
and to handle other cases.
This includes considering instances of a category,
at the entity level, and comparing their set of
categories against the category at hand.

For the third step, {\em graph construction}, 
the method applies transitivity to build a 
multi-rooted graph, eliminates cycles by removing
as few edges as possible, and connects all roots
of the resulting DAG to the universal type \ent{entity}.

An industrial-strength 
(i.e., robust and scalable) 
extension
of the presented method is discussed in
\cite{Deshpande2013kosmix}.
Another Wikipedia component that has been considered as noisy input for taxonomy induction
is {\em infobox templates}: 
mini-schemas with relevant attributes for
particular entity types.
The Wikipedia community has developed a
large number of different templates for
people, musicians, bands, songs, albums etc. 
They are instantiated in highly varying numbers,
and there is redundancy, for example, 
different templates for songs, some used more
than others.
\cite{WuWeld:WWW2008} proposed a learning
approach, using SVM classifiers and
CRF-like graphical models, to infer a clean
taxonomy from this noisy data.

\vspace*{0.2cm}
\noindent {\bf Taxonomies from Catalogs, Networks and User Behavior:}\\
Alternatively to Wikipedia categories, other
catalogs of categories can be processed in a similar manner,
for example, the DMOZ directory of web sites
({\small\url{https://dmoz-odp.org/}})
or the Icecat open product catalog
({\small\url{https://icecat.biz/}}).
Some methods combine
information from catalogs with 
topical networks, for example, connecting
business categories, users and reviews
on sites such as Yelp or TripAdvisor,
and potentially also informative terms
from user reviews.
Examples of such methods are \cite{DBLP:conf/icdm/WangDLDJH13,DBLP:conf/www/ShangZLL020}.
Last but not least, recent methods on this task
start with a high-quality product catalog and
its category hierarchy, and then
learn to extend and enrich the taxonomy with input
from other sources, most notably, logs of customer
queries, clicks, likes and purchases.
This method is part of the {\bf AutoKnow} pipeline
(\citet{DBLP:conf/kdd/DongHKLLMXZZSDM20}),
discussed further in Section \ref{ch9-sec:industrialKG}.

\vspace*{0.2cm}
\noindent {\bf Folksonomies from Social Tags:}\\
The general approach has also been carried over to
build taxonomies from {\em social tagging},
resulting in so-called {\bf folksonomies}
\cite{DBLP:journals/sigkdd/GuptaLYH10}.
The input is a set of items like images or
web pages of certain types that are associated 
with concise {\em tags}
to
annotate items,
such as ``sports car'', ``electric car'', ``hybrid auto''
etc. 
If the number of items and the tagging community 
are very large, the frequencies 
and co-occurrences of (words in) tags provide cues
about proper types as well as type pairs where
one is subsumed by the other. 
Data mining techniques (related to association rules)
can then be applied to clean such a large but noisy
candidate pool, and the subsequent DAG construction
is straightforward
(e.g.,\cite{heymann2006collaborative,DBLP:conf/esws/HothoJSS06,DBLP:journals/ws/JaschkeHSGS08}).

Another target for similar techniques are
fan communities (e.g., hosted at
{\small\url{http://fandom.com}} aka. Wikia),
which have collaboratively
built extensive but noisy category and tagging
systems for entertainment fiction like movie series
or TV series (e.g., Lord of the Rings, Game of Thrones, The Simpsons etc.) \cite{DBLP:conf/www/ChuRW19,HertlingPaulheim:ICBK2018}.

\vspace*{0.2cm}
\noindent{\bf Methods for Web Contents:}\\
Early approaches spotted entity names in web-page collections
and clustered these by various similarity measures
(e.g., \cite{DBLP:conf/aaai/DurmeP08}).
Using Hearst patterns and other heuristics, type labels
are then derived for each cluster. 
Such techniques can be further refined for
scoring and ranking the outputs
(e.g., using 
label propagation
with random-walk techniques \cite{DBLP:conf/emnlp/TalukdarRPRBP08}).

The 
{\bf KnowItAll} project, 
by \citet{Etzioni:ArtInt2005},
advanced this line of research by a suite
of scoring and classification techniques
to enhance the output quality.
It also studied tapping into {\em lists of named entities}
as a source of type cues.
For example, headers or captions of lists may serve
as type candidates, and pairs of lists where one
mostly subsumes the other in terms of elements
(with some tolerance for exceptions) can be
viewed as candidates for hypernymy.

In the {\bf Probase} project
by
\citet{DBLP:conf/sigmod/WuLWZ12}, 
noisy candidates for
hypernymy pairs were mined from the Web index of
a major search engine.
First, Hearst patterns were liberally applied to
this huge text collection.
Then, a probabilistic model was used to prune out
noise and infer likely candidates for hyponym-hypernym
pairs, based on (co-)occurrence frequencies.
The approach led to a huge but still noisy and incomplete
taxonomy. 
Also, the resulting types are not canonicalized,
meaning that synonymous type names may appear as different nodes
in the taxonomy with different neighborhoods of
hyponyms and hypernyms.
Nevertheless, for use cases like 
query recommendation in web search, such a large
collection of taxonomic information can be a valuable asset.

Recent works approached taxonomy induction as
a supervised machine-learning task,
using factors graphs or neural networks, 
or by reinforcement learning
(see, e.g., \cite{DBLP:conf/acl/BansalBMK14,DBLP:conf/acl/ShwartzGD16,DBLP:conf/acl/HanRSMG18}).

\vspace*{0.2cm}
\noindent{\bf Methods for Query-Click Logs:}\\
Search engine companies have huge logs of
query-click pairs: user-issued keyword queries
and subsequent clicks on web pages after seeing
the preview snippets of top-ranked results.
When a sufficiently large fraction of queries
is about types (aka. classes), such as
``American song writers'' or 
``pop music singers from the midwest'', 
one can derive various signals towards
inferring type synonymy and pairs for the IsA relation:
\squishlist
\item {\em Surface cues in query strings:}
frequent patterns of query formulations that
indicate type names.
Examples are queries that start with ``list of''
or noun-phrase query strings with a prefix (or head word)  known to be a type followed by a modifier,
such as ``musicians who were shot'' or
``IT companies started in garages''.
\item {\em Co-Clicks:} pages that are (frequently) clicked upon two different queries. For example, if
the queries ``American song writers'' and
``Americana composers'' have many clicks in common,
they could be viewed as synonymous types.
\item {\em Overlap of query and page title:}
the word-level n-gram overlap between the query string and the title of a (frequently) clicked page.
For example, if the query ``pop music singers from the midwest'' often leads to clicking the page with
title ``baritone singers from the midwest'', this pair is a candidate for the IsA relation
(or, specifically, hypernymy between types
if the two strings are classified to denote types).
\squishend

A variety of methods have been devised to 
harness these cues for inferring taxonomic relations (synonymy and hypernymy, 
or just IsA)
by 
\cite{DBLP:conf/kdd/Baeza-YatesT07,DBLP:conf/emnlp/Pasca13,DBLP:conf/kdd/LiuGNWXLLX19,DBLP:conf/sigmod/LiuGNLWWX20}.
These involve scoring and ranking the candidates,
so that different slices can be compiled depending
on whether the priority is precision or recall.
By incorporating word-embedding-based similarities
and learning techniques, the directly observed cues
can also convey generalizations, for example,
inferring that ``crooners from Mississippi''
are a subtype of ``singers from the midwest''.

Some of the resulting collections 
on types
are 
richer and more fine-grained
than the taxonomies that hinge on Wikipedia-like
sources.
Their strength is that they cover
specific types absent in current KBs,
such as
``musicians who were shot'' (e.g., John Lennon), ``musicians who died at 27'' (e.g., Jim Morrison, 
Amy Winehouse, etc.),
or ``IT companies started in garages''
(e.g., Apple), along with sets of paraphrases
(e.g., ``27 club'' for ``musicians who died at 27'').
Such repositories are very useful for
query suggestions (i.e., auto-completion or
re-formulations) and explorative browsing
(see, e.g., \cite{DBLP:conf/kdd/LiuGNWXLLX19,DBLP:conf/sigmod/LiuGNLWWX20}), but they 
do not (yet) reach
the semantic rigor and near-human quality of
full-fledged knowledge bases.

\vspace*{0.2cm}
\noindent {\bf Discussion:}\\
Overall, the methods presented in this section
 have not yet achieved taxonomies
of better quality and 
much wider coverage than those
built directly from premium sources  (see Chapter \ref{ch2:knowledge-integration}). Nevertheless, 
the outlined methodologies for coping with noisier
input are of interest and value, for example, 
for query suggestion by search engines and
towards constructing
domain-specific KBs (e.g. on health where user queries
could be valuable cues; see \cite{DBLP:conf/wsdm/KoopmanZ19} and references there).

\section{Take-Home Lessons}

The following are key points to remember.
\squishlist
\item The task of entity discovery involves finding more entities for a given type as well as finding more informative types for a given entity.
These two goals are intertwined, and many methods
in this chapter apply to both of them.
\item To discover entity names in web contents,
{\em dictionaries} and {\em patterns} are an easy and very effective
way. 
Patterns can be hand-crafted, such as Hearst patterns,
or automatically computed by
{\em seed-based distantly supervised learning},
following the principle of 
{\em statement-pattern duality}.
For assuring the quality of newly acquired entity-type pairs,
quantitative measures like {\em support} and 
{\em confidence} must be considered.
\item When sufficient amounts of 
{\em labeled training data} are available, in the form
of annotated sentences, 
{\em end-to-end supervised learning} is a powerful approach.
This is typically cast into a sequence tagging task,
known as {\em Named Entity Recognition (NER)}
and {\em Named Entity Typing}.
Methods for this purpose are 
based on probabilistic graphical models
like {\em CRFs}, or 
deep neural networks like {\em LSTMs}, or combinations of both.
\item A useful building block for all these methods are
{\em word and entity embeddings}, which latently encode the degree
of relatedness between pairs of words or entities. 
\item While most of these methods start with a core KB
that already contains a (limited) set of entities and types,
it is also possible to compute IsA relations
by {\em ab-initio taxonomy induction} from text-based
observations only. This has potential for obtaining
more long-tail items, but comes at a higher risk of quality degradation.
\squishend

\clearpage\newpage
\clearpage\newpage

\chapter{Entity Canonicalization}
\label{ch3-sec-EntityDisambiguation}

\section{Problem and Design Space}
\label{subsec:el-designspace}

The entity discovery methods discussed 
in Chapter \ref{ch3:entities}
may
inflate the KB with alias names that refer to the
same real-world entity. For example, we may end up with
entity names such as
``Elvis Presley'', ``Elvis'' and ``The King'', or
``Harry Potter Volume 1'' and ``Harry Potter and the Philosopher's Stone''.
If we treated all of them as distinct entities, we would
end up with redundancy in the KB and, eventually, 
inconsistencies. For example, the birth and death dates
for \ent{Elvis Presley} and \ent{Elvis} could be different,
causing uncertainty about the correct dates.
For some KB applications, this kind of inconsistency
may not cause much harm, as long as humans are satisfied
with the end results, such as finding songs for a search-engine
query about \ent{Elvis}.
However, applications that {\em combine}, 
{\em compare} and
{\em reason} with KB data, such as entity-centric analytics or recommendations,
need to be aware of cases when two names denote the same entity.
For example, counting book mentions for market studies
should properly combine the two variants of the same
Harry Potter book while avoiding conflation with
other book titles that denote different volumes of the series.
Likewise, a user should not erroneously get recommendations
for a book that she already read.

This motivates why a high-quality KB needs to tame
ambiguity by {\em canonicalizing} entity mentions,
creating one entry for all observations of the
same entity regardless of name variants.
The task comes in a number of different settings.

\subsection{Entity Linking (EL)}
\label{ch5:subsec:entity-linking}

\begin{figure}[h!]
  \centering
   \includegraphics[width=0.75\textwidth]{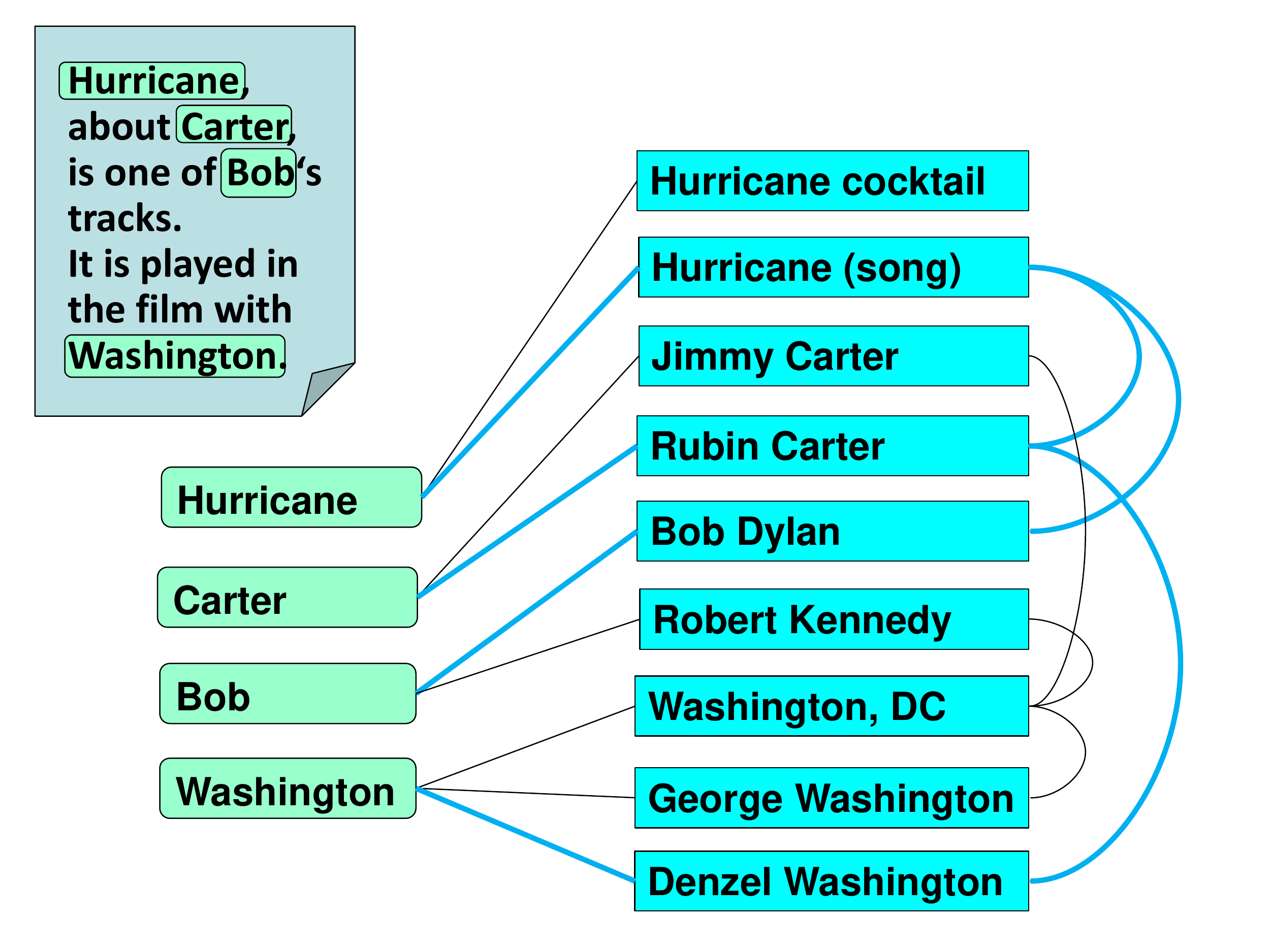}
      \caption{Example for Entity Linking}
      \label{entitylinking-graph}
\end{figure}

The most widely studied case is called
{\bf Entity Linking (EL)}, where we assume 
an existing KB with a rich set of canonicalized
entities (e.g., harvested from premium sources)
and we observe a new set of {\bf mentions} in
additional inputs like text documents or web tables.
When the input is text, the task is also known
as {\bf Named Entity Disambiguation (NED)} in the
computational linguistics community.
Historically, the so-called 
{\em Wikification} task 
\cite{DBLP:conf/cikm/MihalceaC07,DBLP:conf/cikm/MilneW08}  has aimed to
map both named entities and general concepts
onto Wikipedia articles
(including common nouns
such as ``football'', which could mean either
American football or European football aka. soccer,
or the ball itself).
This leads to the broad task of
{\bf Word Sense Disambiguation (WSD)}.
Text with entity mentions also contains general words
that are often equally ambiguous. 
For example,
words like ``track'' and ``album'' 
(cf. Figure \ref{entitylinking-graph})
can have several
and quite different meanings, referring to music
(as in the figure) or to completely different topics.
The WSD task 
is to map these surface words
(and possibly also multi-word phrases) onto their
proper word senses in WordNet or onto Wikipedia articles \cite{Navigli:ComputingSurvey2009,DBLP:journals/tacl/0001RN14,DBLP:series/synthesis/2016Gurevych}. 
For the mission of KB construction, general concepts
and WSD are 
out of scope.

Figure \ref{entitylinking-graph} gives an example
for the EL task. In the input text on the left side,
NER methods can detect mentions, and these need
to be mapped to their proper entities in the KB,
shown on the right-hand side.
Candidate entities can be determined based on
surface cues like string similarity of names.
In the example, this leads to many candidates for
the first name Bob, but also for the 
highly ambiguous mentions
``Hurricane'', ``Carter'' and ``Washington''.
As each mention has so many mapping options, we
face a complex combinatorial problem.
Wikipedia knows more than 700 people with
first name (or nickname) Bob, and Wikidata contains
many more.

To compute, or learn to compute,
the correct mapping, 
all methods consider various signals that relate input
and output:

\begin{samepage}
\begin{mdframed}[backgroundcolor=blue!5,linewidth=0pt]
\squishlist
\item[ ] 
{\bf Mention-Entity Popularity:}\\ 
If an entity is frequently 
referred by the name of the mention, this entity is a likely candidate. For example, ``Carter'' and ``Washington'' most likely
denote the former US president \ent{Jimmy Carter}
and \ent{Washington, DC}.
\squishend
\end{mdframed}
\end{samepage}

\begin{samepage}
\begin{mdframed}[backgroundcolor=blue!5,linewidth=0pt]
\squishlist
\item[ ] 
{\bf Mention-Entity Context Similarity:}\\
Mentions have surrounding text, which can be compared
to descriptions of entities such as short paragraphs
from Wikipedia or keyphrases derived from such texts.
For example, the context words ``tracks'' and ``played''
are cues towards music and musicians, and ``film with''
suggests that ``Washington'' is an actor or actress.
\squishend
\end{mdframed}
\end{samepage}

\begin{samepage}
\begin{mdframed}[backgroundcolor=blue!5,linewidth=0pt]
\squishlist
\item[ ] 
{\bf Entity-Entity Coherence:}\\
In meaningful texts,
different entities do not co-occur uniformly 
at random.
Why would someone write a document about \ent{Jimmy Carter}
drinking a \ent{Hurricane (cocktail)} together with
\ent{Robert Kennedy} and \ent{George Washington}?\\
For two entities to co-occur, a semantic relationship 
should hold between them. The existing KB may have such
prior knowledge that can be harnessed. For example,
\ent{Bob Dylan} has composed \ent{Hurricane (song)},
and its lyrics is about the Afro-American boxer
\ent{Rubin Carter}, aka. Hurricane, who was wrongfully
convincted for murder in the 1970s and later released
after 20 years in prison.
By mapping the mentions to these inter-related entities,
we obtain a highly coherent interpretation.
\squishend
\end{mdframed}
\end{samepage}

In Figure \ref{entitylinking-graph}, 
the edges between mentions and candidate entities
indicate mention-entity similarities, and the edges
among candidate entities indicate entity-entity coherence.
Obviously, for quantifying the strength of similarity
and coherence, these edges should be 
{\em weighted}, which is
not shown in the figure. Some edge weights are stronger
than others, and these are the cues for inferring the
proper mapping. In Figure \ref{entitylinking-graph},
these indicative edges are thicker lines in blue.

Algorithmic and learning-based methods for EL
are based on these three components. 
EL is intensively researched and applied not just
for KB construction, with 
surveys
by 
\citet{Shen:TKDE2015},
\citet{DBLP:journals/tacl/LingSW15} and
\citet{DBLP:journals/semweb/Martinez-Rodriguez20}
and widely used benchmarks (e.g.,
\cite{DBLP:journals/ai/HacheyRNHC13,DBLP:journals/semweb/RoderUN18}).
This entire chapter mostly focuses on EL
and discusses major families of methods
and their building blocks.

\subsection{EL with Out-of-KB Entities}

The EL task has many variations and extensions.
An important case is the treatment of
{\bf out-of-KB entities}: mentions that denote
entities that are not (yet) included in the KB.
This situation often arises with 
{\bf emerging entities},
such as newly created songs or books, people or
organizations that suddenly become prominent, 
and {\bf long-tail entities} such as garage bands or
small startups. 
In such cases, the EL method has an additional option
to map a mention to {\em null}, meaning that none of
the already known entities is a proper match.
This may hold even if the KB has reasonable string matches
for the name itself. For example, the Nigerian saxophonist
Peter Udo, who played with the Orchestra Baobab,
is not included in any major KB (to the best of our knowledge),
but there are many matches for the string ``Peter Udo''
as this is also a German first name.
A good EL method needs to calibrate its linking decisions
to avoid spurious choices, and should map mentions
with low confidence in being KB entities to {\em null}.
Such long-tail entities may become candidates to
be included later in the KB life-cycle.
We will revisit this issue 
in Chapter \ref{chapter:KB-curation}
on KB curation, 
specifically Section 
\ref{ch8-subsec:emergingentities}.

\subsection{Coreference Resolution (CR)}
\label{ch5:subsec:coreference-resolution}

The initial KB against which EL methods operate
is inevitably incomplete, regarding both coverage
of entities and coverage of different names for the
known entities.
The former is addressed by awareness of out-of-KB entities.
The latter calls for grouping mentions into equivalence
classes that denote the same entities.
In NLP, this task is known as
{\bf coreference resolution (CR)};
in the world of 
structured data, its counterpart is the
{\bf entity matching (EM)} problem
(aka record linkage), see
Subsection \ref{subsec:entity-matching}.

\begin{figure} [h!]
  \centering
   \includegraphics[width=0.7\textwidth]{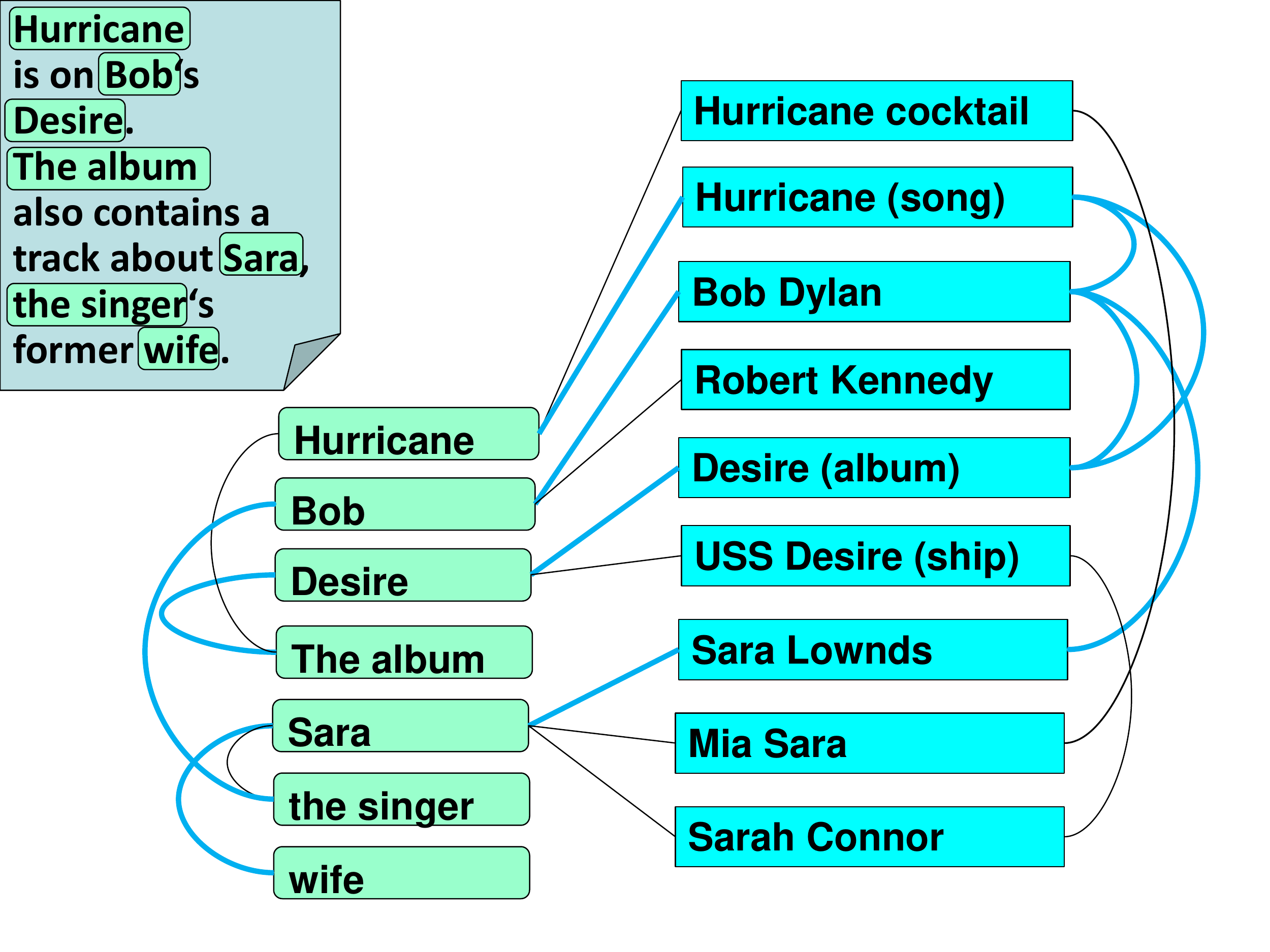}
      \caption{Example for Combined Entity Linking and Coreference Resolution}
      \label{el-cr-graph}
\end{figure}

The CR task is highly related to the EL setting, as illustrated by
Figure \ref{el-cr-graph}.
Here, the input text contains underdetermined
phrases like ``the album'', ``the singer'' and ``wife''
(or say ``his wife''). Longer texts will likely contain
pronouns as well, such as
``she'', ``her'', ``it'', ``they'', etc.
All these are not immediately linkable to KB entities.
However, we can first aim to identify to which other
mentions these coreferences refer, this way computing
{\em equivalence classes of mentions}. 
In Figure \ref{el-cr-graph}, possible groupings are
indicated by edges between mentions, and the correct ones
are marked by thick lines in blue.

The ideal output would thus state that
``Desire'' and ``the album'' denote the same entity,
and by linking one of the two mentions to the Bob Dylan album
\ent{Desire (album)}, EL covers both mentions.
In general, however, the grouping and linking will be partial,
meaning that some coreferences may be missed and some of
the coreference groups may still be unlinkable -- either
because of remaining uncertainty or because the proper entity
does not exist in the KB.
Although this partial picture may look 
unsatisfying,
it does give valuable information for KB construction and
completion:
\squishlist
\item Mentions in the same coreference group linked to a KB entity
may be added
as alias names, or simply textual cues, for an existing entity.
\item Coreference groups that cannot be linked to a KB entity
can be captured as candidates for new entities to be 
added, or at least reconsidered, later.
\squishend
\noindent For example, if we pick up mentions like
``Peter Udo'', ``the sax player'', ``the Nigerian saxophonist''
and ``he'' as a coreference group, not only can we
assert that he is not an existing KB entity, but we already
have informative cues about what type of entity this is
and even a gender cue.

Methods for coreference resolution over text inputs,
and for coupling this with entity linking, can be
rule-based (see, e.g., \cite{DBLP:conf/emnlp/RaghunathanLRCSJM10,DBLP:journals/coling/LeeCPCSJ13,DBLP:conf/emnlp/DurrettK13}),
based on CRF-like graphical models 
(see, e.g., 
\cite{DBLP:journals/tacl/DurrettK14})
or based on neural learning (see, e.g., \cite{DBLP:conf/acl/ClarkM16,DBLP:conf/emnlp/LeeHLZ17,DBLP:conf/emnlp/JoshiLZW19}).
The latter benefits from feeding large unlabeled corpora
into the model training (via embeddings such as BERT (e.g., \cite{DBLP:conf/emnlp/JoshiLZW19},
see also Section \ref{subsec:word-entity-embeddings}).

The CR task mostly focuses on short-distance mentions,
often looking at successive sentences or paragraphs only.
However, the task 
can be extended to compute coreference groups
across distant paragraphs or even across different documents.
This aims to mark up an entire corpus with equivalence
classes of mentions, and partial linking of these classes
to KB entities.
The extended task is known as {\bf cross-document coreference resolution (CCR)}, first studied by \cite{DBLP:conf/acl/BaggaB98}.
Methods along these lines include inference with
CRF-like graphical models (e.g, \cite{DBLP:conf/acl/SinghSPM11})
and hierarchical clustering (e.g., \cite{DBLP:conf/emnlp/DuttaW15}).

\vspace*{0.2cm}
\noindent{\bf Benefit of CR for KB construction:}\\
Partial linking of mentions to KB entities helps 
coreference resolution by capturing distant signals.
Conversely, having good candidates for coreference groups
is beneficial for EL, as it provides richer context
for individual mentions.
In addition to these dual benefits, 
coreference groups can be important for extracting
types and properties of entities, when the latter
are expressed with pronouns or underdetermined phrases
(e.g., ``the album''). 
For example, when given the text snippet\\
\hspace*{1cm}``Hurricane was not exactly a big hit. \\
\hspace*{1cm} It is a protest song about racism.'',\\
\noindent we can infer the type \ent{protest song} and its super-type \ent{song} only with the help of CR.
Likewise, in the example of Figure \ref{el-cr-graph},
we can acquire knowledge about Bob Dylan's ex-wife only
when considering coreferences.
So CR can improve extraction recall
and thus the coverage of knowledge bases.

\subsection{Entity Matching (EM)}
\label{subsec:entity-matching}

Computing equivalence classes over a set of observed
entity mentions is also a frequent task in data cleaning
and integration. For example, when faced with two or more semantically
overlapping databases or other datasets (incl. web tables), 
we need to infer
which records or table rows correspond to the same entity. 
This task of {\bf entity matching (EM)} or {\bf duplicate detection}, historically called
{\bf record linkage}, is a long-standing problem
in computer science
(\citet{dunn1946record,fellegi1969theory}).
The ususal assumption here is that there is no already existing reference repository or core KB of entities. 
Thus, there are no prior entities; the equivalence classes from the EM itself are the normalized entity representations. If one of the input databases is considered more complete and trustworthy than the others, it could be interpreted as the reference repository against which the other data records are matched. In that case, we would be back to an EL task with structured tables as input.

In terms of methodology, EM and EL are highly related.
In a nutshell, all EM methods leverage cues from 
comparing records in a matching-candidate pair:
\squishlist
\item {\bf Name similarity:}
The higher the string similarity
between two names, the higher
the likelihood that they denote
matching entities
(e.g., strings like ``Sara'' and ``Sarah''
being close).
\item {\bf Context similarity:}
As context of a database record or table cell
that denote an entity of interest,
we should consider the full record or row
as the data in such proximity often
denote related entities or
salient attribute values.
Additionally, it can be 
beneficial to {\em contextualize}
the mention in a table cell
by considering the other
cells in the same column,
for example, as signals for 
the specific entity type of 
a cell
(e.g., when all values in a table column are song titles).
\item {\bf Consistency constraints:}
When a pair of rows from two
tables is matched, this rules out
matching one of the two rows to
a third one, assuming that there
are no duplicates within each table.
With duplicates or when we consider
more than two tables as input,
matchings need to satisfy the
transitivity of an equivalence
relation. 
\item {\bf Distant knowledge:}
By connecting highly related entities, using a background KB,
we can establish matching contexts despite low scores
on string similarity.
For example, a singer's name (e.g., Mick Jagger) and the name of his or her band (e.g., Rolling Stones) could be very different,
but there is nevertheless a strong connection
(e.g., in matching a Stones song, using either one or
both of the related names).
\squishend
These ingredients can be fed
into matching rules
generated from human-provided samples
(e.g., \cite{DBLP:journals/pvldb/KondaDCDABLPZNP16,DBLP:journals/pvldb/SinghMEMPQST17}), 
or used
as input to supervised learning or
probabilistic graphical models
(e.g., \cite{DBLP:conf/icdm/SinglaD06,DBLP:journals/tkdd/BhattacharyaG07,
DBLP:conf/kdd/WickRSM08,DBLP:journals/pvldb/RastogiDG11,DBLP:journals/pvldb/SuchanekAS11,DBLP:journals/kais/KoukiPMKG19})
or neural networks (e.g., \cite{DBLP:conf/sigmod/MudgalLRDPKDAR18,DBLP:conf/www/ZhuWSZFD020}).

Similar techniques can be applied also to
spotting matching entity pairs 
across datasets
in the Web of (Linked) Open Data \cite{HeathBizer2011,DBLP:journals/debu/MillerNZCPA18};
see the survey by
\cite{DBLP:journals/semweb/NentwigHNR17}
on this {\em link discovery} problem.
For query processing, the problem takes the form
of finding joinable tables and the respective join columns 
and values
(see, e.g., \cite{DBLP:conf/sigmod/ZhuDNM19,DBLP:journals/pvldb/LehmbergB17}).

\vspace*{0.2cm}
\noindent{\bf Benefit of EM for KB Construction:}\\
\noindent Entity matching between databases or other structured data repositories
is of interest for KB construction when incorporating premium sources
into an existing KB. For example, when adding entities from
GeoNames to a Wikipedia-derived KB, the detection of duplicates
(to be eliminated from the ingest) is essentially an EM task.
Another use case is when an initially empty KB is to be populated from a set of databases which are treated on par (i.e., without singling one out as reference or master DB).

Surveys and best-practice papers on EM include
\cite{DBLP:journals/dke/KopckeR10,DBLP:series/synthesis/2010Naumann,DBLP:books/daglib/0030287,DBLP:series/synthesis/2015Dong,DBLP:journals/corr/abs-1905-06397}.
We will revisit this topic in Section \ref{sec:entity-matching}.

\section{Popularity, Similarity and Coherence Measures}

All EL methods are based on quantitative measures for
mention-entity popularity, mention-entity similarity
and entity-entity coherence.
This section elaborates on these measures.

Throughout the following, we assume that many entities
in the KB, and especially the prominent ones,
are uniquely identified also in Wikipedia.
It is very easy for a KB to align its entities with
Wikipedia articles. This holds for large KBs
like Wikidata, even if the KB itself is not built
from Wikipedia as a premium source.
We will use Wikipedia articles as a background asset
for various aspects of EL methods.

\subsection{Mention-Entity Popularity}

As most texts, like news, books and social media, are about prominent entities, the popularity of an entity determines
a {\em prior likelihood} to be selected by an EL algorithm.
For example, the Web contents about Elvis Presley
is an order of magnitude larger than about the
less known musician Elvis Costello.
Therefore, when EL sees a mention ``Elvis'', 
the probability that this denotes \ent{Elvis Presley}
is a priori much higher than for Elvis Costello.
To measure the {\bf global popularity} of
entities, a variety of indicators
can be used:
\squishlist
\item the length of an entity's Wikipedia article
(or ``homepage'' in domain-specific platforms such
as IMDB for movies or Goodreads for books,
\item the number of incoming links of the Wikipedia page,
\item the number of page visits based on Wikipedia usage
statistics,
\item the amount of user activity, such as clicks or likes,
referring to the entity in a domain-specific platform
or in social media, and more.
\squishend

\noindent While considering global popularity is useful, 
it can be misleading and insufficient
as a prior probability.
For example, for the mention ``Trump'',
the most likely entity is
\ent{Donald Trump}, but for the mention ``Donald'',
albeit Donald Trump still being a candidate,
the more likely entity is \ent{Donald Duck}.
This suggests that we should consider the
{\em combination of mention and entity} for estimating
popularity.

The most widely used estimator for 
mention-entity popularity is based on 
{Wikipedia links}
(\citet{DBLP:conf/cikm/MihalceaC07,medelyan2008topic,DBLP:conf/cikm/MilneW08}),
exploiting the observation that href anchor texts
are often short names and the pages to which they
link are canonicalized entities already.

\begin{samepage}
\begin{mdframed}[backgroundcolor=blue!5,linewidth=0pt]
\squishlist
\item[ ] 
{\bf Link-based Mention-Entity Popularity:} \\
The {\bf mention-entity popularity score}
for mention $m$ and entity $e$ is proportional to
the occurrence frequency of hyperlinks with
href anchor text {\em ``m''} that point to
the main page about $e$.
This includes redirects within Wikipedia
as well as interwiki links between different language editions.
\squishend
\end{mdframed}
\end{samepage}

Obviously, this works only for entities that
have Wikipedia articles, and it hinges
on sufficiently many href links pointing to these articles.
On the other hand, the popularity score is useful only
for prominent entities  and will not be a
good signal for long-tail entities anyway.
Nevertheless, the approach can be generalized
for larger coverage, 
by considering 
all kinds of Web pages with links to Wikipedia
\cite{DBLP:conf/lrec/SpitkovskyC12}.
Further alternatives are to leverage query-click logs,
by considering names in
search-engine queries as mentions and subsequent
clicks on Wikipedia articles or other kinds of
homepages as linked entities.
None of these heuristic techniques is foolproof, though.
For example, web pages may contain anchor texts like
``Wikipedia'' and point to specific entities within Wikipedia.
This would incorrectly associate the name ``Wikipedia''
with these entities. 
A more robust solution may thus have to include hand-crafted
lists of exceptions.

\subsection{Mention-Entity Context Similarity}
\label{ch5-subsec:contextsim4EL}

The most obvious cue for inferring that mention $m$
denotes entity $e$ is to compare their {\em surface strings}.
For $m$, this is given by the input text, for example,
``Trump'' or ``President Trump''.
For $e$, we can consider the preferred (i.e., official
or most widely used) label for the entity, such as
``Donald Trump'' or ``Donald John Trump'', but also
alias names that are already included in the KB,
such as ``the US president''.
{\bf String similarity between names}, like edit distance
or n-gram overlap, can then score how
good $e$ is a match for $m$, this way ranking the
candidate entities. In doing this, we can consider
weights for tokens at the word or even character level.
The weights can be derived from frequency statistics,
in the spirit of IR-style $idf$ weights.
The weights and the similarity measure can
be type-specific or domain-specific, dealing differently with say
person names, organization acronyms, song titles, etc.

Beyond the basic comparison of $m$ and names for $e$,
a fairly obvious approach is to leverage the 
{\bf mention context}, that is, the text surrounding $m$,
and compare it to concise descriptions of entities
or other kinds of {\bf entity contextualization}.
The mention context can take the form of a single sentence,
single paragraph or entire document, possibly in
weighted variants (e.g., weights decreasing with
distance from $m$). 
The entity context depends on the richness of the
existing KB. Entities can be augmented with descriptions
taken, for example, from their Wikipedia articles
(e.g., the first paragraph stating most salient points).
Alternatively, the types of entities, their
Wikipedia categories
and other prominently appearing entities
in a Wikipedia article (i.e., outgoing links) 
can be used for a contextualized
representation. 
In essence, we create a pseudo-document for each
candidate entity, and we may even expand these
by external texts such as news articles about entities,
or highly related
concepts for the entity types using WordNet
and other sources.
Methods for computing context similarity, along these lines, have been devised and advanced by 
\citet{DBLP:conf/www/DillEGGGJKRTTZ03,DBLP:conf/eacl/BunescuP06,DBLP:conf/cikm/MihalceaC07,DBLP:conf/emnlp/Cucerzan07,DBLP:conf/i-semantics/MendesJGB11}.

In the example of Figure \ref{entitylinking-graph},
the mention ``Hurricane'' has words like ``track'' and ``played''
in its proximity, and the mention ``Washington''
is accompanied by the word ``film''.
These should be compared to entity contexts
with
words like song, music etc. for \ent{Hurricane (song)}
versus beverage, alcohol, rum etc. for
\ent{Hurricane (cocktail)}.

The following are some of the widely used 
measures for scoring the 
{\bf mention-entity context similarity}.

\begin{samepage}
\begin{mdframed}[backgroundcolor=blue!5,linewidth=0pt]
\squishlist
\item[ ] 
{\bf Bag-of-Words Context Similarity:} \\
Both $m$ and $e$ are represented as {\em tf-idf} vectors
derived from bags of words (BoW). Their similarity is the
scalar product or cosine between these vectors:
$$sim\left(cxt(m),cxt(e)\right) = 
\overrightarrow{BoW(cxt(m))} \cdot \overrightarrow{BoW(cxt(e))}$$
or analogously for cosine.
Some variants restrict the BoW representation to 
{\em informative
keywords} from both contexts, using statistical and entropy 
measures to identify the keywords.
\squishend
\end{mdframed}
\end{samepage}

A generalization of keyword-based contexts is to
focus on {\bf characteristic keyphrases}
\cite{medelyan2008topic,DBLP:conf/cikm/HoffartSNTW12}:
multi-word phrases 
that are salient and specific for candidate entities.
Keyphrase candidates can be derived from entity types,
category names, href anchor texts in an entity's
Wikipedia article, and other sources along these lines.
For example, \ent{Rubin Carter} (as a candidate for
the mention ``Carter'' in Figure \ref{entitylinking-graph})
would be associated with keyphrases such as
``African-American boxer'', ``people convicted of murder'',
``overturned convictions'', ``racism victim'', 
``nickname The Hurricane'' etc.
Such phrases can be gathered from noun phrases $n$ in
the Wikipedia page (or other entity description) for $e$, and then scored and filtered
by criteria like {\em pointwise mutual information (PMI)}:
$$weight(n|e) \sim \log \frac{P[n,e]}{P[n] \cdot P[e]}$$
or other 
information-theoretic entropy measures,
with probabilities estimated from (co-) occurrence frequencies
(in Wikipedia).
Intuitively, the best keyphrases for entity $e$
should frequently co-occur with $e$, but should not
be globally frequent for all entities.
The context of $e$, $cxt(e)$, then becomes the
weighted set of keyphrases $n$ for which $weight(n|e)$
is above some threshold.

Comparing a set of phrases against the mention context
is a bit more difficult than for the Bag-of-Words
representations. The reason is that exact matches
of multi-word phrases are infrequent, and we need
to pay attention to similar phrases with some words
missing, different word order, and other variations.
\cite{DBLP:conf/cikm/HoffartSNTW12} proposed a word-proximity-aware window-based
model for such approximate matches.
For example, the keyphrase ``racism victim''
can be matched by ``victim in a notorious case
of racism''. 

\begin{samepage}
\begin{mdframed}[backgroundcolor=blue!5,linewidth=0pt]
\squishlist
\item[ ] 
{\bf Keyphrase Context Similarity:} \\
Representing $cxt(e)$ as a weighted set $KP$ of keyphrases $n$
and $cxt(m)$ as a sequence of words
conceptually broken down into a set $W$ of 
(overlapping) small text windows of bounded length,
the similarity is computed by aggregating the
following scores:
\squishlist
\item for each $n \in KP$ identify the best matching window 
$\omega \in W$;
\item for each such $\omega$ compute the (sub-)set of words
$w \in n \cap \omega$ and their maximum
positional distance $\delta$
in $\omega$;
\item aggregate the word matches for $n$ in $\omega$
with consideration of $\delta$ and the
entity-specific weight for $w$ (treating $w$ as if
it were a keyphrase by itself);
\item aggregate these per-keyphrase scores over all
$n \in KP$.
\squishend
This template can be varied and extended
in a number of ways.
\squishend
\end{mdframed}
\end{samepage}

With the advent of latent embeddings (see 
Subsection 
\ref{subsec:word-entity-embeddings}), 
both BoW and keyphrase models
may seem to be superseded by word2vec-like and other
kinds of embeddings, which implicitly capture
also synonyms and other strongly related terms.
However, at the word level alone, the embeddings
are susceptible to over-generalization and drifting focus.
For example, the word embedding for ``Hurricane'' 
brings out highly related terms that have nothing
to do with the song or the boxer (e.g.,
``storm'', ``damage'', ``deaths'' etc.).
So it is important to use entity-centric embeddings
likes the ones for wikipedia2vec 
\cite{yamada2020wikipedia2vec}, and ideally, these
should reflect multi-word phrases as well.

\begin{samepage}
\begin{mdframed}[backgroundcolor=blue!5,linewidth=0pt]
\squishlist
\item[ ] 
{\bf Embedding-based Context Similarity:}\\ 
With embedding vectors $\overrightarrow{cxt(m)}$
and $\overrightarrow{cxt(e)}$, the context similarity
between $m$ and $e$ is $cos\left(\overrightarrow{cxt(m)},\overrightarrow{cxt(e)}\right)$.
\squishend
\end{mdframed}
\end{samepage}

Each of these context-similarity models has its
sweet spots as well as limitations.
Choosing the right model thus depends on the
topical domain (e.g., business vs. music)
and the language style of the input texts 
(e.g., news vs. social media).

\subsection{Entity-Entity Coherence}

Whenever an input text contains several mentions,
EL methods should compute their corresponding
entities {\em jointly}, based on the principle
that co-occurring mentions usually map to
semantically coherent entities.
To this end, we need to define measures for
{\bf entity-entity relatedness} that capture this
notion of coherence.
Early works that explicitly consider coherence 
for joint EL include
\citet{DBLP:conf/cikm/MilneW08},
\citet{DBLP:conf/kdd/KulkarniSRC09},
\citet{DBLP:conf/cikm/FerraginaS10},
\citet{DBLP:conf/sigir/HanSZ11},
\citet{DBLP:conf/acl/RatinovRDA11} and
\citet{DBLP:conf/emnlp/HoffartYBFPSTTW11}.

One of the most powerful and surprisingly simple
measures exploits the rich link structure of Wikipedia.
The idea is to consider two entities as highly related
if the hyperlink sets of their Wikipedia pages
have a large overlap.
More specifically, the incoming links are a good signal.
For example, two songs like ``Hurricane'' and ``Sara''
are highly related because they are both linked to
from the articles about \ent{Bob Dylan}, \ent{Desire (album)}
and more (e.g., the other involved musicians).
Likewise, \ent{Bob Dylan} and \ent{Elvis Presley}
are notably related as there are quite a few
Wikipedia categories that link to both of them.
Intuitively, in-links are considered more informative
than out-links as outgoing links tend to refer to
more general entities and concepts whereas 
incoming links often have the direction from
more general to more specific.
These considerations have given rise to the following
{\em link-based} definition of entity-entity coherence.

\begin{samepage}
\begin{mdframed}[backgroundcolor=blue!5,linewidth=0pt]
\squishlist
\item[ ] 
{\bf Link-based Entity-Entity Coherence:} \\
For two entities $e$ and $f$ with Wikipedia articles
that have incoming-link sets $In(e)$ and $In(f)$,
their coherence score is
$$1-
\frac{\log\left(max\{|In(e)|,|In(f)|\}\right) - 
\log\left|In(e) \cap In(f)|\right)}
{\log(|U|) - 
\log\left(min\{|In(e)|,|In(f)|\}\right)}$$
where $U$ is the total set of known entities
(e.g., Wikipedia articles about named entities).
\squishend
\end{mdframed}
\end{samepage}

Intuitively, the enumerator of the fraction penalizes
entity pairs that have many incoming links in common;
the
denominator is just a normalization term (close
to $\log |U|$) so that the overall value is between 0 and 1.
Then, taking one minus the fraction yields a measure that
increases with the overlap of incoming-link sets between the two entities.
This approach was pioneered by
\citet{DBLP:conf/cikm/MilneW08},
and is thus sometimes referred to as the
Milne-Witten metric.
Similarly to link-based popularity measures,
we can generalize this Wikipedia-centric link-overlap
model to other settings.
In a search engine's query-and-click log,
entities whose pages both appear in the clicked-pages
set for the same queries (so-called ``co-clicks'')
should have a high relatedness score.
In domain-specific content portals and web communities,
such as Goodreads and LibraryThing for books or
IMDB and Rottentomatoes for movies, 
the overlap of the user sets who expressed liking
the same entity is a good measure for
entity-entity relatedness. 
All these can be seen as instantiations of the
{\bf strong co-occurrence principle}
(see Subsection \ref{subsec:dictionary-based-entity-spotting}).

When entities are represented as weighted sets of
keywords (BoW) or weighted sets of keyphrases (KP),
their relatedness can be captured by
measures for the (weighted)
overlap of these sets.

\begin{samepage}
\begin{mdframed}[backgroundcolor=blue!5,linewidth=0pt]
\squishlist
\item[ ] 
{\bf 
BoW-based and KP-based
Entity-Entity Coherence:} \\
Consider two entities $e$ and $f$ with associated
keyword sets $E$ and $F$ whose entries $x$ have
weights $w_E(x)$ and $w_F(x)$, respectively.
The coherence between $e$ and $f$ can be measured
by the weighted Jaccard metric:
$$\sum_{x \in E \cap F} 
\frac{min\{w_E(x),w_F(x)\}}
{max\{w_E(x),w_F(x)\}}$$
The extension to keyphrases is more sophisticated.
It needs to consider also partial matches between
keyphrases for $e$ and for $f$
(e.g., ``rock and roll singer'' vs. ``rock 'n' roll musician''),
following the same ideas as the KP-based context similarity
model (see \cite{DBLP:conf/cikm/HoffartSNTW12} for details).
\squishend
\end{mdframed}
\end{samepage}

Analogously to the context-similarity aspect,
embeddings are a strong alternative for
entity-entity coherence as well.
They are straightforward to apply.

\begin{samepage}
\begin{mdframed}[backgroundcolor=blue!5,linewidth=0pt]
\squishlist
\item[ ] 
{\bf Embedding-based Entity-Entity Coherence:} \\
With embedding vectors $\overrightarrow{e}$
and $\overrightarrow{f}$ for entities $e$ and $f$, 
their relatedness for EL coherence is 
$cos(\overrightarrow{e},\overrightarrow{f})$.
\squishend
\end{mdframed}
\end{samepage}

All these purely text-based coherence measures
-- keywords, keyphrases, embeddings -- have
the advantage that they can be computed solely
from textual entity descriptions. 
There is no need for Wikipedia-style links, and
not even for any relations between entities.
This setting has been called {\em EL with a ``linkless'' KB}
in \cite{DBLP:conf/www/LiTSHRY16}. 
That prior work, which predates embedding-based methods, made use of latent topic models (in the style of LDA \cite{DBLP:journals/jmlr/BleiNJ03}).
Today, word2vec-style embeddings or even BERT-like language models \cite{DBLP:conf/naacl/DevlinCLT19}
(see Section \ref{subsec:word-entity-embeddings})
seem to be the more powerful choice, 
but they all fall into the same architecture
presented above.

Yet another way of defining and computing
entity-entity relatedness is by means of
{\em random walks} over an existing knowledge
graph (e.g., the link structure of Wikipedia),
see, for example, \cite{DBLP:journals/semweb/GuoB18}.
Here each entity is represented by the
(estimated) probability
distribution of reaching related entities by
random walks with restart. The relatedness score
between entities can be defined as the
relative entropy between two distributions.

The above are major cases within a wider space of
 measures for entity-entity coherence.
A good discussion of the broader design space
and empirical comparisons can be found in
\cite{DBLP:conf/aaai/StrubeP06,DBLP:conf/cikm/CeccarelliLOPT13,DBLP:journals/kbs/PonzaFC20}.
Ultimately, however,
the best choice depends on the domain of interest
(e.g., business vs. music) and the style of the
ingested texts (e.g, news vs. social media).

All coherence measures discussed in this section are symmetric.
In general, relatedness scores can also be asymmetric,
for example, based on conditional probability or
conditional entropy.
\citet{DBLP:conf/cikm/CeccarelliLOPT13} discusses
asymmetric options as well, which are useful for
applications outside EL.
Most EL methods treat their input mentions in an unordered manner, though;
hence the prevalent choice of symmetric coherence scores.

\section{Optimization and Learning-to-Rank Methods}
\label{subsec:EL-optimization-and-LTR}

All EL methods aim to optimize a scoring or ranking
function that maximizes a combination of
mention-entity popularity, mention-entity context similarity
and entity-entity coherence. 
This can be formalized as follows.

\begin{samepage}
\begin{mdframed}[backgroundcolor=blue!5,linewidth=0pt]
\squishlist
\item[ ] 
{\bf EL Optimization Problem:} \\
Consider an input text with entity mentions 
$M= \{m_1, m_2 \dots\}$ each of which has
entity candidates, $E(m_i) = \{e_{i1}, e_{i2} \dots\}$,
together forming a pool of target entities 
$E = \{e_1, e_2 \dots\}$.
The goal is to find a, possibly partial, function
$\phi: M \rightarrow E$ that maximizes the objective
$$\alpha \sum_{m} pop(m,\phi(m)) +
\beta \sum_{m} sim(cxt(m),cxt(\phi(m))) +$$
$$\gamma \sum_{e,f} \{coh(e,f) ~|~ \exists m,n \in M:
m \ne n, e = \phi(m), f = \phi(n)\}$$
where $\alpha, \beta, \gamma$ are tunable hyper-parameters,
$pop$ denotes mention-entity popularity,
$cxt$ the context of mentions and entities,
$sim$ the contextual similarity and
$coh$ the pair-wise coherence between entities.
\squishend
\end{mdframed}
\end{samepage}
In principle, coherence could even be incorporated
over the set of all entities in the image of $\phi$,
but the practical sweet spot is to break this down
into pair-wise terms which allow more robust estimators.

Combinatorially, the function $\phi$ is the solution of
a combined {\em selection-and-assignment} problem:
selecting a subset of the entities and assigning the mentions
onto them. 
We aim for the {\em globally best} solution,
with a choice for $\phi$ that bundles the
linking of all mentions together.
Algorithmically, this optimization opens up a wide variety
of design choices: from unsupervised scoring
to graph-based algorithms all the way to 
neural learning. 
The basic choice,
discussed in
the classical works of
\citet{DBLP:conf/www/DillEGGGJKRTTZ03},
\citet{DBLP:conf/eacl/BunescuP06},
\citet{DBLP:conf/cikm/MihalceaC07} 
\citet{DBLP:conf/emnlp/Cucerzan07} and
\citet{DBLP:conf/cikm/MilneW08},
is to view EL as a {\bf local optimization}
problem (notwithstanding its global nature): 
for each mention in the input text,
we compute its best-matching entity from a pool
of candidates. This is carried out for each
mention independently of the other mentions,
hence the adjective {\em local}.
By this restriction, coherence is largely disregarded,
but some aspects can still be incorporated via
clever ways of contextualization.
Most notably, the method of \cite{DBLP:conf/cikm/MilneW08} 
first identifies {\em all unambiguous
mentions} and expands their contexts by (the descriptions
of) their respective entities. 
With this enriched context, the method then
optimizes a combination of popularity and similarity
(i.e., setting $\gamma$ to zero in the general
problem formulation).
Generalized techniques for 
enhancing mention context and entity context 
via document retrieval have been devised
by \cite{DBLP:conf/www/LiTSHRY16}.
The literature on EL contains further techniques
along these lines.

These approaches can be used to score and rank
mention-entity pairs, but can likewise be
broken down into their underlying scoring components
as features for {\em learning}
a ranker. Such methods have been explored
using support vector machines and other kinds
of classifiers, picking the entity with the
highest classification confidence
(e.g., \cite{DBLP:conf/eacl/BunescuP06,DBLP:conf/cikm/MilneW08,DBLP:conf/coling/DredzeMRGF10,DBLP:conf/i-semantics/MendesJGB11,DBLP:conf/www/ShenWLW12,DBLP:journals/tacl/LazicSR015}).
Alternatively, more advanced
{\bf learning-to-rank (LTR)} regression models
\cite{DBLP:books/daglib/0027504,DBLP:series/synthesis/2014Li},
have been investigated
(e.g., \cite{DBLP:conf/naacl/ZhengLHZ10,DBLP:conf/acl/RatinovRDA11,DBLP:journals/semweb/GuoB18}),
for example, with learning
from pairwise preferences
between entity candidates
for the same mention.
Note that these
supervised learning methods hinge
on the availability of a labeled training corpus
where mentions are marked up with ground-truth
entities.

\section{Collective Methods based on Graphs}
\label{subsec:EL-graph-algorithms}

By factoring entity-entity relatedness scores into
the context of candidate entities, context-similarity methods
already go some way in considering global coherence,
an example being the work of
\citet{DBLP:conf/emnlp/Cucerzan07}
\cite{DBLP:conf/tac/CucerzanS13}.
Nevertheless, more powerful
EL methods make decisions on mapping mentions to entities
{\em jointly} for all mentions of the input text.
This family of methods is referred to as 
{\bf Collective Entity Linking}.
Many of these can be seen as operating on an
{\bf EL candidate graph}:

\begin{samepage}
\begin{mdframed}[backgroundcolor=blue!5,linewidth=0pt]
\squishlist
\item[ ] 
For an entity linking task,
the {\bf EL Candidate Graph} consists of
\squishlist
\item a set $M$ of mentions 
and a set $E$ of entities 
as nodes,
\item a set $ME$ of {\em mention-entity edges} with weights derived from
popularity and similarity scores, and
\item a set $EE$ of {\em entity-entity edges} with weights derived from
relatedness scores between entities.
\squishend
\squishend
\end{mdframed}
\end{samepage}

Figure \ref{entitylinking-graph}, in
Section \ref{subsec:el-designspace},
showed an example for
such a candidate graph, with edge weights omitted.

\subsection{Edge-based EL Methods}

\citet{DBLP:conf/cikm/FerraginaS10} 
proposed a
relatively simple but effective and highly efficient approach
for incorporating entity-entity edge weights into the
scoring of mention-entity edges. Given a candidate mapping $m \mapsto e$,
its score is augmented by the coherence of $e$ with
{\em all} candidate entities
for all other mentions in the graph:
$$score(m \mapsto e) = \dots + \sum_{n \ne m} ~
\sum_{f: (n,f) \in ME} sim(n,f) \cdot coh(e,f) $$
The intuition here is that a candidate $e$ for $m$
is rewarded if $e$ has highly weighted edges or paths with
all other entities in the entire candidate graph.

Obviously, the graph still contains spurious entities,
but we expect those to have low coherence with others 
anyway. 
Conversely, entities for unambiguous 
mentions and entities with tight connections to many 
others in the graph have a strong influence on the decisions
for all mentions. Hence the collective flavor of the method.

\subsection{Subgraph-based EL Methods}

Motivated by these considerations, a generalization is
to consider entire {\em subgraphs} that connect the best cues
for joint mappings. This leads to a powerful framework,
although it entails more expensive algorithms.
More specifically, we are interested in identifying 
{\em dense subgraphs} in the candidate graph such that there
is at most one mention-entity edge for each mention 
(or exactly one if we insist on linking all mentions). 
Density refers to high edge weights, where both
mention-entity weights and entity-entity weights are
aggregated over the subgraph.
We assume the weights of the two edge types
are calibrated before
being combined (e.g., via hyper-parameters
like $\alpha,\beta,\gamma$).

\begin{samepage}
\begin{mdframed}[backgroundcolor=blue!5,linewidth=0pt]
\squishlist
\item[ ] 
{\bf EL based on Dense Subgraph:} \\
Given a candidate graph with nodes $M \cup E$ and weighted
edges $ME \cup EE$, the goal is to compute the
densest subgraph $S$, with nodes $S.M \subseteq M$ and $S.E \subset E$ and
edges $S.ME \subset ME$ and
$S.EE \subset EE$, maximizing
$$aggr \{weight(s) ~|~ s \in S.ME \cup S.EE\}$$
subject to the constraint:\\
for each $m \in M$ there is at most one $e \in S.E$
such that $(m,e) \in S.ME$.
In this objective, $aggr$ denotes an aggregation function over
edge weights, a natural choice being 
summation (or the sum normalized by the number of nodes or edges in the subgraph).
\squishend
\end{mdframed}
\end{samepage}

Note that the resulting subgraph is not necessarily
connected, as a text may be
about different groups
of entities, related within a group but unrelated
across.
However, this situation should be rare, and
the method could 
enforce a connected subgraph.
Also, it may be desirable to have identical mentions
in a 
document mapped to the same entity
(e.g., all mentions ``Carter'' linked to the same person),
at least when the text is not too long.
This can be enforced by an additional constraint.

\begin{figure} [h!]
  \centering
   \includegraphics[width=0.6\textwidth]{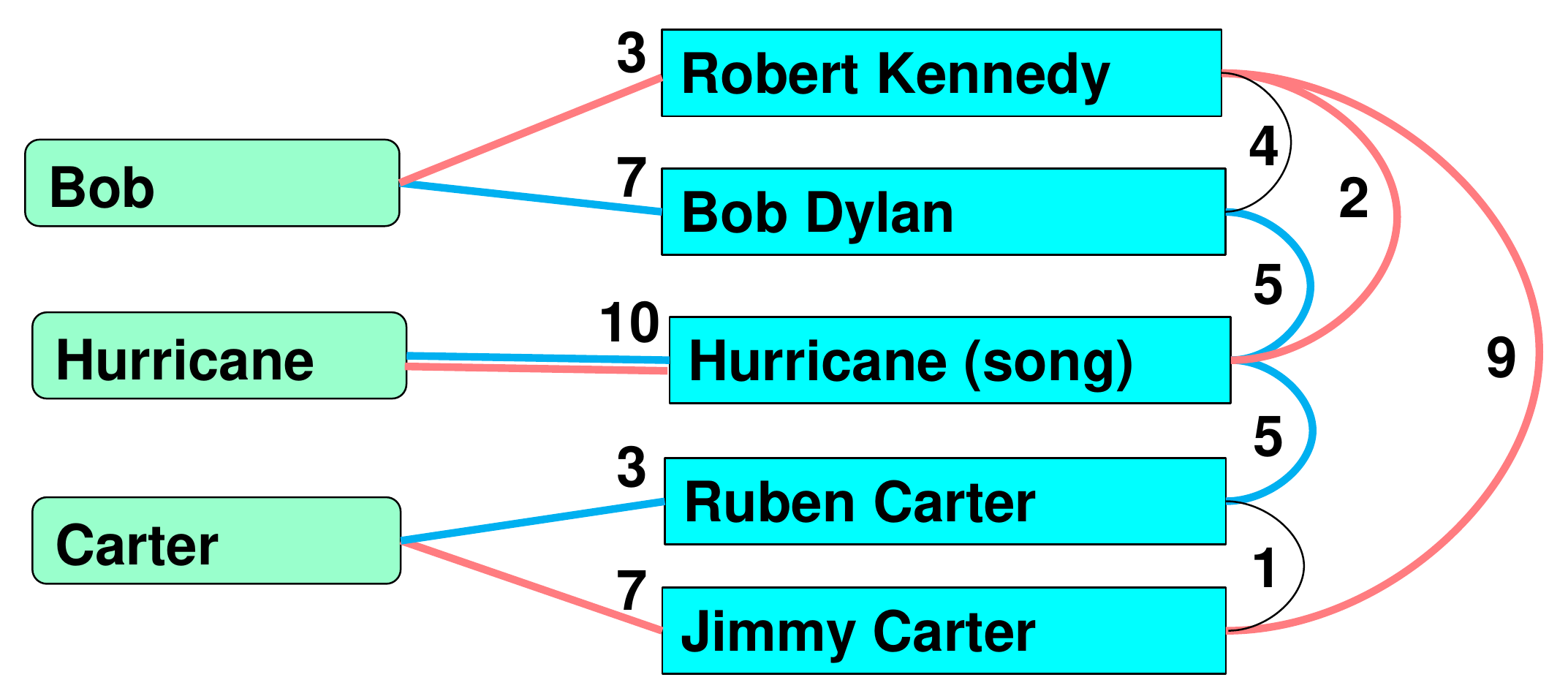}
      \caption{Example for Dense Subgraphs in EL Candidate Graph}
      \label{fig:el-densesubgraph}
\end{figure}

Figure \ref{fig:el-densesubgraph} illustrates this subgraph-based method
with a simple example: 3 mentions and 5 candidate entities,
with one mention being unambiguous. 
There are two subgraphs with high values for
their total edge weight,
shown in blue and red. The one in blue corresponds to the
ground-truth mapping (assuming the context is still the
Bob Dylan song), with a total weight of 30.
The one in red is an alternative solution 
(centered on the two prominent politicians, one of which was
a strong advocate against racism),
with a total edge weight of 31.
Both are substantially better than other choices,
such as mapping the three mentions to Robert Kennedy,
the song and Ruben Carter, which has a total edge weight
of 23. 
However, the best subgraph by using sum for weight
aggregation is the wrong output in red,
albeit by a tiny margin only.

This observation motivates the following alternative
choice for the aggregation function $aggr$:
within the subgraph of interest, instead of paying
attention only to the total weight, we consider
the {\em weakest link}, that is, the edge with the lowest weight.
The goal then is to maximize this minimum weight,
making the weakest link as strong as possible.
Using this objective, the best subgraph in
Figure \ref{fig:el-densesubgraph} is
the one with Bob Dylan, the song and Ruben Carter,
as the lowest edge weight in the respective subgraph is
3, whereas it is 2 for the alternative subgraph
with Robert Kennedy, the song and Jimmy Carter.

\begin{samepage}
\begin{mdframed}[backgroundcolor=blue!5,linewidth=0pt]
\squishlist
\item[ ] 
{\bf EL based on Max-Min Subgraph:} \\
For a given candidate graph, 
the goal is to compute the subgraph $S$
that maximizes
$$min \{weight(s) ~|~ s \in S.ME \cup S.EE\}$$
subject to the constraint:\\
for each $m \in M$ there is at most one $e \in S.E$
such that $(m,e) \in S.ME$.
\squishend
\end{mdframed}
\end{samepage}

Both variants of the dense-subgraph approach are NP-hard, due to
the constraint about mention-entity edges. However, there are
good approximation algorithms, including greedy
removal of weak edges as well as stochastic search to overcome
local optima.
This family of methods has been proposed by
\citet{DBLP:conf/emnlp/HoffartYBFPSTTW11},
and achieved good results on benchmarks
with news articles, in an unsupervised way
(except for tuning a small number of
hyper-parameters).
Methods with similar considerations 
on incorporating coherence have been developed 
by
\citet{DBLP:conf/acl/RatinovRDA11}
and \citet{DBLP:conf/www/ShenWLW12}.

\subsection{EL Methods based on Random Walks}

Using the same EL candidate graph as before,
an alternative approach that 
also has efficient
implementations, is based on 
{\bf random walks with restarts}, essentially
the same principle that underlies 
{\em Personalized Page Rank} \cite{DBLP:journals/tkde/Haveliwala03}.

First, edge weights in the candidate graph are re-calibrated 
to become proper transition probabilities, and a
small restart probability is chosen to jump back
to the starting node of a walk.
Conceptually, we initiate such walks on each of
the mentions, making probabilistic decisions
for traversing both mention-entity and
entity-entity edges many times, and occasionally
jumping back to the origin.
In the limit, as the walk length approaches infinity,
the visiting frequencies of the various nodes
converge to {\em stationary visiting probabilities},
which are then interpreted as scores for mapping
mentions to entities. 
An actual implementation would bound the length
of each walk, but walk repeatedly to obtain
samples towards better approximation.
Alternatively, iterative 
numeric algorithms from linear algebra,
most notably, Jacobi iteration, can be applied 
to the transition matrix of the graph, until
some convergence criterion is reached for the
best entity candidates of every mention.

\begin{samepage}
\begin{mdframed}[backgroundcolor=blue!5,linewidth=0pt]
\squishlist
\item[ ] 
{\bf EL based on Random Walks with Restart:} \\
Given a candidate graph with weights,
re-calibrate the weights into proper transition probabilities.\\
For each mention $m \in M$
\squishlist
\item approximate the visiting probabilities of the
possible target entities $e \in E: (m,e) \in ME$, and
\item map $m$ to the entity $e$ with the highest
probability.
\squishend
\squishend
\end{mdframed}
\end{samepage}

Although these algorithms make linking decisions
one mention at a time, they do capture the
essence of collective EL as the walks involve
the entire candidate graph and the stationary
visiting probabilities take the mutual coherence
into account.
EL methods based on random walks and related
techniques include
\citet{DBLP:conf/sigir/HanSZ11},
\citet{DBLP:journals/tacl/0001RN14} and \cite{DBLP:conf/cikm/GuoB14,DBLP:conf/sigir/PiccinnoF14,DBLP:journals/ws/GrutzeKZN16}.

\subsection{EL Methods based on Probabilistic Graphical Models}
\label{ch5-subsec:probabilisticgraphicalmodels}

The coherence-aware graph-based methods can also be
cast into {\bf probabilistic graphical models},
like CRFs and related models.
They can be seen as reasoning over a joint
probability distribution 
$$P[m_1, m_2 \dots, e_1, e_2 \dots, d]$$ 
\noindent with
mentions $m_i$, entities $e_j$ and the context given
by document $d$. This denotes the likelihood that
a document $d$ contains entities $e_1, e_2 \dots$
and that these entities are textually
expressed in the form of mentions $m_1, m_2 \dots$.
Obviously, this high-dimensional distribution is
not tractable. So it is factorized by making model
assumptions and mathematical transformations, such as
$$P[m_1, m_2 \dots, e_1, e_2 \dots, d] =
~ \prod_{i,j} P[m_i|e_j,d] \cdot 
\prod_{j,k} P[e_j, e_k]$$
where $P[m_i|e_j,d]$ is the probability of $m_i$
expressing $e_j$ in the context of $d$ and
$P[e_j, e_k]$ is the probability of the two entities
co-occurring in the same, semantically meaningful 
document.

This kind of probabilistic reasoning can be
cast into a {\bf CRF model} or {\bf factor graph}
(cf. Section \ref{ch3-subsect-machinelearning}) as follows.

\begin{samepage}
\begin{mdframed}[backgroundcolor=blue!5,linewidth=0pt]
\squishlist
\item[ ] 
{\bf CRF Model for Entity Linking:} \\
For each mention $m_i$ with entity candidates
$E_i = \{e_{i1}, e_{i2} \dots\}$, the model has
a {\bf random variable $X_i$} with values from $E_i$.
These variables capture the probabilities $P[m_i|e_j]$.\\
For each candidate entity $e_k$ (for any of the mentions),
the model has a binary {\bf random variable $Y_k$} that is
true if $e_k$ is mentioned in the document and
false otherwise.
These variables capture probabilities $P[e_k|d]$
of entity occurrence in the document.\\
All variables are assumed to be conditionally
independent, except for the following 
{\bf coupling factors}:
\squishlist
\item $X_i,Y_j$ are coupled if $e_j$ is a candidate for $m_i$,
\item $Y_j,Y_k$ are coupled 
for all pairs $e_j, e_k$.
\squishend
\squishend
\end{mdframed}
\end{samepage}

\begin{figure} [h!]
  \centering
   \includegraphics[width=0.8\textwidth]{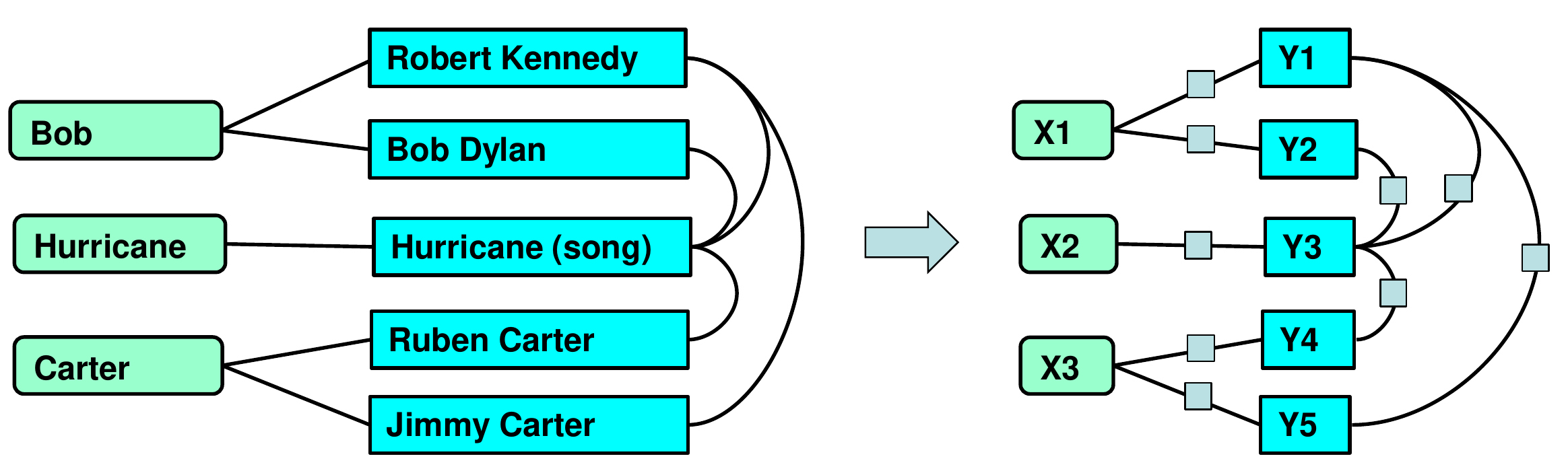}
      \caption{Example for CRF derived from EL candidate graph}
      \label{fig:el-graph-crf}
\end{figure}

Figure \ref{fig:el-graph-crf} depicts an example, showing how
the candidate graph for our running example
can be cast into the CRF structure with variables
for each mention and entity node, and
coupling factors for each of the edges.

Unlike the CRF models for NER, discussed in
Section \ref{ch3-subsect-machinelearning},
this kind of CRF does not operate on token sequences
but on graph-structured input.
So it falls under a more general class of
{\bf Markov Random Fields (MRF)}, but 
the approach is 
nevertheless widely referred to as a CRF
model.
The joint distribution of all $X_i$ and
$Y_j$ variables is factorized, by the Markov
assumption, according to the cliques in the graph.
This way, we obtain one ``clique potential''
or ``coupling factor'' per clique, often with restriction to
binary cliques, coupling two variables (i.e., the
edges in the graph). 
Training such a model entails learning 
per-clique weights
(or estimating conditional probabilities for the coupled 
variables), typically
using gradient-descent techniques.
Alternatively, similarity and coherence scores can
be used for these purposes, at least for a good
initialization of the gradient-descent optimization.
Inference on the values of variables, given a new
text document as input, is usually limited
to joint MAP inference: 
computing the combination of
variable values that has the highest posterior likelihood.
This involves Monte Carlo sampling, 
belief propagation, or variational calculus
(cf. Section \ref{ch3-subsect-machinelearning} on 
CRFs for NER).

CRF-based EL has first been developed by
\citet{DBLP:conf/icdm/SinglaD06},
\citet{DBLP:journals/tkdd/BhattacharyaG07}
and, most explicitly,
\citet{DBLP:conf/kdd/KulkarniSRC09}, 
with a variety of enhancements
in follow-up works such as
\cite{DBLP:journals/pvldb/RastogiDG11,DBLP:journals/tacl/DurrettK14,DBLP:conf/www/GaneaGLEH16,DBLP:journals/tacl/NguyenTW16}.

Many CRF-like models can also be cast into
{\bf Integer Linear Programs (ILP)}:
a discrete optimization problem with constraints,
as discussed by
\citet{DBLP:conf/icml/RothY05}
(see also
\cite{EACL2017:DanRothTutorial}).

\begin{samepage}
\begin{mdframed}[backgroundcolor=blue!5,linewidth=0pt]
\squishlist
\item[ ] 
{\bf ILP Model for Entity Linking:} \\
For each mention $m_i$ with entity candidates
$E_i = \{e_{i1}, e_{i2} \dots\}$, the model has
a binary {\bf decision variable $X_{ij}$} set to 1
if $m_i$ truly denotes $e_j$.\\
For each pair of candidate entities $e_k, e_l$ (for any of the mentions),
the model has a binary {\bf decision variable $Y_{kl}$} 
set to 1 if $e_k$ and $e_l$ are indeed both mentioned
in the input text.\\
The objective function for the ILP is to maximize
the data evidence for the choice of 0-1 values for the
$X_{ij}$ and $Y_{kl}$ variables, subject to constraints:
$$\text{maximize} ~~ \beta \sum_{ij} weight(m_i,e_j) X_{ij}
~+~ \gamma \sum_{kl} weight(e_k,e_l) Y_{kl}$$
with weights corresponding to similarity and coherence
scores and hyper-parameters $\beta,\gamma$.\\
The constraints specify that mappings are functions,
couple the $X_{ij}$ and $Y_{ij}$ variables,
and may optionally capture transitivity among identical
mentions or (obvious) coreferences, if desired:
\squishlist
\item $\sum_j X_{ij} \le 1$ for all $i$; 
\item $Y_{kl} \ge X_{ik} + X_{jl} - 1$,
stating that $Y_{kl}$ must be 1 if both $e_k$
and $e_l$ are chosen as mapping targets;
\item $X_{ik} \le X_{jk}$ and
$X_{ik} \ge X_{jk}$
for all $k$ for identical mentions $m_i, m_j$;
\item $0 \le X_{ij} \le 1$ and
$0 \le Y_{kl} \le 1$ for all $i,j,k,l$.
\squishend
\squishend
\end{mdframed}
\end{samepage}

Solving such an ILP is computationally expensive:
NP-hard in the worst case 
and also costly in practice for large instances.
However, there are very efficient ILP solvers,
such as Gurobi ({\small\url{https://www.gurobi.com/}}),
which can handle reasonably sized inputs such as
short news articles with tens of mentions and
hundreds of candidate entities.
Larger inputs could have their candidate space
pruned first by other, simpler, techniques.
Moreover, ILPs can be relaxed into LPs, linear programs
with continuous variables, followed by
randomized rounding. Often, this yields very good
approximations for the discrete optimization
(see also \cite{DBLP:conf/kdd/KulkarniSRC09}).

\section{Methods based on Supervised Learning}
\label{subsec:EL-supervised-learning}

Early work on EL (e.g., \cite{DBLP:conf/eacl/BunescuP06,DBLP:conf/cikm/MilneW08,DBLP:conf/coling/DredzeMRGF10,DBLP:conf/acl/RatinovRDA11,DBLP:conf/www/ShenWLW12,DBLP:conf/tac/CucerzanS13,DBLP:journals/tacl/LazicSR015})
already pursued machine learning for ranking
entity candidates, building on labeled training
data in the form of ground-truth mention-entity pairs
in corpora (most notably, Wikipedia articles
or annotated news articles).
These methods used support vector machines,
logistic regression and other learners,
all relying on feature engineering,
with features about mention contexts and
a suite of cues for entity-entity relatedness. 
More recently, with the advent of deep neural networks,
these feature-based learners have been superseded
by end-to-end architectures without feature modeling.
However, these methods still, and perhaps even
more strongly, hinge on sufficiently large
collections of training samples in the form
of correct mention-entity pairs.

\subsection{Neural EL}

\begin{figure} [h!]
  \centering
   \includegraphics[width=1.1\textwidth]{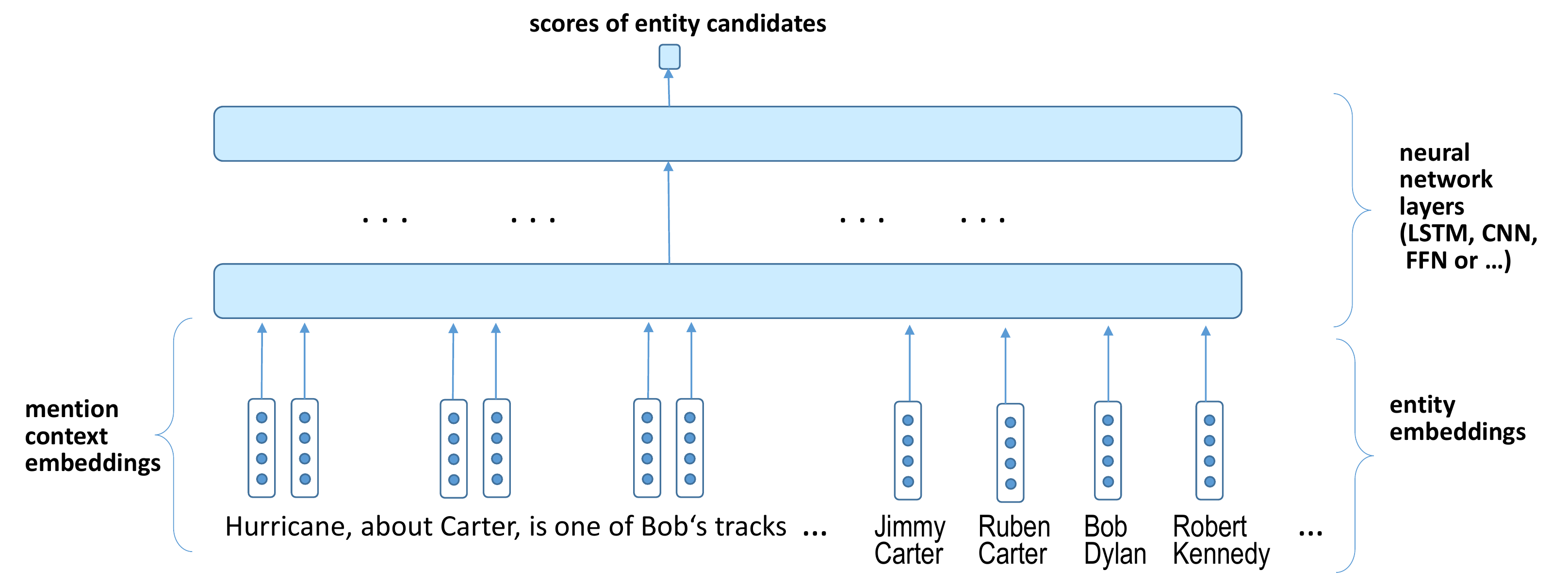}
      \caption{Illustration of Neural EL Architecture}
      \label{fig:el-neural}
\end{figure}

Recall from Section \ref{ch3-subsect-machinelearning} 
that neural networks
require real-valued vectors as input.
Thus, 
a key point in applying neural learning to the
EL problem is the {\em embeddings} of the inputs:
mention context (both short-distance and long-distance),
entity description and most strongly
related entities, and more.
This {\em neural encoding} is already a learning task
by itself, successfully addressed, for example,
by \cite{DBLP:conf/acl/HeLLZZW13,DBLP:conf/ijcai/SunLTYJW15,DBLP:conf/naacl/Francis-LandauD16,DBLP:conf/conll/YamadaS0T16,DBLP:journals/tacl/YamadaSTT17,DBLP:conf/emnlp/GuptaSR17}.
The jointly learned embeddings are fed into 
a deep neural network, with a variety of
architectures like LSTM, CNN, Feed-Forward,
Attention
learning,
Transformer-style, etc.
The output of the neural classifier
is a scoring of entity candidates for
each mention. For end-to-end training, a typical choice
for the
loss function is 
softmax over
the cross-entropy 
between predictions and ground-truth distribution.
Figure \ref{fig:el-neural} illustrates such a neural EL architecture.
As embedding vectors are fairly restricted in size,
mention contexts can be captured at different scopes:
short-distance like sentences as well as long-distance
like entire documents.
By the nature of neural networks, the 
``cross-talk'' between mentions and entities and
among entities is automatically considered,
capturing similarity as well as coherence.

Neural networks for EL are trained 
end-to-end 
on labeled corpora
where mentions are marked up with their proper entities,
using gradient descent techniques
(see, e.g., \citet{DBLP:conf/conll/KolitsasGH18} and
\cite{DBLP:conf/conll/EshelCRMYL17,DBLP:conf/emnlp/MuellerD18,DBLP:conf/acl/SevgiliPB19,DBLP:journals/corr/abs-2006-01969}).
A survey on neural EL is given by \citet{DBLP:journals/corr/abs-2006-00575}.
Some 
methods integrate EL with the NER task, jointly
spotting mentions and linking them.
In the literature, Wikipedia full-text with hyperlink
targets as ground truth 
provides ample training samples. However,
the articles all follow the same encyclopedic style.
Therefore, the learned models, albeit achieving
excellent performance on withheld Wikipedia articles,
do not easily carry over to text with different
stylistic characteristics and neither to domain-specific
settings such as biomedical articles or health
discussion forums.
Another large resource for training
is the {\em WikiLinks} corpus 
\cite{singh2012wikilinks} which comprises Web pages
with links to Wikipedia articles.
This captures a wider diversity of text styles,
but the ground-truth labels have been compiled
automatically, hence containing errors.

Overall, 
neural EL is very promising, but probably not yet as
successful as its neural counterparts for NER
(see Section \ref{ch3-subsect-machinelearning}),
given that it is easier to obtain training data for NER.
Neural EL shines when 
trained
with large labeled collections and the downstream
texts to which the learned linker is applied have
the same general characteristics.
When training samples are scarce or the use-case data characteristics
substantially deviate from the training data, it is much harder
for neural EL to compete with feature-based unsupervised methods.
Very recently, methods for {\em transfer learning} have been integrated
into neural EL (e.g., \cite{DBLP:conf/aaai/KarRBDC18,DBLP:conf/acl/LogeswaranCLTDL19,DBLP:journals/corr/abs-1911-03814}). These methods are trained on one labeled collection,
but applied to a different collection
which does not have any labels and has a disjoint set of target entities.
A major asset to this end is the integration of large-scale language embeddings,
like BERT \cite{DBLP:conf/naacl/DevlinCLT19}, which covers both training and target domains.
The effect is an implicit capability of 
``reading comprehension'',
which latently captures relevant signals about context similarity and coherence.
Transfer learning still seems a 
somewhat brittle approaches,
but such methods will be further advanced, leveraging even
larger language embeddings,
such as GPT-3 based on a neural network with over 100 billion parameters \cite{brown2020language}.

Regardless of future advances along these lines,
we need to realize that
EL comes in many different flavors: for different domains,
text styles and objectives (e.g., precision vs. recall).
Therefore, flexibly configurable, unsupervised methods
 with explicit feature engineering
will continue to play a major role.
This includes methods that require tuning a handful of hyper-parameters, which can be
done by domain experts or using a small set of labeled samples.

\section{Methods for Semi-Structured Data}
\label{sec:el-semistructured}

In addition to text documents, KB construction
also benefits from tapping semi-structured contents
with lists and tables.
In the following, we focus on the case of ad-hoc
web tables as input, to exemplify 
EL over semi-structured data.
Table \ref{table:example-EL-tables} shows an
example with ambiguous mentions such as
``Elvis'', ``Adele'', ``Columbia'', ``RCA'' as well as
abbreviated or slightly misspelled names
(e.g., ``Pat Garrett'' should be the album
\ent{Pat Garrett \& Billy the Kid}).
Note that such tables are usually surrounded
by text -- within web pages, with table captions,
headings etc., which can be harnessed
as additional context.

\begin{table}
\begin{center}
\begin{tabular}{|l|l|l|l|l|}\hline
Name & Title & Album & Label & Year \\\hline\hline
Bob Dylan & Hurricane & Desire & Columbia & 1976 \\\hline
Bob Dylan & Sara & Desire & Columbia & 1976 \\\hline
Bob Dylan & Knockin on Heavens Door & Pat Garrett & Columbia & 1973 \\\hline
Elvis & Cant Help Falling in Love & Blue Hawaii & RCA & 1961 \\\hline
Adele &  Make You Feel My Love & n/a & XL & 2008\\\hline
\end{tabular}
\caption{Example for entity linking task over web tables}
\label{table:example-EL-tables}
\end{center}
\end{table}

From a traditional database perspective, it seems
that the best cues for EL over tables is to 
exploit the table schema, that is, column headers
and perhaps inferrable column types.
However, these tables are very different from
well-designed databases: they are hand-crafted in 
an ad-hoc manner, and their column names are often
not exactly informative (e.g., ``Name'', ``Title'',
``Label'' are very generic).
Thus, it seems that the EL problem is much harder
for tables. However, we can leverage the tabular
structure to guide the search for the proper entities.
Specifically, we pay attention to 
{\bf same-row mentions} and {\bf same-column mentions}:

\squishlist
\item {\bf Same-row mentions} are most tightly related.
So their coherence should be boosted in the objective function.
\item {\bf Same-column mentions} are not directly
related, but they are typically of the same type,
such as \ent{musicians}, \ent{songs}, \ent{music albums}
and \ent{record labels}.
So the objective function should incorporate 
a soft constraint for per-column {\em homogeneity}.
\squishend

By taking these design considerations into account,
the EL optimization can be varied as follows.

\begin{samepage}
\begin{mdframed}[backgroundcolor=blue!5,linewidth=0pt]
\squishlist
\item[ ] 
{\bf EL Optimization for Web Tables:} \\
Consider a table with $c$ columns,
$r$ rows and entity mentions 
$m_{ij}$ where $i,j$ are the row and column
where the mention occurs.
Each $m_{ij}$ has a set of entity candidates
$E(m_{ij})$. 
All mentions together are denoted as $M$, and
the pool of target entities overall as $E$.
For ease of notation, we assume that all table cells
are entity mentions (i.e., disregarding the fact
that some columns are about literal values).
The goal is to find a, possibly partial, function
$\phi: M \rightarrow E$ that maximizes the objective
$$\alpha \sum_{m_{ij}\in M}
pop(m_{ij},\phi(m_{ij})) +$$
$$\beta1 \sum_{m_{ij} \in M}
sim(rowcxt(m_{ij}),cxt(\phi(m_{ij}))) +$$
$$\beta2 \sum_{m_{ij} \in M}
sim(doccxt(m_{ij}),cxt(\phi(m_{ij}))) +$$
$$\gamma \sum_{e,f \in E} \{coh(e,f) ~|~ 
m_{ij},m_{ik} \in M:
j\ne{}k, 
e=\phi(m_{ij}), f=\phi(m_{ik})\} +$$
$$\delta \sum_{j=1..c} hom\{type(e) ~|~ 
e = \phi(m_{\ast{}j}, \ast = 1..r\}$$
where $\alpha, \beta1, \beta2,
\gamma$, $\delta$ are tunable hyper-parameters.
As before,
$pop$ denotes mention-entity popularity,
$cxt$ the context of mentions and entities
in two variants for mentions: 
$rowcxt$ for same-row cells,
$doccxt$ for the entire document.
$sim$ denotes contextual similarity and
$coh$ pair-wise coherence between same-row entities.
$hom$ is a measure of type {\em homogeneity}
and {\em specificity}.
\squishend
\end{mdframed}
\end{samepage}

This framework leaves many choices for the
underlying measures: defining specifics of the
two context models, defining the measures for
type homogeneity and specificity, and so on.
For example, $hom$ may combine the fraction of 
per-column entities that have a common type and
the depth of the type in the KB taxonomy.
The latter is important to avoid overly generic
types like \ent{entity}, \ent{person} or \ent{artefact}.
The inference of column types has been addressed
as a problem by itself (e.g., \cite{DBLP:journals/pvldb/VenetisHMPSWMW11,DBLP:conf/aaai/ChenJHS19}).
The case of {\em lists}, which can be
viewed as single-column tables, has received
special attention, too (e.g., \cite{DBLP:conf/kdd/ShenWLW12,DBLP:conf/esws/HeistP20}).

The literature on EL over tables, most notably
\citet{DBLP:journals/pvldb/LimayeSC10}
and \cite{DBLP:conf/semweb/BhagavatulaND15,DBLP:conf/cikm/IbrahimRW16,DBLP:conf/edbt/RitzeB17}, 
discusses a variety of viable
design choices in depth.
\citet{DBLP:journals/tist/ZhangB20} 
give
a
survey on knowledge extraction
from web tables.

Algorithmically, many of the previously presented
methods for text-based EL carry over to the case of
tables. Scoring and ranking methods can simply
extend their objective functions to the above
table-specific model.
Graph-based methods, including dense subgraphs,
random walks and CRF-based inference, merely
have to re-define their input graphs accordingly.
The seminal work of
\citet{DBLP:journals/pvldb/LimayeSC10}
did this with a probabilistic graphical model,
integrating the column type inference in a joint
learning task.

\begin{figure} [h!]
  \centering
   \includegraphics[width=0.7\textwidth]{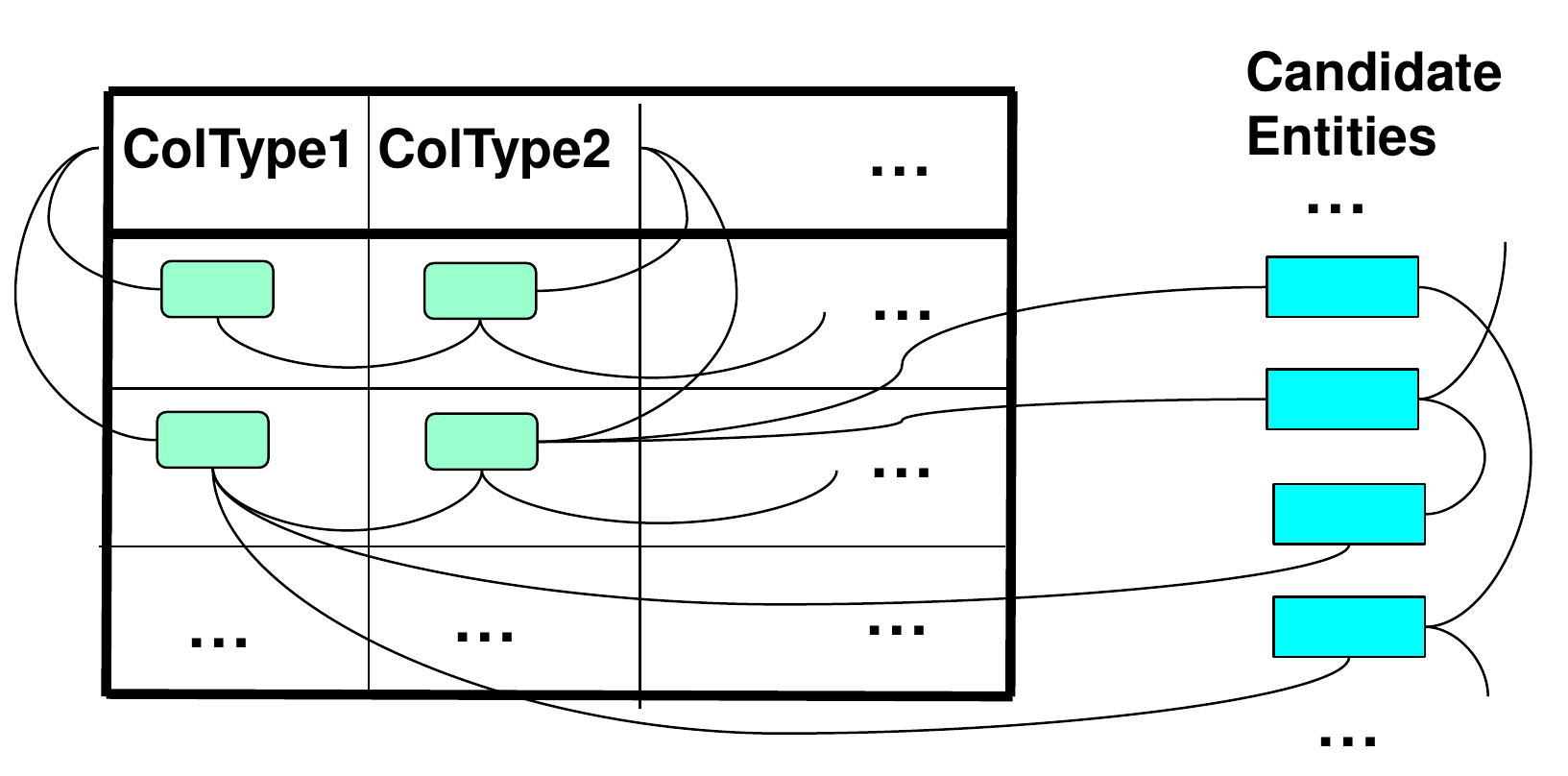}
      \caption{Illustration of EL Graph for Tables}
      \label{fig:el-tables-graph}
\end{figure}

Figure
\ref{fig:el-tables-graph} illustrates
this graph-construction step:
edges denote CRF-like coupling factors
or guide random walks (only some edges are shown).
In the figure, the type homogeneity is depicted
by couplings with the prevalent column type.
Alternatively, the CRF could couple all mention pairs
in the same column including the column header
\cite{DBLP:conf/semweb/BhagavatulaND15}.

\section{EL Variations and Extensions}

\noindent{\bf Iterative Linking:}\\
A simple yet powerful principle that can be
combined with virtually all EL methods is
to make linking decisions in multiple rounds,
based on the mapping confidence
(see, e.g., \cite{DBLP:conf/cikm/GuoB14,DBLP:conf/www/NguyenHTW14,DBLP:conf/sigir/PiccinnoF14}).
Initially, only unambiguous mentions are mapped,
unless there is uncertainty on whether they could
denote out-of-KB entities.
In the next round, only those mentions are mapped
for which the method has high confidence.
After every round, all similarity and coherence
measures are updated triggering updates to the
graph or other model on which EL operates.
As more and more entities are mapped, they
create a more focused context for subsequent rounds.
For the running example, suppose that
we can map ``Hurricane''
to the song with high confidence.
Once this context about music is established, the
confidence in linking ``Bob'' to Bob Dylan, rather
than any other prominent Bobs, is boosted.

\vspace*{0.2cm}
\noindent {\bf Domain-specific Methods:}\\
There are numerous variations and extensions of
EL methods, including {\em domain-specific} approaches,
for example, for mentions of proteins, diseases etc.
in biomedical texts
(see, e.g., \cite{DBLP:journals/jamia/AronsonL10,DBLP:journals/bmcbi/FunkBGRBCHV14,DBLP:conf/sigmod/DaiZ0FNO18,DBLP:journals/corr/abs-1908-03548}
and references given there),
and {\em multi-lingual} approaches
where training data is available only in some 
languages and transferred to the processing of
other languages (see, e.g., \cite{DBLP:conf/acl/SilJRC18} and
references there).
The case for domain-specific methods can also be
made for music (e.g., names of songs, albums, bands --
which include many common words and appear incomplete
or misspelled), 
bibliography with focus on author names and publication titles,
and even business where company acronyms are common
and product names have many variants.

\vspace*{0.2cm}
\noindent {\bf Methods for Specific Text Styles:}\\
There are approaches customized to
specific kinds of {\em text styles},
most notably, social media posts such as tweets
(e.g., \cite{DBLP:conf/wsdm/MeijWR12,DBLP:conf/acl/LiuLWZWL13,DBLP:conf/kdd/ShenWLW13,DBLP:journals/ipm/DerczynskiM0EGTPB15}) with characteristics very
different from encyclopedic pages or news articles.
EL methods that are specifically geared for very short
input texts include the {\em TagMe} tool
\cite{DBLP:conf/cikm/FerraginaS10,DBLP:conf/sigir/PiccinnoF14}.

\vspace*{0.2cm}
\noindent {\bf Methods for Queries and Questions:}\\
Yet another specific kind of input is 
{\em search-engine queries} when users
refer to entities by telegraphic phrases
(e.g., ``Dylan songs covered by Grammy and Oscar winners'').
For EL over queries, 
\cite{DBLP:conf/www/SawantC13}
and 
\cite{DBLP:conf/emnlp/YahyaBERTW12}
developed powerful
methods based on probabilistic graphical models
and integer linear programming, respectively.
Efficiency issues for interactive results are discussed in \cite{DBLP:conf/wsdm/BlancoOM15}.
For fully formulated questions 
with the use case of QA systems, the recent work
\cite{DBLP:conf/emnlp/LiMIMY20} presents an end-to-end
neural learning approach.

\vspace*{0.2cm}
\noindent {\bf Methods for Specific Entity Types:}\\
Finally, there are also specialized EL methods for
disambiguating geo-spatial entities, such as
``Twin Towers'' (e.g., in Kuala Lumpur, or formerly in
New York)
and temporal entities such as ``Mooncake Festival''.
Methods for these types of entities are covered,
for example, by 
\cite{leidner2008toponym,DBLP:journals/cacm/SametSLAFLPST14,DBLP:journals/geoinformatica/ChenVW19} for spatial mentions, aka.
{\em toponyms}, and
\cite{DBLP:journals/coling/StrotgenGHH18,DBLP:conf/www/KuzeySSW16} for temporal expressions, aka. {\em temponyms}. A notable project where spatio-temporal entities have been annotated at scale is the
GDELT news archive ({\small\url{https://www.gdeltproject.org/}}),
supporting event-oriented knowledge discovery
\cite{leetaru2013gdelt}.

\section{Entity Matching (EM)}
\label{sec:entity-matching}

When the input to entity canonicalization takes the form
of {\em data records} 
from two or more database tables and there is
no pre-existing reference repository of entities, we are
facing an {\bf entity matching (EM)} problem, aka
{\em record linkage}, {\em deduplication} or 
{\em entity resolution} in the literature, 
with extensive surveys by
\citet{DBLP:journals/dke/KopckeR10,DBLP:books/daglib/0030287,DBLP:journals/corr/abs-1905-06397}.

\begin{figure} [h!]
  \centering
   \includegraphics[width=0.8\textwidth]{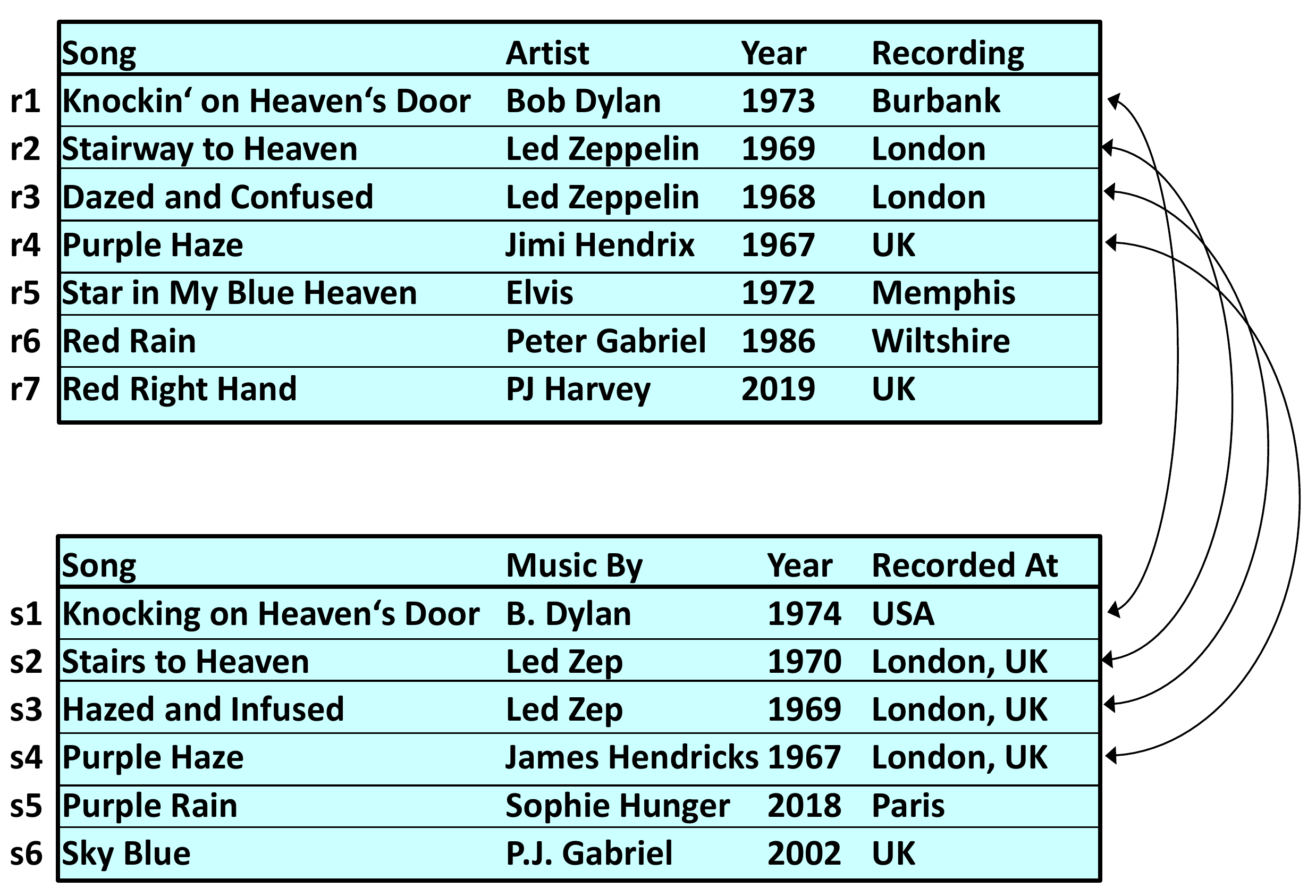}
      \caption{Example Input for Entity Matching}
      \label{fig:em-tables}
\end{figure}

Figure \ref{fig:em-tables} shows a toy example with two small tables, where some attribute values are inaccurate or erroneous.
Here, the entities to be matched are the songs, 
while artists, dates etc.
serve as context features.
The desired output is four pairs of matching records:
$(r1,s1)$, $(r2,s2)$, $(r3,s3)$ and
$(r4,s4)$. 
This partial matching, leaving all other records unmatched,
illustrates that this is again a selection-and-assignment optimization problem
(cf. Section \ref{subsec:EL-optimization-and-LTR}).
For ease of explanation, we assume that none of the input tables 
contains duplicates, that is, none of the tables contains multiple records for the same entity.
The assignment aims at a partial and injective (1:1), but not necessarily surjective mapping.

The desired matching of entities is a crucial step in 
data integration projects and for enterprise KBs.
A typical use case is to combine different databases
of customers, products, etc., to obtain insights from
multiple departments or business branches of a company
or to handle organizational mergers.
The input tables for EM are often complementary in their contents,
each containing some attributes that are not present in the others
(or having foreign keys that point to additional tables with
complementary data). 
Therefore, once matching pairs (or matching sets) of records are determined,
the downstream application typically needs to combine and reconcile the data for each pair (or matching set). This is referred to as {\em entity merging}. 
The specifics of merging are highly application-dependent,
and often involve sophisticated {\em conflict resolution} when
values for the same attribute disagree (e.g., 
\ent{Year} values in Figure \ref{fig:em-tables}).
We do not discuss this out-of-scope aspect in this article,
and refer to textbooks and surveys on data integration \cite{DoanHalevyIves2012,DBLP:journals/debu/StonebrakerI18,DBLP:journals/debu/000118}.

The high-level design considerations for EM solutions are not that
different from those for EL (see Section \ref{subsec:el-designspace}):
exploit i) approximate name matches,
ii) context similarity and 
iii) global consistency.
There is no notion of entity popularity, though.
However, 
the pair-wise nature of entity matchings
and the large scale of input tables mandate a somewhat different
approach. If each table has 100,000 records, 
a quadratic search space of up 10 billion candidate pairs 
is a computational challenge.

\subsection{Blocking}

The key idea to address this scalability issue 
is to partition both tables in a way that 
candidates for matching pairs can be computed 
for each partition separately. This technique is
known as {\bf blocking} 
\cite{newcombe1967record,DBLP:journals/dke/KopckeR10,DBLP:journals/corr/abs-1905-06397}. 
It is usually based on
choosing a {\em blocking key}: one attribute from each table
by whose values we obtain meaningful partitions such that:
\squishlist
\item truly matching pairs likely reside in the same partition (i.e., few false negatives), and
\item partitions are small and computationally tractable
(i.e., not too many false positives).
\squishend
In the example, the values of the \ent{Year} attributes could
serve as blocking keys (e.g., one partition per year or
per interval of a few years).
When the blocking key is a string attribute, such as \ent{Song},
the partitioning usually includes a hash mapping so as
to group similar (but not exactly matching) values.
A widely used technique is to compute character-level
n-grams (e.g., with n=3, aka trigrams) and represent each
string by its bag or set of n-grams.
For efficiency, this can in turn be cast into
so-called {\em minHash sketches} \cite{DBLP:journals/cn/BroderGMZ97,DBLP:journals/jcss/BroderCFM00}, which provide
for very fast estimates of set-overlap sizes. 
Another technique that is often combined with these
considerations is {\em locality-sensitive hashing (LSH)}
\cite{DBLP:conf/stoc/IndykM98,DBLP:conf/vldb/GionisIM99}, giving quick access to all strings that have
high n-gram overlap (or are close by some other distance measure).
The textbook \cite{DBLP:books/cu/LeskovecRU14}
gives in-depth explanations of these hash sketching techniques.
\cite{DBLP:conf/sigmod/ChaudhuriGGM03}
shows how to efficiently and robustly apply 
and customize 
these techniques to
string attributes in databases, such as
person names or addresses.

For the example of Figure \ref{fig:em-tables},
a simple choice for the blocking key could be grouping the 
\ent{Year} values into decades:
[1960,1969], [1970,1979], etc.
This way, $\{$r2,r3,r4,s3,s4$\}$ would end up
in the same block (for the 1960s),
$\{$r1,r5,s1,s2$\}$ in another block (for the 1970s),
and so on.
Note that this gives rise to both 
potential {\em false positives}, such as possibly matching 
$r5$ with $s2$, and {\em false negatives}, such as
missing out on the pair $(r2,s2)$ 
because these records are mapped to different blocks. We will see below how to counter these effects.

An alternative blocking key could be based on
the \ent{Song} attributes: computing n-gram representations and mapping them to hash buckets
using techniques like minHash and LSH (see above).
This way, similar song names would end up in the same blocks, for example:
$\{$r2,r5,s2$\}$ (with common trigrams like ``sta'', ``hea'', ``eav'' etc.),
$\{$r3, r4, s3, s4$\}$  (with trigrams like
``aze'' etc.).
If, as another alternative, we applied this approach to the attributes \ent{Artist} and \ent{Music By},
we should consider that different parts of
the name strings carry different weights.
For example, first and middle names may be
fully spelled out or abbreviated (including nicknames or stage names), and last names are, therefore,
more informative. 
Also, small typos may be present 
(e.g., ``Hendrix'' misspelled as ``Hendricks'').
\cite{DBLP:conf/sigmod/ChaudhuriGGM03}
presents techniques that consider both
word-level and character-level cues, with
smart weighting of different components.
\cite{cohen2003comparison} compares a wide variety
of string-matching measures.

As blocking keys based on single attributes have limited power,
a natural generalization is to combine values from multiple attributes and derived functions into 
{\em blocking functions} or {\em blocking rules}
(see, e.g., \cite{DBLP:conf/sigmod/DasCDNKDARP17,DBLP:journals/corr/abs-1905-06397}).

\subsection{Matching Rules}

The second major ingredient of modern EM methods
is the comparison of candidate pairs of records.
This is cast into a family of {\bf matching rules}
using {\em similarity functions}, with different rules/functions referring to different attributes.
Aligning the attributes from the two tables
(i.e., which column in table $R$ is comparable to which column in table $S$) is assumed to be solved beforehand, either using
techniques for database schema matching 
\cite{DoanHalevyIves2012,DBLP:journals/vldb/RahmB01,DBLP:reference/bdt/RahmP19b}
or simply asking
a domain expert. 
The matching rules can be hand-crafted by experts, or
inferred from labeled record pairs collected via crowdsourcing (with binary labels 
``same'' or ``different'').
A variety of methods for this {\em human-in-the-loop} approach have been investigated (see, e.g., \cite{DBLP:conf/sigmod/DasCDNKDARP17,DBLP:journals/debu/FirmaniGSS18,DBLP:journals/corr/abs-1906-06574}).

Multiple matching rules can be combined
into training a binary {\em classifier}, learned from 
an ensemble of similarity functions
or directly from labeled record pairs.
A popular choice is to train a decision tree or
random forest (i.e., ensemble of decision trees),
and neural learning has been pursued as well
(e.g., \cite{DBLP:conf/sigmod/MudgalLRDPKDAR18}).

For the example of Figure \ref{fig:em-tables},
the following matching rules are conceivable:
\squishlist
\item $r$ and $s$ are considered a matching (candidate) pair if they have the same Year value, or slightly relaxed, if they are at most two years apart.
\item $r$ and $s$ are matching if the locations for 
\ent{Recording} and \ent{Recorded At} are close in terms of geo-coordinates, for example, considering London and Wiltshire and UK to be approximate matches. This may be justified by the assumption that such attributes contain different resolutions, inaccuracies
or data-entry errors.
\item $r$ and $s$ are matching if the two records agree on the same values for the \ent{Artist}
and \ent{Music By} attributes, with leeway for shorthands and other name variations, such as ``Led Zep'' being short for ``Led Zeppelin''.
\item $r$ and $s$ are matching if the \ent{Song} strings have high similarity, in terms of edit distance (Levenshtein, Jaro-Winkler etc.), n-gram overlap
and further 
measures. This could even include word or phrase embeddings (e.g., word2vec, see Section \ref{subsec:word-entity-embeddings}) to quantify similarity.
For example, all strings that contain related colors (e.g., ``red'' and ``purple'') and have informative words in common (e.g., ``rain'')
could be considered similar. 
At the phrase level, we may even deem
``heaven'' and ``sky blue'' as highly related.
\squishend

The pair-wise comparisons of this kind are also known as {\em similarity joins} or
{\em fuzzy joins}.
Obviously, each of the above matching rules 
would be a weak decision criterion.
This is why we need to combine several rules
together, or learn a classifier 
with rule combinations (e.g., a decision tree)
or directly from training samples
(i.e., labeled pairs of records).

\subsection{EM Workflow}

Centered on these two building blocks -- blocking and matching rules -- the {\em workflow for scalable EM} 
typically looks as follows.
\citet{DBLP:reference/bdt/RahmP19a}
and
\citet{DBLP:journals/cacm/DoanKCGPCMC20}
provide further discussion of state-of-the-art
system architectures for entity matching.

\begin{samepage}
\begin{mdframed}[backgroundcolor=blue!5,linewidth=0pt]
\squishlist
\item[ ] 
{\bf Basic Workflow for Entity Matching (EM):}
\squishlist
\item[ ] {\bf Input:} tables $R$ and $S$ 
\item[ ] {\bf Output:} pairs of records $(r \in R, s \in S)$ such that\\
\hspace*{0.75cm} $r$ and $s$ denote the same entity
\item[0.] {\bf Preprocessing:}
\item[ ] {\em Schema Alignment:}
identify comparable attributes of $R$ and $S$
\item[ ] {\em Training Data:} 
obtain positive and negative samples \\
\hspace*{0.75cm} (matching and non-matching record pairs)
\item[ ] {\em Learn Classifier:}
based on training samples and/or \\
\hspace*{0.75cm} hand-crafted matching rules
\item[1.] {\bf Blocking:}
\item[ ]  {\em Blocking Key:} specify grouping attribute (or attribute combination) or\\
\hspace*{0.75cm} derived function of attribute values;
\item[ ] {\em Partitioning:} divide each table into $m$ blocks $R1, R2 \dots Rm, S1, S2 \dots Sm$,\\ 
\hspace*{0.75cm} by grouping records on blocking key;
\item[2.] {\bf Matching:}
\item[ ] for each partition $j=1..m$ do
\item[ ] $~~~~~$ for all pairs $(r,s) \in Rj \times Sj$ 
(above similarity threshold) do
\item[ ] $~~~~~$ $~~~~~$ apply {\em classifier} or 
{\em matching rules} to $(r,s)$,
\item[ ] $~~~~~$ $~~~~~$ obtaining positive candidate pairs;
\item[3.] {\bf Joint Inference:}
\item[ ] {\em Candidate Union:} combine all positive pairs from all partitions,\\
\hspace*{0.75cm} forming a matching-candidate set $C$;
\item[ ] {\em Reconciliation:} apply holistic reasoning to identify matching pairs $(r,s) \in C$\\
\hspace*{0.75cm} (pruning false positives of $C$);
\squishend
\squishend
\end{mdframed}
\end{samepage}

By the nature of partition-based local computations,
the set $C$ of matching candidates, obtained after Phase 2,
typically contains both false positives and false negatives.
Moreover, if the blocking is chosen such that 
partitions overlap, with some records being present in more
than one partition, the output of Phase 2 can even contain
contradictions. 
Therefore, Phase 3 with joint inference over the entire
intermediate output is decisive.
As the data is drastically reduced at this point,
more powerful and computationally more expensive 
methods can be considered here, including
probabilistic graphical models, deep neural networks,
advanced clustering methods, and more
(e.g., \cite{DBLP:journals/pvldb/RastogiDG11,DBLP:conf/adbis/SaeediPR17,DBLP:journals/corr/abs-2004-00584})
This is also the point when consistency constraints and
holistic background knowledge can be utilized, 
at affordable cost.
There are options for data-partitioned parallelism
here as well (e.g., \cite{DBLP:journals/pvldb/RastogiDG11}),
but they scale only to some extent.

\vspace*{0.2cm}
\noindent{\bf Iterative Processing:}\\
Another pragmatic way of improving the output quality of
Phase 2, while avoiding expensive global computations, is
to repeat Phase 1, Phase 2 or both Phases 1 and 2
several times, with different choices of blocking keys
and/or matching rules in each iteration \cite{DBLP:conf/sigmod/WhangMKTG09}.
Iterating over different blockings reduces the risk of
false negatives: missing out on a matching record pair
$(r,s)$ because $r$ and $s$ ended up in different partitions
by the chosen blocking key (e.g., $r2$ and $s2$ in
Figure \ref{fig:em-tables} if we 
blocked by decades of years).
Repeating the matching phase with different configurations
can help to prune false positives, by keeping only record pairs that are deemed the same entity under different kinds of
similarity functions and classifiers.
This is essentially the same effect that an ensemble of classifiers,
like a random forest, would achieve. The idea 
generalizes to all kinds of families of matching algorithms.
Furthermore, iterative EM methods can feed tentative matching results from one round to the next, to achieve more robust final outputs
\cite{DBLP:conf/sigmod/WhangMKTG09}.

\section{Take-Home Lessons}

Key points to remember from this chapter
are the following:

\squishlist
\item {\em Entity Linking (EL)} 
(aka. Named Entity
Disambiguation) is the task of mapping 
entity mentions in web pages 
(detected by NER, see Chapter \ref{ch3:entities}) 
onto
uniquely identified entities in a KB or
similar repository.
This is a key step for constructing 
{\em canonicalized} KBs.
Full-fledged EL methods also consider the
case of {\em out-of-KB} entities, where a mention
should be mapped to null rather than any entity in the KB.
\item The input sources for EL can be {\em text 
documents} or {\em semi-structured
contents} such as lists or tables in web pages. 
In the text case, related tasks
like coreference resolution (CR) for pronouns and
common noun
phrases, or even 
general word sense
disambiguation, may be incorporated as well.
\item {\em Entity Matching (EM)} is a variation
of the EL task where the inputs are structured
data sources, such as database tables. 
The goal here is to map mentions in
data records from one source to
those of a different source and, this way, 
compute equivalence classes.
There is not necessarily a KB as
a reference repository.
\item EL and EM leverage a variety of signals,
most notably: a-priori {\em popularity} measures
for entities and name-entity pairs,
{\em context similarity} between mentions and entities,
and {\em coherence} between entities that are considered
as targets for different mentions in the same input.
Specific instantiations may exploit 
existing links like those in Wikipedia,
text cues like keywords and keyphrases, 
or embeddings for latent encoding of such contexts.
\item EL methods can be chosen
from a spectrum of paradigms, spanning
{\em graph algorithms}, 
{\em probabilistic graphical 
models}, feature-based {\em classifiers}, 
all the way to 
{\em neural learning} without explicit feature modeling.
A good choice depends on the prioritization of
complexity, efficiency, precision, recall,
robustness and
other quality dimensions.
\item None of the state-of-the-art EL methods
seems to universally dominate the others.
There is (still) no one-size-fits-all solution.
Instead, a good design choice depends on
the application setting: 
scope and scale of
entities under consideration (e.g., focus
on vertical domains or entities of specific types),
language style and structure of inputs 
(e.g., news articles vs. social media vs.
scientific papers),
and requirements of the use case (e.g., 
speed vs. precision vs. recall).
\squishend

\clearpage\newpage
\chapter{KB Construction: Attributes and Relationships}
\label{ch4:properties}

\section{Problem and Design Space}
\label{sec:properties-rationale}

Using methods from the previous chapters,
we can now assume that we have a knowledge
base that has a clean and expressive taxonomy of semantic types (aka. classes) and that these
types are populated
with a comprehensive set of
canonicalized (i.e., uniquely identified) entities.

The next step is to enrich the entities with {\em properties} in the form of
SPO triples, covering both
\squishlist
\item {\em attributes} with literal values
such as the birthday of a person, the year when a song or
album was released, 
the maximum speed and energy consumption
of a car model, etc.,
and
\item {\em relations} with other entities
such as birthplace, spouse, composers and
musicians for a song or album,
manufacturer of a car, etc.
\squishend

Regarding terminology, recall that attributes and relations are both {\em mathematical relations},
or exchangeably, {\em predicates} in logic terms.
We distinguish KB attributes from KB relations,
as the former is concerned with literals as arguments
and the latter with entities as arguments.

In this chapter, we present methods for extracting 
SPO triples; most of these can
handle both attributes and relations 
in a more or less unified way.
We will see that many of the principles
(e.g., statement-pattern duality),
key ideas (e.g., pattern learning) and 
methodologies (e.g., CRFs or neural networks)
of Chapter \ref{ch3:entities}
are applicable here as well.

\vspace*{0.2cm}
\noindent {\bf Assumptions:}\\
Best-practice methods build on a number of assumptions that are justified by already having a clean and large KB of entities and types.

\squishlist
\item {\bf Argument Spotting:} 
Given input content in the form of
a text document, Web page, list or table,
we can spot and canonicalize arguments for
the subject and object of a candidate triple.
This assumption is valid because we already
have methods for entity discovery and linking.
As for attribute values, specific techniques
for dates, monetary numbers and other quantities (with units) can be harnessed for spotting and normalization
(e.g., \cite{DBLP:conf/aaai/MadaanMMRS16,DBLP:journals/tacl/RoyVR15,DBLP:conf/sigir/AlonsoS18,DBLP:journals/coling/StrotgenGHH18}).
\item {\bf Target Properties:}
We assume that, for each type of entities, 
we have a fairly good understanding and
a reasonable initial list of which properties
are relevant to capture in the KB. 
For example, we should know upfront that
for people in general we are interested in
birthdate, birthplace, spouse(s), children,
organizations worked for, awards, etc.,
and for musicians, we additionally need to harvest
songs composed or performed, albums released,
concerts given, instruments played, etc.
These lists are unlikely to be complete,
but they provide the starting point for this
chapter. We will revisit and relax the
assumption in Chapter \ref{ch7:open-schema-construction} on
the construction and evolution of open schemas.
\item {\bf Type Signatures:}
We assume that each property of interest has a type signature such that we know the domain and range of the property upfront. 
This is part of the KB schema (or ontology).
By associating properties with types, we already
have the domain, but we require also that
the range is specified in terms of data types
for both attributes (literal values) and
relations (entity types).
This enables
high-precision knowledge acquistion, as we can
leverage type constraints for de-noising.
For example, we will assume the following specifications:
\squishend

{\small
\lstset{language=HTML,upquote=true}
\begin{lstlisting}
     birthdate: person x date
     birthplace: person x location
     spouse: person x person
     worksFor: person x organization
\end{lstlisting}
}%

\vspace*{0.2cm}
\noindent {\bf Schema Repositories of Properties:}\\
Where do these pre-specified properties of interest and their type signatures come from?
Spontaneously, one may think this is a leap of faith, but on second thought, there are major
assets already available.

\squishlist
\item Early KB projects like Yago and Freebase demonstrated that it is well feasible, with limited effort, to manually compile schemas 
(aka. ontologies) for relevant properties.
{\em Freebase} comprised several thousands of properties with type signatures.
\item Frameworks like {\em schema.org} \cite{Guha:CACM2016} have specified 
{\em vocabularies} for types and properties.
These are not populated with entities, but
one can easily use the schemas to drive the KB population.
Currently, schema.org comprises mostly business-related
types (ca. 1000) and their salient properties.
\item There are rich {\em catalogs} and
{\em thesauri}
that cover a fair amount of vocabulary for
types and properties. Some are well organized and clean, for example, the {\small\url{icecat.biz}}
catalog of consumer products. 
Others are not quite so clean, but can still
be starting points towards a schema, an example
being the UMLS thesaurus for the biomedical domain
({\small\url{https://www.nlm.nih.gov/research/umls/}}).
\item {\em Domain-specific KBs}, say on food, health or energy, can start with some of the above repositories and would then require
expert efforts to extend and refine their schemas.
This is manual work, but it is not a huge endeavor, as the KB is very unlikely to require more than
a few thousand types and properties.
For health KBs, for example,
some tens of types and properties
already cover a useful slice
(e.g., \cite{Ernst:BMCbioinformatics2015,waagmeester2020science}).
\squishend

\clearpage\newpage
\section{Pattern-based and Rule-based Extraction}
\label{sec:properties-patterns}

\subsection{Specified Patterns and Rules}
\label{subsec:specifiedpatterns}

The easiest and most effective way of harvesting 
attribute values and relational arguments,
for given entities and a target property,
is again to tap premium sources like Wikipedia
(or IMDB, Goodreads etc. for specific domains).
They feature entity-specific pages and their
structure follows fairly rigid templates.
Therefore, extraction patterns can be 
specified with relatively little effort,
most notably, in the form of
{\bf regular expressions (regex)} over the text and
existing markup of the target pages.
The underlying assumption for the
viability of this approach is:

\begin{samepage}
\begin{mdframed}[backgroundcolor=blue!5,linewidth=0pt]
\squishlist
\item[ ] 
{\bf Consistent Patterns in Single Web Site:} \\
In a single web site, all (or most) pages about 
entities of the same type (e.g., musicians or writers)
exhibit the same 
patterns to express certain properties
(e.g., their albums or their books, respectively).
A limited amount of diversity and exceptions
needs to be accepted, though.
\squishend
\end{mdframed}
\end{samepage}

\begin{figure} [h!]
  \centering
   \includegraphics[width=0.8\textwidth]{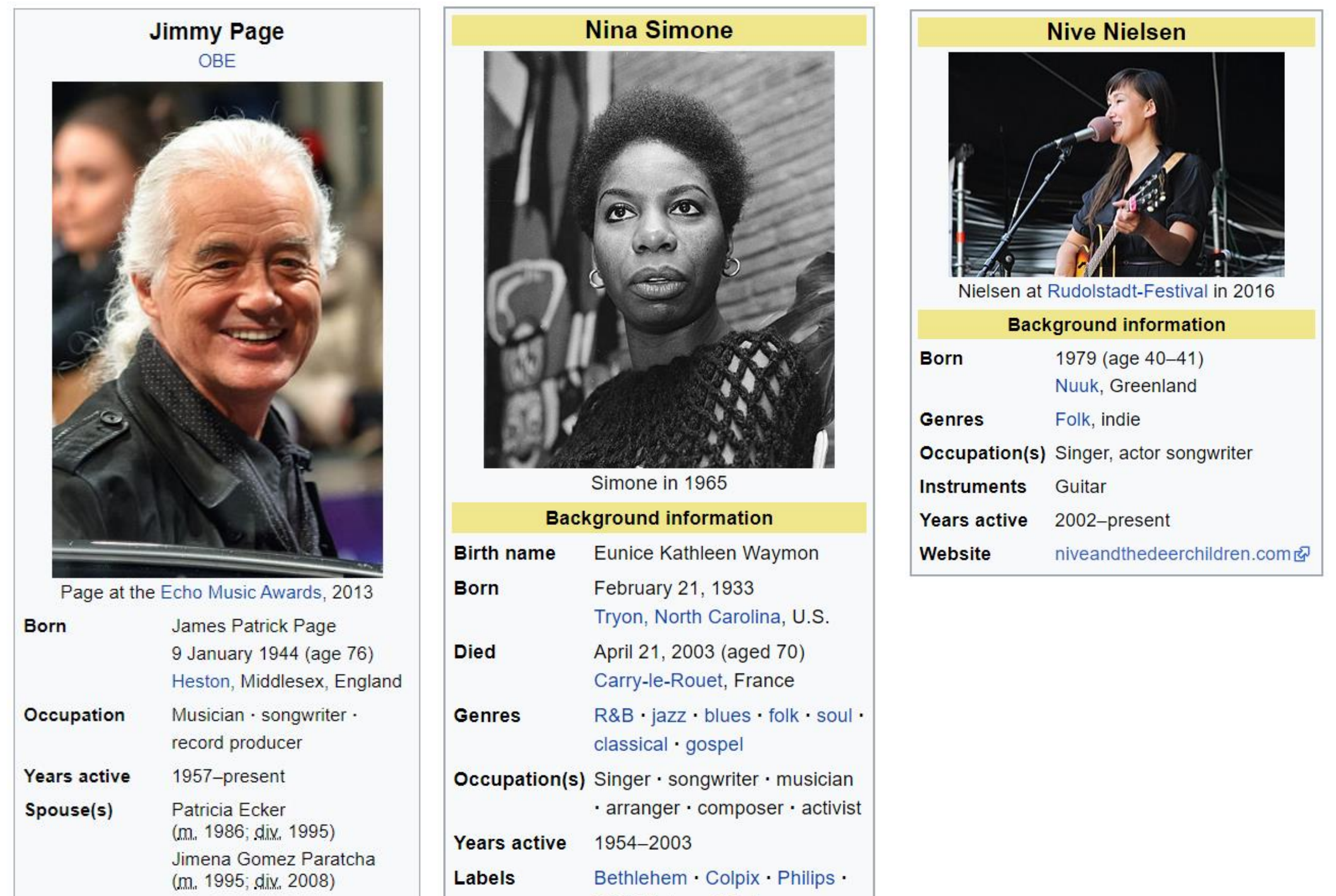}
      \caption{Examples of Wikipedia Infoboxes}
      \label{fig:wikipedia-infoboxes}
\end{figure}

\subsubsection{Regex Patterns}

Within Wikipedia, semi-structured elements like
infoboxes, categories, lists, headings, etc.
provide the best opportunity for harvesting facts by regular expressions.
Consider the infoboxes shown in Figure \ref{fig:wikipedia-infoboxes} for three
musicians (introducing new ones for a change,
to give us a break from Bob Dylan and Elvis Presley). %
Our goal is to extract, say, the 
dates and places of birth 
of these people, to populate the 
\ent{birthdate} attribute and
\ent{birthplace}
relation.
For these examples, the {\em Born} fields
provide this knowledge, with small variations,
though,
such as showing only the year for Nive Nielsen
or repeating the person name for Jimmy Page.
The following regular expressions specify
the proper extractions, coping with the variations.
For simplicity of explanation, we restrict ourselves
to the birth year and birth city.

\begin{lstlisting}
birth year X: Born .* (X = (1|2)[0-9]{3}) .* 
birth city Y:  Born .* ([0-9]{4}|")") (Y = [A-Z]([a-z])+) .*
\end{lstlisting}

In these expressions,
``.*'' denotes a wildcard for any token sequence,
``|'' and ``[...]'' denote disjunctions
and ranges of tokens,
``$\{$...$\}$'' and ``+'' are repetition factors,
and putting ``)'' itself in quotes is necessary
to distinguish this token from the parenthesis
symbol used to group sub-structures in a regex.
Note that the specific syntax for regex varies
among different pattern-matching tools. 

Intuitively, the regex for birth year finds
a subsequence $X$ that has exactly four digits
and starts with 1 or 2 (disregarding,
for simplicity, people who
were born more than 1020 years ago).
The regex for birth cities identifies the first
alphabetic string that starts with an upper-case letter
and follows a digit or closing parenthesis.

This is still not perfectly covering all possible
cases. For example, cities could be
multi-word noun phrases (e.g., New Orleans).
We do not show more complex expressions
for ease of explanation. It is straightforward
to extend the approach for both 
i) completeness,
like extracting the full date rather than merely
the year and the exact place, 
and 
ii) diversity of showing this in infoboxes.
On the latter aspect, the moderation of Wikipedia
has gone a long way towards standardizing 
infobox conventions by templates, but it could still
be (and earlier was the case) that 
some people's infoboxes show fields {\em birth place},
{\em place of birth}, {\em born in},
{\em birth city}, {\em country of birth}, etc.
Nevertheless, it is limited effort to manually specify
wide-coverage and robust regex patterns for
hundreds of attributes and relations.
The YAGO project, for example, did this for
about 100 properties in a few days of single-person
work \cite{yagojournal}.
Industrial knowledge bases harvest 
deterministic patterns 
from Web sites that are fed by back-end databases,
such that each entity page has the very same
structure (e.g., IMDB pages for the cast of movies).

\subsubsection{Inducing Regex Patterns from Examples}
To ease the specification of regex patterns,
methods have been developed that merely require
marking up examples of the desired output in
a small set of pages, sometimes even supported by
visual tools 
(e.g., \cite{DBLP:conf/vldb/SahuguetA99,DBLP:journals/kbs/FerraraMFB14,DBLP:conf/chi/HanafiACL17}).
For restricted kinds of patterns, it is then
possible to automatically learn the regex or,
equivalently, the underlying 
finite-state automaton, essentially inferring 
a regular grammar from examples of the language.
This methodology applied to pattern extraction
has become known as {\bf wrapper induction};
seminal works on this approach include
\citet{DBLP:conf/ijcai/KushmerickWD97,DBLP:journals/ml/Soderland99,DBLP:journals/ai/Kushmerick00,DBLP:journals/aamas/MusleaMK01,DBLP:conf/vldb/BaumgartnerFG01}.
The survey 
by
\citet{sarawagi2008information} 
covers
best-practice methods, with emphasis on CRF-based
learning.
Wrapper induction is a standard building block for
information extraction today.

\subsubsection{Type Checking}
Neither manually specified nor learned regex patterns
are perfect: there could always be an unanticipated
variation among the pages that are processed.
An extremely useful technique to prune out false positives
among the extracted results is
{\bf semantic type checking}, utilizing the
a-priori knowledge of type signatures for the
properties of interest (see Section \ref{sec:properties-rationale}).
If we expect \ent{birthplace} to have cities as
its range rather than countries or even continents,
we can test an extracted argument for this relation
against the specification.
The types themselves can be looked up in the
existing KB of entities and classes, after running
entity linking on the extracted argument.
This technique substantially improves the
precision of regex-based KB population
(see \citet{yagojournal} and \citet{DBLP:journals/ai/HoffartSBW13}. 
It equally applies to literal values of attributes
if there are pre-specified patterns, for example,
for dates, monetary values or quantities with units.

\subsubsection{Operator-based Extraction Plans}
\label{ch6-subsubsec:extractionplans}

Often, a number of patterns, rules, type-checking and
other steps have to be combined into an entire 
execution plan to accomplish some extraction task.
The underlying steps can be seen as operators in
an algebraic language.
The {\bf System T} project
(\citet{DBLP:conf/icde/ReissRKZV08} and \cite{DBLP:conf/acl/ChiticariuKLRRV10,DBLP:conf/naacl/ChiticariuDLRZ18}) 
has developed a declarative
language, called AQL (Annotation Query Language),
and a framework for orchestrating, optimizing
and executing algebraic plans combining such operators.
In addition to expressive kinds of pattern matching,
the framework includes operators for text spans
and for combining intermediate results.
A similar project, with a declarative language
called {\bf Xlog}, was pursued by
\citet{DBLP:conf/vldb/ShenDNR07} \cite{DBLP:conf/sigmod/ShenDMDR08},
and closely related research was carried out
by \cite{DBLP:journals/sigmod/BohannonMYADIJKMRS08}
and \cite{DBLP:journals/sigmod/JainIG08}.

To illustrate the notion and value of operator-based
execution plans, assume that we want to extract from
a large corpus of web pages statements about music bands
and their live concerts,
specifically, the relation between the involved
musicians and the instruments that they played. %
An example input could look as follows:

\vspace{0.2cm}
\begin{mdframed}[backgroundcolor=white!5,linewidth=0pt]
\squishlist
\item[ ] {\footnotesize\bf Led Zeppelin returns with rocking London reunion.}
\item[ ]
{\footnotesize\em
The quartet had a crowd of around 20,000 at London’s 02 Arena calling for more at the end of 16 tracks ranging from their most famous numbers to less familiar fare.
Lead singer Robert Plant, 59, strutted his way through “Good Times Bad Times” to kick off one of the most eagerly-anticipated concerts in recent years.
A grey-haired Jimmy Page, 63, reminded the world why he is considered one of the lead guitar greats, while John Paul Jones, 61, showed his versatility jumping from bass to keyboards.
Completing the quartet was Jason Bonham on drums.
}%
\squishend
\end{mdframed}
\vspace{0.1cm}

\noindent We aim to extract all triples of the form\\
\hspace*{1cm} $playsInstrument: musician \times instrument$,\\
\noindent namely, the five SO pairs\\
\hspace*{1cm} \ent{(Robert Plant, vocals)},
\ent{(John Paul Jones, bass)},\\
\hspace*{1cm} \ent{(John Paul Jones, keyboards)},
\ent{(Jimmy Page, guitar)},
\ent{(Jason Bonham, drums)}.\\
\noindent In addition, we want
to check that this
really refers to a live performance.
The extraction task entails several steps:

\squishlist
\item[1.] Detect all person names in the text.
\item[2.] Check that the mentions of people are indeed
musicians, by entity linking and type checking, or
accept them as out-of-KB entities.
\item[3.] Detect all mentions of musical instruments,
using the instances from the KB type \ent{music instruments},
including specializations such as 
\ent{Gibson Les Paul}
for \ent{electric guitar} and paraphrases from the KB dictionary of labels, such as
``singer'' for \ent{vocals}.
\item[4.] Check that the instrument mentions refer to 
specific mentions of musicians, for example, by 
considering only pairs of musician and instrument that
appear in the same sentence. Here the text proximity 
condition may have to be varied depending on the nature and
style of the text, for example, by testing for co-occurrences
within a text span of 30 words, or by first applying
co-reference resolution.\\
\noindent Sometimes, even deeper analyis is called for,
to handle difficult cases where subject and object
co-occur incidentally without standing in the
proper relation to each other.
\item[5.] Check that the entire text refers to a live performance.
For example, this could require checking for the 
occurrence of a date and specific kinds of location
like concert halls, theaters, performance arenas, 
music clubs and bars, or festivals.
\squishend

There are many ways for ordering the execution of these
steps, each of which involves sub-steps (e.g., for
matching performance locations),
or for running them in parallel or in a pipelined manner
(with intermediate results streamed between steps).
When applying such an entire operator ensemble to
a large collection of input pages, the choice of order
or parallel execution is crucial.
The reason is that different choices
incur largely different costs because they
materialize intermediate results of highly varying sizes.
The System T approach therefore views the entire
ensemble as a declarative task, and invokes
a query optimizer to pick the best execution plan.
This involves estimating the 
selectivities of
operators, that is, the fraction of pages that yield
intermediate results and the number of intermediate
candidates per page. 
For query optimization over text, this 
cost-estimation aspect is still
underexplored, notwithstanding results by
the System T project \cite{DBLP:conf/icde/ReissRKZV08} as well as
\cite{DBLP:conf/vldb/ShenDNR07}
and \cite{DBLP:journals/tods/IpeirotisAJG07}.

\subsubsection{Patterns for Text, Lists and Trees}
\label{ch6:subsubsec:patterns-text-lists-trees}

Extraction patterns can also be specified
for texts or lists as inputs.
This is often amazingly easy and can yield
high-quality outputs.
For example, many Wikipedia articles and
other kinds of online biographies contain
sentences such as
``Presley was born in Tupelo, Mississippi''.
By the stylistic conventions of such 
web sites, there is little variation in
these formulations. Therefore, a regex pattern
like \textit{P * born in * C} with \textit{P} bound
to a person entity and \textit{C} to a city
can robustly extract the birth places of
many people.
Analogously, a pattern like
\begin{lstlisting}
P .* (received|won) .* (award(s)?)? .* A
\end{lstlisting}
applied
to single sentences 
(with \ent{(...)?} denoting optional occurrences)
can extract outputs such as
\begin{samepage}
\begin{lstlisting}
(Bob Dylan, hasWon, Grammy Award)
(Bob Dylan, hasWon, Academy Award)
(Elvis Presley, hasWon, Grammy Award)
\end{lstlisting}
\end{samepage}
Obviously, there are many other ways of phrasing
statements about someone winning an award
(e.g., ``awarded with'', ``honored by'').
Manually specifying all of these would be 
a bottleneck; so we will discuss how to learn
patterns based on distant supervision in
Section \ref{sec:distantlysupervisedpatternlearning}.
Nevertheless, the effort of identifying
a few widely used patterns is modest and
goes a long way in ``picking low-hanging fruit''.

\begin{figure} [h!]
  \centering
   \includegraphics[width=1.0\textwidth]{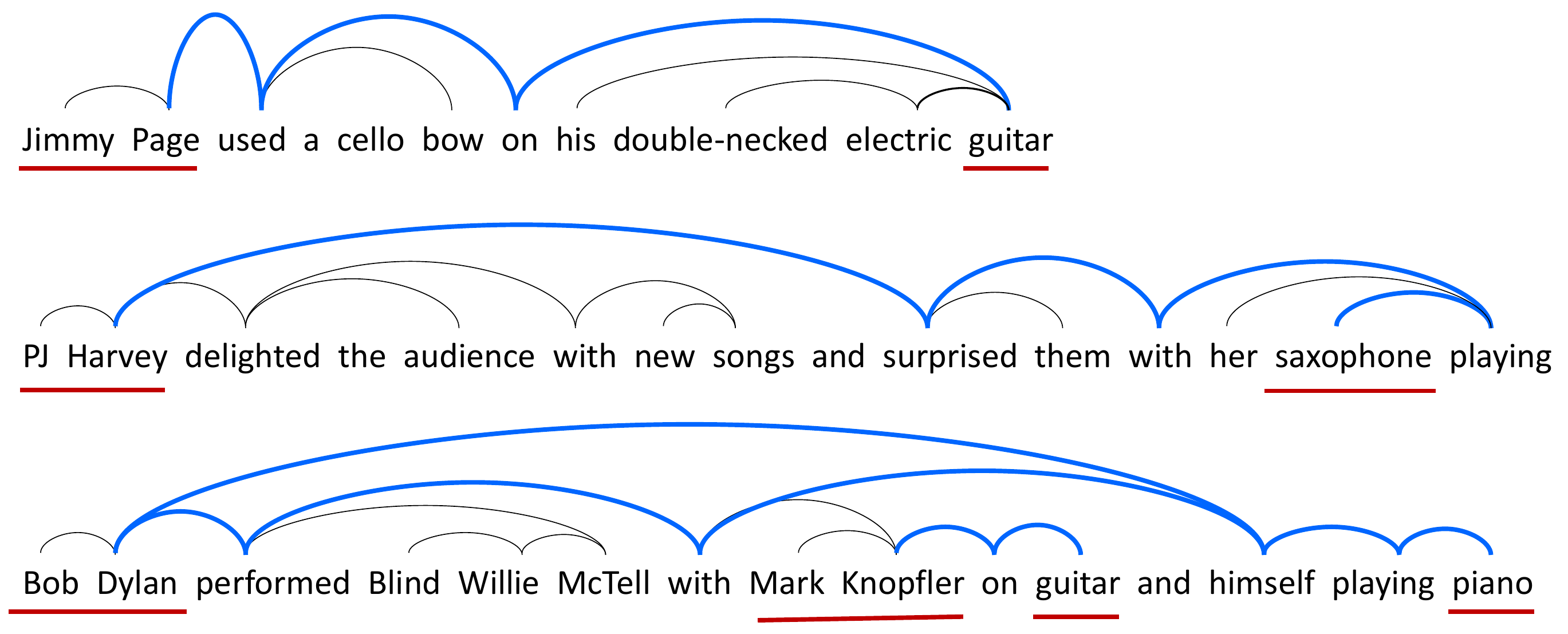}
      \caption{Examples for Relation Extraction based on Dependency Parsing Paths}
      \label{fig:ie-from-dp}
\end{figure}

Simple regex patterns with surface tokens
and wildcards, like the ones shown above,
often face a dilemma of either being too
specific or too liberal, thus sacrificing
either recall or precision.
For example, the pattern {\em P played his X}
with {\em P} and  {\em X} matching a musician
and a musical instrument, respectively,
can capture only a subset of male guitarists,
drummers, etc.
Moreover, it misses out on more elaborate
phrases such as ``Jimmy Page played his
bowed guitar''.
By employing NLP tools for first creating
word-level annotations like 
{\em lemmatization} and {\em POS tags}
(see Section \ref{sec:ch3-category-cleaning}),
more
expressive regex patterns can be specified,
for example
\begin{lstlisting}
P play $PPZ ($ADJ)? X 
\end{lstlisting}
where ``play'' is the lemma for ``plays'', ``played'' etc. and 
\ent{\$PPZ} and \ent{\$ADJ} are the tags
for possessive personal pronouns and adjectives,
respectively.
This pattern would also capture
a triple 
\ent{$\langle$PJ Harvey, playsInstrument, guitar$\rangle$}
from the sentence
``PJ Harvey plays her grungy guitar''.
Further generalization could consider
pre-processing sentences by 
{\em dependency parsing},
so as to capture arguments that are distant
in the token sequence of surface text but
close when understanding the grammatical structure.
Figure \ref{fig:ie-from-dp} shows 
examples where the parsing reveals short-distance
paths, highlighted in blue, between arguments for extracting instances of
\ent{playsInstrument}.
The third example may fail in practice,
as the path between musician and instrument
is not sufficiently short. However, by
additionally running 
{\em coreference resolution} (see Section \ref{subsec:el-designspace}),
the word ``himself'' can be linked back to
``Bob Dylan'', thus shortening the path.

Note, though, that the extra effort of dependency parsing
or coreference resolution
is worthwhile only for
sufficiently frequent patterns and 
properties that cannot be harvested by easier means.
Moreover, parsing may fail on inputs with ungrammatical sentences, like in social media.

Patterns are also frequent in headings of lists,
including Wikipedia categories. The English
edition of Wikipedia contains more than a million
{\bf categories and lists} with informative names
such as ``list of German scientists'',
``list of French philosophers'',
``Chinese businesswomen'' or
``astronauts by nationality''.
As discussed in Section \ref{sec:ch3-category-cleaning} we use
such cues for inferring semantic types,
but we can also exploit them for deriving
statements for specific properties like
\ent{hasProfession}, \ent{hasNationality}
or \ent{bornInCountry} and more.
Especially when harvesting Wikipedia,
judicious specification of patterns with
frequent occurrences yield high-quality outputs
at substantial scale.
This has been demonstrated by the {\em WikiNet} project
\cite{DBLP:journals/ai/NastaseS13}.

\begin{figure} [h!]
  \centering
   \includegraphics[width=1.0\textwidth]{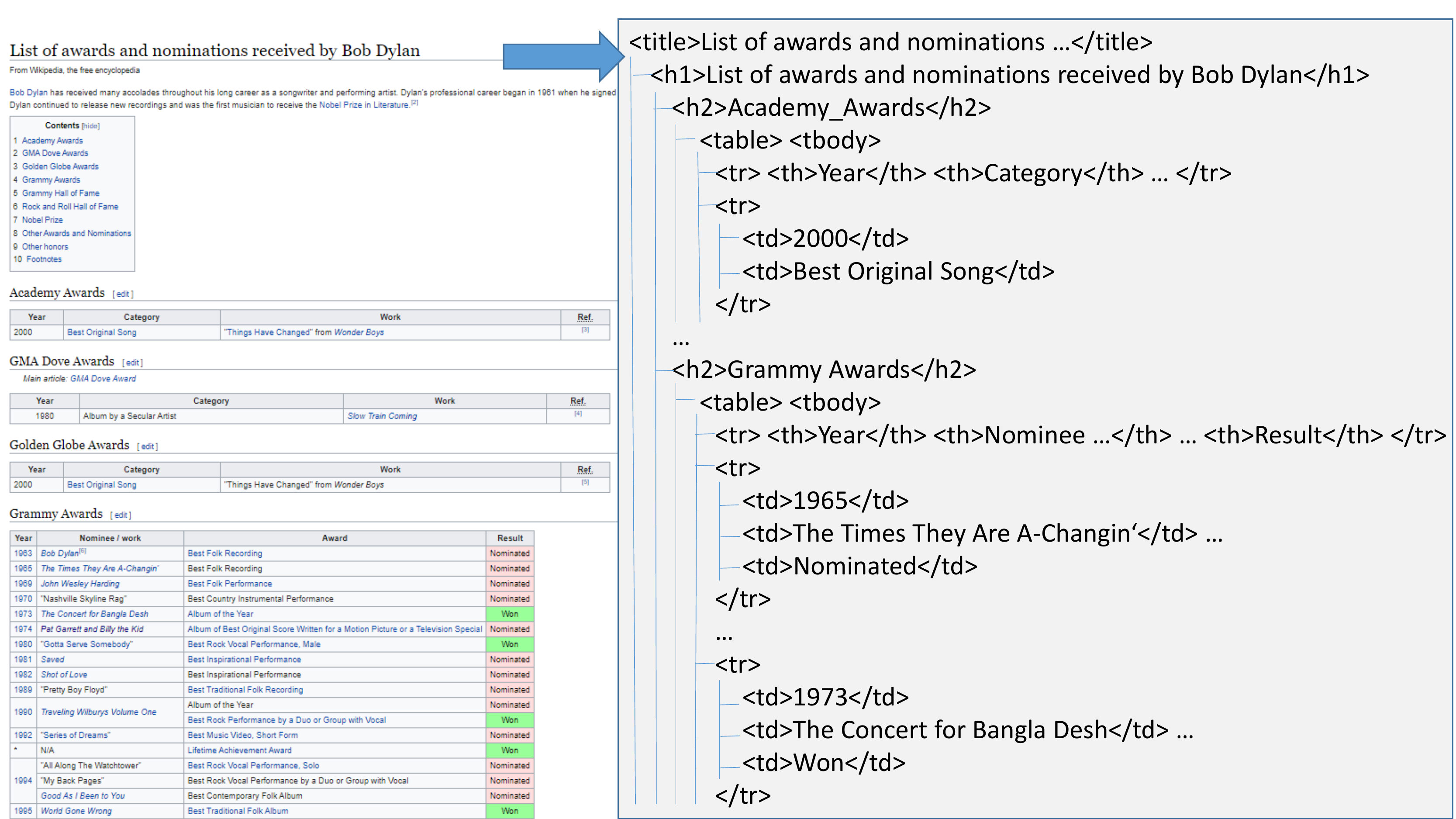}
      \caption{Awards List of Bob Dylan with Resulting DOM Tree}
      \label{fig:BobDylan-awards-DOMtree}
\end{figure}

Last but not least, patterns are also found in
trees, most notably, in the DOM-tree structure
of HTML pages (DOM = Document Object Model, a
W3C standard).
Figure \ref{fig:BobDylan-awards-DOMtree}
shows an excerpt of a web page on Bob Dylan's awards
(from Wikipedia),
and outlines the DOM tree for this page.
The tree has HTML tags like headings (h1, h2),
table rows (tr) and table cells (td) as inner nodes
and the rendered text on the leaf nodes.
This tree model applies also to 
texts with markups in XML/XPath or the Wiki markup
language (which is used by Wikipedia).
Extraction patterns can be specified via rules
over paths and positions of tags.
For example, to identify the fact that
Bob Dylan won the Academy Award in 2000
and a Grammy in 1973, the following 
patterns and rules can be applied:

\squishlist
\item[1.] locate node $\$A$ with path\\
$\phantom{x}~~~~~~ root \rightarrow h1 \rightarrow h2 \rightarrow \$A$\\
such that $\$A$ contains ``Awards''
\item[2.] locate node $\$Y$ with path\\
$\phantom{x}~~~~~~ \$A \rightarrow table \rightarrow tbody \rightarrow tr \rightarrow th[k] \rightarrow \$Y$\\
such that $\$Y$ contains ``Year'' and
$th[k]$ is the $k$-th occurrence of tag $th$
as a child of the $tr$ tag
\item[3.] locate node $\$YY$ with path\\
$\phantom{x}~~~~~~ \$A \rightarrow table \rightarrow tbody \rightarrow tr \rightarrow td[k] \rightarrow \$YY$\\
such that $td[k]$ is the $k$-th occurrence of tag $td$
as a child of the $tr$ tag, with $k$ being the
same as in step 2
\item[4.] locate node $\$R$ with path\\
$\phantom{x}~~~~~~ \$A \rightarrow table \rightarrow tbody \rightarrow tr \rightarrow th[l] \rightarrow \$R$\\
such that $\$R$ contains ``Result'' and
$th[l]$ is the $l$-th occurrence of tag $th$
as a child of the $tr$ tag
\item[5.] locate node $\$RR$ with path\\
$\phantom{x}~~~~~~ \$A \rightarrow table \rightarrow tbody \rightarrow tr \rightarrow td[l] \rightarrow \$RR$\\
such that $td[l]$ is the $l$-th occurrence of tag $td$
as a child of the $tr$ tag, with $l$ being the
same as in step 4
\item[6.] if no $\$R$ node found
then output $\$A$ and $\$YY$
\item[7.] if $\$R$ found and $\$RR$ contains ``Won''
then output $\$A$ and $\$YY$
\squishend

We give this procedurally flavored description
for ease of explanation. There are formal
languages and tools for expressing this
extraction workflow in concise expressions,
generalizing regex from strings to trees
\cite{sarawagi2008information}.
We will discuss advanced methods for
extracting properties from such trees
in Section \ref{ch6-sec:properties-from-semistructured}.

\subsection{Distantly Supervised Pattern Learning}
\label{sec:distantlysupervisedpatternlearning}

\subsubsection{Statement-Pattern Duality}
\label{ch6-subsec:statementpatternduality}

Manually specified patterns go a long way,
but are limited in scope, at least if we aim
for high recall. 
Therefore, analogously to Chapter \ref{ch3:entities}
on discovering entity-type pairs, 
we now discuss methods for automatically
learning patterns.
The key principle of {statement-pattern duality},
introduced in Section \ref{ch3-subsect-patterns},
applies to the machinery for attributes
and relations as well.
This was first formulated by
\citet{DBLP:conf/webdb/Brin98} and 
refined in different contexts by
\citet{DBLP:conf/dl/AgichteinG00},
\citet{DBLP:conf/acl/RavichandranH02}
and
\citet{Etzioni:ArtInt2005}.
We briefly recap this fundamental insight,
generalizing it to triples about properties now:

\begin{samepage}
\begin{mdframed}[backgroundcolor=blue!5,linewidth=0pt]
{\bf Statement-Pattern Duality}
\squishlist
\item[ ] When correct statements 
about \ent{P(S,O)} for property \ent{P}
frequently co-occur with textual pattern $p$
(in snippets that mention \ent{S} and \ent{O}), then $p$ is likely a good pattern for \ent{P}.\\
Conversely, when snippets with two arguments \ent{S} and \ent{O} also
contain a good
pattern $p$ for property \ent{P}, then the statement
\ent{P(S,O)} is
likely correct.
\squishend
\end{mdframed}
\end{samepage}

Following the same rationale as in
Section \ref{ch3-subsect-patterns}, 
this suggests an approach by
{\bf seed-based pattern learning
and statement extraction}
where we
start with a set $T$ of correct statements  $\{P(S_1,O_1)$, $P(S_2,O_2)$, $\dots\}$
for property \ent{P} (and optionally hand-crafted
patterns $P$), and then
iterate the following two steps:
\squishlist
\item[1.] {\bf Pattern Discovery:}\\
find occurrences of 
$(S_i,O_i)$ pairs from $T$ in a web corpus, and
identify new patterns $p_j$ that co-occur with
these pairs with high frequency (and other
statistical measures), and add $p_j$ to $P$;
\item[2.] {\bf Statement Expansion:}\\
find snippets that contain 
i) mentions of new $(S_k,O_k)$
pairs with proper types and 
ii) a good pattern in $P$,
and add new statements $\ent{P(S_k,O_k)}$
to $T$.
\squishend

Note that we assume that entity discovery and
canonicalization is performed on all input snippets,
using methods from Chapters \ref{ch3:entities} and
\ref{ch3-sec-EntityDisambiguation}. 
So we can leverage the pre-existing
KB of entities and types, and can check that
the spotted entity pairs have the right
type signature for the property of interest.
We greatly benefit from our overriding approach
of first acquiring and populating the
entity-type backbone before embarking on
 attributes and relationships.

Table \ref{table:ExampleSeedBasedPatternLearningAndPropertyExtraction} shows a toy example for
the target property\\
\hspace*{1cm} \ent{composed: musician $\times$ song},\\ \noindent using two Dylan songs as seeds.
We shorten entity names for ease of reading.

\begin{table}
\begin{center}
{\footnotesize
\begin{tabular}{|l|l|l|l|}\hline
Round & Seeds & Patterns & New Statements \\ \hline\hline
 1  & (Dylan, Blowin)  & $\$X$ wrote the song $\$Y$ &   (Cohen, Hallelujah) \\
   & (Dylan, Knockin)  & $\$X$ wrote * including $\$Y$ & (Lennon, Imagine) \\ \hline
2  &   & $\$X$'s masterpieces include $\$Y$ &  
(Morricone, Ecstasy) \\
  &   &  &  
(Poe, Tell-Tale Heart) \\
  &   & $\$Y$ performed by $\$X$ &  
(Hardy, Mon Amie) \\ \hline
3  &   & $\$X$ * cover version of $\$Y$ &  (Bono, Hallelujah) \\
   &   & $\$X$ * composer of $\$Y$ & (Beethoven, Elise) \\ \hline
 ... & ... & ... & ... \\ \hline
\end{tabular}
}%
\caption{Toy example for Seed-based Pattern Learning and Property Extraction}
\label{table:ExampleSeedBasedPatternLearningAndPropertyExtraction}
\end{center}
\end{table}

The example shows that the output contains
good as well as bad patterns, and some
mixed blessings such as ``masterpieces include''.
Some of the resulting false positives among
the extracted statements can be detected and
eliminated by type checking (e.g., Poe and his stories). 
Generally, aggressive pattern learning can
boost the recall of the KB, but comes with
high risk of degrading precision.
There are two major ways of mitigating these risks:
\squishlist
\item Corroborate candidates for new statements
based on their spottings in different sources,
using {\em statistical measures of confidence}.
\item Employ {\em consistency constraints} to
reason on the validity of candidate statements,
to prune false positives.
\squishend
We will elaborate on the second approach in the
chapter on {\em KB Curation},
specifically in Section \ref{ch8-sec:consistency-reasoning}.
Here we focus on confidence-driven 
weighting and pruning.

\subsubsection{Quality Measures}

The measures of {\bf support},
{\bf confidence} and {\bf diversity}
introduced in Section \ref{ch4:subsec:PatternLearning},
on pattern learning for entity-type pairs,
carry over to property extraction.
They solely refer to patterns and statements,
so that we can directly compute them for
each property of interest.
For easier reading, we give the definition of
confidence again, in generalized form:

\begin{mdframed}[backgroundcolor=blue!5,linewidth=0pt]
\squishlist
\item[ ] Given positive seeds $S_0$ and negative seeds $\overline{S}_0$,
the {\bf confidence of pattern $p$},
$\textit{conf}(p)$, is  
the ratio of positive occurrences to occurrences with either positive or negative seeds:
$$\textit{conf}(p) = 
\frac
{\sum_{x \in S_0} \textit{freq}(p,x)}
{\sum_{x \in S_0} \textit{freq}(p,x) ~+~ \sum_{x \in \overline{S}_0} \textit{freq}(p,x)}$$
\squishend
\end{mdframed}

\begin{samepage}
\begin{mdframed}[backgroundcolor=blue!5,linewidth=0pt]
\squishlist
\item[ ] The {\bf confidence of statement $x$} is
the normalized aggregate frequency of co-occurring with good patterns,
weighted by the confidence of these patterns:
$$\textit{conf}(x) = \frac{\sum_{p \in P} \textit{freq}(x,p) \cdot \textit{conf}(p)}{\sum_{q} \textit{freq}(x,q)}$$
where $\sum_q$ ranges over all  patterns. 
\squishend
\end{mdframed}
\end{samepage}

This way, we can score and rank both patterns
and statements, and then pick thresholds for
selecting the output depending on whether
we prioritize precision or recall for the KB
population. 
Moreover, we can integrate quality measures
into the iterative algorithm for learning
patterns and extracting statements,
along the same lines that we explained in
Section \ref{ch4:subsec:PatternLearning}.
The Extended Algorithm for 
Seed-based Pattern Learning
applies appropriate
weighting and pruning after each round,
for quality control.

Just like explicitly specified patterns 
discussed in 
Section \ref{subsec:specifiedpatterns},
learned patterns are not limited to
consecutive strings in surface text.
We already used wildcards in our examples,
and we can also learn patterns that have
POS tags as placeholders, or refer to
paths in dependency parsing trees
or
DOM trees 
(see, e.g., \cite{Etzioni:ArtInt2005,DBLP:conf/naacl/BunescuM05,DBLP:conf/kdd/SuchanekIW06}.

\subsubsection{Scale and Scope}

\vspace*{0.2cm}
\noindent{\bf Distant Supervision by Seeds from Prior KB:}\\
The outlined methodology for pattern
learning and statement harvesting relies on
seed statements. These can be manually
compiled, say 10 to 100 for each property of
interest. 
But there is a more intriguing and larger-scale
approach: {\em distantly supervised learning}
from an existing high-quality KB 
(see, e.g., \citet{DBLP:conf/acl/MintzBSJ09}
and
\citet{suchanek2009sofie}):
\squishlist
\item[1.] Build a high-quality KB by using conservative techniques with hand-crafted
patterns applied to premium sources that
have structurally and stylistically 
consistent pages. Give priority to precision,
disregard recall or consider it secondary. For example, harvesting Wikipedia infoboxes 
falls under this regime.
\item[2.] Treat all statements of this
prior KB as correct seeds for pattern learning
and statement expansion.
If needed, generate negative seeds (i.e.,
statements that do no hold) by re-combining
S and O arguments that violate hard constraints
(e.g., wrong birthplaces for people for whom
the KB already has the correct birthplace).
\squishend
Note that this method does not need any
labeling of patterns or features for training, hence the name {\em distant} supervision.

It is straightforward to scale the method to
very large inputs, as we can partition the input pages and process them in parallel
(in addition to handling the properties
independently).
When run over multiple rounds, there is
a need for handshakes regarding
the statistical weights, but this is easy
to implement.
This makes the method perfectly amenable to
Map-Reduce or other kinds of bulk-synchronous
distributed computation
(see, e.g., \cite{nakashole2011scalable}).

\vspace*{0.2cm}
\noindent{\bf Correlated Properties with Identical Type Signatures:}\\
When applying seed-based learning to 
multiple properties, a difficult case arises
whenever two properties have the same type
signature and their instances are correlated.
For example, \ent{bornIn} and \ent{diedIn}
are both of type \ent{person $\times$ city},
and 
many
people die in their birth place,
simply because they are born and spend their
lives in large cities.
In pattern learning, there is a high risk to 
confuse these two relations.

Another notorious example is:\\
\hspace*{1cm} \ent{locatedIn: city $\times$ country}
vs. \ent{capitalOf: city $\times$ country}.\\
\noindent If we learn for the latter with prominent entity pairs as seeds,
such as\\
\hspace*{1cm} \ent{(Paris,France),(Berlin,Germany),(Tokyo,Japan)}, \dots \\
\noindent we will acquire misleading patterns such as\\
\hspace*{1cm} $\$X ~*~ largest ~ city ~ of ~ \$Y$.\\
\noindent Even if the seed set contains some cases where
this does not hold, such as 
\ent{(Canberra,Australia)}, the majority of
seeds would still comply with the pattern.
This is where {\bf negative seeds} can play
a significant role. Explicitly stating
that \ent{(Sydney,Australia)} and
\ent{(Toronto,Canada)} are {\em not} in the \ent{capitalOf} relation
carries much higher weight and can steer
the pattern learner so as to discriminate
the two confusable relations.

\subsubsection{Feature-based Classifiers}
\label{ch6:subsubsec-featurebasedclassifiers}

Instead of learning patterns and then
applying patterns to extract statements,
we can alternatively use the features
that accompany seeds and patterns
to {\em train a classifier} that accepts or rejects
new candidate statements given their
features.
For training, we i) spot the S and O
arguments of seeds in pages 
and ii) observe features from local contexts
such as:
\squishlist
\item words, POS tags, n-grams, NER tags etc. to the left of S,
\item words, POS tags, n-grams, NER tags etc. to the right of O,
\item words, POS tags, n-grams, NER tags etc. between S and O
\item words, n-grams, tags etc. in the root-to-S
and root-to-O paths in DOM trees, tables, lists etc.,
\item and further features that can be observed within local proximity.
\squishend
When the classifier needs negative training samples, too, we use again the technique mentioned above:
generate incorrect statements by replacing S and O
in a correct statement with an alternative
argument that is known  (or at least
very likely) to be incorrect.

This approach to distantly supervised
property extraction has been pursued by
\cite{DBLP:conf/acl/MintzBSJ09}
and \cite{DBLP:conf/acl/HoffmannZW10},
with the latter training property-specific
CRF models for hundreds of properties,
with little effort regarding the compilation
of seeds.
Advances on the TAC competition on
{\em Knowledge Base Population (TAC-KBP)}
have refined and extended this line of
methods (see, e.g., \cite{DBLP:conf/tac/SurdeanuMTBCSM10,ji2010overview}).
A typical task for these approaches is to
populate a set of attributes and relations
for entities of type \ent{person} and
\ent{organization}, with properties typically
found in Wikipedia infoboxes 
(e.g., date of birth, country of birth,
city of birth, cause of death, religion, spouses, children etc. for \ent{person}) -- the sweet spot
for distant supervision.

Another line of feature-based learning
combines relation classification with the
detection of entity mentions, into a
{\em latent topic model} 
(e.g., \cite{yao2011structured}) 
or a 
{\bf collective CRF} or other kind of
probabilistic graphical model (aka. factor graph).
This direction for joint inference
was started by 
\citet{roth2007global},
\citet{DBLP:conf/pkdd/RiedelYM10}
and
\citet{DBLP:conf/emnlp/YaoRM10}, 
using the Freebase KB for distant supervision,
and 
\citet{DBLP:conf/kdd/ZhuNWZM06},
\citet{DBLP:conf/cikm/ZhengSWG09} and
\citet{DBLP:conf/www/ZhuNLZW09} 
for the
{\em EntityCube/Renlifang} project
\cite{Nie:IEEE2012}.
Further advances were made, for example,
by \cite{DBLP:conf/acl/LiJ14,DBLP:conf/www/RenWHQVJAH17,DBLP:conf/www/QuRZ018}.
These joint models
are elegant and powerful,
but so far, they have not been able
to demonstrate extraction accuracy at a level
that allows direct import of statements
into a high-quality knowledge base.

A key issue to consider here is that
stand-alone entity discovery and linking,
as presented in Chapters \ref{ch3:entities} and
\ref{ch3-sec-EntityDisambiguation},
have matured and become so good that
it makes sense to run these steps upfront before
embarking on the task of property extraction.
This is an engineering argument for
quality assurance, suggesting to decompose
the mission of KB construction into
modules that are easier to 
customize, optimize and debug to the
specific goals of the downstream application.

\clearpage\newpage
\noindent{\bf Further Techniques from NLP:}\\
When features are derived from dependency
parsing structures, rather than encoding
them as explicit features, a useful technique
is to integrate {\bf tree kernels}
into the classifier (e.g., using SVM).
This way, similarity comparisons between
parsing structures are carried out 
only on demand.
These techniques have been invented by
\citet{DBLP:conf/acl/CulottaS04,DBLP:conf/nips/BunescuM05,DBLP:conf/naacl/BunescuM05,DBLP:conf/acl/Moschitti04,DBLP:conf/ecml/Moschitti06}.
They have been further advanced for practical usage, for example,
by \cite{DBLP:conf/emnlp/ZhouZJZ07,DBLP:journals/ploscb/TikkTPHL10,liu2013approximate}.

An NLP task highly related to property extraction
is {\bf Semantic Role Labeling (SRL)} where 
a set of property types for an entity
is specified, called a {\em frame}
with {\em slots} to be filled \cite{DBLP:journals/coling/GildeaJ02,DBLP:series/synthesis/2010Palmer}.
Then, given a sentence or text passage, the goal is to infer
the property values for the entity.
A key difference to KB population is
that the input is a short text given upfront and we have the prior knowledge
that the sentence and the target frame are about one and only one context
(e.g., describing an event of a certain type with a fitting frame type).
This is in contrast to large-scale
knowledge harvesting which operates
on a diverse corpus and needs to consider
many targets.
State-of-the-art SRL methods used to be based
on deep syntactic analysis and
constraint-based reasoning; see, for example,
\cite{DBLP:journals/coling/PunyakanokRY08}.
More recently, supervised end-to-end learning
has been brought forward for SRL, using
LSTMs and other neural networks
\cite{DBLP:conf/acl/HeLLZ17a,DBLP:conf/acl/ZettlemoyerFHM18}.
These methods are powerful, but 
how to utilize them for
large-scale KB construction remains an open issue.

\subsubsection{Source Discovery}
\label{ch6:subsubsec:source-discovery}

Information extraction (IE) as a methodology
typically assumed that it would operate on
a given piece of web content. 
This mindset is exemplified in
popular competitions and benchmarks like
TAC (Text Analysis Conference), 
ACE (Automatic Content Extraction), 
SemEval (co-located with premier NLP conferences)
and Semantic Web Challenges.
In constructing large KBs, we have an additional
degree of freedom, though, by judiciously picking
the sources from which we want to extract properties.
While Wikipedia is a universally good choice,
KBs may need to tap additional sources
for in-depth knowledge, especially when covering
vertical domains such as health, food or music.
Several criteria for
{\bf source quality} are important:
\squishlist
\item {\bf Trustworthiness:} The source has
correct information about entities and their
properties of interest.
\item {\bf Coverage:} The source covers many entities and many relevant properties, for the same
type of entities (e.g., medical drugs or songs)
or 
vertical domain
(e.g., health or music).
\item {\bf Freshness:} The source has
(nearly) up-to-date information.
\item {\bf Tractability:} The structural and
stylistic conventions of the source are favorable
for automated extraction, featuring, for example, 
crisp wording and lists, as opposed to,
say, sophisticated essays.
\squishend

The repertoire for identifying good sources
is broad, and appropriate methods highly depend
on the topical domain and the nature of web content
to be tackled. Although it is tempting to think about
universal {\em machine reading} across the entire Web
\cite{DBLP:conf/aaai/EtzioniBC06}, viable engineering needs to 
prioritize carefully.
The following are some of the major choices:

\squishlist
\item {\bf Web directories:} Earlier, the Web offered
good directories, such as {\small\url{http://dmoz.org}}
and the original Yahoo! directory, making it easy 
to identify (some of) the best web sites for 
a category of interest. These directories are mostly
defunct or frozen now, but there are still some
domain-specific sites, such as 
{\small\url{http://wikivoyage.org}} 
on
travel destinations, 
and commercial sites on hotels, restaurants etc.
\item {\bf Topical portals:} For some vertical domains,
highly authoritative ``one-stop'' portals have
emerged. These are easy to find via
search engines and a bit of browsing.
Examples include 
{\small\url{http://mayoclinic.org}} for health,
{\small\url{http://imdb.com}} and
{\small\url{http://rottentomatoes.com}} about movies,
or {\small\url{http://secondhandsongs.com}}
for cover versions of songs (to give a highly
specialized case).
\item {\bf Focused crawling:} If we aim for
long-tail entities or properties that are rarely
featured in web sites, we may have to explore 
a larger fraction of the Web. The strategy
here is {\em focused} crawling \cite{DBLP:journals/cn/ChakrabartiBD99} where we start
with a few seed pages that are known to be good,
and then follow links based on a combination
of classifying sites as topically relevant and
scoring them as authoritative.
In addition to using hyperlinks, a
technique for discovering new sites is to
use informative phrases from the visited pages
as queries submitted to search engines.
Techniques for focused crawling are described in
\cite{DBLP:books/daglib/0012317,DBLP:journals/toit/MenczerPS04,DBLP:journals/sigkdd/SarawagiV04,DBLP:series/synthesis/2012Dragut,DBLP:conf/cidr/SizovTSWGBZ03,DBLP:conf/www/BarbosaF07a,DBLP:journals/www/VieiraBSFM16}.
\squishend

For each identified source of potential interest,
the quality measures outlined above should be
assessed, either via meta-sites or by sampling
some of the source's pages and analyzing them
\cite{Wang2019Midas}.

For {\em trustworthiness}, PageRank-style metrics
based on random walks have been investigated in
great detail (e.g., the Eigentrust method \cite{DBLP:conf/www/KamvarSG03}).
Today, however, traffic-based measures are more
informative, most notably, the Alexa rank of a site
({\small\url{https://www.alexa.com/}}).
Mentioning a site as an external source in a Wikipedia
page can also be seen as an endorsement of
authority.

{\em Freshness} can be judged by metadata or, more reliably,
by sampling pages (see, e.g., \cite{DBLP:conf/www/PandeyO05})
as to whether they contain
known statements of recent origin (e.g., the latest songs
of popular artists).

{\em Coverage} can only be estimated, without
already running the full extraction machinery on
the entire site. This is best done by
sampling, using seeds from the KB.
However, when going for long-tail entities and
infrequently mentioned properties, this may be
treacherous and infeasible. 
Estimating the coverage, or completeness, of
web sources is an open challenge.
We will discuss this further in Chapter \ref{chapter:KB-curation}, specifically
Section \ref{ch8-sec:quality-assessment}.

Finally, assessing the {\em tractability}
of a source is a tricky issue, posing major difficulties.
A pragmatic approach is to sample pages from a site,
run extractions, and then manually inspect and assess
the output (see also Chapter 8).

\section{Extraction from Semi-Structured Contents}
\label{ch6-sec:properties-from-semistructured}

We outlined the basic approach of
using path patterns in trees 
in Subsection \ref{ch6:subsubsec:patterns-text-lists-trees}.
This applies to DOM trees of HTML pages
as well as Web tables and lists. 
For deploying these techniques 
at Web scale,
we face several bottlenecks. Suppose we
tackle a single web site, such as {\small\url{imdb.com}}
on movies or {\small\url{secondhandsongs.com}} on
cover versions of songs. The following issues need
to be addressed.

\vspace*{0.2cm}
\noindent{\bf Richness of Pages:}\\ 
Individual pages can be very rich, if not verbose,
showing many different ``sub-pages'', and it is
all but easy to identify the components that
contain statements of interest for KB construction.
This task has been addressed by techniques for
page segmentation based on 
structural and visual-layout
features (e.g., \cite{DBLP:conf/sigir/CaiYWM04,DBLP:conf/kdd/ZhuZNWH07}).
For major portals such as imdb, hand-crafted 
rules can do this job.
For tapping long-tail sites, though,
feature-based learning is
the method of choice.

\begin{figure} [t!]
  \centering
   \includegraphics[width=1.0\textwidth]{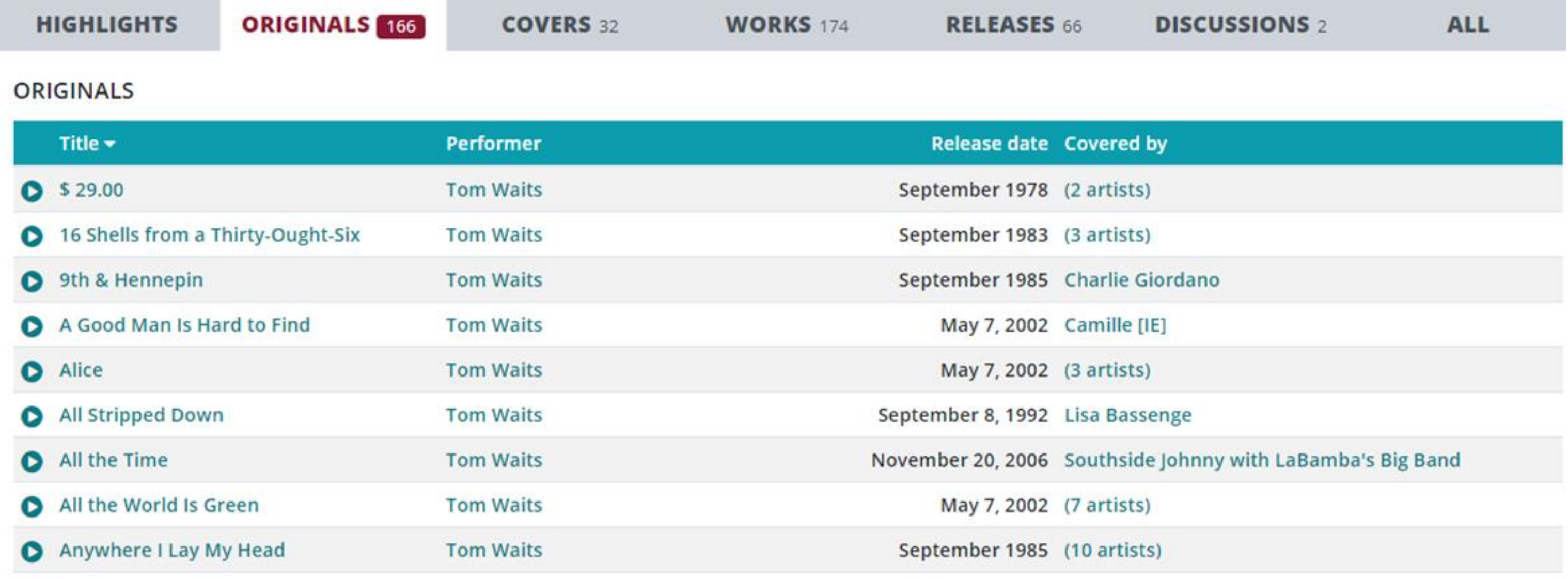}
   \vspace*{0.4cm}
   \par 
   
    \includegraphics[width=0.9\textwidth]{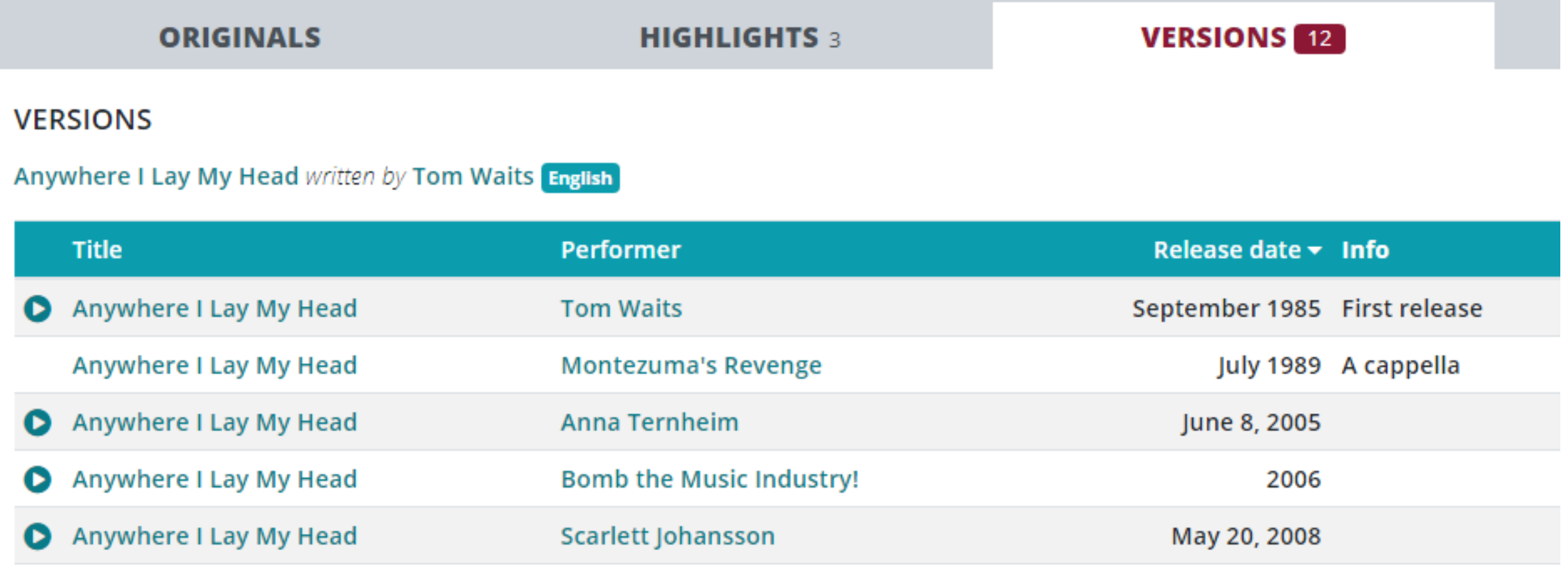}
   \vspace*{0.6cm}
      \par 
   
    \includegraphics[width=1.0\textwidth]{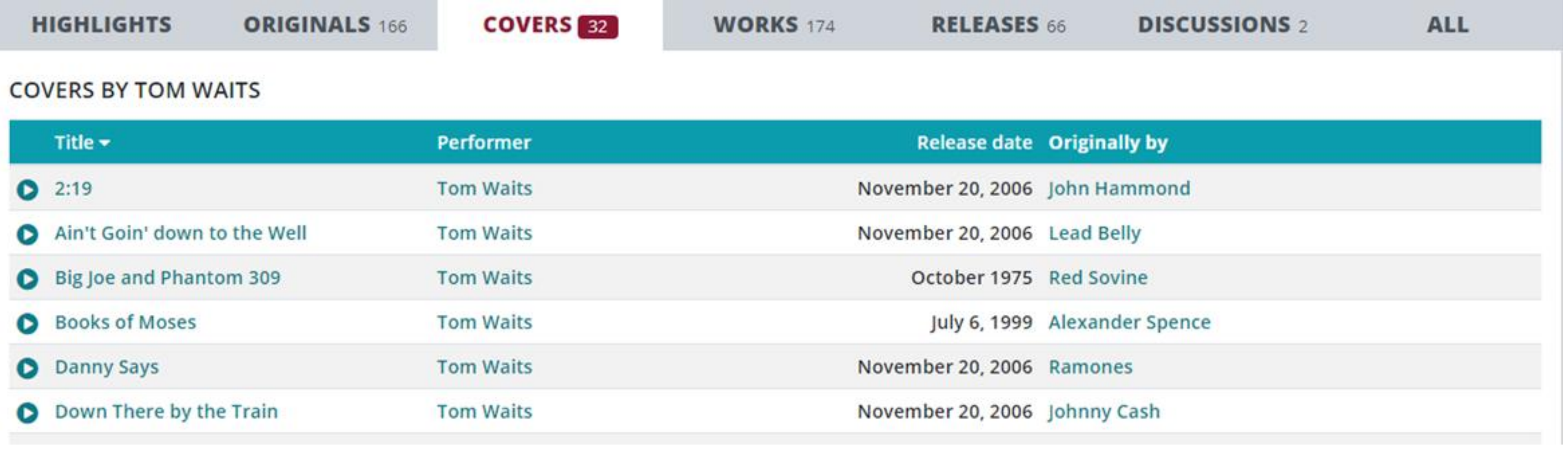}
      \caption{Examples for Three Templates on Original Songs and Cover Versions \\ (Source: \url{https://secondhandsongs.com/})}
      \label{fig:TomWaits-secondhandsongs}
\end{figure}

\vspace*{0.2cm}
\noindent{\bf Diversity across Pages:}\\
Even if all pages of the web site are generated
from a back-end database, they cover different pieces
and aspects of the site content. So there is no
single path pattern that can be applied to
all pages uniformly.
Figure \ref{fig:TomWaits-secondhandsongs} shows an example: three different views
of {\small\url{secondhandsongs.com}} featuring
i) original songs of an artist, 
ii) cover versions of this artist's songs by other artists,
iii) songs by other artists covered by the artist himself.
Each of the three is instantiated for all artists,
and we need three separate patterns to harvest the site.
The problem is that, upfront, without manual
inspection and analysis, we do not know
how many different cases we need to handle.
Therefore, techniques have been devised to
{\em discover templates} among the pages of the site,
based on clustering with tree-alignment and other
similarity models (e.g., \cite{DBLP:conf/kdd/ZhengSWW07,DBLP:conf/icde/GulhaneMMRRSSTT11}).

\vspace*{0.2cm}
\noindent{\bf Cost of Labeling Training Samples:}\\
The biggest potential showstopper is that 
each site and relevant template 
requires
at least one annotated page, so that patterns
and extractors can be properly induced.
These costs may even arise repeatedly for the same
site, namely, whenever the site owner decides to
change the content slicing and layout of its pages.

The cost concern has been mitigated, to some extent, by
\squishlist
\item visual tools for annotation (e.g., \cite{DBLP:journals/kbs/FerraraMFB14,DBLP:conf/chi/HanafiACL17})
and by 
\item hierarchical organization of patterns into
{\em libraries} (e.g., \cite{DBLP:conf/cikm/ZhengSWG09,DBLP:conf/icde/GulhaneMMRRSSTT11,DBLP:conf/sigir/HaoCPZ11})
for easy re-use and light-weight
adaptation to different templates or even across sites.
\squishend
Nevertheless, for harvesting the long tail of
smaller sites within a vertical domain, it would
be desirable to completely remove the dependency
on manual labeling.
Next, we describe a recent approach to this goal.

\subsection{Distantly Supervised Extraction from DOM Trees}
\label{ch6:subsubsec:extraction-from-dom-trees}

Like with distant supervision for extraction from text,
we start with seed pairs $(S,O)$ of entities for a given
property $P$ and make two key assumptions.
First, it is assumed that the web site under consideration
contains many {\em entity detail pages} where 
most information on the page refers to the same entity,
for example, the IMDB page on a movie
or the Goodreads page on
a book. 
Second, when processing a page about a seed entity $S$,
we assume that the property $P(S,O)$ occurs on that page
with high probability. In other words, a significant
fraction of the entity detail pages contain 
seed statements. This is key to the viability of
distant supervision.
We then learn extraction patterns from these
seed-matching pages, and apply them to all other
pages of the web site to acquire new statements.

This high-level approach needs to overcome
several difficulties, and this leads to the
following processing stages, following
the {\em Ceres} method by
\citet{DBLP:journals/pvldb/LockardDSE18}.

\begin{samepage}
\begin{mdframed}[backgroundcolor=blue!5,linewidth=0pt]
\squishlist
\item[ ] 
{\bf Distantly Supervised Method for Property Extraction from DOM Trees of Entity Detail Pages:}
\squishlist
\item[Input:] set of entity detail pages $x$, \\
seed statements $P(S,O)$, \\
other statements about $S$ from the KB
\item[Output:] set of objects $T$ such that $P(S,T)$ for some $S$ with page $x$\\
DOM-tree paths to $T$ for all these
objects $T$
\item[ ] For all pages $x$:
\squishlist
\item[1.] Locate subject entity $S$ in page $x$.
\item[2.] Locate the dominant path to $S$ across all pages.
\item[3.] Prune doubtful entities and pages.
\item[4.] Locate the path to $T$ for $P(S,T)$ in page $x$.
\item[5.] Compute the dominant path to $P(S,T)$ across all pages.
\squishend
\squishend
\squishend
\end{mdframed}
\end{samepage}

\vspace*{0.2cm}
\noindent{\bf Locating 
Subject
Entity in Page:}\\
There are easy cases where the URL, page title
or top-level headings give proper cues.
However, in cases like 
Figure \ref{fig:TomWaits-secondhandsongs},
it is not obvious where the central entity
is located for the target property 
\ent{covered: $Musician \times Musician$}.
In fact this varies across the three pages:
\ent{Tom Waits} in the {\em Performer} column
for {\em Originals},
\ent{Tom Waits} next to the text ``$\langle song \rangle$ written by'' below the heading
for {\em Versions}, 
and \ent{Tom Waits} in the header {\em Covers By Tom Waits}
for the third page.
Note that the third page can be interpreted as
a source for extracting the inverse relation
of \ent{covered} so that Tom Waits is still the
$S$ entity.

To identify the central entity on each page,
we first spot all 
$S$ entities
in the seed set for property $P$.
For each of these candidates $e$ 
(e.g., all musicians on the page) 
we look up its 
related
object set $Obj(e)$ in the KB 
(i.e., all $O$ entities that have some relation
with the $e$ musician), and compute {\em overlap} measures,
like Jaccard coefficients, of $Obj(e)$ and all
entities on the page.
The $e$ candidate with the highest overlap score is
considered the winner.
For example, for the first page of Figure
\ref{fig:TomWaits-secondhandsongs}, if half of
the shown songs are listed in the KB as
Tom Waits songs, this is high indication that
we have identified the Tom Waits detail page in
the web site.

\vspace*{0.2cm}
\noindent{\bf Dominant Path Across Pages:}\\
The previous step may still yield too many false positives.
By assuming consistent patterns across all
proper pages with entity detail, we can 
compute {\em global} evidence for the
preliminary choices.
Given a set of pairs $(x,e)$ with page $x$
being the identified candidate for the detail page
on entity $e$, we derive the DOM-tree paths for
each pair, resulting in a pool of path patterns.
The most frequent pattern is considered as 
the proper one. Optionally, we may relax the pattern a bit,
allowing for minor variations, or we can pick
multiple patterns based on a frequency threshold.

\vspace*{0.2cm}
\noindent{\bf Pruning Entities and Pages:}\\
Having identified the dominant path(s) 
allows us to eliminate all entities and pages
that do not exhibit a strong pattern.
Additional heuristics can be used to 
further prune doubtful pages.
For example, pages $x$ that contain only
very few 
entities related (by the KB) to
the candidate entity $e$
may be discarded, too.
Likewise, if the same entity $e$ is identified
as the central entity for a large number
of pages, this could be an erroneous case
(e.g., due to an entity linking error)
to be pruned.
The implementation for this 
prioritization of precision 
requires setting threshold parameters.

\vspace*{0.2cm}
\noindent{\bf Locating Paths to Objects:}\\
With the consolidated set of
proper $(x,e)$ pairs for seed subjects, 
we can locate the objects $O$ from
seeds $P(e,O)$ in page $x$.
This may lead to multiple matches of the same
object in the same page, though,
posing ambiguity as to whether some
object mentions may refer to entities
other than $e$ (even if $x$ is the true detail page for $e$)
or refer to properties other than $P$.
This is not the case in Figure \ref{fig:TomWaits-secondhandsongs},
but could easily happen with these kinds of pages.
Consider the third page about cover songs 
performed by Tom Waits.
If Waits had covered multiple songs
originally by Johnny Cash, we may consider
the object match \ent{Johnny Cash} as doubtful
(although it is correct in this case).
If the page were extended by further columns
about the {\em singer} of the original song --
which could be different from the song {\em writer} --,
we could accidentally match a seed object
in the singer column and thus pick a wrong path pattern.
In this example, such cases would be rare, as
song writers are often also the original singers,
but the risk of spurious paths can be
considerable in other settings.

To mitigate these risks, one can simply prune object
matches with multiple occurrences in the same page.
This sacrifices recall for the benefit of higher
precision.
An alternative is to check for additional evidence
that the located $o$ in page $x$ is indeed the
match for seed statement $P(e,o)$, using other
objects for the same $e$ and $P$ in the KB.
If the other objects (mostly) share the same
DOM-tree path prefix as $o$, then all these
siblings are likely in the same relation $P$.
For example, if the KB gives us a set of
song writers covered by Tom Waits, most of these
should be in the same column {\em originally by}
and only some may occur in other columns as well.
The path prefix to the most frequent column wins.

\vspace*{0.2cm}
\noindent{\bf Dominant Path Across All Pages:}\\
Finally, for given property $P$,
the best paths for every page can be compared
to strengthen the {\em global} evidence.
To this end, we define a similarity measure
between paths, like edit distance scores,
and all paths are clustered based on such a metric.
Only clusters with high density and clear
delineation from less focused clusters are
selected, to retain a subset of
high-quality DOM-tree path patterns.

\vspace*{0.3cm}
The final set of path patterns are utilized
by applying them to all identified entity detail pages
of the entire web site.
Comprehensive experiments by \cite{DBLP:journals/pvldb/LockardDSE18}
have shown that this distant-supervision
method is viable
at Web-scale and yields high-quality output,
while completely avoiding manual annotation of web pages.

\subsection{Web Tables and Microformats}
\label{ch6-subsec:webtables}

Web tables can be viewed as a special case of DOM trees.
Nevertheless they have received special attention
for knowledge extraction in the literature,
with recent surveys by
\citet{DBLP:journals/pvldb/CafarellaHLMYWW18} and
\citet{DBLP:journals/tist/ZhangB20},
and original works including
\cite{DBLP:journals/pvldb/LimayeSC10,
DBLP:conf/edbt/RitzeB17,DBLP:conf/edbt/OulabiB19,DBLP:conf/semweb/KruitBU19,VinhThinhHo:WWW2021}.
We already discussed methods for
entity linking (EL) and column typing
in Section \ref{sec:el-semistructured}.

By applying EL for tables, we identify 
$S$ entities in column $E$, and we hypothesize that 
another 
table column $C$ contains the entities (or
attribute values) for the $O$ argument
of a KB property $P$.
The key task then is to scrutinize whether
$C$ does indeed correspond to $P$.
The method devised by \cite{DBLP:conf/edbt/OulabiB19}
applies an ensemble of matchers to test this
hypothesis. This includes computing the
overlap between the already known $O$ entities
for $P$ in the KB against the entities mentioned
in column $C$. 
The column header, table caption and further context
serve as cues for other matchers.

By applying this technique to a large collection
of Web tables, a huge number of candidate
statements are collected. 
As the 
candidates
come from many different sites,
they 
can be 
consolidated
by similarity-based
clustering to distill the candidates into
the highest-evidence statements
(see \cite{DBLP:conf/edbt/OulabiB19} for details).

\clearpage\newpage
\noindent{\bf Higher-arity Relations:}\\
Tables naturally exhibit ternary and higher
n-ary relations.
As discussed in Chapter \ref{ch:foundations},
some of these cannot be decomposed without losing
information.
Recall the example that someone wins the Nobel Prize
in a certain field and year. This is a ternary
relation 
\ent{winsNobelPrize: $person \times field \times year$}
that cannot be properly reconstructed
by having only the binary projections
\ent{winsNobelPrizeInField: $person \times field$} and
\ent{winsNobelPrizeInYear: $person \times year$}.
A number of methods have been developed in the 
literature for extracting higher-arity relations
from both text and tables (e.g., \cite{DBLP:conf/semweb/KrauseLUX12,DBLP:journals/ws/KrauseH0WXUN16,DBLP:conf/www/ErnstSW18,DBLP:conf/wims/LehmbergB19,DBLP:conf/cikm/KruitBU20}).

One of the difficulties that goes beyond
the case of binary relations is that observations in
web contents are often partial:
many pages mention only the field of the Nobel Prize
winner while others mention only the year,
and only few pages if any have all arguments in
the same spot.
This calls for reasoning over sets of
extraction candidates. We will come back to this issue
in Chapter \ref{chapter:KB-curation},
specifically, Section 
\ref{ch8-sec:consistency-reasoning}.

\vspace*{0.2cm}
\noindent{\bf Microformat Annotations:}\\
An increasing number of Web pages contain
microformat markup embedded in HTML \cite{DBLP:conf/semweb/Bizer0MMSV13}, using
standards like RDFa, hcard or XML- or JSON-based vocabularities.
For specific properties, such as addresses of organizations, authors of publications or product data,
this can be a rich source, and there are specific
techniques to tap microformat contents
\cite{DBLP:journals/semweb/YuGFLRD19}.

\section{Neural Extraction}
\label{ch6-sec:neuralextraction}

\subsection{Classifiers and Taggers}

The most straightforward way of harnessing deep neural learning
for the extraction of properties is by viewing the task as a classification problem, or alternatively as a
sequence-tagging problem.

For the {\em classifier}, the input is a sentence or a short passage of text,
and the output is a binary decision on whether the sentence
contains an entity pair or entity-value pair
for a given property of interest. 
Often, the input already contains markup
for candidate entities, and these 
text spans are fed into the classifier as well.

For the {\em tagger}, the input is the sentence
or text snippet,
and the output is a sequence of tags that
identify the entity pair or entity-value pair
if the network decides to accept the input
as a positive case.
The tags are pretty much the same as for
neural NER, using BIO labels to identify the
begin and inner part of tagged sub-sequences
and O for irrelevant tokens; see Section \ref{subsec:deep-neural-networks}.
For example, for extracting the property
\ent{playsInstrument(Dylan,piano)}, the input

\lstset{language=HTML,upquote=true}
\begin{lstlisting}
Dylan performed Blind Willie McTell on this rare occasion
with Mark Knopfler on guitar and himself playing piano.
\end{lstlisting}

would be tagged
\begin{lstlisting}
B O O O O O O O O O O O O O O O O B
\end{lstlisting}

Both classifier and tagger methods are trained by
giving them sample sentences, positive as well as negative samples. The training is for a given
property type; each property calls for a separate learner.
In this regard, there is no difference from any
other -- simpler -- machine-learning method,
based, for example, on logistic regression or
random forests.
However, the neural learners do not take 
explicitly modeled features as input (e.g., dependency-parsing
tags) but instead expect an embedding vector for
each word (e.g., using word2vec), and nothing else.
Some methods further consider word positions
by encoding them into per-word vectors as well.

With these assumptions, the neural architectures that can be readily applied are the same as for NER,
discussed in Section \ref{subsec:deep-neural-networks}.
Most notably, recurrent neural networks (RNNs),
including {\bf LSTMs}, and 
convolutional neural networks (CNNs)
have been intensively studied
in the literature for this purpose of property extraction
from sentences.
Early work 
by
\citet{DBLP:conf/emnlp/SocherHMN12},
\citet{DBLP:conf/emnlp/HashimotoMTC13}
and
\citet{DBLP:conf/coling/ZengLLZZ14}
tackled the task of classifying coarse-grained
lexical relations between pairs of common nouns
or phrases (e.g., partOf, origin and destination of motion, etc.),
and encoded word embeddings and
positional information (i.e., relative
distances between words) to this end.
Later work integrated syntactic structures
(e.g., dependency parsing), lexical
knowledge such as hypernymy, and
entity descriptions
into their learned representations
(e.g., \cite{DBLP:conf/acl/LiuWLJZW15,DBLP:conf/emnlp/GormleyYD15,DBLP:conf/emnlp/XuMLCPJ15,DBLP:conf/acl/SantosXZ15,DBLP:conf/acl/ShwartzGD16,DBLP:conf/acl/LinSLLS16,DBLP:conf/aaai/Ji0H017,DBLP:conf/emnlp/ZhangZCAM17,DBLP:conf/emnlp/SorokinG17,DBLP:conf/lrec/Christodoulopoulos18,DBLP:conf/acl/YaoYLHLLLHZS19}).
Major advances and releases of large data resources for experimental research,
such as
{\em TACRED} and {\em DOCRED}, 
include works by 
\citet{DBLP:conf/acl/LinSLLS16},
\citet{DBLP:conf/emnlp/ZhangZCAM17},
\citet{DBLP:conf/acl/SoaresFLK19}
and
\citet{DBLP:conf/acl/YaoYLHLLLHZS19}.

More recent methods integrate
property extraction with NER
(e.g., \cite{DBLP:conf/acl/MiwaB16,DBLP:journals/corr/abs-1912-01070}),
and some additionally incorporate coreference
resolution into joint learning
(e.g., \cite{DBLP:conf/emnlp/LuanHOH18}).
The assumption that the input
is a single sentence has also been relaxed
in a few approaches that can cope with
arbitrary text passages as input sequence
(of limited length) (e.g., \cite{DBLP:journals/tacl/PengPQTY17,DBLP:conf/acl/YaoYLHLLLHZS19,DBLP:conf/acl/SahuCMA19}).
This way, properties can be detected where
subject and object occur in separate sentences
within close proximity.
On the other hand, 
entity linking (EL) is usually
disregarded. It is either assumed that the 
sub-sequences that denote entities in the input
are already canonicalized, or EL is postponed
and applied only to the entity mentions in the
positive outputs of
the neural extractor.
Only few works have attempted to integrate EL
into this kind of neural extractors
(e.g., \cite{DBLP:conf/acl/TrisedyaWQZ19}).

\subsection{Training by Distant Supervision}

Since labeled training data is, once again, 
the bottleneck for supervised deep learning,
training the outlined neural networks
is pursued by distant supervision.
To this end, statements from a prior KB
can be used to gather sentences that 
contain a subject-object pair for a given property. For example, Wikipedia is a rich
source of such sentences and even comes
with partial entity markup, in the form
of hyperlinks to entity articles or
entity mentions that match article titles
of prominent entities.
Other corpora, such as news collections,
could be considered, too. 

As an example, assume the KB contains the
pairs \ent{$\langle$Bob Dylan, guitar$\rangle$}
and \ent{$\langle$Bob Dylan, harmonica$\rangle$}
for the property \ent{playsInstrument}.
We may spot these pairs in the following sentences:
\begin{samepage}
\begin{lstlisting}
Bob Dylan sang and played acoustic guitar.
The audience cheered when Dylan took out his harmonica.
Dylan was accompanied by Knopfler on guitar.
The harmonica was played by Robertson, not Dylan.
\end{lstlisting}
\end{samepage}

For \ent{playsInstrument}, only the first two of these four sentences are positive samples, and
the other two would be misleading.
However, we do not know which are the good ones upfront.
This is a case for 
{\bf multi-instance learning} with 
{\em uncertain labels}.
The classifier has to cope with this noise
in its training data, 
and this involves learning to (implicitly) distinguish
proper positive samples (the first two in the example) from positive samples with spurious labels
(the latter two).

The simplest technique for multi-instance learning for property extraction is
to learn the best sentence from a given set,
as part of the classifier training.
However, this could overly focus on
a single sentence, underutilizing the
potential of the entire set. 
Therefore, a better approach is to learn
weights that capture the relative
influence of individual sentences towards the learned model.
This is known as a {\bf sentence-level
attention mechanism} in the literature
 \cite{DBLP:conf/emnlp/ZengLC015,DBLP:conf/acl/LinSLLS16,DBLP:conf/aaai/Ji0H017,DBLP:conf/aaai/YuanLTZZPWR19}, a 
 component in CNNs, LSTMs and other
 networks for selecting or prioritizing 
 input regions.
 The classifier could thus be enabled
 to properly distinguish between 
 newly seen sentences such as
 
{
\lstset{language=HTML,upquote=true}
\begin{lstlisting}
Bob Dylan performed the song with a tight rhythm on his piano.
Bob Dylan's accompanying band included an alto sax.
\end{lstlisting}
}

\noindent where only the former leads to a correct extraction.

Another difficulty that distantly supervised learners need to handle is that 
the same subject-object pair may appear
in more than one relation.
For example, the pair 
\ent{$\langle$Mark Knopfler, Bob Dylan$\rangle$}
could be a training sample for all three of the properties
\ent{accompanied}, \ent{has role model}
and \ent{friend of}. 
Moreover, several properties could even
co-occur in the same sentence, such as:

{
\lstset{language=HTML,upquote=true}
\begin{lstlisting}
Knopfler played guitar with Bob Dylan, his admired role model.
Robertson and his late friend Manuel were on Dylan's band.
\end{lstlisting}
}

In principle, state-of-the-art neural learners
for property extraction are geared for 
all these difficulties.
Moreover, they can be combined with other techniques like CRF-based
inference to enforce consistent outputs (e.g., \cite{DBLP:conf/kdd/ZhengMD018}).
However, it is not yet clear how
robust these methods perform in
the presence of limited supervision
and complex inputs.
Recent research therefore explores
also leveraging BERT embeddings 
(e.g., \cite{DBLP:conf/acl/SoaresFLK19}),
{\em reinforcement learning} and
other methods that relax reliance
on training data (e.g., \cite{DBLP:conf/emnlp/WuBR17,DBLP:journals/access/SunZJH19a,DBLP:conf/aaai/TakanobuZLH19}).

{\color{purple}
\citet{DBLP:journals/corr/abs-2004-03186}
}%
review
state-of-the-art methods and 
outlines challenges and opportunities
in this line of neural extraction.
Key issues identified for future research
include:
\squishlist
\item utilizing more data and background knowledge for better de-noising of training samples with distant supervision;
\item improving the scope and scale of neural learners, to cope with longer inputs and
for faster training;
\item robustness to coping with complex inputs
in sophisticated contexts (e.g., conversations
and narrative texts, or with cues 
spread across wider distances);
\item transfer learning to handle new property
types and
discover properties
never seen before. We will elaborate on this theme
of Open Information Extraction 
in Chapter \ref{ch7:open-schema-construction}.
\squishend

\subsection{Transformer Networks}

Instead of viewing property extraction as a classification or tagging task, a different paradigm is to
approach it as a {\bf machine translation} problem.
The input, or source language, is English sentences,
and the output, or target language, is the 
formal language of SPO triples.
This perspective has been pursued in recent
literature, mostly building on
neural learning with general 
{\bf encoder-decoder} architectures (which subsume LSTMs and CNNs,
although this is often not made explicit).
The encoder learns a latent representation of
the source language, and the decoder generates
output from this learned representation into
the target language.

Among the best performing and most popular
models for this setting
are {\bf Transformer networks}
\cite{DBLP:conf/nips/VaswaniSPUJGKP17}
(see also \cite{AnnotatedTransformer,IllustratedTransformer} for coding and illustrations).
In a nutshell, a Transformer consists
of a stack of encoders and, on top of this,
a stack of decoders. Each of these components
comprises a {\bf self-attention} layer
and a feedforward network. 
The self-attention mechanism allows the
model to inform the latent
representation of a word (as learned by the encoder)
with signals about all other words in the
input sequence. 
This way, Transformers capture
``cross-talk'' between words more directly
than LSTMs where latent states accumulate
distant-word influence.
Moreover, Transformer networks are
designed such that both training and
application can be highly parallelized.
The language embedding model BERT
\cite{DBLP:conf/naacl/DevlinCLT19} is an example of a very powerful
and popular Transformer network.\\

Recent works that leverage Transformers
(including BERT) for neural extraction
of properties include
\cite{DBLP:conf/naacl/VergaSM18,DBLP:conf/acl/AltHH19,DBLP:conf/acl/WangTYCWXGP19,DBLP:conf/cikm/WuH19a,DBLP:journals/corr/abs-1911-12753}.
Another recent trend is to leverage
neural methods for 
{\em Machine Reading Comprehension (MRC)}
\cite{DBLP:conf/emnlp/RajpurkarZLL16}
to extract relational pairs of entity mentions
from a given text passage.
One way of achieving this is by
casting the target property into 
one or more questions
(e.g., ``Who performed the song \dots?'')
to be answered by the MRC model
(e.g., \cite{DBLP:conf/conll/LevySCZ17,DBLP:conf/acl/LiYSLYCZL19,DBLP:conf/iclr/DasMYTM19}).
These techniques build on the asset that
their underlying neural networks have been
trained with huge amounts of texts 
about many topics of our world.

\section{Take-Home Lessons}

Major observations and recommendations from
this chapter are the following:
\squishlist
\item Tapping into {\em premium sources} 
(ideally, with
semi-structured cues like DOM trees, lists and tables) and using 
{\em highly accurate specified patterns} is the best
for high-precision extraction, to keep the
KB at near-human quality.
For large-scale extraction, this approach
has benefits also regarding
declarative programming and optimization
as well as scrutable quality assurance.
\item For recall, and especially to acquire
properties of {\em long-tail} entities or 
infrequently mentioned properties, 
pattern learning is a viable option.
Still, harvesting high-quality web sites
with {\em semi-structured content} by
{\em distantly supervised extractors}, is the
practically most viable way.
For less prominent vertical domains,
discovering these sources may be a challenge
by itself.
\item Extractors with {\em deep neural learning}
have been greatly advanced, based on
distant supervision from prior KB statements.
This allows tapping textual contents on 
a broader scale, with great potential for
higher recall at acceptable precision.
However, the reliance on sufficiently large
and clean training data (for distant supervision)
is a potential obstacle,
and needs further research.
\squishend

\clearpage\newpage
\chapter{Open Schema Construction}
\label{ch7:open-schema-construction}

\section{Problems and Design Space}

In Chapter \ref{ch4:properties}, we assumed that the property types of interest are explicitly specified.
For example, for entities of type \ent{musician}, we would gather triples for attributes and relationships like
{
\lstset{language=HTML,upquote=true}
\begin{lstlisting}
bornOn, bornIn, spouses, children, citizenOf, wonAward,
creatorOf (songs), released (albums), musicalGenre,
contractWith (record label), playsInstrument
\end{lstlisting}
}

This approach goes a long way towards building densely populated and expressive knowledge bases.
However, it is bound to be incomplete by missing out on the {\em ``unknown unknowns''}: properties beyond
the specified ones that are of potential interest but cannot be captured as we do not know about them yet.
For example, 
suppose we want to expand the population of a KB about musicians such as Bob Dylan. 
Which properties should this KB cover, in addition to the ones listed above?
Here are some candidates that quickly come to mind:

\begin{samepage}
{
\lstset{language=HTML,upquote=true}
\begin{lstlisting}
performedAt (location), performedAtAct (event),
coveredArtist (other musician), coveredByArtist,
workedWith (producer), performedWith (band or musician), 
accompaniedBy (instrumentalist) duetWith (singer),
composed (song), wroteLyrics (for song), 
songAbout (a person), etc. etc.
\end{lstlisting}
}
\end{samepage}

\noindent For the Bob Dylan example, noteworthy instances of these properties would include
(with \ent{Bob Dylan} abbreviated as \ent{BD}):

{\small
\lstset{language=HTML,upquote=true}
\begin{lstlisting}
< BD, performedAt, Gaslight Cafe >, 
< BD, performedAtEvent, Live Aid Concert >,
< BD, coveredArtist, Johnny Cash >, 
< BD, coveredByArtist, Adele >,
< BD, coveredByArtist, Jimi Hendrix >, 
< BD, performedWith, Grateful Dead >, 
< BD, duetWith, Patti Smith >,
< Sad Eyed Lady of the Lowlands, songAbout, Sara Lownds >,
< Hurricane, songAbout, Rubin Carter >, and many more.
\end{lstlisting}
}

\vspace*{0.2cm}
\noindent{\bf Limitations of Hand-Crafted Schemas:}\\
Although it is conceivable that a good team of knowledge engineers come up with all these property types,
there are limitations in manual schema design \cite{DoanHalevyIves2012} or ontology engineering \cite{StaabStuder2009} 
To underline this point, consider the following examples.
Schema.org ({\small \url{http://schema.org}}) is an
industry standard for microformat data within web pages \cite{Guha:CACM2016}. 
As of April 2020, 
it comprises
ca. 800 entity types with a total of ca. 1300 property types.
This is highly incomplete. For example, 
the type \ent{CollegeOrUniversity}, despite having 67 specified properties, misses out on \ent{numberOfStudents} or \ent{degreesOffered}. %
Wikidata ({\small\url{http://wikidata.org}}) \cite{Vrandecic:CACM2014}
builds on collaborative KB building; so one would expect better coverage from its large online community
(further discussed in Section~\ref{subsec:wikidata}).  %
As of April 2020, it specifies about 7000 property types, 
but still lacks many of the interesting ones for musicians and songs, such as 
\ent{performedAt}, \ent{coveredArtist}, \ent{duetWith}, \ent{songAbout} etc. 
Finally, for knowledge about movies, even the most authoritative IMDB website ({\small\url{http://imdb.com}})
misses many of the attributes and relations found across a variety of semi-structured sites:
a study by \cite{DBLP:conf/naacl/LockardSD19} manually identified interesting properties about movies
and found that IMDB comprises only about 10\% of the ones jointly covered by eight domain-specific websites.

\vspace*{0.2cm}
\noindent{\bf Key Issues:}\\
The following sections address three key points about handling these ``unknown unknowns'':
\squishlist
    \item Given solely a corpus of documents or web pages, automatically {\em discover all predicates of interest},
using {\em Open Information Extraction}.
\item Given a domain of interest, such as music or health, and an existing KB with a subset of specified properties,
{\em discover new attributes and relationships} for existing entity types, using {\em distant supervision}.
\item Given a set of discovered properties, with noise and redundancy, organize them into a clean system
of {\em canonicalized attributes and relations} with clean type signatures, by means of 
clustering, matrix/tensor
factorization and other data mining algorithms.
\squishend
Note that once we have identified a new property of interest and have gathered at least a few of its subject-object pairs as instances,
the task of further populating the property falls into the regime of Chapter \ref{ch4:properties},
most importantly, using seed-based distant supervision.

\section{Open Information Extraction for Predicate Discovery}
\label{ch7-sec:openIE}

To discover property types in text, without any
assumptions about prior background knowledge,
the natural and best approach is to exploit
{\bf syntactic patterns} in natural language.
More precisely, we aim at 
{\em universal patterns}, or 
{\em ``hyper-patterns''}, 
that capture relations and
attributes of entities regardless of their types.
The simplest pattern is obviously the 
basic and ubiquitous grammatical structure of
the English language: 
{\em noun -- verb -- noun}, where nouns, or more
generally, noun phrases, should denote entities in their roles as subject and object of a sentence.
A simple example is ``Bob Dylan sings Hurricane''
with the verb ``sings'' being the newly detected
property. The verbal part could also be a phrase,
often of the form {\em verb $+$ preposition},
like in sentences such as 
``Bob Dylan sings with Joan Baez'' or
``Bob Dylan performed at the Live Aid concert''.
The output of this approach is not just the
property itself, but an entire
{\bf predicate-argument structure}.
The verbal phrase becomes a 
(candidate for a)
binary predicate
in a logical sense, and its arguments are
the subject and object extracted from the sentence.
In full generality, there could be more than
two arguments, because of adverbial modifiers
(e.g., ``Dylan performed Hurricane at the Rolling Thunder Revue'')
or verbs that require two objects
(to express higher-arity predicates).

\begin{mdframed}[backgroundcolor=blue!5,linewidth=0pt]
\squishlist
\item[ ] \textbf{Open Information Extraction (Open IE)}\\ 
is the task of extracting predicate-argument structures from natural language sentences.\\
{\em Input:} a propositional sentence
(i.e., not a question or mere exclamation)
that can be processed with generic methods
for syntactic analysis (incl. POS tagging,
chunking, dependency parsing, etc.).\\
{\em Output:} a predicate-argument tuple,
as a proto-statement, where the predicate
and two or more arguments take the form
of short phrases extracted from input sentence.
\squishend
\end{mdframed}

To further illustrate the task, consider the sentences:
\begin{samepage}
{\small
\lstset{language=HTML,upquote=true}
\begin{lstlisting}
Bob Dylan sang Hurricane accompanied by guitar and violin.
In this concert Dylan sings together with Baez.
Bob Dylan performed a duet with Joan Baez.
Dylan performed the song with Joan Baez.
Bobby's and Joan's voices beautifully blend together.
Hurricane is about Rubin Carter and criticizes racism.
\end{lstlisting}
}
\end{samepage}
State-of-the-art Open IE 
methods produce the following proto-statements
as output
(e.g., 
{\small \url{https://nlp.stanford.edu/software/openie.html} 
or \url{https://demo.allennlp.org/open-information-extraction/}}):

\begin{samepage}
{\small
\lstset{language=HTML,upquote=true}
\begin{lstlisting}
< Bob Dylan, sang, Hurricane >
< Dylan, sings, together with Baez >
< Bob Dylan, performed, a duet >
< Dylan, performed, the song >
< Bobby's and Joan's voices, blend, [null] >
< Hurricane, is about, Rubin Carter >
< Hurricane, criticizes, racism >
\end{lstlisting}
}
\end{samepage}

A few remarks are in order:
\squishlist
\item The S and O arguments are surface mentions
in exactly the same form as they appear in the input sentence. There is no canonicalization yet,
but an Entity Linking (EL) steps could be
added for post-processing.
Alternatively, one could first run entity recognition
and EL on the input sentences as pre-processing.
\item Some of the output triples appear incomplete and uninformative, missing crucial information.
Open IE methods tend to add this as additional
{\em modifier arguments}. For example, the
sentence about the duet with Joan Baez would
return:\\
predicate = ``performed'', 
arg1 = ``Bob Dylan'', 
arg2 = ``a duet'', modifier = ``with Joan Baez''.\\
So the crucial part is not lost, but it is not that easy to properly recombine arguments and modifiers.
For example, for the first sentence, predicate, arg1 and arg2 are perfect, and the modifier ``accompanied by guitar and violin'' is indeed auxiliary information that could be disregarded for
the purpose of discovering the predicate
\ent{sings} between singers and songs.
\item Some sentences do not easily convey the subject-object arguments, an example being
``Bobby's and Joan's voices'' collapsed into
the subject argument. As a result, the output
for this sentence has no object, denoted by
\ent{[null]} (see, e.g., \cite{DBLP:conf/coling/SahaM18}
for handling this).
\item Some sentences, like the last one, contain information that is properly captured by multiple triples.
This is a typically case for sentences with
conjunctions, but there are other kinds of
complex sentences as well.
\squishend

\subsection{Pattern-based Open IE}
\label{ch7-subsec:OpenIE-patterns}

Until recently, most methods for Open IE
crucially relied on patterns and rules
\cite{DBLP:conf/ijcai/BankoCSBE07,DBLP:conf/acl/WuW10,fader2011identifying,DBLP:conf/emnlp/MausamSSBE12,DBLP:conf/ijcai/EtzioniFCSM11,DBLP:conf/www/CorroG13,DBLP:conf/ijcai/Mausam16} 
(with a minor role of learning-based
components).
Recent methods that build on neural learning
will be discussed in Subsection \ref{ch7-subsec:OpenIE-neural}.
As an exemplary representative for pattern-based
Open IE, we discuss the
{\bf ReVerb} method 
by \citet{fader2011identifying},
which focuses on verb-mediated predicates
and builds on a single but powerful
{\bf regular expression pattern} over POS tags,
to recognize predicate candidates:

\begin{mdframed}[backgroundcolor=blue!5,linewidth=0pt]
\squishlist
\item[ ] {\bf Regex Pattern for Open IE:}
\squishlist
\item[ ] predicate = V | VP | VW*P
\item[ ] V = verb particle? adverb?
\item[ ] W = (noun | adjective | adverb | pronoun | determiner)
\item[ ] P = (preposition | particle | infinitive marker)
\squishend
\item[ ] where determiners are words like
``the'', ``a'' etc., particles subsume
conjunctions and auxiliary verbs, and
infinitive markers capture the word ``to''
in constructs such as ``Bob and Joan reunited 
\underline{to} perform Farewell Angelina''.
\squishend
\end{mdframed}
  
The above expression matches, for instance, %
  \textit{``sang''} (V), \textit{``sang with''} (VP), or \textit{``performed a duet with''} (VWP).
  
This way ReVerb gathers a pool of predicate candidates, and subsequently corroborates them
into a cleaner set by the following steps:
\squishlist
  \item[1.] Filter predicate candidates  based on frequency in a large corpus, discarding rare ones.
  \item[2.] For each retained predicate, 
  consider the nearest noun phrase to the left as subject and the nearest noun phrase to the right as object (assuming active-voice sentences; reverse for passive voice).
  \item[3.] Use a supervised model to run
  these (proto-)statement candidates through
  a classifier, yielding a confidence on the
  plausibility of the candidate triple.
\squishend

For step 1, the huge diversity of 
predicates is reduced by removing adjectives, adverbs, and similar components, so that, 
for example, ``performs a duet with'' and
``performs a beautiful duet with'' are combined,
and so are ``sings with'' and ``occasionally sang with''. 

For step 3, a logistic regression classifier is learned over a hand-crafted set of training data,
based on manually compiled features.
 Example features include \textit{sentence begins with statement subject} (positive weight),  
 \textit{sentence is longer than 20 words} (negative weight), 
 \textit{last preposition in the predicate is a particular word, like
 ``on'', ``of'', ``for'' etc.}
 (different positive weights), 
\textit{ presence of noun phrases other than those
 for subject and object} (negative weight), and more.
 Training this classifier to reasonable accuracy required manual labeling of 1000 sentences \cite{fader2011identifying}.

\subsection{Extensions of Pattern-based Open IE}
\label{ch7-subsec:OpenIE-extensions}

The basic principles of ReVerb can be extended in many different ways, most notably, by incorporating richer syntactic cues from dependency parsing
\cite{DBLP:conf/ijcai/BankoCSBE07,DBLP:conf/acl/WuW10,DBLP:conf/emnlp/MausamSSBE12,DBLP:conf/acl/AngeliPM15}
or from analyzing the clauses that constitute
a complex sentence \cite{DBLP:conf/www/CorroG13}.
Another theme successfully pursued is to
run a conservative method first, such as ReVerb,
and use its output as pseudo-training data
to bootstrap the learning of more sophisticated extractors. An example for the latter is the {\em OLLIE} method 
by \citet{DBLP:conf/emnlp/MausamSSBE12}.

In the following, we discuss some of the less obvious extensions that together make Open IE 
a powerful tool for discovering predicates. 

\vspace*{0.2cm}
\noindent{\bf Non-contiguous and Out-of-order Argument Structure:}\\
The regex-based approach is elegant, 
but falls short of the full complexity of natural language. For example, the following predicates 
comprise non-contiguous words:

\squishlist
\item Dylan \underline{was covered} among many others \underline{by} Adele.
\item Hurricane \underline{made} the case of Rubin Carter widely \underline{known}.
\squishend

\noindent Similarly, statements do not necessarily follow the standard subject-verb-object order:
\squishlist
\item When \underline{winning} ({\bf P}) the \underline{Nobel Prize in Literature} ({\bf O}),
\underline{Bob Dylan}
({\bf S}) announced that he would not attend the award ceremony.
\squishend

There are two ways to improve the coverage of Open IE: 1) by increasing the expressiveness of patterns, and 2) by using machine learning to 
generalize the pre-specified patterns.
Examples of the former include
the {\em WoE} system \cite{DBLP:conf/acl/WuW10}
the {\em ClausIE} tool \cite{DBLP:conf/www/CorroG13}
and the {\em Stanford Open IE} tool \cite{DBLP:conf/acl/AngeliPM15},
using dependency parsing and clause structures.
Learning was already considered
by the first major system, {\em TextRunner}
(\citet{DBLP:conf/ijcai/BankoCSBE07}),
but merely leveraged a small amount of
labeled samples for bootstrapping a classifier.
More recent systems such as OLLIE
\cite{DBLP:conf/emnlp/MausamSSBE12}
{\em Stanford Open IE} \cite{DBLP:conf/acl/AngeliPM15}
and {\em OpenIE 4.0}
\cite{DBLP:conf/ijcai/Mausam16}
systematically
combine supervised learning with hand-crafted patterns.

To generate training data, the 
paradigm of {\bf distant supervision} (see Section~\ref{sec:distantlysupervisedpatternlearning}) can be leveraged. 
For example, the method of \cite{DBLP:conf/acl/WuW10}
matches Wikipedia infobox entries against sentences
in the same articles, and treats the matching
sentences as positive training samples.
In contrast to learning extractors for 
a-priori known property types, this Open IE approach
puts all properties together into a single pool
of samples. This way, it can learn features
that indicate new property types in text
which are not covered in any of the infoboxes at all.
We will come back to this form of seed-based supervision
in Section \ref{ch7-sec:seed-based-discovery}.

\vspace*{0.2cm}
\noindent{\bf Noun-mediated Properties:}\\
In addition to expressing properties by verbs or verbal phrases, sometimes relations are also
expressed in the form of modifiers in noun phrases.
For example, the sentence 
``Grammy winner Bob Dylan also received an Oscar for \dots'' gives a strong cue for the property
\ent{$\langle$Bob Dylan, winner of, Grammy$\rangle$}.
Wikipedia category names are a prominent case
of this observation (see Section \ref{ch6:subsubsec:patterns-text-lists-trees}).
Especially long-tail property types such as
\ent{wroteLyricsFor} benefit from tapping all
available cues, hence the need for including noun phrases
such as ``Dylan's \underline{lyrics for} Hurricane \dots''.

The works of \cite{yahya2014renoun,DBLP:conf/akbc/PalM16} are examples for tackling this issue.
They operate by using seed facts to learn
extraction patterns from noun phrases based
on a variety of features, and
then combine the learned classifier with additional rules.

\paragraph*{Attribution and Factuality:} %
Another limitation of the basic method is its ignorance of utterance context, such as attribution of claims.  For example, from the sentence
``Blogger Joe4Pacifism demanded that Dylan will also receive the Peace Nobel Prize'', 
a simple pattern-based Open IE method would
extract the triple 
\ent{$\langle$Bob Dylan, will receive, Peace Nobel Prize$\rangle$} -- which is wrong when taken out
of the original context.
Open IE extensions such as OLLIE \cite{DBLP:conf/emnlp/MausamSSBE12} handle this
by adding {\em attribution fields} in their
combination of pattern-based rules and supervised
classification. This would yield the following output, denoted here in the form of nested tuples:
\begin{samepage}
\lstset{language=HTML,upquote=true}
\begin{lstlisting}
<  Blogger Joe4Pacifism,  demanded, 
   <  Bob Dylan,  will receive,  Peace Nobel Prize  >
>
\end{lstlisting}
\end{samepage}

Another notorious case where simple Open IE
often fails is the presence of negation cues
in sentences. 
For example, the sentence ``None of Dylan's songs was ever covered by Elvis Presley'', a straightforward
extractor would incorrectly yield
\ent{$\langle$Dylan's songs, covered by, Elvis Presley$\rangle$}, ignoring the crucial word
``None''.
{\em MinIE} \cite{DBLP:conf/emnlp/GashteovskiGC17}
fixes this issue by adding a {\em polarity field}
that is toggled when observing a negation cue
from a list of keywords (e.g., not, none, never, etc.).

The {\em NestIE} method \cite{DBLP:conf/emnlp/BhutaniJR16}
and the {\em StuffIE} method \cite{DBLP:conf/cikm/PrasojoKN18}
generalize these approaches by casting 
sentences into {\em nested-tuple structures}
to represent facets like attribution, 
location, origin, destination, 
time, cause, consequence, etc.
These facets are handcrafted based on
lexical resources such as 
Wiktionary ({\small\url{https://www.wiktionary.org/}}),
PropBank ({\small\url{https://propbank.github.io/}}) \cite{DBLP:journals/coling/PalmerKG05},
FrameNet ({\small\url{https://framenet.icsi.berkeley.edu/}}) \cite{DBLP:conf/paclic/FillmoreWB01}
and OntoNotes ({\small\url{https://github.com/ontonotes/}}) \cite{DBLP:journals/ijsc/PradhanHMPRW07},
reflecting especially the
semantic roles of prepositions (``at'', ``for'', ``from'', ``to'' etc.) and conjunctions (``because'', ``when'', ``while'' etc.).
Semi-automatically annotated sentences
are used to train a logistic regression
classifier for casting complex sentences into
nested tuples.

\subsection{Neural Learning for Open IE}
\label{ch7-subsec:OpenIE-neural}

Distilling propositional sentences into predicate-argument
structures has been pursued in computational linguistics
for the task of {\em Semantic Role Labeling} (SRL, see Section \ref{ch6:subsubsec-featurebasedclassifiers}).
A typical setting is to consider a small set of {\em frame types}
with specific slots to be filled. For example,
for frame type \ent{creates} they would include 
creator, created entity, components, co-participant, circumstances, means, manner, and more.
An SRL method takes a sentence as input, uses a classifier to
assign it to one of the available frame types (or none),
and then applies a sequence tagger to fill all (or most) of
the frame slots. 
This setup with arguments for {\em frame-specific roles}
does not make sense for {\em Open} IE, since we do not want to limit
ourselves to pre-specified frame types.
Instead, we settle for more generic arguments, like arg0, arg1, arg2, etc.
-- an ordered list of arguments, typically with subject as arg0,
object as arg1, and context-capturing modifiers as further arguments. 
This eliminates the need for a type classifier and focuses on
the slot-filling tagger.

\begin{figure} [h!]
  \centering
   \includegraphics[width=1.1\textwidth]{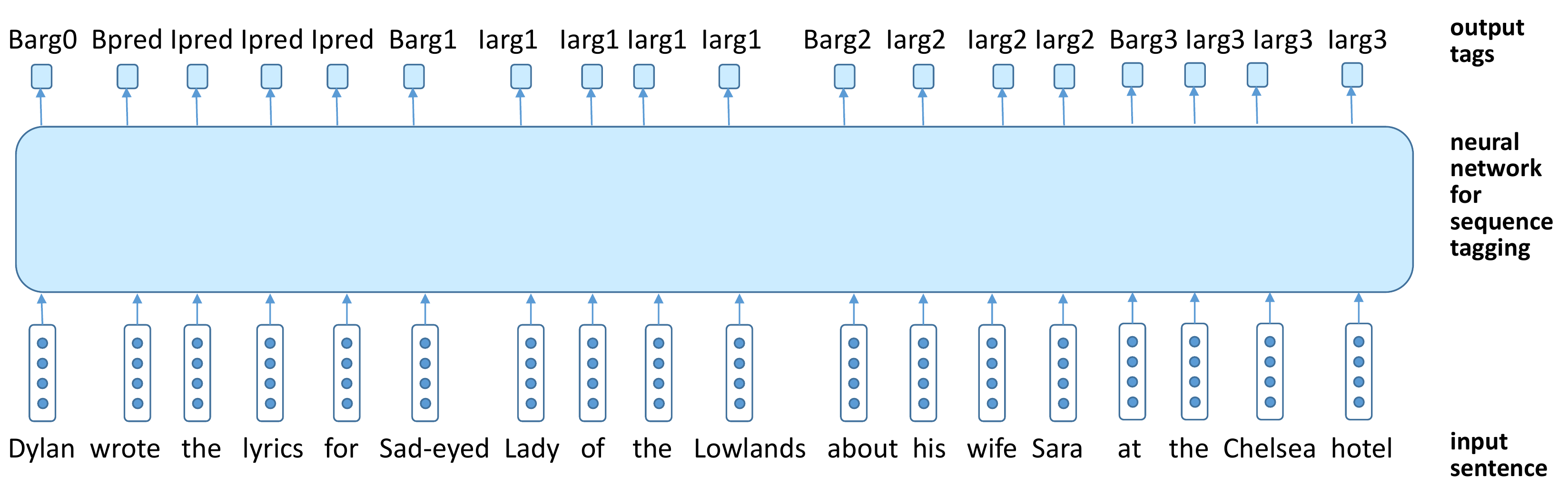}
      \caption{Sequence-Tagging Neural Network for Open IE}
      \label{fig:lstm4openIE}
\end{figure}

Not surprisingly, the state-of-the-art
methodology for this sequence tagging task
is deep neural networks, more specifically,
bi-LSTM transducers and other recurrent networks 
or Transformer-based networks
(see Section \ref{subsec:deep-neural-networks}).
\citet{DBLP:conf/naacl/StanovskyMZD18}
present
a full-fledged solution that encodes
sentences with word and position
embeddings as initial inputs. 
The learned representation is decoded
into a sequence of BIO tags to mark up
the begin and inner part of predicate and
argument spans (using O for all other, irrelevant
tokens). To distinguish the
predicate and the different arg-i's,
the BIO tags are separately instantiated for
each of these constituents.
An example of input and output is shown 
in Figure \ref{fig:lstm4openIE}.
The learning architecture itself pretty much
follows those that have been developed
for entity recognition (NER) and entity typing
(see Chapter \ref{ch3:entities}).
Therefore, we do not elaborate on this any further.

An interesting aspect of   \cite{DBLP:conf/naacl/StanovskyMZD18} is
the way that training data is compiled.
This includes labeled sentences specifically
for Open IE, but also a clever way of
leveraging annotated collections for
QA-style machine reading comprehension
\cite{DBLP:conf/emnlp/HeLZ15}.
The latter data comprises a large set of
triples with a propositional sentence,
a question that someone could ask about the
sentence, and an answer that could be 
derived from the sentence.
Questions are generated by templates and
other means; they provide a user-friendly
way to annotate sentences and are thus
more amenable to crowdsourcing 
than other forms of labeling \cite{DBLP:conf/naacl/MichaelSHDZ18}.

Another approach to mitigate the training bottleneck is to combine sequence-tagging methods with reinforcement learning
(see, e.g., %
\cite{DBLP:conf/www/LiuLWSL20}).

\section{Seed-based Discovery of Properties}
\label{ch7-sec:seed-based-discovery}

In the previous section, we assumed that there
is no prior knowledge base of properties: Open IE taps into
any given text collection and aims at maximally broad coverage.
In this section, we exploit the fact that
we already have a pre-existing KB that contains
many triples about a limited set of 
specified properties. 
This prior knowledge can be leveraged for
distant supervision, by spotting patterns
for the limited-properties statements
and generalizing them into {\em hyper-patterns}
that are applicable to discover previously
unknown property types.
For example, semi-structured web sites
about commercial products may have many pages
where salient attributes and their values are
rendered in a certain style.
A frequent case is list elements
with the
attribute name in boldface or other 
highlighted font followed by a colon or tab
followed by the attribute value.
This is a hyper-pattern.

The methods presented in this section are based
on the following key intuitions:
\squishlist
\item Property names often share similar 
presentation patterns, like language style
(for text) or layout (for lists and table).
\item Property names are much more frequent in
a topically focused text corpus or
semi-structured
web site than their values
for individual entities.
\squishend

\subsection{Distant Supervision for Open IE from Text}
\label{ch7-subsec:openie-fromtext}

The {\bf Biperpedia} project \cite{DBLP:journals/pvldb/GuptaHWWW14}
took a large-scale approach to discovering
new properties in web pages and query logs,
by learning patterns and classifiers based
on statements for pre-specified properties
from a large KB. 
Specifically, \cite{DBLP:journals/pvldb/GuptaHWWW14}
leveraged Freebase; the {\em WoE} system 
\cite{DBLP:conf/acl/WuW10} pursued similar techniques at smaller scale, based on
Wikipedia infoboxes.
The key idea is as follows.
Suppose the KB has many SPO triples for
properties like 

{\small
\lstset{language=HTML,upquote=true}
\begin{lstlisting}
     createdSong: musician x song
     createdMovieScript: writer x movie
     performedWith: musician x musician
     playedOnAlbum: song x album
     playedAtFestival: song x event
\end{lstlisting}
}%
\noindent We can spot these SO pairs in a large text corpus
and check that there is contextual evidence
for the property P as well, by good pattern,
such as:
{\small
\lstset{language=HTML,upquote=true}
\begin{lstlisting}
     song O written by S,   script for O written by S,
     song S played on album O,   song S played live at O
\end{lstlisting}
}
\noindent Next, we aim to identify commonalities among
patterns for multiple properties, such as
{\small
\lstset{language=HTML,upquote=true}
\begin{lstlisting}
     NN * O written by S
     song S played PREP NN O
\end{lstlisting}
}
\noindent with noun phrases {\em NN} and prepositions {\em PREP}.
These can now serve as {\em hyper-patterns}
to be instantiated, for example,
by ``lyrics for O written by S''
and ``song S played in movie O''.
This is the idea to discover the previously
unknown property types
{\small
\lstset{language=HTML,upquote=true}
\begin{lstlisting}
    wroteLyricsFor: artist x song
    playedInMovie: song x movie
\end{lstlisting}
}
\noindent where the type signatures would be learned from
co-occurring S-O pairs via entity linking
(assuming the KB is already richly populated
with entities and types).

Generally, the method runs in the following steps:

\begin{samepage}
\begin{mdframed}[backgroundcolor=blue!5,linewidth=0pt]
\squishlist
\item[ ] \textbf{Distantly Supervised Discovery of Properties in Text:}\\
\noindent Input: text corpus and prior KB\\
\noindent Output: new property types
\squishlist
\item[1.] Spot text snippets with occurrences of SPO triples from the KB. Use coreference resolution to capture also passages that span more than one sentence.
\item[2.] Compute frequent hyper-patterns by relaxing specific words and entities into
syntactic structures (regex on part-of-speech tags,
dependency-parsing tags etc.).
\item[3.] Assume that new properties are
always expressed by concise nouns or noun phrases
next to the parts that match S or O.
Gather these noun phrases as candidates for
new properties.
\item[4.] Prune noisy candidates by means of
statistics and classifiers,
and group candidates into synonymy-sets 
via co-occurrence mining or classifiers
(e.g., ``lyrics of O written by S''
is synonymous to ``text of O written by S'').
\item[5.] Train a feature-based classifier
to assign semantic types to the arguments of
the new properties (e.g., {\em artist $\times$ song}),
and run this classifier on the retained
property candidates.
Features include informative words in textual
proximity (e.g., ``rise'' or ``drop'' as cues for
datatype {\em number} of a discovered attribute).
\squishend
\squishend
\end{mdframed}
\end{samepage}

\noindent The implementation of Biperpedia
focused strongly on attributes rather than
relations, such as 
{\small
\lstset{language=HTML,upquote=true}
\begin{lstlisting}
     salesOfAlbum: album x integer
     songIsAbout: song x text
\end{lstlisting}
}
\noindent where the O argument could be a person,
location, event (before entity linking)
or any general topic --
hence the generic datatype {\em text}.
Here, the type inference covers datatypes
like numeric, date, money, text.
For attributes, a fairly compact subset
of hyper-patterns already gives high yield,
including {\em NN of S is O}, {\em S and PREP NN O},
etc.
The method itself applies to
relationships between entity pairs 
equally well.
For extracting complete SPO triples,
entity linking must be run in a post-processing
step or incorporated into one of the earlier steps
(potentially even as pre-processing before
the first step).

\clearpage\newpage
\vspace*{0.2cm}
\noindent{\bf Tapping Query Logs:}\\
Similar principles of distant supervision
and the learning of hyper-patterns that
generalize from specified properties to
a-priori unknown properties, also
apply to {\em query logs} as an input source
\cite{DBLP:conf/ijcai/PascaD07,DBLP:journals/pvldb/GuptaHWWW14,DBLP:conf/cikm/Pasca15,DBLP:conf/kdd/LiuGNWXLLX19}.
For example, many users ask search engines
about ``sales of $\langle$album$\rangle$'',
``lyrics writer of $\langle$song$\rangle$'',
``topic of $\langle$song$\rangle$'',
``meaning of $\langle$song$\rangle$'',
``what is $\langle$song$\rangle$ about?'',
and so on.
This is a huge source for property discovery.
However, it is accessible only by the
search-engine providers, and it faces
the challenge that user queries contain
an enormous amount of odd inputs (e.g., misspelled,
nonsensical or heavily biased).

\subsection{Open IE from Semi-Structured Web Sites}
\label{ch7-openie-semistructured}
Semi-structured content is a prime resource for KB construction (see Chapter~\ref{ch2:knowledge-integration} 
Section~\ref{ch6-sec:properties-from-semistructured}), and this applies also to open information extraction.
In particular, many web sites with content
generated from back-end databases organize a major
part of their information into {\bf entity detail pages}.
In Section \ref{ch6-sec:properties-from-semistructured},
we exploited this fact to identify pages that are
(more or less) exclusively about a single entity
from the prior KB.
As we dealt with populating pre-specified property types
in Section \ref{ch6-sec:properties-from-semistructured},
this meant that we already knew S and P when tapping
a web page about possible O values for an SPO triple.
This greatly simplified the task, and methods like
Ceres 
(\citet{DBLP:journals/pvldb/LockardDSE18})
showed the way to extraction with both 
high precision and high recall.
In our current setting, with the goal of discovering
new property types, we do not have the luxury
anymore that we already know which P we are after,
but we can still exploit that S is known for 
most pages from a semi-structured web site
(if sites and pages are chosen carefully).

The \textit{OpenCeres} method, by
\citet{DBLP:conf/naacl/LockardSD19}, 
extends Ceres %
to discover new properties. Earlier work
along similar lines is \cite{DBLP:journals/pvldb/BronziCMP13}
with focus on regex-based rules.

OpenCeres starts with distant supervision from
the prior KB to extract and generalize patterns.
This is similar to Biperpedia (see 
Section \ref{ch7-subsec:openie-fromtext}), but focuses on DOM-tree patterns.
For example, many instances for a known property type
could be spotted by paths in the DOM tree that end
in a leaf of the form {\it NN:O} with a noun phrase
{\it NN} in highlighted font followed by an entity mention
$O$ that can be linked to the KB.
If {\it NN} does not correspond to any of the
KB properties, we can consider it a newly
discovered property, holding between $S$, the subject
of the entity detail page, and $O$.
This way, a large number of candidate properties
can be gathered, from cues across all pages of the
web site.

The key idea of OpenCeres lies in the subsequent
{\em corroboration stage}, where it uses
the semi-supervised method of
{\bf label propagation} \cite{Zhu2002LearningFL} to clean the candidates.

\begin{samepage}
\begin{mdframed}[backgroundcolor=blue!5,linewidth=0pt]
\squishlist
\item[ ] \textbf{Input Graph for Label Propagation over New Property Candidates:}\\
The method operates on a graph where each
node is either 
\squishlist
\item a distant-training sample on
a PO pair from the KB (S is assumed to be certain from the entity
detail page), or
\item a candidate pair of PO with P as a text phrase
and O in text or entity form (i.e., before or after entity linking).
\squishend
The graph connects nodes by similarity, using
two kinds of features to compute edge weights:
\squishlist
\item distance measures with regard to the DOM tree, and
\item cues about visual layout such as font size, color etc.
\squishend
\squishend
\end{mdframed}
\end{samepage}

\begin{figure} [h!]
  \centering
   \includegraphics[width=0.8\textwidth]{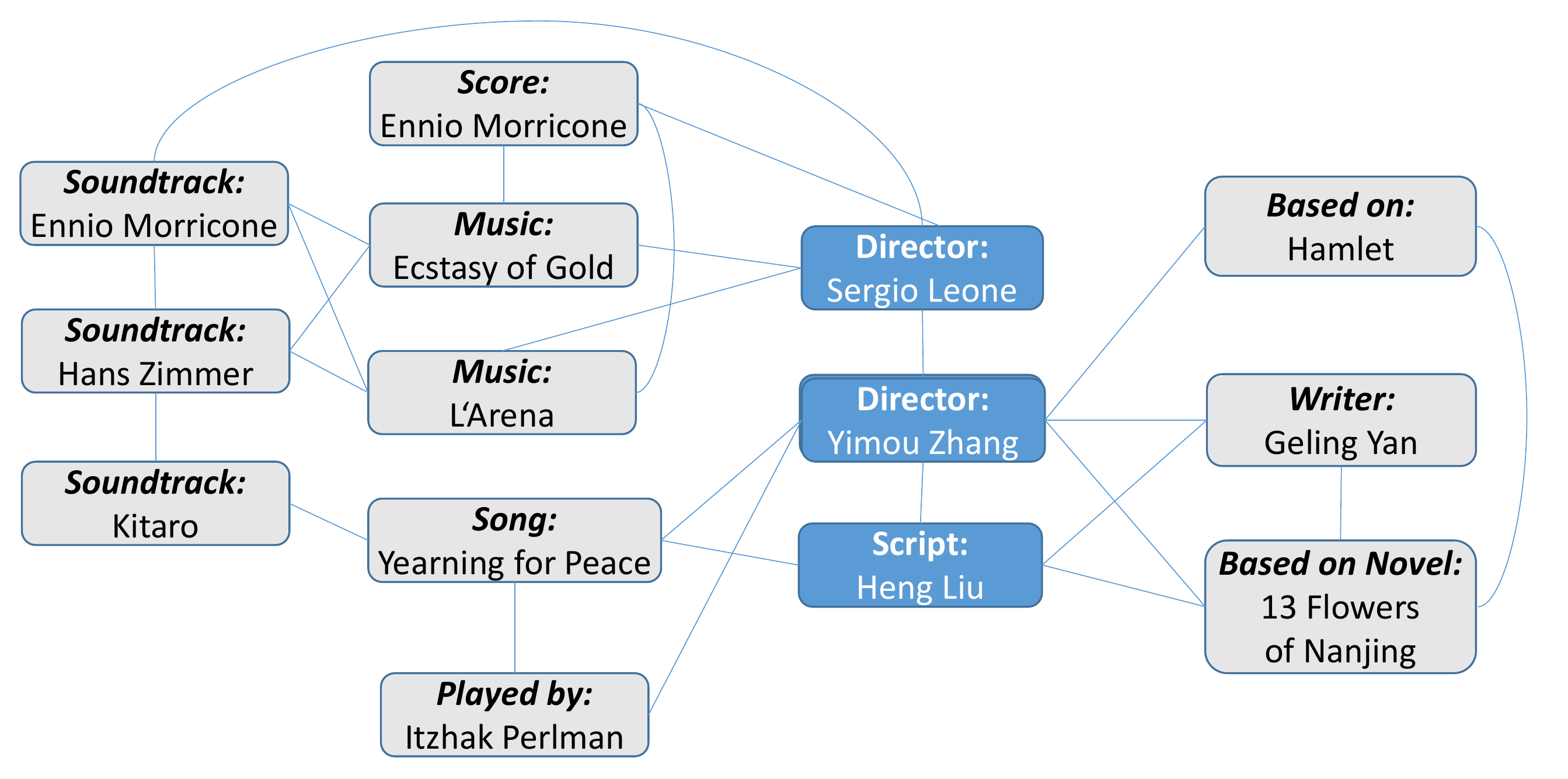}
      \caption{Candidate Graph for Label Propagation to Discover New Properties}
      \label{fig:lp4openie}
\end{figure}

\noindent An example graph is shown in Figure \ref{fig:lp4openie},
showing PO pairs as nodes, with S being movies (not shown)
from entity detail pages of the same web site.
Seed nodes are in blue; edge weights are omitted
for simplicity.
The figures shows that entities like music pieces and musicians are connected with film directors, because
they co-occur in detail pages about the respective movies.
Likewise, there are movies based on books or theater plays,
and these lead to edges between these literature works
and the respective film director
(i.e., Yimou Zhang who directed \ent{The Banquet}
and \ent{The Flowers of War}).

Label propagation aims to assign a binary truth label
to each node with a certain probabilistic confidence. To this end, it propagates the initial perfect-confidence
truth of its positive seeds to neighbors in proportion
to the edge eights. An underlying assumption is
that the variations in truth confidence should be
gradual and smooth within neighborhoods.
This resembles the computation of PageRank for
web-page authority, and indeed it is the same
family of nicely scalable algorithms (e.g.,
Jacobi iteration) that can be used here.
In the end, when the algorithm converges and 
each node has a confidence score, a threshold is
applied to select the most likely proper nodes
as output.

An extension
called {\em ZeroShotCeres}
\cite{Lockard2020ZeroShotCeres}
has further advanced this methodology by
relaxing the need to have at least some 
seeds per web site in the distant supervision.
The new method uses visual and structural cues
from any domain-specific web site to 
learn how to tap a previously unseen web site of
the same domain (e.g., movies).
Technically, this is based on a
{\em graph neural network} with graph embeddings
for the visual and structural cues.

\vspace*{0.2cm}
\noindent{\bf Tapping Web Tables:}\\
The InfoGather project \cite{DBLP:conf/sigmod/YakoutGCC12}
tackled the problem of property discovery by
focusing on web tables (cf. Section %
\ref{ch6-subsec:webtables}).
A key idea is again using SPO triples from a prior KB
for distant supervision:

\squishlist
\item[1.] Spot SPO in a table with SO in the same row
and column header P for the column where O occurs.
\item[2.] Identify additional columns with headers
Q1, Q2, \dots as candidates for new properties that
apply to S. 
\item[3.] Aggregate these cues over all tables in the data collection, and apply statistics and probabilistic
inference to compute high-confidence properties.
\squishend

\noindent In addition, the discovered properties are
organized into synonymy groups, using
overlap measures and schema-matching techniques.

A particular challenge with web tables is that
headers may be very generic and lack enough information
to meaningfully identify them as informative and new.
For example, column headers like {\em Name}
or {\em Value} do not add any benefit to a KB schema.
Even more specific headers such as {\em Sales} or
{\em Growth} are doubtful,
if we miss the relevant reference dimensions like
currency and year.
\cite{DBLP:conf/sigmod/ZhangC13}
extended this framework by propagating such information across tables.
\cite{DBLP:conf/cikm/IbrahimRW16,DBLP:conf/icde/IbrahimRWZ19}
enhanced
the interpretation of web tables by
considering information from the text
that surrounds tables in web pages,
matching table cells against text phrases
and, this way, picking up more context.

\subsection{View Discovery in the KB}
\label{ch7-subsec:viewdiscovery}

There are cases where the KB itself exhibits patterns
that can suggest additional property types that
are not explicitly specified yet.
For example, suppose the KB contains properties
\ent{created:}{\em musician $\times$ song}
and
\ent{performed:}{\em musician $\times$ song}.
Then the composition of \ent{created} and \ent{performed}$^{-1}$ denotes the
property \ent{coveredBy} between musicians.
Methods for rule mining (see Section \ref{sec:rule-mining})
and path mining over knowledge graphs can
automatically discover such interesting predicates.
Appropriately naming them would be a modest
manual effort afterwards.
The {\em Path Ranking Algorithm (PRA)}, by
\citet{lao2011random}\cite{lao2012reading},
has advanced this direction in the context of the
NELL project.
Note, though, that such methods can only discover
what is implicitly already in the KB -- there is
no way to find ``unknown unknowns'' that have
no cues at all in the KB.

\section{Property Canonicalization}
\label{ch7-sec:property-canonicalization}

OpenIE-style methods, as discussed in the previous
sections, are good at coverage, aiming to discover
as many new property types as possible.
However, this comes at the expense of redundancy
and inconsistency. Despite some steps to
clean their outputs, a method could yield
two seemingly distinct properties
\ent{playedIn} and \ent{heardIn}, both between
songs and movies and both denoting the very
same semantic relationship.
Similarly, the output could have both
\ent{composed} and \ent{wroteLyricsFor}
between artists and songs; these are highly
correlated but not semantically equivalent.
To avoid all these pitfalls and arrive at
unique representations of properties without
redundancy and risks of inconsistencies,
we need additional methods for {\em canonicalization}
-- the property-focused counterpart to what
entity linking does for the property arguments S and O.

Techniques for this purpose include
clustering, matrix or tensor factorization,
itemset mining, and more.
We start the discussion by first introducing
the construction of {\bf paraphrase dictionaries}
for properties, as a building block for other techniques.
This includes also inferring {\bf type signatures}
for properties, and organizing all properties into
a {\bf subsumption hierarchy}.

\subsection{Paraphrase Dictionaries}
\label{ch7-subsec:paraphrasedictionaries}

Recall that for constructing rich type taxonomies,
as a key part of the KB schema,
we leveraged existing resources like
WordNet and the Wikipedia category system
(see Chapter\ref{ch2:knowledge-integration}).
Such assets were available for types/classes,
that is, unary predicates in a logical sense,
but not for properties, the binary predicates
for the KB.
If we had a comprehensive dictionary of
binary-predicate names, along with type signatures,
and each name associated with a set of synonyms,
then we could use this as 
{\em complete schema} where all collected properties
are positioned. 
In other words, there would be no
more ``unknown unknowns'' if the dictionary
were truly complete.

Needless to say that a perfect dictionary,
with complete coverage and 
absolutely
accurate synonymy-sets, is wishful thinking.
Nevertheless, we can approximate this goal
by constructing a reasonable dictionary and
then using it as an asset in the property discovery.

The dictionary building is itself a property
discovery task, though. 
As properties are expressed by names and phrases,
such as {\em romance with}, {\em is dating} and
{\em loving couple}, the NLP community views this
as the task of {\bf paraphrase detection}.
In addition to inferring {\bf synonymy} among such phrases
(i.e., one paraphrasing the other), we also
want to understand their {\bf subsumption} relationships
(i.e., one entailing the other).
For example, {\em romance with} is a specialization
of {\em friendship with} (at least when it is fresh),
and {\em duet with} (between singers) is a 
sub-property of {\em performed with} (between musicians).

\subsubsection{Property Synonymy}

To infer the synonymy of two property phrases, all methods make use of
the {\em distributional hypothesis}: 
the meaning of a word or phrase can by understood 
by observing the contexts in which it occurs.

\vspace*{0.2cm}
\paragraph*{Textual Contexts:} %
The first methods for paraphrase detection were built directly on the distributional hypothesis applied to text corpora. 
The {\em DIRT} method
(Discovering Inference Rules from Text), by 
\citet{DBLP:conf/kdd/LinP01a},
used dependency parsing to identify subject and object of
property phrases (in general text, not necessarily about KB entities).
Typically, the property phrase would be a verb or verbal phrase.
IDF weights (inverse document frequency) were assigned to the
co-occurring subject and object, to compute an informativeness score
for the property phrase.
Finally, pairs of phrases could be ranked as paraphrase candidates
by weighted overlap scores of their two arguments.

The approach was extended by the {\em OceanVerb} method
(\citet{DBLP:conf/emnlp/ChklovskiP04}) 
to consider also antonyms
(i.e., opposite meanings).
For example, {\em ``in love with''} and {\em ``enemy of''}
would have high distributional similarity in DIRT, because
the subject-object arguments would include common names such as
Joe and Jane and even personal pronouns (as opposed to crisp entities).
OceanVerb uses additional compatibility checks to infer that the example
consists of opposing phrases.
In principle, OceanVerb even infers semantic (meta-) relations between properties
like \ent{happened before} and \ent{enables}.
The former would hold between
{\em ``married with''} and {\em ``divorced from''},
an example of the latter is 
{\em ``composed song''} and {\em ``played in movie''}.
However, this early work was carried out at small scale,
and lacked robustness.

\vspace*{0.2cm}
\paragraph*{Co-Occurrences of Entity Pairs:}%
By restricting the subject-object arguments of property phrases
to be canonicalized entities, based on a prior KB and using entity linking,
the distributional-similarity signals can be made sharper.
This was investigated in the {\em PATTY} project %
(\citet{DBLP:conf/emnlp/NakasholeWS12})
at large scale, with extraction of candidate phrases from the full text of
Wikipedia and web crawls. 
Each property phrase is associated with a {\bf support set}
of co-occurring entity pairs. Then, distributional similarity can be
defined by overlap measures between support sets.
To this end, PATTY devised techniques for {\bf frequent itemset mining}
with generalizations along two axes: 
\squishlist
\item[i)] lifting entity pairs to
type pairs and super-type pairs, and 
\item[ii)] lifting words to POS tags (to replace overly specific words, e.g.,
a personal-pronoun tag instead of ``his'' or ``her'') including
wildcards for short subsequences.
\squishend

A key advantage of this approach is that the entities are typed,
so that we can automatically infer type signatures for property phrases
(e.g., person $\times$ person, or song $\times$ movie).
In fact, the type signatures are harnessed in
computing synonymy among phrases, by checking if different
type signatures are compatible (e.g., musician $\times$ musician
and person $\times$ person) or not
(e.g., song $\times$ movie and person $\times$ movie).
The PATTY method compiled a large resource of paraphrases,
but still suffered from sparseness and a fair amount of noise.
Most notably, its reliance on sentences that contain a property phrase
and two named entities was a limiting factor in its coverage.
It did not consider any coreference resolution, or other means for
capturing more distant signals in text passages.

The {\em DEFIE} project \cite{DBLP:journals/tacl/BoviTN15} took this methodology further,
specifically tapping into concise definitions, 
like the first sentences in Wikipedia articles or entries in Wiktionary
and similar resources.

\vspace*{0.2cm}
\noindent{\bf Multilingual Alignments:}\\
Collections of multilingual documents with sentence-wise alignments,
referred to as {\em parallel corpora} in NLP,
are another asset, first considered in works by \cite{DBLP:conf/acl/BarzilayM01}
and \cite{DBLP:conf/acl/BannardC05}.
The key idea is to use pairs of sentences in language A, and their translations into language B: A1, A2, B1, B2.
If A1 and A2 contain different property phrases that are translated onto
the same phrase in B1 and B2, and if this happens frequently,
then A1 and A2 give us candidates for synonymy.

For example, consider the English-to-German translations:

\begin{tabular}{lcl}
A1: \dots composed the soundtrack for \dots %
& $\rightarrow$ & \hspace*{0.2cm} B1: \dots schrieb die Filmmusik f\"ur \dots \\
A2: \dots wrote the score for \dots %
& $\rightarrow$ & \hspace*{0.2cm} B2: \dots schrieb die Filmmusik f\"ur \dots \\
\end{tabular}

\noindent These are cues that ``composed the soundtrack'' and ``wrote the score''
are paraphrases of each other.

One of the largest paraphrase dictionaries, {\em PPDB}
(Paraphrase Database), by
\citet{DBLP:conf/naacl/GanitkevitchDC13}\cite{DBLP:conf/acl/PavlickRGDC15}
({\small\url{http://paraphrase.org/#/download}}), 
has been constructed by similar methods
from multilingual corpora with alignments.
This comprises more than 100 million paraphrase pairs, with
rankings and further annotations, covering both unary predicates
(types/classes and WordNet-style senses) and binary predicates (relations and
attributes).
However, the property phrases have no type signatures attached to them.

The methodology is not limited to intellectually aligned corpora,
but can be extended to leverage large collections of {\em machine-translated
sentences}. This has been investigated by \cite{DBLP:conf/emnlp/GrycnerW16}
using multilingual data constructed by \cite{DBLP:conf/naacl/FaruquiK15} 
from Wikipedia sentences.
By integrating entity linking (easily feasible for Wikipedia text), the
method also infers type signatures for all paraphrase-sets.
The resulting dictionary, called {\em POLY} and extending PATTY (see above),
is available at {\small\url{https://www.mpi-inf.mpg.de/yago-naga/patty}}.

Note that all these methods for constructing large paraphrase dictionaries
date back a number of years. With today's advances in machine learning
and entity linking and the availability of much larger datasets,
similar methodology re-engineered today would have potential
for strongly enhanced dictionaries with both much better coverage and accuracy.

\vspace*{0.2cm}
\paragraph*{Web Tables and Query Logs:}%
Another source for detecting distributional similarity of property phrases
is web tables and query logs \cite{DBLP:conf/www/HeCCT16}.
Both are customized cases of the {\bf strong co-occurrence principle}
introduced already in Section \ref{subsec:dictionary-based-entity-spotting}.
For web tables, if two columns with headers $P$ and $Q$ in different tables
contain approximately the same entities or values and both tables
have high row-wise overlap of key entities (the $S$ entities),
then $P$ and $Q$ are good candidates for synonymy.
As for query logs, the method exploits that 
search-engine queries that share an entity
but vary in another keyword may yield highly overlapping sets of {\em clicked pages}.
For example, the two queries ``movie soundtracks by Ennio Morricone'' and
``film scores by Ennio Morricone'' may return, to a large degree, the same good web pages that are clicked on by many users. Therefore, ``soundtrack by'' and
``film score by'' are likely synonymous.
These indirect co-occurrence signals are even more pronounced for
attributes such as ``Paris population'', ``Paris inhabitants'' and
``how many people live in Paris''.

\subsubsection{Property Subsumption}

For semantically organizing large sets of property phrases,
synonymy grouping is not sufficient. In addition, we want to 
identify cases where one property subsumes another, this way
building a taxonomic hierarchy of binary predicates.
For example, \ent{performedWith} subsumes \ent{duetWith} between
musicians, so the former should be a super-property of the latter.
The PATTY, DEFIE, PPDB,
and POLY projects (discussed above) addressed this issue
of inferring subsumptions (or linguistic entailments) as well
\cite{DBLP:conf/emnlp/NakasholeWS12,DBLP:journals/tacl/BoviTN15,DBLP:conf/acl/PavlickRGDC15,DBLP:conf/emnlp/GrycnerW16}.

The key idea in PATTY was to compare the 
{\em support sets} of different
phrases (see above), computing scores for {\em soft inclusion}
(i.e., set inclusion that tolerates a small amount of exceptions).
When property phrase $A$ has a support set that mostly includes
the support set of phrase $B$, we can conclude that $A$ subsumes $B$.
This way, a large set of candidates for subsumption pairs can be
derived, each with a quantitative score.

Since phrases can be generalized also at the lexico-syntactic level
(e.g., replacing words by POS tags or by wildcards),
this subsumption mining is carried out also for generalized phrases.
For example, the phrase {\em VB soundtrack PREP} is more general
than {\em wrote soundtrack for} and would thus have a larger support set,
which in turn affects the mined subsumptions.

PATTY integrates all these considerations into an efficient and scalable
sequence mining task, to arrive at set of candidate subsumptions.
As these candidates do not necessarily form an acylic DAG, 
a final step breaks cycles using a minimum-cost graph cleaning technique.
The taxonomic hierarchy of property paraphrases is available at 
{\small\url{https://www.mpi-inf.mpg.de/yago-naga/patty}}.
Despite its large size, its coverage and refinement are limited, though,
by the limitations of the underlying input corpora.
Reconsidering the approach with modern re-engineering
for Web-scale inputs and also leveraging embeddings
could be a promising endeavor.

\subsection{Clustering of Property Names}

Suppose we have run an Open IE method (see Section \ref{ch7-sec:openIE})
on a large corpus and obtained a set of proto-statements
in the form of SPO triples where all three components are
short text phrases. This may give high recall but comes
with noise and diversity: many variants of P phrases potentially
denoting the same property and ambiguous phrases for S and O.
An obvious idea to clean this data and organize it towards
crisper triples for KB population, KB schema construction 
or schema extension with new properties, is by {\bf clustering}
entire triples or P components.
Several methods have been developed to this end 
\cite{DBLP:journals/jair/YatesE09,DBLP:conf/emnlp/MinSGL12,DBLP:conf/semweb/PujaraMGC13,DBLP:conf/cikm/GalarragaHMS14};
they differ in specifics but share the same key principles,
as discussed in the following.

Consider the following example output of Open IE:
{\small
\lstset{language=HTML,upquote=true}
\begin{lstlisting}
< Bob Dylan, sang with, Patti Smith >
< Dylan, sang with, Joan Baez >
< Bobby, duet with, Joan >
< Dylan, performed with, Baez >
< Dylan, performed with, violinist Rivera >
< Dylan, performed with, his band >
< Dylan, performed with, Hohner harmonica >
< Bobby, shot by, assassinator >
< Hurricane, is about, Rubin Carter >
< Hurricane, criticizes, racism >
< Sara, is about, Dylan's wife >
< Green Mountain, is about, civil war >
< Green Mountain, critizes, civil war >
< Masters of War, criticizes, war >
\end{lstlisting}
}%

\vspace*{0.2cm}
\noindent{\bf Pre-processing of S and O Arguments:}\\
To have a clearer focus on canonicalizing the properties,
most methods first process the S and O arguments towards
a cleaner representation. An obvious approach, if a prior
KB about entities and types is available, is to run 
{\em entity linking} on both S and O.
If this works well, we could then group the triples
by their proper S and/or O arguments.
For example, the first 5 triples all have \ent{Bob Dylan}
as S argument. The sixth triple, which actually
refers to \ent{Robert Kennedy}, may produce a linking error
and would then be spuriously added to the previous group.
The KB could further serve to lift the entities into
their most salient types, like \ent{musician} and \ent{song}
for the S arguments. These types can further annotate the
groups as features in downstream processing.

If no KB exists a priori, then the method of choice is
to cluster all S arguments and all O arguments based on
string similarity and distributional features like
co-occurrence of P and O phrases and tokens in the underlying sentences
or passages. 

The result of this stage would be a set of {\em augmented triples}
where each unique P phrase is associated with a group of S arguments
and a group of O arguments (where non-entity phrases are kept as strings):

{\small
\lstset{language=HTML,upquote=true}
\begin{lstlisting}
<  {Bob Dylan},  sang with,  {Patti Smith, Joan Baez}  >
<  {Bob Dylan},  duet with,  {Joan Baez}  >
<  {Bob Dylan},  performed with, 
   {Joan Baez, Scarlet Rivera, his band, Hohner harmonica}  >
<  {Bob Dylan},  shot by,  {assassinator}  >
<  {Hurricane, Sara, Green Mountain},  is about, 
   {Rubin Carter, Sara Lownds, civil war}  >
<  {Hurricane, Masters of War},  criticizes, {racism, civil war, war}  >
\end{lstlisting}
}%

\vspace*{0.2cm}
\noindent{\bf Similarity Metrics for Clustering}\\
The main task now is the clustering of the P phrases themselves.
The crucial design decision here is to choose features
and a similarity metric that guides the clustering algorithms.
For this purpose, prior works investigated combinations of the
following feature groups:

\begin{samepage}
\begin{mdframed}[backgroundcolor=blue!5,linewidth=0pt]
\squishlist
\item[ ] {\bf Features for Property Clustering}
\squishlist
\item {\bf String similarity} of property phrases P and Q,
potentially incorporating words embeddings, so that, for example,
{\em sang with} and {\em performed with} would be similar.
\item {\bf Support sets overlap} where the support sets for phrase P
are the groups of S and O arguments associated with P (see example above).
\item {\bf Type signature similarity} between P and Q, with the 
semantic types of S and O arguments derived from the support sets,
such as \ent{musician}, \ent{singer}, \ent{song} etc.
The scores for how close two types are to each other can be based
on word embeddings or on other relatedness measures derived from
the KB´s taxonomy or entity-linked data collections
(cf. Section \ref{sec:el-semistructured}).
\item {\bf Relatedness in property dictionary}, to leverage an existing
dictionary such as PPDB or PATTY (see Section \ref{ch7-subsec:paraphrasedictionaries}) where 
a subset of property synonyms or near-synonyms (e.g., ``plays with'', ``performs with''
and perhaps also ``sings with'') are already known, as are highly
related properties in the subsumption hierarchy.
\item {\bf Context similarity} based on the sentences or passages
from where the triples were extracted, again with the option of word embeddings
(cf. Section \ref{ch5-subsec:contextsim4EL}).
\squishend
\squishend
\end{mdframed}
\end{samepage}

\clearpage\newpage
\noindent{\bf Algorithmic Considerations:}\\
The algorithm itself could be as simple as {\em k-means clustering}, or could be as advanced as learning clusters via graphical models or neural networks.
The output for our toy example could be as follows, with $k$ set to three:
{\small
\lstset{language=HTML,upquote=true}
\begin{lstlisting}
cluster 1: {sang with, duet with, performed with}
cluster 2: {shot by}
cluster 3: {is about, criticizes}
\end{lstlisting}
}%
This would be a fair set of clusters.
Still it confuses duets of singers with musicians performing with each other,
and it does not discriminate the different semantics of 
\ent{is about} and \ent{criticizes}.
A generally critical issue is the choice of the number of clusters, $k$.
If we had set $k$ to 4, it may perhaps separate \ent{is about} and \ent{criticizes},
but the output could also get worse, for example, by combining
\ent{performed with} and \ent{shot by} into one cluster (which could well be
due to ``shot by´´ also having a meaning for movies and videos).

For more robust output, one should adopt more advanced algorithms
like {\em soft clustering} or {\em hierarchical clustering}. The latter has been
used by prior works \cite{DBLP:journals/jair/YatesE09,DBLP:conf/emnlp/MinSGL12,DBLP:conf/cikm/GalarragaHMS14}. 
It allows picking a suitable number of
clusters by principled measures like the Bayesian Information Criterion.
For efficiency, the actual clustering can be preceded by simpler
{\em blocking techniques} (cf. Section \ref{sec:entity-matching}), 
such that the more expensive
computations are run only for each block separately.

Another, more far-reaching, extension is to cluster properties P
and SO argument pairs {\em jointly} by using {\em co-clustering}
methods. Such approaches have been developed, for example, by \cite{DBLP:conf/semweb/PujaraMGC13,DBLP:conf/comad/PalHW20}.
These methods do not perform the pre-processing for SO pairs,
but instead integrate the clustering of entities and/or SO phrases
with the clustering of properties.
The factorization method discussed in the following section can also
be seen as a kind of co-clustering.
Another variation of this theme is the work of 
\cite{DBLP:journals/pvldb/NguyenATTW17} where Open IE triples are
first organized into a graph and then refined towards canonicalized
representations by iteratively merging nodes.
This approach has been further enhanced by
integrating property canonicalization with
entity linking \cite{DBLP:journals/pvldb/LinLXLC20}.

\subsection{Latent Organization of Properties}
\label{ch7-subsec:latentschema}

An alternative to explicit grouping of property phrases,
with type signatures, synonymy and subsumption,
is to organize all triples from Open IE into a latent
representation, using {\bf matrix or tensor factorization}.
These approaches take their inspiration from the
great success of such models for recommender systems \cite{DBLP:journals/computer/KorenBV09,DBLP:reference/sp/KorenB15,DBLP:journals/ftir/ZhangC20}.
A typical recommender setting is is to start with a user-item matrix
whose entries capture the purchases or likes of online users regarding
product items, or numeric ratings from product reviews.
Naturally, this matrix is very sparse, yet it latently reflects
the users' overall tastes; hence its usage to generate recommendations
for new items. The input matrix is factorized into
two low-rank matrices such that their product approximates
the original data but in a much lower-dimensional space of rank $k$.
One factor matrix maps users into this latent space, and the other one maps items.
By re-combining user and item representations in the joint latent space,
we compute how well a user may like an item that she has never seen before.
This is also referred to as {\em matrix completion} as it fills in the
initially unknown values of the sparse input matrix.

An approach to property organization along similar lines has been
developed by 
\citet{DBLP:conf/naacl/RiedelYMM13} 
and \citet{DBLP:conf/cikm/YaoRM13}
under the name {\bf Universal Schema}.
The basic idea is to arrange the collected Open IE triples into 
a matrix and then apply low-rank factorization to map both SO pairs
and P phrases into the same latent space. 

\begin{mdframed}[backgroundcolor=blue!5,linewidth=0pt]
\squishlist
\item[ ] {\bf Latent Organization of SO-P Data:}\\
The input data for computing a latent representation is an $m \times n$ matrix $M$
whose rows are SO pairs (phrases, as proxies of entity pairs) and whose columns
are P phrases.\\
The output is a pair of factor matrices $U:m \times k$ and $W:k \times n$
with a specified low rank $k$, such that their product $\tilde{M} = U \times W$
minimizes a distance norm (e.g., Frobenius norm) between $M$ and $\tilde{M}$.
\squishend
\end{mdframed}

This low-rank factorization 
reduces the original data to its gist, capturing cross-talk and removing noise.
We may think of this as a {\em soft clustering} technique, as it leads
to representing each SO pair and each P phrase as a $k$-dimensional
vector of soft-membership scores. 
The rows of $U$ are latent vectors of SO pairs, and the columns of
$W$ are the vector for P phrases.
Computing the factorization requires
non-convex optimization
as 
it typically involves additional constraints and regularizers. Therefore, 
gradient descent methods are the algorithms of choice.

\vspace*{0.2cm}
\noindent{\bf Inference over Latent Vectors:}\\
The latent form of 
{Universal Schema} makes it difficult to interpret for human users,
including knowledge engineers. It still remains unclear how the Universal Schema data
can be incorporated for KB curation and long-term maintenance.
Nevertheless, there are valuable use cases for harnessing the latent data.

To determine if two P phrases denote the same property, %
we compare their latent vectors, for example, by the scalar product or cosine
of the respective columns of $W$.
Analogously, we can test if two SO pairs denote the same entity pairs.
This is not a crisp true-or-false solution, yet it is a mechanism for
interpreting the Open IE data under a {\em universal but soft ``schema''}.
Moreover, it is a building block for other downstream tasks, particularly,
testing or predicting if a given property would likely hold between
two entities.
This resembles and is closely related to {\em Knowledge Graph
Embeddings} \cite{DBLP:journals/pieee/Nickel0TG16,DBLP:journals/tkde/WangMWG17,DBLP:journals/corr/abs-2002-00388,DBLP:conf/iclr/RuffinelliBG20}, to be discussed in Section \ref{ch8-sec:kgembeddings}.

A drawback of matrix factorization models is that they are
inherently tied to the underlying input data. So any inference we perform
in the latent space is, strictly speaking, valid only for the original triples
and not for any newly seen data that we discover later.
However, there are workarounds to this issue. Most widely used is the
so-called fold-in technique: when we spot a previously unknown P phrase
in a new set of triples, we organize it as if it were an $n+1^{st}$ column 
for the data matrix $M$, capturing all its co-occurrences with SO pairs. 
The multiplication $W \times M_{\ast,n+1}$ of the factor matrix $W$ with
this additional vector maps the vector into the $k$-dimensional space.
From here, now that we have a latent vector for the new phrase, we can
compute all comparisons as outlined above.

There is still a problem if the new P phrase comes with SO pairs where
an S or an O are not among the entities captured in the original matrix $M$.
The ``row-less'' Universal Schema method \cite{DBLP:conf/eacl/McCallumNV17} adds techniques to
handle property phrases whose observations include newly seen
entities (``out-of-KB'' entities if $M$ is viewed as a pseudo-KB; see also \cite{DBLP:conf/ijcai/JainMMC18}).

\vspace*{0.2cm}
\noindent{\bf Extensions:}\\
The Universal Schema method has been extended in a variety of ways.
Instead of treating SO pairs as one side of an input matrix, one
can arrange all data into a  {\em third-order tensor} with modes
corresponding to S, P and O phrases. This tensor is decomposed
into low-rank matrices which yield the latent representations of
S arguments, P phrases and O arguments (see, e.g., \cite{DBLP:conf/cikm/ErdosM13,DBLP:conf/emnlp/ChangYYM14}).

Another extension 
is to combine inputs from Open IE triples and
a pre-existing KB, where the KB is confined to pre-specified properties,
the triples may contain new property types unknown so far, and both
inputs overlap in the 
entities 
that they cover \cite{DBLP:conf/eacl/McCallumNV17}.
In a similar vein, multilingual inputs can be combined into a joint matrix
or tensor, for higher coverage \cite{DBLP:conf/naacl/VergaBSRM16}. 
This exploits
co-occurrences of SO pairs with property phrases in different languages.

Finally, more recent works on Universal Schema have devised 
probabilistic and neural models
for learning the latent representations of S, P and O phrases \cite{DBLP:conf/eacl/McCallumNV17,DBLP:conf/naacl/ZhangMLDM19},
replacing the matrix/tensor factorization machinery.
Essentially, a classifier is trained for P phrases to denote one of
$n$ existing property types or a new, previously unseen, property.
The input for learning the classifier is the vectors (and additional
embeddings) for the S, P and O components of triples from both
prior KB and Open IE collection.
The objective is to learn an ``entity-neighborhood-aware'' encoder
of property phrases.
When the trained model receives a new triple of SPO phrases, it 
makes the classification decision, as mentioned above.
P phrases which are most likely none of the KB properties
become (still noisy) candidates for extending the KB schema
at a later point.

\clearpage\newpage
\section{Take-Home Lessons}

Major conclusions from this line of research are the following:
\squishlist
\item Open information extraction, from any given text collection
without prior KB,
is a powerful machinery to {\em enhance recall}
towards better coverage of KB construction. It is able to discover new properties,
but this often comes with substantial noise, adversely affecting precision.
\item Using {\em distant supervision} from known properties in a {\em prior KB}
can help to improve the precision of property detection,
especially when tapping semi-structured web sites.
\item {\em Canonicalizing} the resulting SPO triples,
especially the newly seen property phrases, for inclusion in a high-quality KB
still poses major challenges. Methods based on distributional-similarity cues
employ {\em clustering} or {\em data mining} techniques, either to construct {\em paraphrase
dictionaries} or for 
cleaning of Open IE collections.
\item Organizing SPO triples from Open IE in a {\em latent low-rank space} is an
interesting alternative to explicit schemas, but latent ``schemas''
are not easily interpretable. The role of such approaches in 
KB curation and long-term maintenance 
is open.
\squishend

\clearpage\newpage
\chapter{Knowledge Base Curation}
\label{chapter:KB-curation}

Although the KB construction methods presented in
the previous chapters aim for high-quality output,
no KB will be anywhere near perfect.
There will always be some errors that result
in incorrect statements, and complete coverage
of everything that the KB should have 
is 
an elusive goal.
Therefore, additional {\bf quality assurance} and
{\bf curation} is inevitable for a KB to 
provide,
maintain and enhance its value.
This aspect is of utmost importance as
the KB grows and evolves over time, with deployment
in applications over many years.

We start this chapter by introducing 
quantitative metrics for {\em KB quality}
(Section \ref{ch8-sec:quality-assessment}) 
and the degree of {\em completeness}
(Section \ref{ch8-sec:knowledgebase-completeness}).
A key instrument for quality assurance is
logical {\em constraints} and {\em rules} 
(Section \ref{sec:intensional}), as well as their latent
representations via {\em graph embeddings} 
(Section \ref{ch8-sec:kgembeddings}).
Consistency constraints, in particular,
are crucial for detecting and pruning false positives
(i.e., spurious statements) and thus
facilitate the {\em KB cleaning} process
(Section \ref{ch8-sec:consistency-reasoning}).
We conclude this chapter by discussing
key aspects in the long-term {\em life-cycle} of
a KB, including provenance tracking,
versioning and emerging entities
(Section \ref{ch8:sec:kb-life-cycle}).

\section{KB Quality Assessment}
\label{ch8-sec:quality-assessment}

\newcommand{\GT}{\mathit{GT}}

\subsection{Quality Metrics}
The degrees of correctness and coverage of a KB are 
captured by the metrics of 
{\bf precision} and {\bf recall}, analogously to
the evaluation of classifiers.
Precision and recall matter on several levels: 
for a set of entities that populate a given type, 
for the set of types known for a given entity, 
for the set of property statements about a given entity,
or for all entities of a given type. 
For all these cases, assume that the KB
contains a set of statements $S$
whose quality is to be evaluated, 
and a ground-truth set $\GT$
for the respective scope of interest.

\begin{mdframed}[backgroundcolor=blue!5,linewidth=0pt]
\squishlist
\item[ ] \textbf{Precision} is the fraction of elements in $S$ that are also in $\GT$:
$$precision(S) = 
\frac
{S\cap \GT}
{S}$$
\squishend
\end{mdframed}
Precision may also be referred to, in the literature, 
by the
terms accuracy, validity or correctness
(with some differences in the technical detail).

\begin{mdframed}[backgroundcolor=blue!5,linewidth=0pt]
\squishlist
\item[ ] \textbf{Recall} is the fraction of elements in $\GT$ that are also in $S$:
$$recall(S) = 
\frac
{S\cap \GT}
{\GT}$$
\squishend
\end{mdframed}
Recall may also be referred to by the terms completeness or coverage.

Knowledge extraction methods, from both semi-structured
and textual contents, 
usually yield {\em confidence scores} for
their output statements.
To compute precision and recall for confidence-annotated statements, one needs to choose a 
threshold above which statements are retained and below which statements are discarded. 
For example, for a binary classifier, we could
use the odds of its scores for accepting versus
rejecting a candidate statement as a 
confidence value. 
For advanced machine-learning methods,
this can be much more sophisticated, though.
Properly calibrating confidence scores for
human interpretability poses difficulties
depending on the specific choice of the learner
(see, e.g., \cite{DBLP:conf/icml/CaruanaN06}).
Moreover, some methods, such as MAP inference
for probabilistic graphical models \cite{KollerFriedman2009} (cf. Section \ref{ch5-subsec:probabilisticgraphicalmodels}),
produce only joint scores
for entire sets of statements, which cannot
be easily propagated to individual statements.
Computing marginal probabilities 
is often prohibitively expensive
for such joint inference methods.

To avoid the difficult and highly
application-dependent choice of an adequate
threshold, a widely used approach is
 to compute precision and recall for 
 a range of threshold values, and to inspect the resulting \textbf{precision-recall-tradeoff-curve}. 
This curve can be aggregated into a single metric
as the \textbf{area under the}
\textbf{curve (AUC)}, also known as
{\bf average precision}, or alternatively, the area under the receiver-operating-characteristic curve (ROC).
Also, for each of the points where we know
precision and recall, we can compute their
harmonic mean, and use the best of these values
as a quality metric. This is known as\\
\vspace*{0.1cm}
\hspace*{1.0cm}{\bf F1 score} = {\Large %
$\frac{2}{(1/precision) + (1/recall)}$}
\vspace*{0.1cm}

Besides these standard measures, 
further metrics of interest are the 
{\em density} and {\em connectivity} of the KB:
the number of statements per entity and the
average
number of links to other entities.
Also, the number of distinct property types for
entities (of a given type) can be interpreted
as a notion of {\em semantic richness}.
{\em Freshness} with regard to capturing
up-to-date real-world facts is another important dimension.
More discussion on KB quality measures
can be found in a survey by 
\citet{DBLP:journals/semweb/Paulheim17}
and in
\cite{DBLP:journals/semweb/FarberBMR18,DBLP:conf/wikis/PiscopoS19,DBLP:series/ssw/HeistHRP20}. 
For taxonomic knowledge, that is, the hypernymy
graph of types (aka. classes), specific measures
have been studied
(e.g., \cite{bordea2016semeval}).

\subsection{Quality Evaluation}

The definitions of precision and recall require ground truth $\GT$, such as 
all songs by Bob Dylan or all artists who covered
Bob Dylan.
Since obtaining comprehensive ground truth is often impossible, 
evaluating precision and recall usually resorts to \textbf{sampling}.

\vspace*{0.2cm}
\noindent{\bf Evaluating Precision:}\\
For precision, this entails selecting a 
(random) subset of elements from the KB, and for each of them, deciding whether it is correct or not. 
For example, we could pick a few hundred statements
for a given property type uniformly at random
and have them assessed \cite{DBLP:journals/ai/HoffartSBW13}.
More sophisticated sampling strategies for KBs
are discussed in \cite{DBLP:journals/pvldb/GaoLXSDY19}.

The assessment is by human judgement, which
may come from crowdsourcing workers or, if needed, domain experts
(e.g., for biomedical knowledge). 
Such gathering of ground-truth data
requires care and effort in terms of annotation guidelines and quality control, especially if relying on laypeople on crowdsourcing platforms.
Also, assessments often have inherent uncertainty, 
as the annotators may misjudge some statements
depending on their personal background and 
(lack of) thoroughness.
By evaluating different subsets of the sampled
data points,
the process can also serve to approximate
precision-recall curves \cite{sabharwal2017good}.

A common scheme is to have multiple annotators judge the same samples, and to consider 
assessments only as ground-truth if the
inter-annotator agreement is sufficiently high.
In this process, some annotators may be dismissed
and the agreement and outcome could be based on
annotator-specific confidence weights, performing
a form of weighted voting
(\citet{DBLP:conf/wsdm/TanAIG14}).
The
overviews  
\cite{DBLP:journals/cacm/DoanRH11,DBLP:journals/tkde/ChittilappillyC16,DBLP:series/synthesis/2019Alonso} give
more background on handling crowdsourcing tasks.

\vspace*{0.2cm}
\noindent{\bf Evaluating Recall:}\\
For recall, evaluation based on sampling is 
more difficult, 
as the key point here is coverage with the goal
of capturing {\em all} knowledge of interest.
A common approach is to 
have human annotators hand-craft small collections
of ground-truth, for example, all songs by Bob Dylan
or all movies in which Elvis Presley starred.
In contrast, for evaluating the quality of an automated extraction
method and tool, annotators have to read an entire
text or web page and mark up all statements that
would ideally be captured for a KB.

Obviously, both tasks are labor-intensive and do not scale
well. Therefore, a proxy technique is to 
evaluate a KB or extractor output relative to
another pre-existing KB (or dataset)
which is known to have
high recall with regard to a certain scope
(e.g., songs and movies of famous musicians)
\cite{DBLP:journals/semweb/Paulheim17}.

\subsection{Quality Prediction}
\label{ch8-subsec:qualityprediction}

Often, the sources from which a KB is built 
cannot be easily characterized in terms of confidence.
This holds, for example, for collaboratively
created knowledge bases where many users enter and
edit statements (e.g., Wikidata). 
For automatic extraction, on the other hand, it
is often desirable to estimate the quality that
a certain extractor obtains from a certain source.
This is a building block for assessing the
resulting KB statements, and it is also useful
as an a-priori estimator 
before paying the cost of actually
running the extractor at scale. 
These settings call for {\bf predicting}
the output quality based on features of the
underlying sources and methods.

For the case where KB content is created
by online users,
three kinds of signals can be used: 
\squishlist
\item[1.]  {\em user features}, such as general expertise and the number of previous contributions, 
\item[2.] features of the {\em KB subject or statement} under
consideration, such as popularity or controversiality, and 
\item[3.]
{\em joint features} of the user and the subject 
\cite{DBLP:conf/wsdm/TanAIG14,heindorf2016vandalism}
(e.g., user living in a country on which she adds KB statements). 
\squishend
A finding in \cite{DBLP:conf/wsdm/TanAIG14} is that the third group of features is most important: quality of contributions correlates more with topic-specific user expertise 
than with general user standing or the popularity of the topic. 
These findings are based on empirical data from
editing the Freebase KB.
On the other hand, \cite{heindorf2016vandalism} reports that for the Wikidata KB,
user features, such as the number of edits and the
user status, are indicative for quality.
Other studies have looked at the edit history of
Wikipedia (e.g., \cite{DBLP:conf/icwsm/LiebermanL09,DBLP:conf/www/Kumar0L16,DBLP:conf/emnlp/HuaDTTSD18}).
All the above features can contribute to 
counter vandalism and other kinds of quality-degrading
edits.

\vspace*{0.2cm}
\noindent{\bf Knowledge Fusion:}\\
For the case of using automatic extractors,
a key issue is to compare and consolidate
outputs from different sources and methods,
like statements obtained from DOM trees, from web tables,
from query logs, and from text sources,
or statements from different web sites on the same topic.
By quantifying the quality of each 
{\em source/extractor
combination}, their outputs can be given adequate
weights in the corroboration process.
The {\bf knowledge fusion} method of
\citet{Dong:KDD2014,DBLP:journals/pvldb/DongGMDHLSZ15}
modeled and quantified the mutual 
reinforcement between output quality and source/extractor quality. The output statements have higher quality
if they come from a better source and extractor,
and the source and extractor have higher quality
if they yield more correct statements.
This line of arguments is related to 
the principle of statement-pattern duality
(see Sections \ref{ch4:subsec:PatternLearning} and \ref{ch6-subsec:statementpatternduality})
as well as to
joint learning over probabilistic factor graphs.
We discuss this direction further in Section \ref{ch8-sec:consistency-reasoning}.
A key finding from \cite{Dong:KDD2014,DBLP:journals/pvldb/DongGMDHLSZ15} is that the best quality was obtained
from DOM trees (see Sections \ref{ch6:subsubsec:extraction-from-dom-trees} and \ref{ch7-openie-semistructured}), whereas both query logs and web tables provided
rather poor quality due to high noise.

\vspace*{0.2cm}
\noindent{\bf Evidence Collection:}\\
The knowledge fusion techniques can also be
used to validate or refute KB statements collected
from multiple sources.
This is particularly challenging for knowledge
about long-tail entities.
To this end, \cite{DBLP:journals/pvldb/LiDLL17}
devised strategies for collecting evidence
in support of candidate statements or
counter-evidence for falsification.
The compiled pieces of evidence are then subject
to jointly assessing the quality of sources
and statements, following the outlined idea
of knowledge fusion.

\cite{DBLP:journals/pvldb/LiDLL17} developed
a complete fact-checking tool called {\em FACTY},
successfully used for KB curation.
Section \ref{ch9-sec:industrialKG} 
gives more details on FACTY
as part of an industrial KB infrastructure.
Other fact-checking methods have been studied,
for example, by
\cite{DBLP:journals/tkde/YinHY08,DBLP:conf/icde/LiMY11,DBLP:conf/acl/NakasholeM14,DBLP:conf/wsdm/Gad-Elrab0UW19}; 
an overview on this topic is given
by \cite{DBLP:journals/sigkdd/LiGMLSZFH15}.

\vspace*{0.2cm}
\noindent{\bf Predicting Recall:}\\
Again, recall is more tricky and tedious to estimate.
Prior work on text-centric query processing
\cite{DBLP:journals/tods/IpeirotisAJG07} investigated
selectivity estimators, but these do not carry over
to KB quality.
Viable approaches exist for special cases, though.
For predicting the \textbf{recall of
types}, that is, the instances of a semantic class
such as folk songs or Grammy award winners,
statistical methods for {\em species sampling} can be used~\cite{DBLP:journals/tkde/TrushkowskyKFSR15,luggen2019non}:
given a set of birdwatchers' partial counts of
different species in the same habitat, what is
(a good statistical estimator for)
the population size for each of the species?
In the KB setting, we would have to sample counts
for class instances from different sources, and
then estimate the total number of distinct instances.
\cite{luggen2019non} shows that
the growth history of the Wikidata KB can be used to derive multiple samples, and the overlap between these samples yields estimates for the size of semantic classes.
Other statistical techniques along similar lines
include using Benford's Law on the
distribution of class cardinalities
\cite{DBLP:conf/semweb/SouletGMS18}.

An alternative to these intrinsic predictions
are extrinsic studies and coverage models
on how well KBs support
typical workloads of queries and questions
\cite{DBLP:conf/naacl/HopkinsonGPM18,Razniewski:CIKM2020}.

\vspace*{0.2cm}
\noindent{\bf Textual Cues for Cardinalities and Recall:}\\
Sometimes, there are ways to estimate the cardinality
of a semantic class or the cardinality of the
distinct objects for a given subject-property pair,
even without being able to extract the actual entities.
In particular, text sources often contain cues such as:

\begin{samepage}
{\small
\lstset{language=HTML,upquote=true}
\begin{lstlisting}
Bob Dylan released 51 albums
Bob Dylan's works include 39 studio and 12 live albums
On his 51st album Dylan covered Sinatra songs
Frank Sinatra and his first daughter Nancy
Frank Sinatra and his second daughter Tina
There are more than two thousand Grammy award winners
\end{lstlisting}
}
\end{samepage}

These sentences contain 
numbers, numerals (i.e., text expressions that
denote numbers) and ordinal phrases (e.g., ``51st'', ``second'').
From these we can infer cardinalities or at least
lower and upper bounds. 
Adding this knowledge, in the form of
{\bf counting quantifiers}, enhances the KB.
The extraction is not straightforward, though.
On one hand, there is the diversity of surface expressions;
on the other hand, there are overlapping and noisy cues.
For example, we must not over-count albums by
adding up studio albums, live albums, albums from
the 20th century and albums from the 21st century.
\cite{DBLP:conf/acl/MirzaRDW17} presents techniques for extraction and
consolidation of such cardinalities.

Counting quantifiers can be used to assess recall by comparing a confirmed count (e.g., for the number of
Bob Dylan albums) against the KB instances of the respective set. 
This can in turn reveal gaps in the KB, and can
steer the knowledge gathering process to specific
enrichment of the KB contents 
\cite{DBLP:conf/semweb/MirzaRDW18}.

The textual form of expressing lists of objects for
a given subject-property pair can also give cues
about recall.
For example, the sentence
``Dylan's favorite instruments used to be guitar
and harmonica'' suggests that there could be more instruments,
whereas the sentence ``Bob Dylan has played
guitar, harmonica, banjo, piano and organ'' implies
high recall or even completeness.
\cite{razniewski2019coverage} presents methods for
leveraging such textual cues towards recall estimation,
based on linguistic theories of communication.

\section{Knowledge Base Completeness}
\label{ch8-sec:knowledgebase-completeness}

No knowledge base can ever be fully complete.
This suggests that, unlike database, which are
traditionally interpreted under a {\em Closed-World Assumption (CWA)},
we should treat KBs as following an 
{\em Open-World Assumption (OWA)}:

\begin{mdframed}[backgroundcolor=blue!5,linewidth=0pt]
\squishlist
\item[ ] The \textbf{Open-World Assumption (OWA)}
postulates that if a statement is not in the KB, it may
nevertheless be true in the real world.
Its truth value is unknown.
In other words, absence of evidence is not evidence of absence.
\squishend\end{mdframed}

Whenever we probe a statement and the KB does not have that
statement (and neither a perfectly contradictory statement), we assume that
this statement may be true or may be false -- we just do not know
its validity 
\cite{DBLP:conf/akbc/RazniewskiSN16}.
For example, even when a rich KB contains 500 songs by Bob Dylan and
1000 artists who covered him, Boolean queries such as 
``Has Bob Dylan written the anthem of Vanuatu?'' or
``Has Bob Dylan ever been covered by Woody Guthrie?''
have to return the answer ``maybe'' (or ``don't know''). 

This is at least
in the absence of knowing a different composer of the Vanuatu anthem,
and not being able to rule out a covering artist because he or she
died already before Dylan was born.
The answer cannot be ``no'' even if both statements are sort of absurd.

\subsection{Local Completeness}

Notwithstanding the general OWA, we are often able
to observe and state that a KB is {\em locally complete}.
For example, a human knowledge engineer may assert that the
instances of a specific type, such as \ent{Nobel laureates}
or \ent{Grammy winners}, are complete. Then all queries of the
form ``Did Elvis Presley win the Nobel Prize?'' can faithfully
return the answer ``no''.

This idea can be generalized to other localized settings, most notably,
to the set of objects that would be in the range of a given subject-property pair.
For example, the KB could have a complete list of Bob Dylan albums,
perhaps even a complete list of his songs, but only a partial set of artists
who covered him.
Formally, we specify that the sets\\
\hspace*{0.5cm} $\{O | \langle \ent{Bob~Dylan}, \ent{released Album}, O\}$
and\\
\hspace*{0.5cm} $\{O | \langle \ent{Bob~Dylan}, \ent{composed Song}, O\}$\\
are complete. 
This helps querying under CWA semantics for the feasible queries, and it helps
the KB curation process, as we know that further triples for these SO pairs
that human contributors or automatic extractors may offer
should not be accepted. 

The principle that underlies these considerations is the following
(\citet{DBLP:conf/www/GalarragaTHS13,Dong:KDD2014}):

\begin{mdframed}[backgroundcolor=blue!5,linewidth=0pt]
\squishlist
\item[ ] The \textbf{Local Completeness Assumption (LCA)},
aka. {\em Partial Completeness Assumption} or 
{\em Local Closed-World Assumption}, 
for entity $S$ in the KB asserts that if, for some
property $P$, the KB contains at least one 
statement $\langle S,P,O \rangle$, then it contains
all statements for the same $S$ and $P$.
That is, there exists no object $X$ for which 
$\langle S,P,X \rangle$ holds in the real world
but is not captured in the KB.\\
For example, if we know one of a person's children,
then we know all of them; $S$ is complete with
regard to $P$.
\item[ ] A generalized form of the LCA that
applies to {\em entire classes} asserts that
for all entities $S$ of a given type $T$
and a given property $P$, the set of objects
for $SP$ is complete.
\squishend
\end{mdframed}

The rationale for the LCA is that salient
properties of important entities will be 
covered completely, throughout the KB maintenance process,
or not at all. 
Hence the LCA for properties like children,
types like Nobel laureates, and subjects like
premier league football clubs.
\cite{DBLP:conf/wsdm/GalarragaRAS17}
conducted empirical studies on how
well the LCA holds for large KBs like Wikidata.
The studies did find gaps (e.g., only subsets of children known), but by and large, the LCA is often
a valid assumption. 
More precisely, this holds for relations in the direction that is closer to being functional
(i.e., with smaller range of distinct objects per subject).
For example, the LCA is empirically supported for
\ent{hasCitizenship:} {\em person $\times$ country},
but not for its inverse relation
\ent{hasCitizen:} {\em country $\times$ person}.

Of course, for $SP$ object sets that may still grow over time, the LCA has to be
checked again 
every now and then.
But there are many cases with immutable sets
where a {\em completeness guarantee} would indeed freeze the relevant part of the KB.
Examples are the parents of people, 
the founders of companies,
or 
movies by directors who are already dead.
Instead of the complete set being the objects for an SO pair, we can also
consider the situation with a set of subjects for a given PO pair.
Furthermore, the local completeness could be conditioned on an additional
property, most notably, on the type of the S entities.
For example, we could assert completeness of parents and children for
all politicians or all EU politicians, but perhaps not for artists
(who may have a less documented life -- at least by the clich\'e).

\vspace*{0.2cm}
\noindent{\bf Completeness Assertions:}\\
Starting from seminal database work \cite{motro1989integrity}, proposals have been made to extend knowledge bases with \emph{completeness statements}, to assert which parts of the data should be treated under LCA semantics.
By default, all other parts would then be potentially
incomplete, to be treated under OWA semantics.
Formal languages and patterns for
expressing such assertions have been investigated,
for example, by \cite{DBLP:conf/sigmod/RazniewskiKNS15,DBLP:journals/tweb/DarariNPR18,langnehme}
for relational databases and for the RDF data model.

The RDF standard itself can not express these assertions, though. Instead, we can turn to the Web Ontology Language
OWL 2 \cite{W3C:OWL2}, which has different ways of specifying
local completeness.
The following are OWL examples for asserting that i) there are only fourteen eight-thousander mountains,
and that
ii) people have at most one birthdate:

{\small
\lstset{language=HTML,upquote=true}
\begin{lstlisting}
i) :classOfEightthousanders owl:oneOf 
     ( :Everest :K2 :Kangchenjunga ... :Shishapangma )
ii) :people rdfs:subclassOf [rdf:type owl:Restriction;
     owl:maxcardinality "1"^^xsd:nonNegativeInteger;
     owl:onProperty :birthdate]
\end{lstlisting}
}

The second example can be generalized.
Asserting local completeness
is often possible through additional properties about the
{\em cardinalities} for $SP$ object sets.
By adding statements such as\\
\hspace*{0.5cm} $\langle \ent{Bob~Dylan}, \ent{number~of~ albums}, 51 \rangle$\\
it is easy to compare the cardinality of\\
\hspace*{0.5cm} $\{O ~|~ \langle \ent{Bob~Dylan},  \ent{released ~album}, O \rangle \}$\\
against the asserted number.
A perfect match indicates local completeness.
If the object set is smaller, some knowledge is missing;
if it is larger than the stated number, some
of the statements are wrong (or the statement
about \ent{number of albums} is incorrect or stale).
A systematic study of such counting statements and their
underlying object sets has been performed by
\cite{ShresthaGosh:JWS2020}.

\subsection{Negative Statements}

The design philosphy of KBs is to store
positive statements only: facts that do hold.
However, it is sometimes also of interest
to make negative statements explicit:
statements that do not hold, despite common belief
or when they are otherwise noteworthy.
For example, we may want to explicitly assert
that {\em Ennio Morricone 
has never lived in the USA}, despite
his great success and influence in Hollywood,
or that {\em Elvis Presley has not won an Oscar}
despite starring in a number of movies and
having been the idol of an entire generation.
In addition to such fully grounded statements,
the absence of objects for a given property could be of interest, too.
For example, {\em Angela Merkel has no children}.
Having this knowledge at hand can help
question answering (e.g., to correct high-scoring
spurious answers when the proper result is the empty set)
as well as KB curation (e.g., to refute improper
insertions based on common misbeliefs).

Negative assertions are straightforward to express
in logical terms:\\
\hspace*{1.5cm} $\lnot ~ livedIn ~(Ennio~Morricone, USA)$,\\
\hspace*{1.5cm} $\lnot ~ wonPrize ~(Elvis~Presley, Academy~Award)$,\\
\hspace*{1.5cm} $\lnot ~ \exists~ O: ~ hasChild ~(Angela~Merkel, O)$.\\
The OWL 2 standard \cite{W3C:OWL2}
provides
syntax to capture such formulas.
However, specific KBs like Wikidata 
have only limited ways of expressing negative
statements in a principled and comprehensive manner.
Wikidata, for example, has statements of the form\\
\hspace*{1.5cm} $\langle$ \ent{Angela Merkel}, \ent{child}, \ent{no value}$\rangle$\\
where \ent{no value} has the semantics that
no object exists (in the real world) for this
$SP$ pair. Essentially, this asserts the LCA for
this local context, confirming that the absence
of objects is indeed the truth (as opposed to
merely not knowing any objects).
In addition, statements about counts such as\\
\hspace*{1.5cm} $\langle$ \ent{Angela Merkel}, \ent{number of children}, 0$\rangle$\\
can capture empty sets as well.
The negation of fully grounded statements
is not expressible in RDF, whereas it is straightforward in the OWL 2 language \cite{W3C:OWL2}. 

Obviously, it does not make sense to add
{\em all} (or too many) negative statements
even if they are valid.
For example, Elvis Presley did not win the
Nobel Prize, the Turing Award, the Fields Medal etc.
But these statements are not interesting, as nobody
would expect him to have these honors.
So a key issue is to identify
{\em salient negative statements}
that deserve being made explicit.

Works by 
\cite{DBLP:conf/www/BalaramanRN18,Arnaout:AKBC2020}
developed a number of
techniques to this end.
One of the approaches is to compare an
entity against its peers of similar standing,
for example, Ennio Morricone against other
famous contributors of Hollywood productions.
If we see that many of these peers have lived
in the USA, then Morricone's counterexample
is a remarkable negative statement.
Of course, the negative statement could as well be
wrong; that is, Morricone has lived in the USA
but we have missed out on this fact so far.
To rectify this situation, the LCA can be asserted
for properties of interest, allowing only the prediction of negative statements if at least one positive statement is present.
This high-level idea can be made quantitative
based on statistical arguments, leading to
a ranked list of salient negative statements
for a given entity (see \cite{Arnaout:AKBC2020}
for details).

\section{Rules and Constraints}
\label{sec:intensional}

So far, we have considered only atomic SPO statements such as {\ent{married (ElvisPresley, PriscillaPresley)}}. An important asset for curation are 
logical patterns, or ideally {\bf invariants}
in the KB, such as:\\
\hspace*{0.5cm} {\em Every child has two parents (at the time of birth).}\\
\hspace*{0.5cm} {\em Mothers are (mostly) married to the fathers of their children.}\\
\hspace*{0.5cm} {\em Only humans can marry.}\\
Such invariants are called \emph{intensional knowledge}. There are different ways to make use of intensional knowledge.

\begin{mdframed}[backgroundcolor=blue!5,linewidth=0pt]
\squishlist
\item[ ] {\bf Constraints} define invariants that
must be satisfied by the KB to be 
{\em logically consistent} and fulfilling a necessary
condition for being {\em correct}.
For example, whoever is married must be a human person. 
If the KB knows about Elvis's marriage but does not have him in the class \ent{person} yet, we flag the KB as inconsistent. This is the purpose of constraint languages such as SHACL, discussed in Section~\ref{sec:constraints}.
\squishend
\end{mdframed}

\begin{mdframed}[backgroundcolor=blue!5,linewidth=0pt]
\squishlist
\item[ ] {\bf Rules} define a calculus to deduce
additional statements, to complete the KB 
and make it logically consistent.
For example, if the KB misses out on Elvis being a 
\ent{person}, a rule could add him to this class
upon the premise that there is a marriage statement
about him.
This is the purpose of logical rule languages such as OWL, discussed
in Section~\ref{sec:rules}. These languages can also be used to enforce constraints, although with some restrictions.
\squishend
\end{mdframed}

\begin{mdframed}[backgroundcolor=blue!5,linewidth=0pt]
\squishlist
\item[ ] {\bf Soft rules} express plausibility 
restrictions, holding for most cases but tolerating
exceptions. For example, the pattern that
mothers are married to the fathers of their children
certainly has exceptions, but would still capture
the common case.\\
Such rules can serve as either {\em soft constraints},
to detect implausible content in the KB and
flag it for curation, or
{\em soft deduction rules}, to infer additional
statements that are likely valid and could
expand the KB.\\
Soft rules can be derived from logical patterns
in the KB.
This is the purpose of {\bf rule mining}
algorithms, discussed in Section~\ref{sec:rule-mining}.
\squishend
\end{mdframed}

Hard and soft constraints are a powerful instrument
for KB cleaning: identifying implausible and dubious
statements, and resolving conflicts and gaps in the
KB to establish logical consistency.
Methods for this purpose are discussed in
Section~\ref{ch8-sec:consistency-reasoning}.

\subsection{Constraints}\label{sec:constraints}

Consistency constraints {\em prescribe}
invariants in the KB: 
people must have a birth date, people cannot have more than one birth date, etc.

There is a wide spectrum of invariants that we could consider. 
The following lists some useful templates, by examples.
\begin{itemize}
\item {\bf Type constraints}, e.g.:\\
$\forall x,y: composed(x,y) \Rightarrow type(x,musician)$ \\
(domain of the \ent{composed} relation) and\\
$\forall x,y: composed(x,y) \Rightarrow type(y,song)$ \\
(range of the \ent{composed} relation)
\item {\bf Value constraints}, e.g.:\\
$\forall x,v: type(x,basketballer) \land height(x,v) \Rightarrow v < 250cm$
\item {\bf Relation restrictions} like symmetry or transitivity, e.g.:\\
$\forall x,y: spouse(x,y) \Rightarrow spouse(y,x)$ \\
(symmetry of the \ent{spouse} relation)
\item {\bf Functional dependencies}, e.g.,:\\
$\forall x,y,z: birthplace(x,y) \land birthplace(x,z) \Rightarrow y = z$ \\
(the \ent{birthplace} relation is a function)
\item {\bf Conditional functional dependencies}, e.g.:\\
\mbox{$\forall x,y,z: citizen(x,y) \land citizen(x,z) \land y=Germany \Rightarrow y=z$} \\
(Germany does not allow dual citizenships)
\item {\bf Inclusion dependencies}, e.g.: \\
$\forall x,y: type(x,singer) \Rightarrow type(x,musician)$
\end{itemize}
More advanced kinds of constraints include:
\begin{itemize}
\item {\bf Disjointness constraints}, e.g.: \\
$\forall x,y: type(x,weightlifter) \Rightarrow \lnot type(x,ballerina)$ and\\
$\forall x,y: type(x,ballerina) \Rightarrow \lnot type(x,weightlifter)$
\item {\bf Existential dependencies}, e.g.:\\
$\forall x: type(x,scholar) \Rightarrow$ \\ 
$\exists y: \left( published (x,y) \land type(y,\textit{scientific}~article) \right)$
\item {\bf Temporal dependencies}, e.g.:\\
$\forall x,y,z,s,t: marriage(x,y,s) \land marriage(x,z,t) \Rightarrow$ \\
$( y=z \lor \lnot overlaps(s,t))$ \\
where \ent{marriage} is a ternary relation and
$s$ and $t$ are time intervals during which the marriages last. %
 \end{itemize}

Some of these invariants may appear too strict, given that real life holds many
surprises. For example, there could be exceptions for 
the mutual exclusion of being a weightlifter and being a ballerina.
It is up to the scope and purpose of the KB whether we want to rule out such exceptions or not.
Constraints that cover the prevalent case but do tolerate
exceptions are very useful; they can be used to flag suspicious
statements in a KB and prompt a human curator
(see Section \ref{ch8-sec:consistency-reasoning}).

Database languages like SQL support the declarative specification of
constraints (see textbooks such as \cite{DBLP:books/daglib/0011318,DBLP:books/mk/ONeilN00}). 
However, knowledge bases have mostly adopted Web standards.
The W3C standard language for expressing constraints is the {\bf Shapes Constraint Language (SHACL)} \cite{shacl}. Here is an example:

\begin{lstlisting}
:Person
	rdf:type owl:Class, sh:NodeShape ;
	:property [
		sh:path :hasMother ;
		sh:maxCount 1 ;
		sh:Class :Person ;
	] .
\end{lstlisting}
This code specifies that every instance of the
type \ent{Person} has at most one entity in the
range of the \ent{hasMother} relation, disallowing
more than one biological mother.
To enforce exactly one mother in the KB,
an analogous shape constraint with \ent{minCount}
could be added. However, a SHACL validation tool
would then raise an inconsistency flag if
some people miss the respective statements
for \ent{hasMother}.
Therefore, to allow for leeway in the KB growth
process, with people being added without
complete properties, the \ent{maxCount}
constraint only would be a typical 
design choice.

More advanced features of SHACL allow specifying constraints on strings (such as length restrictions
and compliance with regex patterns) for attributes,
value ranges for numerical properties, 
and composite constraints via Boolean operators
\cite{shacl}.
An alternative to SHACL is %
the {\bf Shape Expressions Language (ShEx)} \cite{shex}. ShEx 
comes with a validation algorithm to test whether all constraints hold in a given KB~\cite{shexsemantics}.

\subsection{Deductive Rules}\label{sec:rules}

A deductive rule can infer statements 
for addition to a KB, to make it consistent
and to enhance its coverage. 
For example, a rule can codify that the entities in a
 \ent{marriedTo} relationship must be human people;
 so \ent{Elvis Presley} would be added to the class
 \ent{person}. 
Rules may produce contradictions among their deduced statements, though.
For example, an additional rule could state that entities must not belong to both the class of persons and the class of fictitious characters.
If the KB already identifies Elvis as an instance
 of \ent{fictious character} %
 then would derive a contradiction. 
 However, unlike constraints discussed in Section~\ref{sec:constraints}, 
 this would not automatically result 
 in the explicit removal of 
 contradictory statements.

Deduction rules are expressed in appropriately chosen
fragments of first-order predicate logics, typically
trading off expressiveness versus computational complexity.
In the Semantic Web world, several logics are widely used. {\bf RDFS (= RDF Schema)} is the simplest formalism \cite{W3C:RDFS}.
It allows only limited kinds of rules: domain and range rules, subclass-of rules, and sub-property rules. 
In RDFS, we can specify that everyone who is married is a person:
\begin{lstlisting}
:marriedTo rdfs:domain :Person
\end{lstlisting}

RDFS cannot express the disjointness of classes. For this, a more powerful language has been developed: the 
{\bf Web Ontology Language OWL 2} \cite{W3C:OWL2},
which allows asserting that certain classes rule each other out.
OWL exists in several flavors, from the least expressive (and more computationally benign) to the most expressive (and computationally expensive) variant. For our purpose, the OWL 2 QL flavor is sufficient to 
specify that real-life persons and fictitious characters
are disjoint:
\begin{lstlisting}
:Person owl:disjointWith :FictitiousCharacter
\end{lstlisting}
OWL reasoners can ingest sets of
formulas to perform deduction, and flag the KB as inconsistent if a contradiction is found.
The theoretical foundation of OWL is 
{\bf description logics} 
\cite{StaabStuder2009,DBLP:books/daglib/0041477}. 
Our example can be expressed as follows:
\vspace*{0.1cm}

\noindent \hspace*{0.5cm} $FictitiousCharacter~(Elvis)$\\
\hspace*{0.5cm} $marriedTo~(Elvis, Priscilla)$\\
\hspace*{0.5cm} $\exists ~ marriedTo \sqsubseteq Person$\\
\hspace*{0.5cm} $Person \sqcap FictitiousCharacter \sqsubseteq \bot$

\noindent The OWL reasoner would detect that this has a
logical contradiction. There are many approaches that aim to repair such contradictions \cite{DBLP:series/synthesis/2011Bertossi,DBLP:conf/pods/Bertossi19}. 
One option is to identify a minimal set of axioms to remove. Alternatively, we can compute the maximal subset of instance-level statements that are still compatible with all the given axioms. In the example, we can remove either the first statement or the second. 
There are different ways to prioritize the statements, for example, by preferably keeping those with high confidence \cite{bienvenu2020short}.

\subsection{Rule Mining}\label{sec:rule-mining}
\label{ch8-subsec:rulemining}

Rule mining is the task of identifying logical 
patterns and invariants in a knowledge base. 
For example, we aim to find that people who are married usually live in the same city:
\vspace*{0.1cm}

\noindent \hspace*{0.5cm} $\forall x,y,z: ~ marriedTo(x, y) \wedge livesIn(x,z) \Rightarrow livesIn(y,z)$

\noindent or that mayors of cities have the citizenship of the respective countries:
\vspace*{0.1cm}

\noindent \hspace*{0.5cm} $\forall x,y,z: ~ mayor(x,y) \land 
locatedInCountry(y,z) \Rightarrow citizen(x,z)$
\vspace*{0.1cm}

\vspace*{0.1cm}
Rule mining can be used either for learning 
{\em constraints} to prune out spurious statements,
or for {\em rule-based deduction} to fill gaps in the KB.
Typically, the same rule is used for either one of
two purposes; this is the choice of the KB architect,
depending on where the pain points are: precision or recall.

Rules are often of soft nature, holding for
a large fraction of statements and tolerating
``minorities''. 
Thus, rule mining needs to consider metrics
for the degree of validity: 
{\em support} and {\em confidence},
as discussed below.
In addition to playing a vital part in the KB
curation process, rules can also give insight into
potential biases in the KB content and, possibly,
even the real world.
For example, a rule saying that\\
\hspace*{0.5cm} {\em actors are (usually) millionaires}\\
may arise from the
KB emphasis on successful stars and lack of
covering long-tail actors, and a rule saying that\\
\hspace*{0.5cm} {\em Nobel laureates are (mostly) men}\\ 
reflects the gender imbalance in our society.
Yet another potential use case for rule mining is to generate {\em explanations}
for doubtful facts -- in combination with
retrieving evidence or counter-evidence
from text corpora (see, e.g., 
\cite{DBLP:conf/wsdm/Gad-Elrab0UW19}).

Rule mining, as discussed in the following, is related to the task of discovering {\em approximate functional dependencies} and {\em approximate inclusion dependencies}
in relational databases (e.g., \cite{DBLP:conf/sigmod/IlyasMHBA04,DBLP:journals/pvldb/ChiangM08,DBLP:journals/pvldb/SaxenaGI19}), 
as part of the 
data cleaning pipeline or for data exploration.
The textbook by 
\citet{DBLP:books/acm/IlyasC19}
covers this topic,
including references to state-of-the-art algorithms.
Note that functional and inclusion dependencies
are special cases of logical invariants.
KB rule mining also aims at a broader, more expressive
class of logical patterns, such as Horn rules. 

\vspace*{0.2cm}
\noindent{\bf Horn Rules and Inductive Logic Programming:}\\
Unless we restrict rules to suitable fragments
of first-order logics, rule mining is bound to
run into computational complexity issues and
intractability. Therefore, the rules under
consideration are usually restricted to the
following shape:

\begin{mdframed}[backgroundcolor=blue!5,linewidth=0pt]
\squishlist
\item[ ] A {\bf Horn rule} is a first-order
predicate logic formula restricted to the format
\squishlist
\item[ ] $\forall x_1 \forall x_2 \dots \forall x_k:$\\
         $P_1(args_1) \land P_2(args_2) \land \dots \land P_m(args_m) 
         \Rightarrow Q(args)$
\squishend
where $x_1$ through $x_k$ are variables,
$P_1$ through $P_m$ and $Q$ are KB properties, and
their arguments $args1$ through $args_m$
and $args$ are ordered lists of either
variables from $\{x_1 \dots x_k\}$ or constants,
that is, entities or literal values from the KB.
The properties are often binary (if the KB adopts RDF)
but could also have higher arities.\\
The conjunction of the terms $P_1(args_1)$
through $P_m(args_m)$ is called the {\bf rule body},
and $Q(args)$ is called the {\bf rule head}.
The terms $P_i(args_i)$ and
$Q(args)$ are called {\bf atoms} (or
{\em literals}).
\\
As all quantifiers are universal (i.e., no existential
quantifiers) and prefix the propositional-logic part
of the formula (i.e., using prenex normal form),
we often drop the quantifiers and write\\
\hspace*{0.5cm} $P_1(args_1) \land P_2(args_2) \dots \land P_m(args_m)  \Rightarrow Q(args)$.\\
When normalized into {\bf clause} form
with disjunctions only,
the propositional-logics part looks like\\
\hspace*{0.5cm}$\lnot P_1(args_1) \lor \lnot P_2(args_2) \dots \lor \lnot P_m(args_m) \lor Q(args)$.\\
{\bf Horn clauses} restrict formulas to have
at most one positive 
atom.
The two examples at the begin of this subsection are Horn rules. 
\squishend
\end{mdframed}

Automatically computing such rules from extensional data
is highly related to the task of
{\bf Association Rule Mining}
\cite{DBLP:conf/vldb/AgrawalS94,DBLP:books/mit/fayyadPSU96/AgrawalMSTV96,DBLP:books/mk/HanKP2011},
for example, finding rules such as:\\
\hspace*{0.5cm} {\em Customers who buy white rum and mint leaves}\\
\hspace*{0.5cm} {\em will also buy lime juice
(to prepare Mojito cocktails).}\\
or\\
\hspace*{0.5cm} {\em Subscribers who like Elvis Presley and Bob Dylan}\\
\hspace*{0.5cm} {\em will also like Nina Simone.}\\
However, association rules, mined over transactions
and other user events, do not have the full
logical expressiveness. Essentially, they are restricted
to single-variable patterns, capturing customers,
subscribers, users etc., but no other variables.
The logics-based generalization of rules has
been addressed in a separate community, known
as {\bf Inductive Logic Programming} \cite{DBLP:journals/jlp/MuggletonR94,DBLP:journals/pvldb/ZengPP14,DBLP:journals/corr/abs-2002-11002}.
Both association rule mining and inductive logic programming face a huge combinatorial space
of possible rules. To identify the interesting ones,
quantitative measures from data mining need
to be considered:

\begin{mdframed}[backgroundcolor=blue!5,linewidth=0pt]
\squishlist
\item[ ] Given a rule $B \Rightarrow H$ over a knowledge base $KB$ and a substitution $\theta$
that instantiates the rule's variables with entities and values from $KB$, the resulting
instantiation of the rule head, $\theta(H)$, is called
a {\bf prediction} of the rule.
A prediction is 
called {\em positive}
if the grounded head is
contained in the KB, and 
{\em negative}
otherwise. 
We refer to these also as {\em positive and negative samples}, respectively.
\item[ ] The {\bf support} of a Horn rule
is the number of 
positive
predictions: \\
\hspace*{0.5cm} $support (P_1 \dots P_m \Rightarrow Q(args)) =$\\
\hspace*{0.5cm} $|\{\theta(args) | P_1 \dots P_m, Q ~and~ Q(\theta(args)) \in KB\}|$
\item[ ] The {\bf confidence} of a Horn rule is
the fraction of 
positive
predictions relative to
the sum of positive and negative predictions:\\
\hspace*{0.5cm} $confidence (P_1 \dots P_m \Rightarrow Q(args)) = ~ support ~ /$\\
\hspace*{0.5cm} $\left( support ~+~ 
|\{\theta(args) | P_1 \dots P_m, Q ~and~ Q(\theta(args)) \notin KB\}| \right)$
\squishend
\end{mdframed}

\vspace*{0.2cm}
\noindent{\bf Rule Mining Algorithms:}\\
The goal of KB rule mining is to compute
high-confidence rules, but to ensure significance,
these should also have support above a specified
threshold.
Finding rules can proceed in two ways:
bottom-up or top-down. 

{\em Top-down algorithms} start with the head of a rule and aim to
construct the rule body by incrementally adding and
refining atoms -- so that rules gradually become more specific.
{\em Bottom-up algorithms} start from the data, the KB content,
and select atoms for the body of initially special rules
that are gradually generalized.
Both of these paradigms resemble key elements of the
A-Priori algorithm for association rules \cite{DBLP:conf/vldb/AgrawalS94}, namely,
iteratively growing the atoms sets for rule bodies
(like in itemset mining), while checking for sufficiently
high support and scoring the rule confidence.
In particular, all algorithms exploit the anti-monotonicity 
of the support measure, so as to prune candidate rules
that cannot exceed the support threshold.

The following algorithmic skeleton 
outlines the {\bf AMIE}
method, by 
\citet{DBLP:conf/www/GalarragaTHS13} \cite{DBLP:journals/vldb/GalarragaTHS15}\cite{DBLP:conf/esws/LajusGS20},
for top-down rule mining
with binary predicates (i.e., over SPO triples),
using a queue for atom sets that can form rule bodies. %

\begin{samepage}
\begin{mdframed}[backgroundcolor=blue!5,linewidth=0pt]
\squishlist
\item[ ] {\bf Top-down mining of KB rules:}\\
${\rm Input:}$ head predicate Q(x,y) with variables x,y,
and the  KB\\
${\rm Output:}$ rules with support and confidence
\squishlist
\item[0.] Initialize a queue with empty set
\item[1.] Pick an entry ${\bf P}=\{P_1 \dots P_l\}$ from the queue and
          extend it by generating an additional atom $P_{l+1}$ where
\squishlist
\item[a.] $P_{l+1}$ shares one argument with those of ${\bf P}$ or $Q$
          and has a new variable for the other argument, or
\item[b.]  $P_{l+1}$ has a variable as one argument shared with those of ${\bf P}$ or $Q$
          and introduces a new constant (entity or value) for the other argument, or
\item[c.] $P_{l+1}$ shares both of its arguments with those of ${\bf P}$ or $Q$
\squishend
\item[2.] Compute confidence, support and other measures for
          the rule $P_1 \land \dots \land P_{l+1} \Rightarrow Q$ %
\item[3.] Insert atom set $\{P_1 \dots P_l, P_{l+1}\}$ into the queue \\
if support above threshold
\item[4.] Output the rule $P_1 \land \dots \land P_{l+1} \Rightarrow Q$
\item[5.] Repeat steps 1--4 \\
until
         enough rules or priority queue exhausted
\squishend
\squishend
\end{mdframed}
\end{samepage}

\begin{figure} [tbh!]
  \centering
   \includegraphics[width=0.9\textwidth]{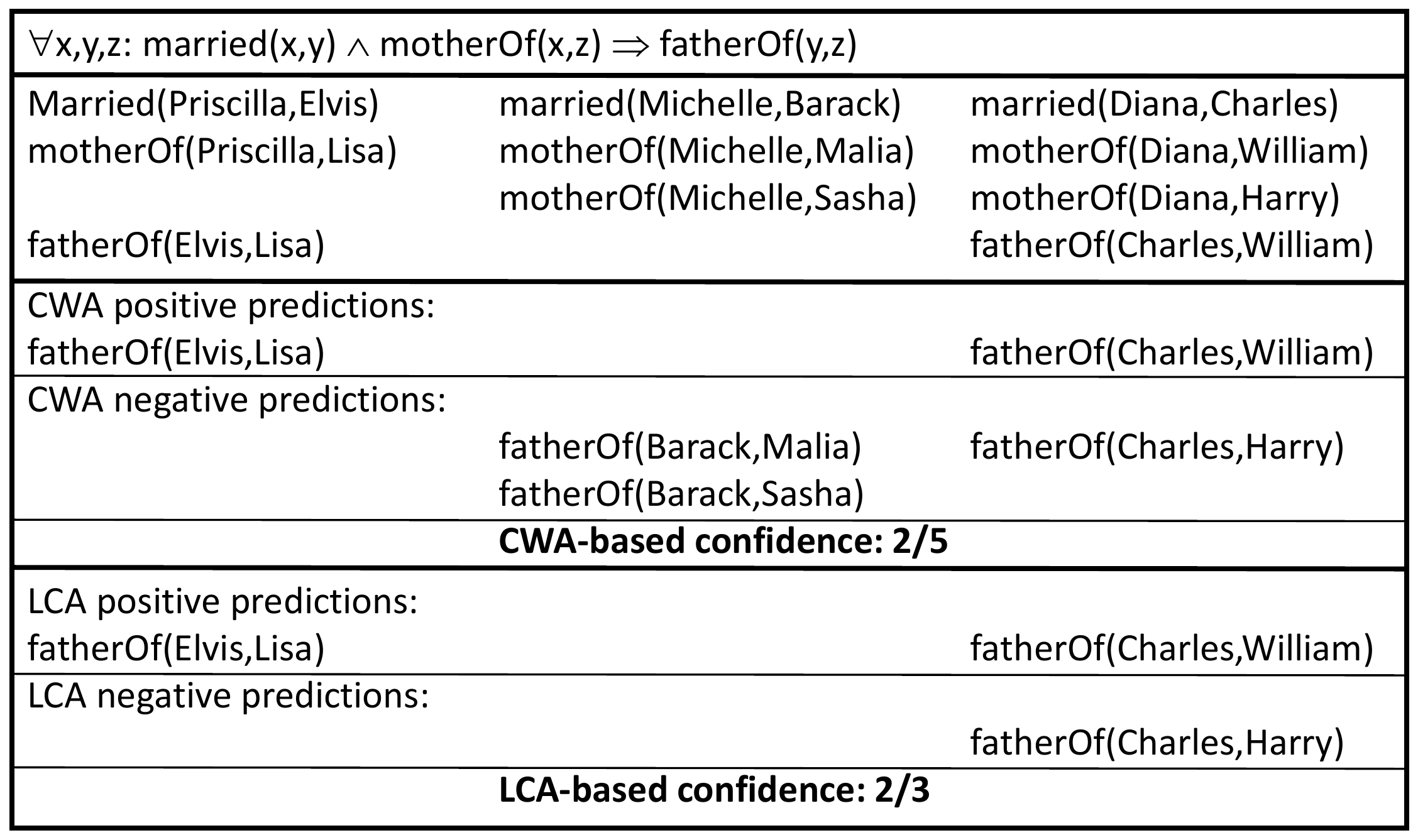}
      \caption{Example for rule confidence under CWA and LCA}
      \label{fig:rule-confidence}
\end{figure}

Instead of a simple queue, we can also use a priority queue 
by confidence score,
a combination of confidence and support,
or other measures of interestingness,
so as to arrive at the most insightful rules for the final
output. 
Recall that confidence is the ratio of correct predictions and the sum of correct and incorrect predictions,
where a prediction is an instantiation of the rule head
in line with the variable bindings in the rule body.
Confidence can be calculated in different ways in a KB setting. 
The difference lies in selecting negative cases
(i.e., what counts as an incorrect prediction).
\squishlist
\item[1.] By postulating the {\bf Closed-World Assumption (CWA)},
we identify instantiations of the rule body such that the
accordingly instantiated head is not in the KB.
Figure \ref{fig:rule-confidence} shows an example
for deducing fathers of children.
All predicted fathers for which no statement is in the KB count as negative samples.
This way, the number of negative samples is typically high,
as most KBs are far from being complete. So the confidence tends
to be inherently underestimated.
\item[2.] To counter the biased influence of absent statements
being counted as negative samples, we can alternatively adopt
the {\bf Local Completeness Assumption (LCA)}, as defined in
Section \ref{ch8-sec:knowledgebase-completeness}. Missing statements %
are considered as negative evidence only if they have at least
one statement for the property in the respective rule atom
(e.g., \ent{fatherOf}). Otherwise, the absent statements are
disregarded, as their validity is uncertain under the default
Open World Assumption.
Figure \ref{fig:rule-confidence} shows an example.
\squishend

As shown by \cite{DBLP:conf/www/GalarragaTHS13,DBLP:journals/vldb/GalarragaTHS15}, the LCA-based confidence measure
is more effective in identifying interesting rules
by confidence scoring.
Note that the mined rules may include constants
for some arguments, for salient values or entities.
Examples are:
\vspace*{0.1cm}

\noindent \hspace*{0.5cm} $\forall x,y: type (x,musician) ~\land~ livesIn (x,Nashville) \Rightarrow citizenOf (x,USA)$\\
\hspace*{0.5cm} $\forall x,y: type (x,politician) ~\land~ citizenOf (x,France) \Rightarrow livesIn (x, Paris)$

\vspace*{0.1cm}
By swapping the positive examples and the negative examples, it is also
possible to learn rules with negation. %
For example, if
two people are married, then one cannot be the child of the other. 
An issue here is that if LCA
were used to generate negative samples,
it would yield
an infinite number of negative statements: Elvis is not the child of Madonna, Elvis is not the child of Barack Obama, etc. 
To consider only a finite number of meaningful cases,
the solution is to generate a
negative sample
$Q(x,y)$ only if $x$ and $y$ 
appear together in at least one SPO statement in the KB~\cite{DBLP:journals/pvldb/OrtonaMP18,DBLP:conf/icde/OrtonaMP18}. 
More precisely,
we generate the statement $Q(x,y)$ 
as an incorrect prediction
if
\squishlist
\item[1.] $Q(x,y)$ is not in the KB.
\item[2.] There is $y'$ with $Q(x,y')$ in the KB or there is $x'$ with $Q(x',y)$.
\item[3.] There is a predicate $R$ with $R(x,y)$ in the KB.
\squishend
\vspace*{0.1cm}

A variety of works have developed algorithmic
alternatives, extensions and optimizations, most importantly,
for pruning the search space and for avoiding
expensive computations of support and confidence 
to the best possible extent.
Relevant literature includes 
\cite{lao2011random,DBLP:conf/acl/LaoMC15,DBLP:conf/ijcai/Tanon0RMW18,DBLP:conf/ijcai/MeilickeCRS19,DBLP:conf/esws/LajusGS20}.
A particularly noteworthy method is the
{\bf Path Ranking Algorithm (PRA)} by
\citet{lao2011random,DBLP:conf/acl/LaoMC15},
which operates on a graph view of the KB,
with entities as nodes and properties as edges.
The algorithm computes frequent edge-label sequences
on the paths of this graph; these form candidates
for the atom sets of rules.
Random walk techniques are leveraged for efficiency.

Several works have investigated generalizations beyond Horn rules: mining rules with exceptions~\cite{gad2016exception}, OWL constraints~\cite{volker2011statistical}, rules with numerical constraints~\cite{minaei2013mining}, 
rules in combination with text-based evidence and
counter-evidence \cite{DBLP:conf/wsdm/Gad-Elrab0UW19},
and rules that take into account the confidence scores of other rules
\cite{DBLP:conf/ijcai/MeilickeCRS19}. 
Surveys on KB rule mining are given by \cite{DBLP:conf/rweb/0001GH18}
and \cite{DBLP:conf/rweb/SuchanekLBW19}.

\section{Knowledge Graph Embeddings}
\label{ch8-sec:kgembeddings}

So far, we have looked at symbolic representations of entities, types and properties, founded on computational logics. Recently, another kind of 
knowledge representation has gained popularity:  mapping entities, properties and SPO triples into real-valued vectors in a low-dimensional latent space. 
Such {\bf knowledge graph embeddings},
or {\bf KG embeddings} for short, are motivated by the great success of word embeddings in machine learning for NLP. %
Note that the setting here is different
from those for word embeddings and
entity embeddings 
such as
Wikipedia2vec \cite{yamada2020wikipedia2vec},
as covered in Section \ref{subsec:word-entity-embeddings}.
The focus here is on capturing structural 
patterns of an entity in its graph neighborhood
alone, as an asset for machine learning and
predictions.

\begin{figure} [tbh!]
  \centering
   \includegraphics[width=0.8\textwidth]{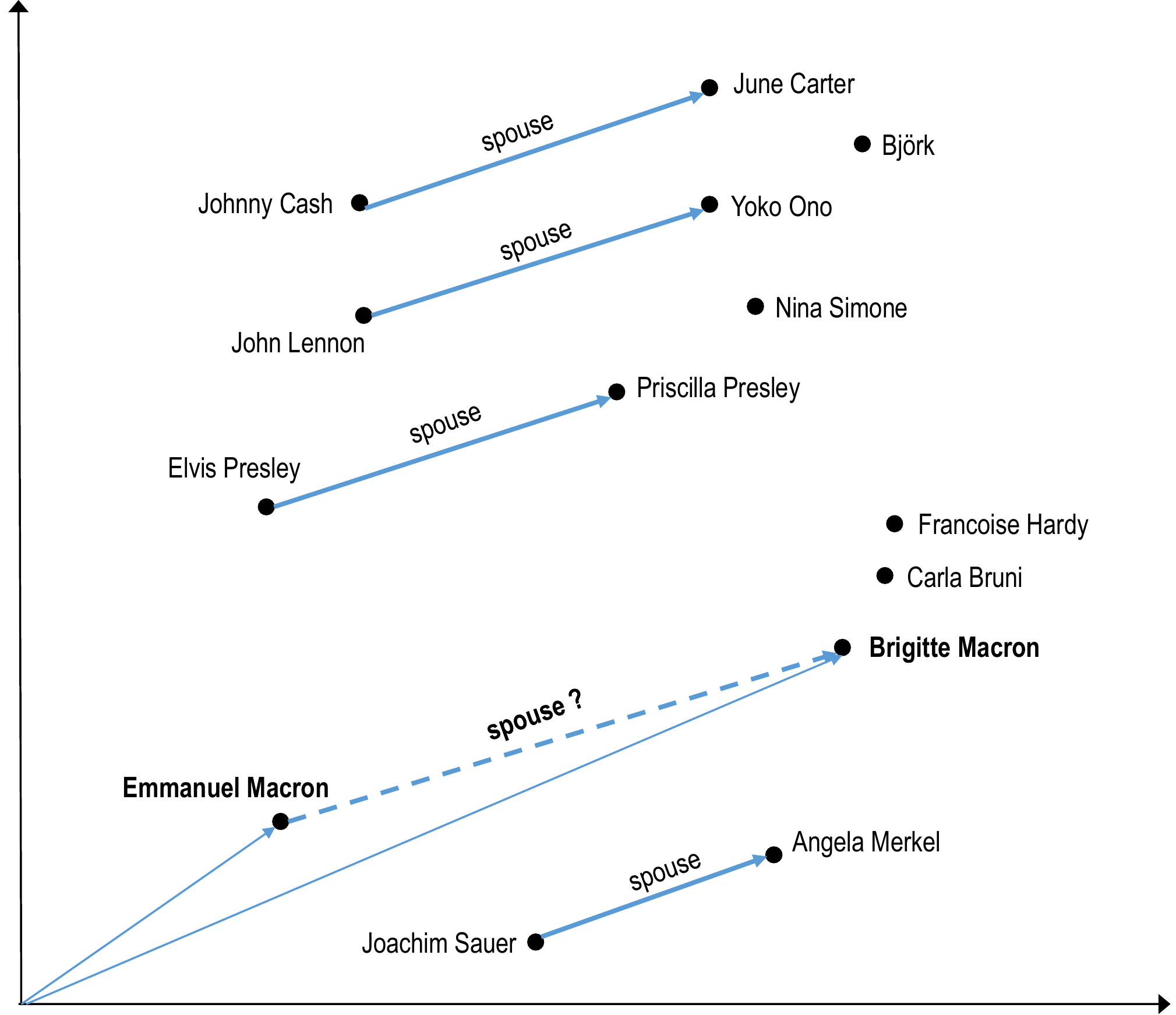}
      \caption{Embeddings of entities and the spouse relation}
      \label{fig:kg-embeddings}
\end{figure}

Imagine a KB where 
a set of entities of type \ent{person} and
the relation \ent{spouse} have all been mapped
to latent vectors, as illustrated in 
Figure \ref{fig:kg-embeddings} with the simplified case of a
two-dimensional latent space.
The dots represent the coordinates of these vectors.
The \ent{spouse} vector has been shifted and scaled
in length,
while retaining its direction, to connect
entities which are known to be spouses.
Now consider the French president 
\ent{Emmanuel Macron} and a set of potential wives
such as \ent{Carla Bruni},
\ent{Francoise Hardy} and \ent{Brigitte Macron}.
We can identify \ent{Brigitte Macron} as his
proper spouse by adding up the vectors for
\ent{Emmanuel Macron} and the \ent{spouse} relation,
as illustrated in the figure.

\vspace*{0.2cm}
\noindent{\bf Use Cases of KG Embeddings:}\\
The vector representation has several use cases. 
It is a good way of quantifying similarities
between entities, and thus enables statistical
methods like clustering. Also, deep learning
requires real-valued vectors as input.
The most prominent application is
to {\bf predict missing statements} for the KB.
The literature has referred to this as
{\bf knowledge graph completion} 
(\citet{DBLP:conf/www/NickelTK12,DBLP:conf/nips/BordesUGWY13,DBLP:conf/nips/SocherCMN13,DBLP:conf/aaai/LinLSLZ15}).
By operating on structural patterns of the graph alone,
this is actually a form of \emph{link prediction},
similar to predicting (and recommending) ``friends''
in a social network.
Specifically, for a given target entity and relation,
KG completion would compute a ranked list of
candidates for the respective object(s) to suggest
a new SPO triple.
This could support a KB curator in filling gaps.
However, state-of-the-art methods for KG embeddings
are still far from consistently recommending
the correct entities at the top ranks. 
So human curators do not run out of work that soon.

A dual case is to scrutinize doubtful statements
in the KB, or in a pool of candidates considered for
addition.
If none of the say top-100 predictions for a
given SP pair produces the O that appears in a 
candidate statement, we should consider
discarding the candidate completely.
However, this should still be taken with a big grain of salt
when maintaining a production-level KB.

\vspace*{0.2cm}
\noindent{\bf Learning of KG Embeddings:}\\
The objective for learning embedding vectors is
as follows: 
\begin{samepage}
\begin{mdframed}[backgroundcolor=blue!5,linewidth=0pt]
\squishlist
\item[ ] {\bf KG Embedding Vectors:}\\
Given a KB, find a mapping $v$ from 
its entities and relations onto real-valued vectors, such that\\ 
\hspace*{0.5cm} $v(s)+v(p)\approx{}v(o)$ \\
for each statement $\langle{}s, p, o\rangle$ in the KB.\\
The application then is, for given $s$ and $p$ with
unknown $o$, to predict $o$ by finding objects
close to $v(s)+v(p)$.
\squishend
\end{mdframed}
\end{samepage}

Embeddings can be computed by algorithms for
tensor factorization (e.g., \cite{DBLP:conf/www/NickelTK12})
or by training a neural network 
(e.g., \cite{DBLP:conf/nips/BordesUGWY13,DBLP:conf/nips/SocherCMN13}).
At training time, the neural network takes as
input $\langle{}s, p, o\rangle$  statements
that are one-hot encoded as preliminary vectors.
The network learns latent vectors 
$v(s), v(p), v(o)$ and outputs a score
how well these learned embeddings approximate
the existing facts. So the key term in the loss function is $||v(s)+v(p)-v(o)||$ summed up over all training samples.
To minimize the loss function, the learner
performs backpropagation with gradient descent,
and this way computes the best embedding vectors.
At deployment time, one can simply add or
subtract learned vectors to compute similarities
between vector expressions.

To avoid that the network just returns 0,
that is, perfect matches, for all inputs, it has to 
include also \emph{negative samples} for its training. %
These are usually obtained by perturbing 
positive statements $\langle{}s,p,o\rangle$ of the KB into statements $\langle{}s,p,o'\rangle$ that are not in the KB. 
For negative training samples, the network should return a large output value (i.e., mismatch). 

This basic method is known by the name {\bf TransE}
(\citet{DBLP:conf/nips/BordesUGWY13}),
for ``translating'' entities into latent-space vectors.
A major limitation of TransE
is that all objects of a one-to-many relation end up in the same spot in the vector space. In our example, if Emmanuel Macron ever marries someone else
and the marriage triples are the only statements about his spouses, then that other person would have the same vector as Brigitte Macron (much to her annoyance, certainly). %
To overcome this and other limitations, 
 more advanced models have been developed.
Major works, overviews and experimental comparisons
include
\cite{DBLP:conf/www/NickelTK12,DBLP:conf/nips/SocherCMN13,DBLP:conf/nips/BordesUGWY13,DBLP:conf/aaai/LinLSLZ15,DBLP:journals/pieee/Nickel0TG16,DBLP:journals/tkde/WangMWG17,DBLP:journals/corr/abs-2002-00388,DBLP:conf/iclr/RuffinelliBG20}.

\section{Consistency Reasoning for Knowledge Cleaning}
\label{ch8-sec:consistency-reasoning}

Constraints allow us to scrutinize each statement candidate,
one at a time.
For example, a candidate statement that Donald Trump has composed
the US national anthem ``Star Spangled Banner'' could be
refuted by violating the type constraint that pieces of
music must be created by musicians, or at least artists.
For doing this, we only need to check this {\em individual candidate}
against the applicable constraints, without looking
at any of the other candidates.

However, the situation is often more complex.
Suppose, for example, that we have three candidates
for the place where Elvis Presley died:
\squishlist
\item[a)] $\langle$\ent{Elvis Presley, deathplace, 
Baptist Hospital Memphis}$\rangle$,
\item[b)] $\langle$\ent{Elvis Presley, deathplace, Graceland Ranch}$\rangle$,
\item[c)] $\langle$\ent{Elvis Presley, deathplace, Mare Elysium (Mars)}$\rangle$\\
         (where he lived until February 2020, according to some beliefs).
\squishend
Each of them individually could be accepted, but together they violate
the constraint that the deathplace of a person is unique.
So we must inspect the evidence and counter-evidence for the
three statements {\em jointly}, to arrive at a conclusion on which of these
hypotheses is most likely to be valid.
The following subsections present various methologies for the
necessary {\em holistic} consistency checking.

\subsection{Data Cleaning and Knowledge Fusion}
\label{ch8-subsec:datacleaningandknowledgefusion}

\vspace*{0.2cm}
\noindent{\bf Data Cleaning with Minimal Repair:}\\
Although relational databases usually undergo a careful
process for schema design and data ingest, the case
where some data records contain erroneous values is
a frequent concern. 
Humans may enter incorrect values,
and, most critically, database tables 
may be the result
of some imperfect data integration, for example,
merging two or more customer tables with discrepancies
in their addresses.
A major part of the necessary {\em data cleaning} \cite{DBLP:journals/debu/RahmD00,DBLP:journals/ftdb/IlyasC15,DBLP:books/acm/IlyasC19}, therefore,
is {\em entity matching (EM)}
\cite{DBLP:journals/dke/KopckeR10,DBLP:series/synthesis/2010Naumann,DBLP:books/daglib/0030287,DBLP:series/synthesis/2015Dong,DBLP:journals/corr/abs-1905-06397}: inferring which records truly
correspond to the same entities and dropping or correcting
doubtful matches.
We discussed this issue in Section \ref{sec:entity-matching}.

Entity-matching errors are not the only issue in data cleaning, though.
A typical setting involves a relational table and a set of integrity
constraints that the data should satisfy. 
These constraints could be manually specified, but could also be
automatically discovered from the data itself -- by data-mining
algorithms that reveal {\em approximate invariants} 
\cite{DBLP:conf/sigmod/IlyasMHBA04,DBLP:journals/pvldb/ChiangM08,DBLP:journals/pvldb/SaxenaGI19}.
These include functional dependencies (FDs), inclusion dependencies (IDs),
conditional FDs, and more.
The
family of {\em denial constraints} subsumes a variety of these kinds of invariants,
and has been used in industrial-strength data-cleaning tools
\cite{DBLP:journals/pvldb/RekatsinasCIR17,DBLP:books/acm/IlyasC19}.
The key idea for cleaning is:
when a constraint holds for almost all of the data instances,
the remaining violations are likely errors.
To remove these errors, we either need a way of obtaining the correct values,
possibly by some human-in-the-loop procedure, or we remove the incorrect records.
Often, there are multiple ways of choosing a set of erroneous records for
removal, in order to restore the integrity of the remaining data.
The guiding principle in making good choices is the following 
(\citet{DBLP:conf/pods/ArenasBC99,DBLP:conf/sigmod/BohannonFFR05,DBLP:conf/pods/Bertossi19}):

\begin{samepage}
\begin{mdframed}[backgroundcolor=blue!5,linewidth=0pt]
\squishlist
\item[ ] {\bf Minimal-Repair Data Cleaning:}
\squishlist
\item Input: Relational table $R$ with errors, and a set of constraints
\item Output: Clean table $S \subset R$\\
Choose tuples $\{t_1, t_2 \dots\}$ such that\\
$S = R - \{t_1, t_2 \dots\}$ satisfies all constraints, and\\
$|\{t_1, t_2 \dots\}|$ is mimimal.
\squishend
\squishend
\end{mdframed}
\end{samepage}

The repair mechanism can vary: instead of
solely considering discarding entire tuples,
there are alternative options, including
the replacement of erroneous attribute values.
The literature on database (DB) cleaning has developed a suite of powerful
algorithms for minimum repair 
\cite{DBLP:books/acm/IlyasC19}.
They include both combinatorial optimization methods, 
with smart
pruning of the underlying search space (e.g., \cite{DBLP:conf/icde/ChuIP13,DBLP:journals/pvldb/ProkoshynaSCMS15}),
and machine-learning methods, such as probabilistic graphical models
(e.g., \cite{DBLP:journals/pvldb/RekatsinasCIR17}).
We will come back to the latter in Section \ref{ch8-sec:probgraphmodels}.

In principle, DB cleaning methods are applicable to knowledge bases
as well, and could contribute to KB curation.
However, this has not been explored much, for several reasons.
First, data cleaning typically tackles a single relational table,
whereas KBs cover hundreds or thousands of relations.
Second, at this magnitude of the schema size, the
number of constraints can be huge and
DB methods are not geared for this scale (at the schema level).
Third, the assumption for DB cleaning is that most records are correct
and errors are spread among a few incorrect values of single attributes.
In contrast, KB curation may start with a huge number of uncertain statements,
all collected by potentially noisy extraction algorithms. 
It is well possible that a KB-cleaning process would have to remove
ten, twenty or more percent of its inputs.\\

\vspace*{0.2cm}
\noindent{\bf Multi-Source Fusion:}\\
Another related topic in database research is
{\bf data fusion} 
(\citet{DBLP:journals/pvldb/DongN09,DBLP:books/sp/13/DongBS13}):
given a set of value-conflicting
tuples from different databases or structured web sources, the task is to
infer the correct value.
For illustration, consider the data about people's 
current residence, with values obtained from four different sources.

\begin{table}[h!]
\begin{center}
{\small
\begin{tabular}{|l||l|l|l|l|}\hline
 & Source A & Source B & Source C & Source D \\ \hline\hline
Fabian & Paris & Paris & Mountain View & Paris \\ \hline
Gerhard & Seattle & Saarbruecken & Kununurra & Saarbruecken \\ \hline
Luna & Mountain View & Seattle & Mountain View & Seattle \\ \hline
Simon & Saarbruecken & Bolzano & Saarbruecken & Bolzano \\ \hline
\end{tabular}
}%
\end{center}
\end{table}

The goal is to choose the correct residence values for each of
the four people. 
A simple approach could aim to identify the {\em best source}, by
some authority or trust measure (e.g., Alexa rank for web sites), 
and pick all values from this source.
In the example, Source B could be the best choice. However, this alone
would yield an incorrect value for Simon. 
Therefore, an alternative is to compute a {\em voting} from
all sources, and let the majority win. However, this would yield ties
for Luna and Simon. 
So the best approach is to combine {\em source quality} and
{\em weighted voting}, with source quality cast into weights.

The key point now is how to measure source quality.
Suppose, for ease of explanation,
that we have ground-truth values
for a 
subset of the tuples,
for example, by considering some of the sources
as a priori trustworthy, and applying
the fusion algorithm to all other sources
and their additional data values.
Then, the source quality can be defined as the fraction of its values
that match the ground-truth. 
In the example, if we knew all correct values, source A would have
weight $2/4$, source B $3/4$, source C $1/4$ and source D $3/4$.

The solution devised in \cite{DBLP:books/sp/13/DongBS13}
is based on this intuition, but does not require
any ground-truth knowledge.
The approach uses voting to estimate a prior for 
trustworthy sources, and then applies
Bayesian analysis with all observed data
to infer the most likely valid values.
This method has been generalized and carried over to {\bf knowledge fusion}
\cite{DBLP:journals/pvldb/DongGHHMSZ14,DBLP:journals/pvldb/DongGMDHLSZ15}:

\begin{samepage}
\begin{mdframed}[backgroundcolor=blue!5,linewidth=0pt]
\squishlist
\item[ ] {\bf Multi-Source Fusion for KB Statements:}
\item[ ] Input: set of uncertain SPO statements $\{t_1,t_2 \dots t_n\}$\\
from various sources $\{S_1 \dots S_k\}$ with $k \ll n$,\\
where each $S_j$ stands for a combination of
website and extraction method
(e.g., regex-based extraction from lists in {\small musicbrainz.org})
\item[ ] Output: jointly learned probabilities (or scores) for
\squishlist
\item the correctness of statement $t_i$ ($i=1..n$) and
\item the trustworthiness of source $S_j$ ($j=1..k$).
\squishend
\squishend
\end{mdframed}
\end{samepage}

The joint learning of statement validity and source trustworthiness
takes into account {\em mutual reinforcement dependencies}.
A specific instantiation, developed in \cite{DBLP:journals/pvldb/DongGMDHLSZ15}, 
is based on a sophisticated equation system
and an iterative EM algorithm (EM = Expectation Maximization) for an approximate solution.

Methods along these lines have been utilized
in the 
{\em Knowledge Vault} project
\cite{Dong:KDD2014}.
Section \ref{subsec:KV} provides further insights
on this work.

\subsection{Constraint Reasoning and MaxSat}
\label{ch8-subsec:MaxSat}

Knowledge fusion and other techniques for
ensemble learning still treat each candidate statement in isolation.
For example, they make independent decisions about the truth of \ent{$\langle$Bob Dylan, citizenship, USA$\rangle$} and \ent{$\langle$Bob Dylan, citizenship, UK$\rangle$}.
However, it is often beneficial to perform \emph{joint inference} over a set of 
candidates, based on {\em coupling constraints}.
In the following, we present a major approach
to do this, 
using so-called
{\em Weighted MaxSat} inference, which is
an extension of the {\em Maximum Satisfiability}
problem, MaxSat for short \cite{DBLP:series/faia/2009-185}.
A broader perspective would be to harness
methods from {\em constraint programming}
\cite{DBLP:reference/fai/2},
but we focus on MaxSat here as it has
a convenient way of incorporating the
{\em uncertainty} of the inputs.

Consider, for example, a set of candidate statements \emph{hasWon(Bob\-Dylan,Grammy), hasWon(Bob\-Dylan,Literature\-Nobel\-Prize)}, and (by erroneously extracting the type of Nobel Prize) \emph{hasWon(Bob\-Dylan,Physics\-Nobel\-Prize)},
along with two type statements
\emph{type(Bob\-Dylan,musician)} and
\emph{type(Bob\-Dylan,scientist)}.
Imposing consistency constraints is a good way of
corroborating such uncertain candidates. 
For the simplified example, we postulate that people are either scientists or musicians (i.e., never both), and that it is unlikely for a musician to win a Nobel Prize in Physics. These (soft) constraints
can be either specified by a knowledge engineer or 
automatically learned (e.g., by rule mining, see Section~\ref{ch8-subsec:rulemining}). 
The following table shows the five noisy candidate statements
and the constraints (with numbers in brackets denoting weights, explained below).

\begin{center}
\begin{tabular}{| l |}\hline
\ent{$\langle$BobDylan hasWon Grammy$\rangle$} [0.7] \\
\ent{$\langle$BobDylan hasWon LiteratureNobelPrize$\rangle$} [0.5] \\
\ent{$\langle$BobDylan hasWon PhysicsNobelPrize$\rangle$} [0.3] \\ 
\ent{$\langle$BobDylan type musician$\rangle$} [0.9] \\ 
\ent{$\langle$BobDylan type scientist$\rangle$} [0.1] \\
\hline
$\forall x (hasWon(x, Grammy)) \Rightarrow type(x, musician))$ [0.9]\\
$\forall x  (hasWon(x, PhysicsNobelPrize)) \Rightarrow type(x, scientist))$ [0.9]\\
$\forall x  (type(x, scientist)) \Rightarrow \neg type(x, musician))$ [0.8]\\
$\forall x  (type(x, musician)) \Rightarrow \neg type(x, scientist))$ [0.8]\\ \hline
\end{tabular}
\end{center}

All of these logical formulas together are unsatisfiable;
that is, they are inconsistent and imply {\em false}.
However, if we choose a proper subset of them, we may arrive
at a perfectly consistent solution and would thus identify
the statements that we should accept for the KB.
Typically, this would only involve discarding some of the
atomic statements while retaining all constraints. 
However, it is also conceivable to drop constraints,
as they are soft and could be violated by exceptions.
Obviously, we want to sacrifice as few of the inputs as possible,
maximizing the 
number of surviving statements.
This problem of computing a large consistent subset of the
input formulas is the {\bf Maximum Satisfiability}
problem, {\bf MaxSat} for short.

Both candidate statements and constraints often come with weights;
these are shown in brackets in the above table. 
For statements, the weights are usually the confidence scores
returned by the extractor, potentially with re-scaling and
normalization.
For constraints, the weights would reflect confidence scores
for underlying logical invariants (see Section \ref{ch8-subsec:rulemining}),
or the degrees to which they should be fulfilled in a proper KB.
Another way of interpreting weights is that they denote
{\em costs} that a reasoner has to pay when dropping a statement
or violating a constraint.
The goal is now refined as follows:

\begin{samepage}
\begin{mdframed}[backgroundcolor=blue!5,linewidth=0pt]
\squishlist
\item[ ] {\bf Principle of Weighted Maximum Satisfiability}:\\
For a given set of weighted candidate statements and weighted consistency constraints, identify a subset of formulas that is
logically consistent and has the highest total weight.
\squishend
\end{mdframed}
\end{samepage}

 Intuitively, in the example, we can discard the statement that Dylan won the Nobel Prize in Physics and the statement that he is a scientist, which come at a cost of 0.3 and 0.1, respectively. The remaining formulas are then logically consistent.

\vspace*{0.2cm}
\noindent{\bf Grounding into Clauses:}\\
Our input so far is heterogenous, mixing apples and oranges:
candidate statements are propositional-logic formulas
without variables, whereas constraints are predicate-logic
formulas with variables (and quantifiers like $\forall$
and perhaps also $\exists$).
To cast the consistency reasoning into a tangible 
algorithmic problem, we need to unify these two constituents.
First, we limit constraints to be of the {\em Horn clause}
type, in prenex normal form with universal quantifiers only
(see Section \ref{ch8-subsec:rulemining}).
By being more permissive on the 
number of positive atoms,
we can 
further relax this
into arbitrary {\em clauses}. 

Second and most importantly, we {\em instantiate}
the constraints by substituting variables with constants from
the statements. For example, we plug in \ent{BobDylan} for
the $x$ variable in all constraints. 
If we had more candidate statements, say also about
\ent{ElvisPresley} and \ent{EnnioMorricone}, we would
generate more instantiations.
This renders all formulas into propositional logic without
any variables, and all formulas become clauses.
In computational logics, this process is referred to as {\bf grounding}.
For our simple example, the grounding would produce the
following set of clauses:

\begin{center}
\begin{tabular}{| l |}\hline
\ent{$\langle$BobDylan hasWon Grammy$\rangle$} [0.7] \\
\ent{$\langle$BobDylan hasWon LiteratureNobelPrize$\rangle$} [0.5] \\
\ent{$\langle$BobDylan hasWon PhysicsNobelPrize$\rangle$} [0.3] \\ 
\ent{$\langle$BobDylan type musician$\rangle$} [0.9] \\ 
\ent{$\langle$BobDylan type scientist$\rangle$} [0.1] \\
\hline
$\lnot$\ent{$\langle$BobDylan hasWon Grammy$\rangle$} $\lor$ \ent{$\langle$BobDylan type musician$\rangle$} [0.9] \\
$\lnot$\ent{$\langle$BobDylan hasWon PhysicsNobelPrize$\rangle$} $\lor$ \ent{$\langle$BobDylan type scientist$\rangle$} [0.9] \\
$\lnot$\ent{$\langle$BobDylan type scientist$\rangle$} $\lor$ $\lnot$\ent{$\langle$BobDylan type musician$\rangle$} [0.8] \\
\hline
\end{tabular}
\end{center}
\noindent The mutual exclusion between types 
\ent{scientist} and \ent{musician} was written in the form of two implication constraints before, but both result in the same clause which is thus stated only once.

Obviously, a less simplified case could have constraints with
multiple variables -- so this process has a potential for
``combinatorial explosion''. 
The full grounding happens only conceptually, though,
and does not have to be fully materialized.
Also, if entities have been canonicalized upfront and are
associated with types in the KB, then only those combinations
for variable substitutions need to be considered that 
match the type signatures of the constraint predicates.
There are further optimizations for lazy computation of
groundings.

This way, the consistency reasoning task has been translated
into propositional logics, with weighted clauses.
We now treat the atoms of all clauses together as
statements for which we want to infer truth values.
For clauses with more than one atom (i.e., the ones for instantiated
constraints), the entire clause becomes satisfied if at least
one of its atoms has a positive truth value
(with consideration of whether an atom itself is positive
or negative, i.e., prefixed by $\lnot$).

\begin{samepage}
\begin{mdframed}[backgroundcolor=blue!5,linewidth=0pt]
\squishlist
\item[ ] {\bf Objective of Weighted MaxSat}:\\
Given a set of propositional-logic clauses, each with a positive weight, the objective of Weighted MaxSat is to compute
a truth-value assignment for the underlying atoms such that
the total weight of the satisfied clauses is maximal.
\squishend
\end{mdframed}
\end{samepage}

In the example, with 
8 clauses
and a total of 5 atoms,
the optimal solution is to assign 
{\em false} to the atoms \ent{$\langle$BobDylan hasWon PhysicsNobelPrize$\rangle$} and
\ent{$\langle$BobDylan type scientist$\rangle$},
with a 
total weight of 4.7.
A different truth-value assignment, which is sub-optimal but
consistent, would be to assign 
{\em false} to 
\ent{$\langle$BobDylan type musician$\rangle$} and 
\ent{$\langle$BobDylan hasWon Grammy$\rangle$},
and {\em true} to the other three atoms.
This would have 
a total weight of 3.5.

\vspace*{0.2cm}
\noindent{\bf Weighted MaxSat Solvers:}\\
Relating this approach to MaxSat reasoning for  tasks
in computational logics, we treat the atoms of our clauses
as variables for truth-value assignment; this way, we can directly
use state-of-the-art solvers for weighted MaxSat.
The problem is NP-hard, though, 
as it generalizes the classical SAT problem. 
Nevertheless, there are numerous approximation algorithms
with very good approximation ratios (see, e.g., \cite{li2009maxsat}).
Many of these have been designed for use cases like
theorem proving and reasoning
about program correctness. 
In these settings, the structure of the input sets is rather
different from the KB environment. Theorem proving often
deals with a moderate number of clauses where each clause
could have a large number of atoms. These are automatically
derived from declarative specifications in more expressive logics.
For reasoning over KB candidates, where all clauses are
either single-atom or generated by grounded constraints,
the situation is very different.
The grounding can lead to a huge number of clauses, but
the atom set per clause is fairly small as the formulas
mostly express type constraints, functional dependencies,
mutual exclusion,
and other compact design patterns.

This design consideration has motivated the development
of {\em customized MaxSat solvers} for KB cleaning.
Two simple but 
powerful 
heuristics have been
used by 
\citet{suchanek2009sofie},
originally developed
to augment the YAGO knowledge base.

The first technique leverages 
{\em Unit Clauses}, that is, 
clauses that have exactly one variable whose truth value has not yet been determined. We compute, for each unassigned variable $x$, the difference between the combined weight of the unit clauses where $x$ appears positive and the combined weight of the unit clauses where $x$ appears negative. We choose the variable $x$ with the highest absolute difference, and set it to {true} if the difference is positive and to false otherwise. This technique is a greedy heuristic, which assumes that the variable with the highest value brings a high gain for the final solution. 

The method can be combined with another technique known as 
{\em Dominant Unit Clause (DUC) Propagation}. DUC propagation fixes the values of variables that are already so constrained that only one truth value can be part of the optimal solution. 
Specifically, it sets the value of variables that appear with one polarity in unit clauses that have a higher combined weight than all clauses (regardless of unit clause or not) where the variable appears with the opposite polarity.  
In combination, a SAT solver with these two heuristics 
has an approximation guarantee of 1/2: solutions have a weight
that is at least 50\% of the optimal solution. 
In practice, the approximation is much better,
often reaching 90\% or better \cite{suchanek2009sofie}.
Also, the algorithm is very efficient and can run on 
large sets of input clauses -- automatically generated
from uncertain candidate statements and consistency constraints.
For huge inputs, it is even feasible to {\em scale out} the
computation, by partitioning the input via
graph-cut algorithms and running the reasoner on all
partitions in parallel \cite{nakashole2011scalable}.

\vspace*{0.2cm}
\noindent{\bf Extensions:}\\
The Weighted MaxSat method has been used in the 
by \citet{suchanek2009sofie},
to extract information from noisy text. SOFIE reformulates the 
statement-pattern duality of Section \ref{ch4:subsec:PatternLearning}
as a soft constraint between patterns and relations: 
if a pattern expresses a relation, 
then two co-occurring 
entities with this pattern are an instance pair of the relation. 
Conversely, every sentence that contains two entities that are known to be connected by the relation
becomes a candidate for a pattern. 
In addition, SOFIE has integrated
{\em entity canonicalization} (see Chapter \ref{ch3-sec-EntityDisambiguation})
into its joint reasoning,
by additional constraints that couple surface names and entities.

The HighLife project \cite{DBLP:conf/www/ErnstSW18} has extended
MaxSat reasoning to higher-arity relations.
Texts can express a relationship that holds between more than two entities. If only some of these entities are mentioned, the resulting statement amounts to a formula with existential quantifiers for the unmentioned entities. HighLife devised clause systems specifically for this case, used Weighted MaxSat
reasoning over all the partial observations,
and could thus infer non-binary statements relating more than two entities.

\subsection{Integer Linear Programming}
\label{ch8-subsec:ILP}

Another way of operationalizing the reasoning
over uncertain statements and consistency constraints
is by means of {\em integer linear programs (ILP)}
(see, e.g., \cite{DBLP:books/daglib/0090562}).
These models have been extensively studied for all
kinds of industrial optimization problems, such as
production planning for factories, supply chains and
logistics, or scheduling for
airlines, public transportation, and many other applications.
ILP is a very mature methodology, hence a candidate
for our setting.
In general, an ILP consists of 
\squishlist
\item a set of decision variables,
often written as a vector $x$, allowed to take only non-negative
integer values, 
\item an objective function $c^T \cdot x$ to be maximized,
with a vector $c$ of constants, and
\item a set of inequality constraints over the decision variables,
written in matrix form as $M x \le b$ with matrix $M$ and vector
$b$ holding constants as coefficients.
\squishend

To map the consistency reasoning problem onto an ILP,
we associate each uncertain statement $A_i = \langle \ent{S P O} \rangle$, that is, a logical atom,
with a decision variable $X_i$. Their Boolean nature, deciding
on whether to accept a statement or not, is realized by
restricting $X_i$ to be a {\bf 0-1 variable}, by adding constraints
$X_i \le 1$ and $X_i \ge 0$.
The clauses that we construct by grounding logical constraints
(as explained in Subsection \ref{ch8-subsec:MaxSat}) are encoded into a set of
inequalities that couple the decision variables.
Weights of candidate statements become coefficients for the ILP
objective function, and weights for grounded constraints become
the coefficients for a big inequality system.

\begin{samepage}
\begin{mdframed}[backgroundcolor=blue!5,linewidth=0pt]
\squishlist
\item[ ] {\bf ILP for Constraint Reasoning:}\\
Input:
\squishlist
\item Uncertain candidate statements $A_1 \dots A_n$,
each a logical atom of the form $\langle \ent{S P O} \rangle$
or $\lnot \langle \ent{S P O} \rangle$.
The weight of $A_i$ is $w_i$.
\item A set of clauses with more than one atom, one clause for
each grounded constraint: $C_1 \dots C_m$
where each $C_j$ consists of positive atoms (without $\lnot$)
and negative atoms (with $\lnot$).
We denote these subsets of atoms as $C^+_j$ and $C^-_j$.
The weight of $C_j$ is $u_j$.
\squishend
Construction of the ILP:
\squishlist
\item For each statement $A_i$, there is a 0-1 decision variable $X_i$.
\item The objective function of the ILP is to maximize $\sum_{i=1..n} w_i X_i$.
\item For each grounded constraint $C_j$ we create an inequality constraint:\\
$\sum_{\mu \in C^+_j} X_\mu ~+~ \sum_{\nu \in C^-_j} (1-X_\nu) ~ \ge ~ 1$\\
which is equivalent to the condition that at least one atom in the clause should be satisfied
(taking the polarities, positive or negative, into account).
\squishend
\squishend
\end{mdframed}
\end{samepage}

The above ILP enforces all grounded constraints to hold. In other words, it does not
consider the weights of the original constraints, which would allow slack for exceptions.
To achieve this interpretation of {\em soft constraints}, we need to extend the
objective function. Essentially, we add a cost term, proportional
to the constraint weight, each time we violate a grounded constraint.
The overall objective function would then take this form:\\
\hspace*{0.5cm} maximize 
$\lambda ~ \sum_{i=1..n} w_i X_i$\\
\hspace*{0.5cm} 
$~-~ (1-\lambda) ~ \sum_{j=1..m} u_j 
\left( \sum_{\mu \in C^+_j} (1-X_\mu) ~+~ \sum_{\nu \in C^-_j} X_\nu \right)$\\
where $\mu$ and $\nu$ range over the subscripts
of the respective atoms, and $\lambda$ is a tunable hyper-parameter.
The sum over statements is the {\em benefit} from accepting many candidates,
and the sum over grounded constraints is the {\em penalty} to be paid
for constraint violations.
This basic form can be varied in other ways. 

The outlined approach has shown how to incorporate soft constraints, whereas our
original ILP formulation allowed only hard constraints.
By combining both, via inequality constraints as well as penalty terms
in the objective function, we have a choice about which of the
original consistency constraints should be treated as strict invariants
and which ones can be treated in soft form with 
allowance for exceptions.

Computing the optimal solution for an ILP is, not surprisingly, also NP-hard.
On the positive side,
ILPs are very versatile and widely used in all kinds of mathematical optimizations;
therefore, a vast array of algorithms exist for making ILPs tractable
in many practical cases.
Also, there are mature software packages that are very well engineered, most notably,
the Gurobi solver ({\small\url{https://www.gurobi.com/}}).
A common optimization technique is to {\em relax} the ILP by dropping
the requirement that the variables can take only integer values.
Instead we allow real-valued solutions between 0 and 1, this way treating the ILP
as a standard linear program (LP). This technique is known as the
{\bf LP relaxation} of the ILP.
LPs can be tackled much more efficiently than ILPs, with polynomial algorithms
and various accelerations.
To obtain a valid solution for the ILP, the solution must be rounded to
one of the neighboring integers (e.g., 0.65 can be rounded to 1 or 0).
A principled method is to {\em randomly round} by tossing a coin
that falls on 1 with a probability proportional to the real value (e.g., with probability
0.65 becoming 1 and with probability 0.35 becoming 0).
This randomized algorithm has good approximation guarantees with high probability
\cite{DBLP:books/cu/MotwaniR95,DBLP:books/daglib/0004338}.

ILP models have been applied to various tasks of KB cleaning 
as well as for the underlying knowledge extraction. %
\citet{DBLP:conf/icml/RothY05}\cite{EACL2017:DanRothTutorial} 
provide general discussion on using ILP for 
KB curation and for NLP tasks.

\subsection{Probabilistic Graphical Models}
\label{ch8-sec:probgraphmodels}

The prior uncertainty in the weighted MaxSat problem can also be modeled in a probabilistic fashion, such that the Boolean decision variables become {\em random variables} with
probabilities for being true or false.
This line of models often starts with
a {\em declarative specification} of
candidate statements and logical constraints,
just like we did in the previous subsections
on MaxSat and ILP.
For probabilistic inference, the high-level specification is translated into a
{\bf Markov Random Field (MRF)}, the general
framework for probabilistic graphs that couple
a large number of random variables (see, e.g., the textbook \cite{KollerFriedman2009}).
Conditional Random Fields (CRF) that are
widely used for entity discovery (Chapter 
\ref{ch3:entities}) also fall into this regime
(see Section \ref{ch4-subsec:CRF}).
For tractability, these models have to make
assumptions about conditional independence,
thus becoming {\bf probabilistic factor graphs}
with factors expressing the local coupling of 
(small) subsets of (non-independent) random variables.

\vspace*{0.2cm}
\noindent{\bf Markov Logic Networks:}\\
In the following, we focus on one prominent 
and powerful model
from this broad family:
{\em Markov Logic Networks (MLN)}, by
\citet{DomingosLowd2009}.

\begin{samepage}
\begin{mdframed}[backgroundcolor=blue!5,linewidth=0pt]
\squishlist
\item[ ] {\bf Markov Logic Network (MLN):}
\item[ ] Input:
\squishlist
\item a set of grounded clauses $C_1,...,C_n$, derived from
uncertain statements and soft constraints, with weights $w_1,...,w_n$, and 
\item binary random variables $X_1,...,X_k$ for the %
atoms that appear in the clauses.
\squishend
\item[ ] Construction of the MLN:\\
The corresponding MLN is an undirected graph with
random variables $X_1 \dots X_k$ as nodes,
and edges between nodes $X_i,X_j$ if these variables
are considered to be coupled.
Conversely, if we assume
conditional independence\\
\hspace*{0.5cm} $P[X_i~|~X_j, \text{all}~X_\nu \ne X_i ~\text{and}~ \ne X_j]$\\
\hspace*{0.5cm} $= P[X_i~|~ \text{all}~X_\nu \ne X_i ~\text{and}~ \ne X_j]$\\
then there is no edge between $X_i$ and $X_j$.
Otherwise, $X_i$ and $X_j$ are coupled by an edge between them.\\
The MLN constitutes a joint probability distribution
for the variables $X_1 \dots X_k$, further explained below.
\squishend
\end{mdframed}
\end{samepage}

As an example, reconsider the set of 8 clauses over 5 atoms discussed in Section \ref{ch8-subsec:MaxSat}.
To make it more interesting, let us add another
constraint that leads to a 9th clause:
\vspace*{0.1cm}

\noindent
\hspace*{1.0cm} $\lnot$\ent{$\langle$BobDylan hasWon Grammy$\rangle$}\\ 
\hspace*{1.0cm} $\lor$ 
$\lnot$\ent{$\langle$BobDylan hasWon LiteratureNobelPrize$\rangle$}\\ 
\hspace*{1.0cm} $\lor$
$\lnot$\ent{$\langle$BobDylan type scientist$\rangle$} [0.6] 
\vspace*{0.1cm}

\begin{figure} [htb!]
  \centering
   \includegraphics[width=0.9\textwidth]{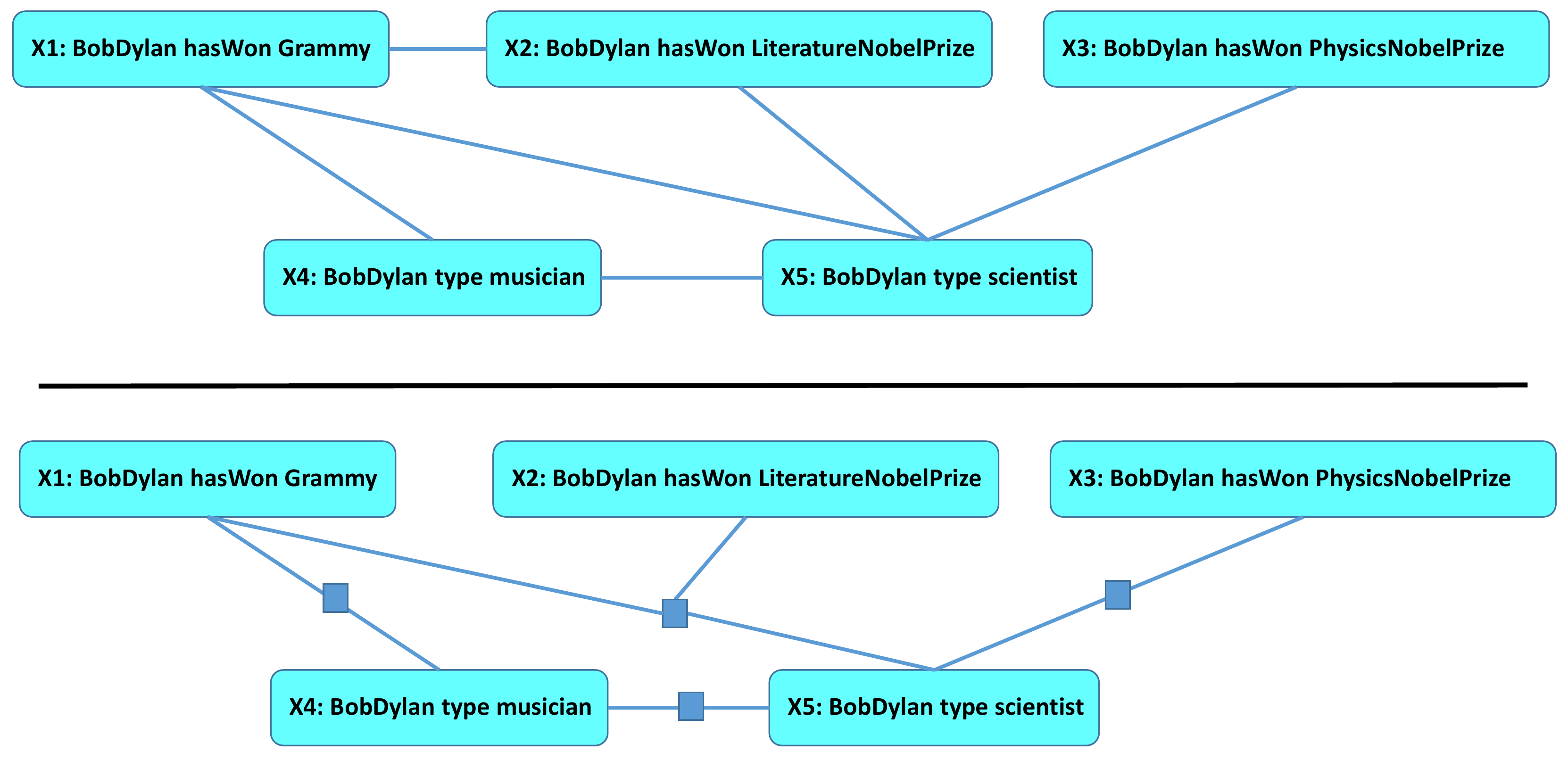}
      \caption{Example for MLN Graph}
      \label{fig:ch8-mln}
\end{figure}

\noindent (if someone has won a Grammy and a Literature Nobel Prize, she/he cannot be a scientist).\\
The constraints are the mechanism for coupling the
random variables for the candidate statements;
variables that do not share any constraints are conditionally independent.
This leads to the grounded MLN in Figure \ref{fig:ch8-mln}. The upper part shows the actual
graph with binary edges; the lower part shows the
same graph with cliques in the graph 
made explicit by the small
blue-rectangle connectors. 
So this MLN of five atoms
has three cliques of size 2 and one clique of size 3.
We added the new constraint for this very 
illustration of a clique with more than two nodes.

\begin{samepage}
\begin{mdframed}[backgroundcolor=blue!5,linewidth=0pt]
\squishlist
\item[ ] {\bf Factorized Distribution of MLN:}\\
The MLN graph induces a {\em joint probability distribution} for all random variables together.
By the assumptions about which variables are coupled and which ones are conditionally independent,
and by the Hammersley-Clifford Theorem from MRF theory
\cite{KollerFriedman2009,DomingosLowd2009},
the joint distribution takes the following factorized form:\\
\hspace*{0.5cm} $P[X_1 \dots X_k] \sim \prod_{cliques~j} \Phi_j \left(X_{j_1} \dots X_{j_l} ~\text{forming clique} ~j\right)$\\
where $j$ ranges over all cliques in the graph
and $\Phi_j$ are so-called 
{\em clique potential functions}
for capturing local probabilities.
These per-clique terms are the {\em factors} of the 
{\bf factor graph}.\\
For our KB curation setting, the cliques correspond to the constraint clauses. This leads to the product form\\
\hspace*{0.5cm} $P[X_1 \dots X_k] = \frac{1}{Z} \prod_{C_j}
~e^{w_j} ~~
\text{with} ~C_j~ \text{ranging over}$\\
\hspace*{0.5cm} $\text{all clauses satisfied by} ~ X_1 \dots X_k$\\
with clause weights $w_j$ and normalization constant $Z$. 
\squishend
\end{mdframed}
\end{samepage}

An MLN  defines a probability distribution over all 
{\em possible worlds}, that is, over all possible joint assignments of truth values to the variables $X_1,...,X_k$. 
Computing the world
with the maximum joint probability is equivalent to
solving a weighted MaxSat problem. 
Finding this world is known as 
{\bf MAP inference}, 
where MAP stands for \emph{maximum a posteriori}. 
Like MaxSat, this computation is NP-hard
(or, more precisely, \#P-complete, a related 
and potentially harder complexity class for 
counting problems) \cite{DBLP:journals/ai/Roth96}.
Practical methods for MAP inference, therefore,
resort to 
{\bf Monte Carlo sampling}, most notably, 
{\em Gibbs sampling}, or to variational calculus (see also Section \ref{ch4-subsec:CRF}
on CRFs). 
Approximation algorithms for MaxSat and ILP
have been used as well (see, e.g., \cite{DBLP:journals/corr/abs-1206-3282}).
In addition to the joint MAP, MLN inference can also
compute the {\em marginal probability} for each variable
which can be interpreted as the confidence in an
individual statement being valid.
Unfortunately, this task comes at even higher
computational complexity
than MAP, and is hardly supported by any software tools
for probabilistic graphical models.

The weights for the MLN clauses
can be automatically learned from
training data, without reliance
on human inputs or external statistics.
However, this learning is itself
an expensive task, as it involves
non-convex optimization. It is
typically addressed by gradient
descent methods (see also
Section \ref{ch4-subsec:CRF}).

MLNs have been used for a 
variety of 
probabilistic reasoning
tasks~\cite{DBLP:journals/ml/RichardsonD06,DomingosLowd2009,DBLP:journals/cacm/DomingosL19}, including entity linking
(e.g., \cite{DBLP:conf/icdm/SinglaD06}) and the extraction of is-a and part-of relationships from text sources
(e.g., 
\cite{DBLP:conf/acl/PoonD10}).
Also, advanced methods for
{\em minimum-repair database cleaning} have made clever use
of MLN inference \cite{DBLP:journals/pvldb/RekatsinasCIR17}.

A prominent case of using MLN models for KB construction is the {\bf DeepDive} project
({\small\url{http://deepdive.stanford.edu/}}), by
\citet{Shin:VLDB2015,DBLP:journals/cacm/ZhangRCSWW17,DBLP:conf/sigmod/Zhang0RCN16}.
.

It comprises a framework and software suite for constructing KBs from scratch as well as augmenting them in an incremental manner.
DeepDive has been applied to build
domain-specific KBs, for example,
for paleobiology, geology and crime fighting
(fighting human trafficking \cite{kejriwal2017knowledge}).
Usability for knowledge engineers is
boosted by having a declarative
interface for inputs,
with automatic translation into
MLNs and, ultimately, MRFs.
Nevertheless, 
high-quality KB construction
requires substantial care and effort for
proper configuration, training, tuning and
other human intervention \cite{DBLP:conf/sigmod/Zhang0RCN16}.
Like with MaxSat, a major concern
for scalability is that the grounding
step -- moving from logical constraints with quantified variables to fully instantiated clauses -- comes with the risk of combinatorial explosion. To mitigate this issue,
the DeepDive project has developed
effective techniques 
for {\bf lazy grounding}, avoiding unnecessary instantiations \cite{DBLP:journals/pvldb/NiuRDS11}.

Another prominent system that makes use of MLNs is {\bf StatSnowball}, by 
\citet{DBLP:conf/www/ZhuNLZW09} and~\cite{Nie:IEEE2012}.
This system
uses both predefined MLN constraints and logical patterns that are learned at run-time. The former express prior knowledge, such as 
$hasMother(x,y) \Rightarrow hasChild(y,x)$.
The learned kind of soft invariants, on the other hand, capture the relationship between textual patterns and 
the relations and their arguments, similar to 
the statement-pattern duality in DIPRE \cite{DBLP:conf/webdb/Brin98} 
(see Section \ref{ch4:subsec:PatternLearning})
and the reasoning for extraction in SOFIE \cite{suchanek2009sofie} (see Section \ref{ch8-subsec:MaxSat}).

\clearpage\newpage
\noindent{\bf Probabilistic Soft Logic (PSL):}\\
MLNs relax the Weighted MaxSat model by admitting many sub-optimal worlds with lower probabilities. In each of these worlds, a Boolean variable is true or false, with certain probabilities. This setting can be relaxed further
by allowing different {\em degrees of belief} in
a variable being true 
(\citet{DBLP:journals/jmlr/BachBHG17}).
Essentially, the discrete optimization
problem of MaxSat and MLNs is relaxed into a continuous
optimization by making all random variables real-valued.
This technique is analogous to relaxing ILPs into LPs
(see Section \ref{ch8-subsec:ILP}).
Note that these real-valued degrees of truth are different
from the probabilities (which are real-valued anyway, for
discrete and continuous models alike). 
Every combination of real-valued belief degrees, on a continuous scale
from 0 to 1, is associated with a probability density. %

For making this approach tractable, 
the truth value, or
{\em interpretation} $I$, of a conjunction of variables is defined as the Lukasiewicz t-norm: $I(v \wedge w) = max~(0, ~ I(v)+I(w)-1)$. Each clause $a_1 \vee ... \vee a_n$ is treated as a rule $\neg a_1 \wedge ... \wedge \neg a_{n-1} \Rightarrow a_n$, with body $\neg a_1 \wedge ... \wedge \neg a_{n-1}$ and head $a_n$
(see Section \ref{ch8-subsec:rulemining}). 
The \emph{distance from satisfaction} of this rule in an interpretation $I$ is defined as $d_I(body \Rightarrow head) = max~ (0, ~ I(body)-I(head))$. This yields the following setting:

\begin{samepage}
\begin{mdframed}[backgroundcolor=blue!5,linewidth=0pt]
\squishlist
\item[ ] {\bf Probabilistic Soft Logic (PSL) Program:}\\
For a set of rules $r_1,...,r_n$
(including atoms for statement candidates)
with weights $w_1,...,w_n$, a PSL program computes the probability distribution over interpretations $I$:\\
\hspace*{0.5cm} $P[I] = \frac{1}{Z} exp(-\sum_{i=1..n} w_i d_I(r_i))$\\
where $Z$ is a normalizing constant. 
\squishend
\end{mdframed}
\end{samepage}

Like for MLNs, the central task in PSL is 
MAP inference, for the value assignment (degrees of being true)
that maximizes the joint probability over all variables together.
\citet{DBLP:journals/jmlr/BachBHG17}
have
shown that, unlike for discrete MLNs, this can be computed in polynomial time.
By using a Hinge-loss objective
function, MAP inference becomes a convex optimization problem
-- still not exactly fast and 
not easily scalable, but no longer NP-hard.
PSL has been used for a variety of tasks, from entity linking
\cite{DBLP:conf/icdm/KoukiPMKG17} and paraphrase learning
\cite{DBLP:conf/emnlp/GrycnerWPFG15}
to KB cleaning \cite{DBLP:conf/aaaifs/Pujara0GC13}.

\vspace*{0.2cm}
\noindent{\bf Further Works on Probabilistic Factor Graphs:}\\
Numerous other works have leveraged variants of probabilistic graphical models for different aspects of KB construction and curation. These include {\em constrained conditional models} by
\citet{DBLP:journals/ml/ChangRR12}, 
{\em factor graphs} for joint extraction of entities and
relations \citet{DBLP:conf/pkdd/RiedelYM10,DBLP:conf/emnlp/YaoRM10} (see also Section \ref{ch6:subsubsec-featurebasedclassifiers}),
{\em coupled learners} for the NELL project \citet{nell,nell-2018}
(see also Section \ref{sec:nell}), and others.

\section{KB Life-Cycle}
\label{ch8:sec:kb-life-cycle}

Construction and curation of a knowledge base
is a never-ending task.
As the world keeps evolving, knowledge acquisition
and cleaning needs to follow these changes,
to ensure the freshness, quality and utility of the KB.
This life-cycle management, over years and decades,
entails several challenges.
Surprisingly, these issues have received
little attention in the KB research community.
There are great opportunities for novel and
impactful work.

\subsection{Provenance Tracking}
\label{ch8-subsec:provenance}

While provenance of data is a well-recognized concern for database systems, with long-standing research
(e.g., {\color{purple} \citet{DBLP:conf/icdt/BunemanKT01} and~\cite{DBLP:journals/sigmod/BunemanT18}}),
it is largely underrated and much less explored
for knowledge bases.
In a KB, each statement should be annotated with \textbf{provenance metadata} about:
\squishlist
\item the {\bf source} (e.g., web page)  from where the statement was obtained, or sources when multiple inputs are combined,
\item the {\bf timestamp(s)} of when the statement was acquired, and
\item the {\bf extraction method(s)} by which it was acquired, for example, the rule(s), pattern(s) or classifier(s) used.
\squishend
This is the minimum information that a high-quality KB
should capture for manageability (see, e.g., \cite{Carlson:AAAI2010,DBLP:journals/ai/HoffartSBW13}). 
In addition, for learning-based extractions, the underlying {\em configuration} (e.g., hyper-parameters)
and {\em training data} should be documented.
For human contributions, the assessment and
approval steps, involving {\em moderators/curators},
needs to be documented (see, e.g., \cite{DBLP:conf/semweb/PiscopoKPS17,DBLP:conf/socinfo/PiscopoPS17}).

This metadata is crucial for being able to trace errors back to their root cause, when they show up later: the spurious statements, the underlying sources and the
extraction methods.
This way, provenance tracking supports removing
incorrect KB content or correcting the errors.
For query processing,
provenance information can be propagated through query operators. This allows tracing each query
result back to the involved sources \cite{DBLP:conf/icdt/BunemanKT01},
an important element of explaining answers to
users in a human-comprehensible way.

Provenance information 
can be added as additional arguments to relational 
statements, in the style of this example:\\
\hspace*{0.5cm} \ent{$\langle$Bob Dylan, won, Nobel Prize in Literature},\\ 
\hspace*{0.5cm} \ent{source: {\small\url{www.nobelprize.org/prizes/literature/2016/dylan/facts}}, \\
\hspace*{0.5cm} extractor:lstm123$\rangle$},\\
or by means of reification 
and composite objects
in RDF format (see Section \ref{ch2-subsec:properties}).
Another option is to group an entire
set of statements into a  
{\bf Named Graph} \cite{hartig2009provenance}, a W3C-approved way of considering a set of statements as an entity~\cite{sparql,nquads}. Then, provenance statements can be made about such a group entity. 
Yet another option is RDF*~\cite{rdf-star,rdf-star-poster}, a proposed mechanism for making statements about other statements. The most common solution, however, is what is known as {\bf Quads}: triples that have an additional  component that serves as an identifier of the triple. Then, provenance statements can be attached to the identifier. Conceptually, this technique corresponds to a named graph of exactly one statement; it is used, for example, in YAGO~2~\cite{DBLP:journals/ai/HoffartSBW13} and 
in Wikidata~\cite{Vrandecic:CACM2014}.
Various triple stores support quads.

\subsection{Versioning and Temporal Scoping}
\label{ch8-subsec:versioning}

Nothing lasts forever: people get divorced and marry again, and even capitals of countries change once
in a while. For example, Germany had Bonn as its
capital, not Berlin, from 1949 to 1990.
Therefore, 
{\em versioning} and {\em temporal scoping} 
of statements are
crucial for proper interpretation of KB contents.
In addition, versioning rather than
over-writing statements is also useful
for quality assurance and long-term maintenance.
For database systems this is an obvious issue,
and industrial KBs have certainly taken this into account
as well. However, academic research has
often treated KBs as a one-time construction effort,
disregarding their long-term life-cycle.

\vspace*{0.2cm}
\noindent{\bf Versioning:}\\
Keeping all versions of statements,
as relationships between
entities are established and dissolved, supports
{\em time-travel search}: 
querying for knowledge as of a given point in the past
(e.g., for Germany's capital in 1986).
KB projects like DBpedia and YAGO
approximated version support by periodically releasing new versions of the entire KB. 
In addition, YAGO~2 
\cite{DBLP:journals/ai/HoffartSBW13}
introduced systematic annotation
with temporal scopes to many of its statements
(see below).
The Wikidata KB keeps histories of individual items
(e.g., {\small \url{https://www.wikidata.org/w/index.php?title=Q392&action=history}} about Bob Dylan) %
and supports convenient access to earlier versions
on specific entities.
Also, SPARQL queries can be performed over 
the Wikidata edit history \cite{tanon2019querying}.

\vspace*{0.2cm}
\noindent{\bf Temporal Scopes:}\\
We aim for temporally scoped statements, by
annotating SPO triples with their {\bf validity times},
which can be timepoints or time intervals.
This can be expressed either via higher-arity
relations, such as

{\small
\lstset{language=HTML,upquote=true}
\begin{lstlisting}
     wonPrize (EnnioMorricone, Grammy, 11-February-2007)
     wonPrize (EnnioMorricone, Grammy, 8-February-2009)
     capital (Germany, Bonn, [1949-1990])
     capital (Germany, Berlin, [1991-now])
\end{lstlisting}
}%

\noindent or by means of reification and composite objects
(see Section \ref{ch2-subsec:properties}).

Note that the temporal scopes can have different
granularities: exact date, month and year, or only the year. The choice depends on the nature of the event
or temporal fact, and also on the precision of
how they are reported in content sources.

\vspace*{0.2cm}
\noindent{\bf Discovering and Inferring Temporal Scopes:}\\
Given a statement, such as
\ent{married (Bob Dylan, Sara Lownds)},
how can we determine the begin and end of
its validity interval,
and how do we go about assigning timepoints to
events?
There is a variety of methods for spotting
and normalizing
{\bf temporal expressions} in text and semi-structured
content, based on rules, patterns or
CRF/LSTM-like learning -- see, for example, 
\cite{DBLP:conf/acl/CunninghamMBT02,DBLP:conf/acl/VerhagenMSLKJRPP05,DBLP:conf/acl/FinkelGM05,DBLP:journals/ai/HoffartSBW13,DBLP:conf/cikm/KuzeyVW14,DBLP:journals/coling/StrotgenGHH18}.
Methods for property extraction (see Chapter \ref{ch4:properties})
can be applied to linking these time points or
intervals to respective entities, and thus 
assigning them to SPO triples, at least
tentatively.
In addition, special pages, categories and lists
in Wikipedia, such as monthly and daily events
(e.g., {\small\url{https://en.wikipedia.org/wiki/Portal:Current_events/2020_July_14}})
and annual chronologies
(e.g., {\small\url{https://en.wikipedia.org/wiki/2020}})
provide great mileage
\cite{DBLP:conf/www/KuzeyW12,DBLP:conf/cikm/KuzeyVW14,DBLP:conf/conll/SilC14}.

The resulting extractions of temporally scoped
statements may result in
noisy and conflicting outputs, though.
For example, we may obtain
statement candidates:

{\small
\lstset{language=HTML,upquote=true}
\begin{lstlisting}
     married (Charles, Diana, [1981-1996]): [0.7]
     married (Dodi, Diana, [1995-1997]): [0.6]
     married (Charles, Camilla, [1990-2020]): [0.5]
     married (Andrew, Camilla, [1973-1997]): [0.4]
\end{lstlisting}
}%

\noindent where the numbers in brackets, following the statements, denote 
confidence scores.
Together, these statements imply some cases of illegal
bigamy: being married to more than one spouse
at the same time.
This is again a case for imposing
consistency constraints and reasoning
to clean this space of candidates.
We can use the repertoire from Section \ref{ch8-sec:consistency-reasoning},
including MaxSat, integer linear programming (ILP), or probabilistic graphical models.
For example, the cleaning task could be cast into
an ILP as follows:

\begin{samepage}
\begin{mdframed}[backgroundcolor=blue!5,linewidth=0pt]
\squishlist
\item[ ] {\bf ILP for Temporal Scoping:}\\
Input: Candidates of the form $\langle{}S,P,O,T\rangle:w$,\\
with time interval $T$ and confidence score $w$\\
Decision Variables: 
\squishlist
\item $X_i$ = 1 if candidate $i$ is accepted, 0 otherwise
\item $P_{ij}$ = 1 if candidate $i$ should be ordered before $j$
\squishend
Objective Function: maximize $\sum_i w_i \cdot X_i$\\
Constraints:
\squishlist
\item $X_i + X_j \le 1$ if candidates $i$ and $j$
overlap in time and are conflicting
(e.g., violating the monogamy law)
\item $P_{ij} + P_{ji} \le 1$ for all $i,j$, for acyclic ordering
\item $(1-P_{ij}) + (1-P_{jk}) \ge (1-P_{ik})$ for
all $i,j,k$, for transitivity if $i,j,k$ must be ordered
\item $(1-X_i) + (1-X_j) + 1 \ge (1-P_{ij}) + (1-P_{ji})$
for all $i,j$ that must be totally ordered -- in order to couple the $X_i$ and $P_{ij}$ variables
\squishend
\squishend
\end{mdframed}
\end{samepage}

This ILP model can be seen as a template
for all kinds of temporal-scope constraint
reasoning, generalizing the anti-bigamy case at hand.

Unfortunately, by the temporal overlap 
of our four candidate statements, 
this would conservatively accept  only the
first statement about Charles and Diana
being married from 1981 through 1996
and the last statement about Andrew and Camilla.
A technique to improve the recall of such
reasoning is to decompose the temporal scopes
of the candidates
into disjoint time intervals, cloning
the non-temporal parts of the statements.
For example, the marriage of Charles and Camilla
is split into one statement for
 [1990,1996] and another one for [1997,2020].
This would allow the reasoner to reject the
scope [1990,1996] while accepting [1997,2020].

Such techniques and the application of
consistency reasoning for temporal scoping
of KB statements have been investigated by
\cite{DBLP:conf/aaai/LingW10,DBLP:conf/wsdm/TalukdarWM12,DBLP:conf/cikm/TalukdarWM12,DBLP:conf/cikm/WangYQSW11, DBLP:conf/acl/WangDSW12}.

A variation of this theme is to discover and infer
the {\bf relative temporal ordering}
between events and/or the validity of statements.
This aims to detect relationships like
\ent{happened-before}, \ent{happened-after},
\ent{happened-during} etc.
Methods along these lines include
\cite{DBLP:conf/semeval/UzZamanLDAVP13,DBLP:journals/jbi/JindalR13,DBLP:conf/coling/MirzaT16,DBLP:conf/emnlp/DasguptaRT18,DBLP:conf/emnlp/NingSR19,DBLP:conf/emnlp/JainRMC20}
(see also the tutorial \cite{MuhaoChen:AAAI2021} and further references given there).

\subsection{Emerging Entities}
\label{ch8-subsec:emergingentities}

An important aspect of evolving knowledge is to cope with
newly \textbf{emerging entities}.
When discovering entity names in Web sources,
we aim to disambiguate them onto already known entities in the KB; see Chapter \ref{ch3-sec-EntityDisambiguation}. 
However, even if there is a good match in the KB, it is not necessarily the proper interpretation.
For entity linking, this is the 
{\bf out-of-KB entity} problem %
\cite{DBLP:journals/tacl/LingSW15}.

For example, when the documentary movie {\em ``Amy''} 
was first mentioned a few years ago, it would have been tempting to link
the name to the soul singer \ent{Amy Winehouse}. 
However, although the movie is about the singer’s life, the two entities must not be confused. 
Today, after having won an Oscar, the movie is, of course,
a registered entity in all major KBs.
The general situation, though, is that there will
be a delay between new entities coming into
existence and becoming notable.
Likewise, existing entities that are not notable enough
to be covered by a KB may become prominent overnight,
such as indie musicians getting popular
or startup companies getting successful.
Recognizing these as early as possible can
{accelerate} the KB growth, and most importantly,
it is crucial to avoid confusing them with
pre-existing entities that have the same or similar names.

To address the problem, 
the methods for entity linking (EL) (see Chapter
\ref{ch3-sec-EntityDisambiguation}) always 
have to consider
an additional candidate \ent{out-of-KB}
for the mapping of an observed mention.
Whenever the EL method has higher confidence
in this choice than in any 
candidate entity
from the KB, the mention should be flagged accordingly.
As confidence scores are not always
well-calibrated,  
more sophisticated scoring and score-calibration
methods have been investigated 
by
\citet{DBLP:journals/ai/HacheyRNHC13} and \citet{hoffart2014discovering}.

These techniques are conservative, in that they
avoid incorrectly mapping an emerging entity onto
a pre-existing one. However, this alone is not sufficient, because we do want to add the 
emerging entity to the KB at some point.
Moreover, there could be multiple out-of-KB
entities with the same or similar names
in the discovered text mentions.
For example, in addition to the movie ``Amy''
(about Amy Winehouse), the character ``Amy''
(Farrah Fowler) from the TV series
``Big Bang Theory'' could also become
a candidate for addition to the KB.

An approach to handle such cases has been
proposed in \cite{DBLP:conf/www/HoffartMWAS16}.
For each mention mapped to \ent{out-of-KB},
a {\em contextual profile} is created and
maintained.
This comprises the mention itself
and keyphrases from the surrounding contexts
or latent models derived from contexts
(cf. Section \ref{ch5-subsec:contextsim4EL}).
The profile is gradually enhanced as we observe
more mentions of what is likely the same out-of-the-KB
entity.
After a while, we obtain a repository of emerging entity names with their
contextual profiles. 
When the profile of a name is rich enough
to infer its semantic type(s), such as
\ent{documentaryMovie} or \ent{fictitiousCharacter},
we may consider adding the emerging entity into the KB,
properly registered with its type(s).

There are still two caveats to consider.
First, although there is initial evidence for
an emerging out-of-KB entity, it may turn out later
that this actually denotes an already known entity
in the KB. So the method has to periodically
reconsider and possibly revise the EL decision.
Second, as the same name may denote multiple
out-of-KB entities, the contextual profile for
this name could improperly conflate more
than one emerging entity. The ``Amy'' scenario
is an example.
To handle such cases, the method needs to
consider splitting a profile, to identify
strongly coherent contexts -- one for each
emerging entity.
In the long run, human-in-the-loop curation 
may still be required for keeping the KB
at its high quality.

Further methods for early discovery
of emerging entities have been developed,
for example, by \cite{DBLP:conf/bigdataconf/JanssonL17,DBLP:journals/tkde/YeoCPH17,DBLP:conf/ijcai/Akasaki0T19,DBLP:conf/edbt/OulabiB19,DBLP:conf/www/0006MBR20},
with emphasis on social media or web tables as sources.
For entities of type \ent{event}, news digests
such as daily pages in Wikipedia
(e.g., {\small\url{https://en.wikipedia.org/wiki/Portal:Current_events/2020_July_14}})
are valuable assets, too. 
Note that these daily pages often contain headlines
that are not yet covered by any of the
encyclopedic articles in Wikipedia.

Finally, a methodological related issue is
to correct existing KB statements where
the object is simply a string literal,
such as ``Amy -- Oscar-winning documentary movie, 2015''
in a triple like 
$\langle$\ent{AmyWinehouse featuredIn}
``Amy \dots 2015''$\rangle$.
This is a frequently arising case as
the initial knowledge acquisition may
only pick up strings but miss out on
proper entity linking.
\cite{DBLP:conf/www/ChenCHMJ20} has
developed a general framework for
correcting these omissions, utilizing
EL techniques and consistency constraints.

\section{Take-Home Lessons}

We highlighted that knowledge bases require continuous
curation, from quality assessment to quality assurance.
Key lessons are the following:

\squishlist
\item Quality measures for correctness and coverage are
computed by {\em sampling} statements with human judgements, 
by crowdsourcing or, if necessary, experts.
\item No KB is ever complete: assessing and predicting
{\em completeness} often builds on the 
Local Completeness Assumption (LCA). 
\item Logical {\em invariants} about the KB content serve twofold roles:
as {\em constraints} they can detect erroneous statements to keep
the KB consistent;
as {\em rules} they can deduce additional statements to fill gaps in the KB.
\item Logical patterns in the KB can be automatically discovered, to {\em learn rules}
and to {\em analyze bias}. 
\item For cleaning candidate statements at scale, {\em constraint-based reasoning}
is a best-practice approach, with a suite of models and methods from
MaxSat and ILP to probabilistic factor graphs.
\item For the long-term {\em life-cycle} of a KB, tracking the provenance of
statements as well as versioning with temporal scopes are essential components.
\squishend

\clearpage\newpage
\chapter{Case Studies}
\label{chapter:casestudies}

\section{YAGO}\label{sec:yago}

The YAGO project ({\small\url{https://yago-knowledge.org}}, since 2006),
by \citet{Suchanek:WWW2007}
(see also \cite{yagojournal,DBLP:journals/ai/HoffartSBW13}),
created one of the first large knowledge bases that were automatically extracted from Wikipedia, 
in parallel to DBpedia, discussed in Section~\ref{sec:dbpedia}. 
YAGO has 
been maintained and advanced by
the Max Planck Institute for Informatics in Germany and Télécom Paris University in France.
The KB %
has been
used in many projects world-wide, most notably for semantic type checking
in the IBM Watson system~\cite{DBLP:journals/ibmrd/Ferrucci12,DBLP:journals/ibmrd/MurdockKWFFGZK12} (which went on to win the
Jeopardy quiz show).

\subsection{Design Principles and History}

\vspace*{0.2cm}
\noindent{\bf Initial YAGO: Core Knowledge}\\
\noindent The key observation 
of YAGO was that Wikipedia contains a large number of 
individual entities,
such as singers, movies or cities, but does not organize them in a semantically clean type system.
Wikipedia's hierarchy of categories was not suitable as a taxonomy. WordNet, on the other hand, has a very rich and elaborate taxonomy, but is hardly populated with instances. 
YAGO aimed to combine the two resources to get the best of both worlds.

The first version of YAGO 
\cite{Suchanek:WWW2007}
converted every Wikipedia article into an entity, and extracted its classes from the categories of the article. To distinguish between thematic categories for human browsing (e.g., \emph{Rock 'n Roll music} for Elvis Presley) and properly taxonomic categories (e.g., \emph{American singers}), YAGO developed the heuristics discussed in Section~\ref{sec:ch3-category-cleaning}.
If the head noun of a category name is in plural form (as in \emph{American singers}), then it is a taxonomic class. 
These leaf-level categories were then linked to WordNet with the methodology 
presented in Chapter \ref{ch2:knowledge-integration}.

The first version of YAGO also extracted selected kinds of facts from Wikipedia categories. A small set of
relations was manually identified (including \emph{has\-Won\-Award}, \emph{is\-Located\-In}, etc.), and 
regular expressions were specified for the corresponding Wikipedia categories (e.g., \emph{Grammy Award winners} or \emph{Cities in France}). 
YAGO extracted labels for entities from Wikipedia redirects, and attached provenance information to each statement. 
The hand-crafted specification included domain and range constraints to eliminate spurious statements.
The objective was to focus YAGO entirely on precision, even if this meant a loss in recall. The rationale was that a KB with  5 million facts and 95\% precision is more useful than a KB with 10 million facts and 80\% precision.

The quality of the extracted statements was evaluated manually
by a sampling technique. A random sample was drawn for each relation, and the statements were manually compared to 
Wikipedia as ground truth. 
The number of samples was chosen so as to bring the Wilson confidence interval of the estimated precision to
$95\%\pm 5\%$.
Refining by relation, some properties 
including the
\ent{type} property 
even reached nearly $99\%$
precision \cite{yagojournal,DBLP:journals/ai/HoffartSBW13}.
For a long time, YAGO was the only major KB that came with such statistical guarantees about its correctness.

In 2008, YAGO was extended with facts from infoboxes of Wikipedia \cite{yagojournal}. 
These were extracted by hand-crafted regular expressions 
(see Section \ref{subsec:specifiedpatterns}) for around 100 selected relations. This process also extracted the validity times of statements
when applicable. 
The 
SPO triple notation was slightly extended to allow
statements with temporal scope, such as:
\begin{lstlisting}
Germany hasGDP "$3,667 trillion" inYear "2008"
\end{lstlisting}
By reification (see Section \ref{ch2-subsec:properties}), this shorthand notation was 
mapped into pure triples (or quads, see 
Section \ref{ch8-subsec:provenance}).

The data model was extended by a taxonomy of types for literal values. For example, the literal  ``\$3,667 trillion'' was linked by the property \ent{has\-Value} to ``3,667,000,000,000'', which is an instance of the class \ent{integer}, a subclass of \ent{number}, which is a subclass of \ent{literal}. 
Likewise, the property \ent{has\-Unit}
could capture the proper currency US dollars.
This further strengthened the ability for early type checking and ensuring near-human quality.

The YAGO KB was the focal point of the broader
{\em YAGO-NAGA} project that included methods and tools for exploring and searching the KB
\cite{DBLP:journals/sigmod/KasneciRSW08}.

\vspace*{0.2cm}
\noindent{\bf YAGO 2: Spatial and Temporal Scoping}\\
\noindent In 2010, YAGO was systematically extended with %
temporal and spatial knowledge~\cite{yago2demo,DBLP:journals/ai/HoffartSBW13}. Entities and statements had time intervals assigned to denote when entities existed and statements were valid. Timestamps were extracted from the Wikipedia infoboxes when possible, and propagated to other statements and entities by a limited form of Horn rules
(see Section \ref{sec:rules}).
For example, if we know the birth date and the death date of Frank Sinatra, then we can deduce the validity interval for the fact that he was a person. Such rules were systematically applied to people, artifacts, events, and organizations, thus giving validity times to about one third of the facts.
This temporal scoping was in turn beneficial as a 
consistency check when extending the KB or building
a new major version.

The spatial dimension of YAGO came from GeoNames
{\small\url{https://www.geonames.org/}}),
a large repository of geographical entities, with 
coordinates and informative types. 
To avoid duplicates, the entities from Wikipedia were matched to the entities in GeoNames by comparing names and geographical coordinates with thresholds on similarity.
This simple entity-matching technique (cf. 
Section \ref{sec:entity-matching}) 
preserved the high quality of
canonicalized entities.
The type taxonomy of GeoNames was  mapped to the class taxonomy of YAGO, by taking into account the name of the GeoNames class, the head noun of that name, the most frequent meaning in WordNet, and the overlap of the glosses in the two resources. 
Similar to the temporal scoping,
Horn rules 
were applied to propagate locations from entities to their statements, and vice versa. %

This KB was accompanied by a query engine~\cite{yago2demo,DBLP:journals/ai/HoffartSBW13}  that allowed searching entities not just by their facts, but also by  validity times, spatial proximity
and space-time combinations. For example, the query\\
\hspace*{0.5cm} \emph{George\-Harrison created ?s after John\-Lennon}\\
\noindent would find songs written by George Harrison after John Lennon’s death, and
the query\\
\hspace*{0.5cm} \emph{guitarists bornIn ?p near Seattle}\\
\noindent would return Jimi Hendrix, Kurt Cobain, Carla Torgerson and more.
Recent work on augmenting YAGO2
with spatial knowledge is the 
{\em Yago2geo} project
\cite{DBLP:conf/semweb/KaralisMK19},
which integrated content from OpenStreetMap
and other sources,
and supports a very expressive query language
called GeoSPARQL.

The extraction patterns
for YAGO2 were specified declaratively.
An example pattern is:
\begin{lstlisting}
"Category:(.+) births" pattern "\$0 wasBornOnDate Date(\$1)"
\end{lstlisting}
\noindent This pattern extracts birth dates from Wikipedia category names. 
The next version of YAGO, YAGO2s~\cite{yago2s} 
further advanced
the principle of declarative specifiction and
modularization. 
The system was factored into 30 \emph{extractors},
each for a 
specific scope.
The extractors were orchestrated by a dependency graph,
where each module depends on inputs from other modules.
A scheduler could run these extractors largely in parallel or in pipelined mode.

This DB-engine-like declarative machinery proved very valuable 
for debugging and quality assurance, and could build
new KB versions very efficiently
(cf. Section \ref{ch6-subsubsec:extractionplans}).

\vspace*{0.2cm}
\noindent{\bf YAGO 3: Multilingual Knowledge}\\
\noindent In 2014, YAGO 3~\cite{yago3} started extracting from
Wikipedia editions beyond English, covering editions like
German, French, Dutch, Italian, Spanish, Romanian, Polish, Arabic, and Farsi
(based on the authors' language skills, for validation of results).
The goal was to 
construct a single, consolidated KB from these 
multilingual sources,
with more entities and statements (as many appear only
in specific editions) but without any duplicates.
The extractors harnessed inter-wiki links between
Wikipedia editions and 
the inter-language links in Wikidata 
({\small\url{https://wikidata.org/}}).
For example, 
 the French Wikipedia article about \emph{Londres} is about the same entity as the English article about \emph{London}. 

A difficulty to address was the extraction from 
non-English infoboxes, without manually specifying
patterns for each different edition.
For this purpose, 
the YAGO 3 system employed distant supervision from
the English edition (see Sections \ref{ch6-sec:properties-from-semistructured}
and \ref{ch7-openie-semistructured}).
By statistically comparing English seeds for 
subject-object pairs against those observed in
non-English infoboxes, the system could learn the
correspondences between properties from different sources.

To construct the taxonomy, the foreign category names were mapped to their English counterparts, by
harnessing inter-language links from Wikidata.
The modular architecture developed for YAGO2s allowed 
all this with just a handful of new extractors:
translation of non-English entities, mapping of infobox attributes to relations, and taxonomy construction.
Downstream extractors were not affected by these extensions.

Another major project on multilingual knowledge,
with even larger coverage, is BabelNet \cite{Navigli:ArtInt2012,DBLP:conf/lrec/EhrmannCVMCN14} 
({\small\url{https://babelnet.org/}}).

\vspace*{0.2cm}
\noindent{\bf YAGO 4: Alignment with Wikidata}\\
\noindent YAGO 3 has been continuously improved, 
and
the software became open source \cite{yago-resource}. 
However, the KB was bound to Wikipedia and the entities featured there. Thus, it became clear that the KB could never reach the scale of entity coverage
that Wikidata achieved in the meantime,
with nearly 100 million entities. 
On the other hand, Wikidata follows the principle of
including claims as statements
(see Section \ref{subsec:wikidata}), with potential diversity of
perspectives, rather than undisputed factual statements
only. Therefore, semantic constraints cannot be rigorously enforced. 
Furthermore, the large number of contributors 
to the Wikidata community has led to a convoluted 
and cluttered taxonomy
of classes, where an entity such as \emph{Paris} is buried under 60, mostly uninformative, classes, 20 of which are called ``object'', ``unit'', ``seat'', ``whole'', etc.

The cleaning of contradictory statements and the transformation of convoluted and noisy taxonomies 
into clean type systems have been key competences of the YAGO project from its very start. 
In this spirit, the latest YAGO version, YAGO 4 \cite{yago4}, 
abandoned Wikipedia as input source, and set out to 
tap into Wikidata as a premium source,
applying the same principles (cf. Chapter \ref{ch2:knowledge-integration}).
The higher-level types of the taxonomy are no longer 
based on WordNet; instead YAGO 4 adopts the
type system of
{\em schema.org} ({\color{purple}\citet{Guha:CACM2016}}),
an industry standard for semantic markup in web pages.

The directly populated leaf-level classes of YAGO 4 are carried over
from Wikidata. In this regard, Wikidata is fairly clean
and coarse-grained; most entities belong to only one or two types directly. For example, all people have type \ent{human} and
none of the conceivable fine-grained types such as 
singer, guitarist etc. Wikidata expresses the latter by means of various properties.
Since schema.org has only ca. 1000 classes and the overlap
with the immediate types of Wikidata instances is even smaller, 
it was best to manually align the relevant types from the
two sources.
Properties are also adopted from schema.org, with the advantage
that they come with clean type signatures for domain and range.
Again, the alignment with Wikidata properties required 
a reasonably limited amount of manual work.
The entities, Wikidata's best asset, were transferred from
Wikidata to populate the newly crafted KB schema.
This data was complemented by hand-crafted consistency constraints for class disjointness, functional dependencies and inclusion dependencies (see Section \ref{sec:intensional}),
expressed in the SHACL language (see Section~\ref{sec:constraints}).

The resulting KB comprises ca. 60 million entities with
2 billion statements, organized into a clean and
logically consistent taxonomic backbone.
For this high quality, the KB construction ``sacrificed''
about 30 million Wikidata entities that had to be
omitted for consistency.
However, these affect only the long tail of less notable
entities which have very few properties.
As a result of this constraint-aware construction process, 
the YAGO 4 knowledge base 
is ``reason-able'' \cite{yago4}:
provably consistent and amenable to OWL reasoners.
Metadata, about provenance, is represented in the RDF* format~\cite{rdf-star}.

\subsection{Lessons Learned}

YAGO was one of the first large KBs automatically constructed from web sources. 
Its 
unique traits are
high precision, semantic constraints, and a judiciously constructed and fairly comprehensive type taxonomy. 
Over YAGO's 15-year history,
several major lessons were learned:

\begin{itemize}
\item \textbf{Harvest low-hanging fruit first:} YAGO was  successful because it focused on premium sources that were comparatively easy to harvest and could yield high-quality output: the category system of Wikipedia and semi-structured content like infoboxes. 
This resulted in near-human precision that was previously
unrivaled by automatic methods for information extraction
at this scale.
\item \textbf{Focus on precision:} YAGO has  focused on precision at the expense of recall. The rationale is that every KB is incomplete, and that applications are thus necessarily prepared to receive incomplete information. Under this regime, a user is not surprised if some song or city is missing from the KB. 
Conversely, users are confused or irritated
when they encounter wrong statements. 
Therefore, YAGO focused on premium sources and relatively
conservative extration methods, reserving more aggressive
methods for KB augmentation. The project investigated and
developed a variety of such advanced methods as well,
including the SOFIE tool \cite{suchanek2009sofie} (see Section \ref{ch8-subsec:MaxSat}), but 
these had limited impact on the more conservative releases of the YAGO KB.
Nevertheless, SOFIE and its scalable
parallelization \cite{nakashole2011scalable}
were successfully used in another project
on building a health KB called 
{\em KnowLife} \cite{Ernst:BMCbioinformatics2015,Ernst:ACL2016}.
\item \textbf{Limited-effort manual contributions:} 
The project identified sweet spots where limited manual effort had a very large positive impact. 
The specification of properties and their type signatures is a case in point. It is not much effort to define hundreds of properties 
and constraints by hand: 
a small price when these can guarantee the 
cleanliness and tangiblity
of 
many million statements.
\item \textbf{Modularized extractors:} 
There is enormous value in unbundling the KB construction code
into smaller extractor modules, each with a specific scope.
This allows declarative orchestration of an entire 
extractor ensemble, and
greatly simplifies maintenance, debugging, and 
project life-cycle.
\item \textbf{Use open standards:} 
The adoption of open standards (like RDF, RDFS, RDF*~\cite{rdf-star}, and SHACL)
boosts the usage and utility of
KB resources and APIs by a wider community of researchers and developers.
This 
includes
the representation of statements.
In the early years of YAGO, many questions arose about the 
syntax of KB statements, character encodings,
reserved characters, escape conventions, etc.
These could have been avoided by an earlier full adoption of RDF.  
\end{itemize}

\section{DBpedia}\label{sec:dbpedia}

DBpedia ({\small\url{https://dbpedia.org}}, starting in 2007), by
\citet{Auer:ISWC2007},
was 
the other early project
to construct a large-scale knowledge base from Wikipedia contents 
(see also \cite{DBLP:conf/esws/AuerL07,DBLP:journals/semweb/LehmannIJJKMHMK15}).
DBpedia spearheaded the idea of 
{\em Linked Open Data (LOD)} 
(\citet{HeathBizer2011}):
a network of data and knowledge bases in which equivalent entities are interlinked
by the {\em sameAs} predicate.
The LOD ecosystem has grown to ten thousands of sources, with DBpedia as a central hub
(see also Section \ref{subsec:webofdata}).

DBpedia targeted infobox attributes right from the beginning (cf. Section \ref{subsec:specifiedpatterns}):
every distinct attribute was cast into a property type, without manual curation. 
Thus, DBpedia could not apply domain and range constraints and other steps for canonicalization and cleaning. In return, DBpedia captured all information of the Wikipedia infoboxes, and thus provided much larger coverage than YAGO. 
Over the next years~\cite{DBLP:journals/ws/BizerLKABCH09}, the project also extracted 
further contents; 
abstracts,  images, inter-wiki links between different language editions,  redirect labels, 
category names, geo-coordinates, external links and more, becoming a ``Wikipedia in structured form''.

In 2009, DBpedia started organizing the entities into a small hand-crafted taxonomy~\cite{DBLP:journals/ws/BizerLKABCH09}, driven by the most frequent infobox templates on Wikipedia. 
It also specified clean property types for these classes, and manually mapped the infobox templates onto properties. To keep its high coverage, DBpedia stored both the raw infobox
attributes
(which cover all infoboxes) and the
selectively curated ones. 
All in all, DBpedia offered four different taxonomies: the Wikipedia categories, the YAGO taxonomy, the UMBEL taxonomy (an upper-level ontology derived from Cyc, see {\small\url{https://en.wikipedia.org/wiki/UMBEL}}), and the hand-crafted DBpedia taxonomy. 
This shows the difficulty of reaching agreement on a universal class hierarchy.
Another novelty in 2009 was ``DBpedia live'', a system that continuously processes the change logs of Wikipedia and feeds them into the KB as incremental updates~\cite{DBLP:conf/otm/HellmannSLA09}. 

\begin{figure}\centering
\includegraphics[height=6cm]{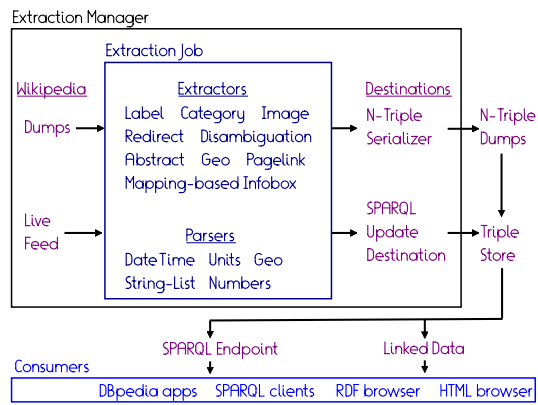}
\caption{Architecture of DBpedia live, simplified from~\cite{DBLP:conf/otm/HellmannSLA09}.}\label{dbpedia}
\end{figure}

Figure~\ref{dbpedia} shows the architecture of DBpedia live. The pivotal element of the system is the triple store for SPO triples (bottom right). By using a triple store instead of plain files, DBpedia can handle live updates directly via SPARQL.
The triple store can also 
provide
the data in different formats to different consumers: as a SPARQL endpoint to Semantic Web applications, as Linked Data to RDF clients, or as HTML for humans. Wikipedia is ingested in two forms (left): by dumps and by the live feed of Wikipedia. The Wikipedia pages are handled by a number of extractors, much like described previously for YAGO (Section~\ref{sec:yago}), in tandem with parsers for dates, measurement units, geographical coordinates, and numbers. Depending on their origin, the extracted statements are then sent to one of two destinations: simple triple dumps (to bootstrap the triple store), or SPARQL updates (to add, delete, ore modify existing statements). This %
system could
sustain
up to 2.5 Wikipedia page updates per second, while Wikipedia had an update rate of 1.4 pages per second, at that time.

Over the next years, the project
built up an international community of contributors~\cite{DBLP:conf/lrec/MendesJB12,DBLP:journals/ws/KontokostasBAHAM12,DBLP:journals/semweb/LehmannIJJKMHMK15}, and added many
useful tools including SPARQL endpoint and other APIs
as well as the {\em Spotlight} tool for
named entity recognition and disambiguation \cite{DBLP:conf/i-semantics/MendesJGB11} (cf. Section \ref{subsec:EL-optimization-and-LTR}).
Separate DBpedia editions were created for each of ca. 100 language editions of Wikipedia. 
These DBpedias are independent, but volunteers in different countries mapped infobox attributes of non-English Wikipedias to the common DBpedia schema.
Finally, DBpedia dealt with the rise of Wikidata
(see Section \ref{subsec:wikidata}) by incorporating its entities and properties~\cite{DBLP:journals/semweb/IsmayilovKALH18} while keeping its
genuine taxonomy. Mappings between
Wikidata properties and the DBpedia schema
were manually specified by the community.

Since 2014, DBpedia is run by the
DBpedia Association,
an 
organization 
({\small\url{https://wiki.dbpedia.org/dbpedia-association}})
with 
regional chapters in 15 countries.

\section{Open-Web-based KB: NELL}
\label{sec:nell}

\subsection{Design Principles and Methodology}

The Never-Ending Language Learner NELL ({\small\url{http://rtw.ml.cmu.edu/}}, starting in 2010), 
by \citet{Carlson:AAAI2010} 
(see also \cite{DBLP:conf/wsdm/CarlsonBWHM10,nell,nell-2018}),
is a project at Carnegie Mellon University to build a knowledge base ``ab initio'' from any kinds of web sources. 
NELL distinguishes itself from other KB projects %
by its paradigm of {\em continuously running}
over many years, the idea being that
the KB is incrementally grown and that
the underlying learning-based extractors 
would gradually improve in both precision and recall.
The key principle to tackle this ambitious goal is \emph{coupled learning}:
NELL has learners for several tasks, and these tasks are coupled to support each other. For example, learning that {\em Elvis} is a singer reinforces the confidence in an extraction that Elvis released a certain album, and vice versa.
Likewise, learning that {\em Paris} is the capital of France  from a textual pattern and learning it from a web table strengthens the belief that this is a correct fact.
This is highly related to harnessing
soft constraints for consistency
and to the factor coupling of
probabilistic graphical models
(see Section \ref{ch8-sec:consistency-reasoning}).

NELL %
starts with
a manually created schema, with ca. 300 classes
and ca. 500 binary relations with type signatures. 
NELL 
boostraps its learners
with a few tens of
labeled training samples for each class and each relation, for example, \emph{guitar} for the class \emph{music\-Instruments}, 
\emph{Colorado} for the class \emph{rivers},
\emph{Paris/Seine} for the relation
\emph{cityLiesOnRiver},
and \emph{Elvis/Jailhouse\-Rock} for the relation \emph{released}. 
The extractors run on a large pre-crawled Web corpus, with the following central learning tasks: 
\begin{itemize}
\item \textbf{Type classification:} Given a noun phrase 
such as ``Rock en Seine'' or
``Isle of Wight Festival'',
classify it 
into 
one or more of the 300 classes, 
like \ent{musicfestival} and \ent{event}
(cf. NER and entity-name typing in
Section \ref{ch4-subsec:CRF}).
NELL uses different learners for this task, based on: string features (e.g., learning that
the suffix ``City'' in a compound noun phrase
often identifies cities, such as ``New York City''), textual patterns (such as ``mayor of X''), appearance in web tables (e.g., appearance in a column with other entities that are known to be cities),
image tags and visual similarities,
and embedding vectors. 
\item \textbf{Relation classification:} Given a pair of noun phrases, classify it into one or more of the relations, this way gathering instances of
the relations.
Again, NELL uses an ensemble of several learners, some of which are similar to the type classifiers. Features include textual patterns, the DOM-tree structure of web pages, and word-level embeddings (cf. Chapter \ref{ch4:properties}).
\item \textbf{Synonymy detection:} Given a pair of noun phrases, detect whether they denote the same entity (e.g., ``Big Apple'' and ``New York City''). 
Several supervised classifiers are employed, using features like string similarity or
co-occurrence with pairs of noun-phrase entity mentions (cf. Section \ref{subsec:EL-optimization-and-LTR}).
\item \textbf{Rule mining:} NELL can learn a restricted form of Horn rules as soft constraints, such as: if two people are married they (usually) live in the same city. These rules are used to predict new statements and to constrain noisy candidate statements
(cf. Section \ref{sec:intensional}),
\end{itemize}

\begin{figure}\centering
\includegraphics[height=6cm,trim=0mm 0mm 0mm 0mm]{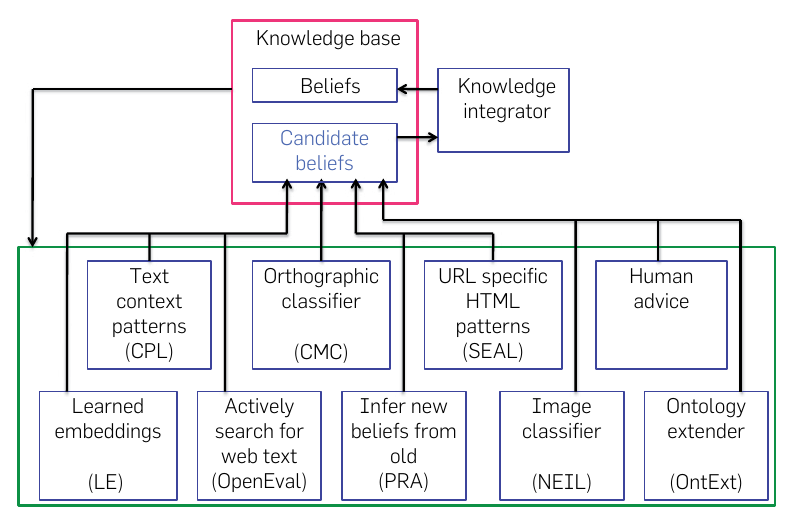}
\caption{Architecture of NELL, adapted from~\cite{nell-2018}.}\label{nell-fig}
\end{figure}

These learning and inference tasks are coupled: results of one task serve as training samples, counter-examples or (soft) constraints for other tasks. 
For example, one coupling constraint is that 
in an ensemble of type classifiers for
noun phrases, all classifiers should agree on the predicted label. Another constraint is that instances of a class should also be instances of the class's superclasses, that instances cannot belong to mutually exclusive classes, 
and that domain and range constraints must
be satisfied (cf. Section \ref{sec:intensional}).
In the same spirit, classifiers receive positive feedback if their classification corresponds to the predictions of NELL's learned rules.
Overall, NELL 
has 
more than 4000 (instantiations of)
learning tasks and more than a million
(instantiations of) constraints. Figure~\ref{nell-fig} shows a high-level view of NELL's architecture. Just like YAGO and DBpedia, NELL hosts different types of extraction modules (bottom). Unlike the two other systems, though, the NELL modules extract from sources of completely different nature: natural language text, tables, 
embeddings, images, and also human input. The central 
element
of the architecture is the feedback loop between the KB and the extraction modules: The modules deliver belief candidates, these are consolidated into beliefs (statements with high confidence), and these serve to re-train the modules.

To implement this retraining, NELL runs an infinite loop of two alternating steps, which are loosely modeled after the EM (Expectation Maximization) algorithm.
In the E-step, NELL re-estimates the probability of each statement in its KB, by combining and reconciling the inputs from the different learners. In the M-like step, this 
confidence-refined
KB is used to re-train the learners. To avoid semantic drift, humans intervene from time to time, and correct aberrant patterns or statements. Overall, NELL has learned 
more than 100 million
confidence-weighted statements, 
including ca. 3 million with high confidence.
Interestingly, in a sequence of such EM-style epochs, NELL can ``unlearn'' statements and rules that were accepted in previous rounds
(i.e., lower its confidence in these).
Nevertheless, subsequent epochs may learn these again (i.e., increase confidence).

As for its schema, NELL has also some means
for discovering new relations, to be added
to the ontology
\cite{DBLP:conf/emnlp/MohamedHM11}.
This is based on the 
{\em Path Ranking Algorithm} by
\cite{DBLP:conf/emnlp/LaoMC11,DBLP:conf/acl/LaoMC15}, which computes frequent edge-label
sequences on paths of the KB graph for
view discovery and
rule learning (see Sections
\ref{ch7-subsec:viewdiscovery} and
\ref{ch8-subsec:rulemining}).
The head of a newly learned rule can
be interpreted as a new predicate.
For example, NELL could potentially discover
the relation \ent{coveredBy} between musicians 
from path labels
\ent{created} (between musicians and songs) and \ent{performed}$^{-1}$ (between songs and musicians).

The NELL website allows visitors  to give feedback about statements, which not
only corrects errors but also provides
cues for future learning rounds.
NELL could even proactively solicit feedback on uncertain statements via Twitter.

Last but not least, 
the NELL project also considered a limited
form of {\em introspection}: automatically
self-reflecting on the weak spots in the KB.
By its ensemble learners, the confidence of
statements and rules can be estimated in a
calibrated way. This way, the learning
machinery can be steered towards obtaining
more evidence or counter-evidence on
low-confidence beliefs and weak rules.

\subsection{Lessons and Challenges}

The overview article \cite{nell-2018} provides specific references for all these methods, the learners, and the potential extensions. It also discusses lessons learned.
The overarching insight is that
NELL's principle of coupling different learners allows it to achieve good results with a small number of training samples for bootstrapping. %
Constraints are the key to tame noise and
arrive at high-confidence statements.

NELL is also facing a number of open challenges:
\begin{itemize}
\item \textbf{Beware of the long tail:}
Prominent instances of classes and relations are learned fast, but less frequently mentioned instances are more difficult to extract.
This leads to the problem of when to cut off the extractions, and more generally, how to cope with the inevitable trade-off between precision and recall. The NELL KB is essentially
a probability distribution over statements and rules with a long tail of low-confidence beliefs. Interpreting this 
{\em uncertain KB} is a challenge for
downstream applications like querying and
reasoning.
\item \textbf{Learning convergence:} 
Another challenge is caused by NELL's 
never-ending learning, which makes it hard to
tell when some task is 
completed (cf. Section \ref{ch8-sec:quality-assessment}).
For example,
the world has only ca. 200 countries, but NELL will continue trying to find more, learning spurious statements and rules, unlearning them later, and so on. 

\item \textbf{Entity canonicalization:} 
Although NELL can learn to identify synonymous entity names, it has 
insufficient support for entity linking (EL,
see Chapter \ref{ch3-sec-EntityDisambiguation})
to ensure unique representation of entities.
For example, its learner for entity detection and typing yields noun phrases
labeled ``Elvis Presley'', ``Elvis Aaron Presley'', ``legendary Elvis Presley'',
``Elvis the King'', 
``Elvis Presley 1935-1977'', ``Elvis lives'',
``Elvis Joseph Presley'' and many more,
without understanding that all these
(except for the last one) are just 
different names for the same entity.
This lack of canonicalization is an
obstacle for some downstream applications
(e.g., entity-centric analytics) and
also makes it hard to maintain the
KB in a consistent manner.
For example, some of the 
seemingly different Elvises
above have died in Memphis, others have died
in Nashville, and others are 
(or are rumored to be)
still alive.

\end{itemize}

\section{Knowledge Sharing Community: Wikidata}
\label{subsec:wikidata}

Wikidata ({\small\url{https://www.wikidata.org}}) is a collaborative community to build and maintain a 
large-scale encyclopedic KB,
initiated by
\citet{Vrandecic:CACM2014}
(see also \cite{DBLP:conf/semweb/MalyshevKGGB18}).
It is currently the most comprehensive endeavor on 
publicly accessible knowledge bases.

Wikidata operates under the auspices of the Wikimedia foundation, and has close ties with other Wikimedia projects like Wikipedia, Wiktionary, and Wikivoyage. 
As of August 2020, the KB contained 88 Million entities, and
has 23,000 active contributors.

For this survey article, three aspects of Wikidata are especially relevant: i) the way how knowledge in Wikidata is organized, 
ii) the way how the schema evolves in this collaborative setting, and iii) the role that Wikidata plays as a hub for entity identification and interlinkage between datasets.

\subsection{Principles and History}

Wikidata was launched in 2012, following earlier experiments with collaborative data spaces such as Semantic MediaWiki \cite{DBLP:conf/semweb/KrotzschVV06}.
The motivation for collecting and organizing
structured data in the Wikimedia ecosystem was twofold: (i) centralizing inter-language linking
and (ii) centralizing infobox data. 
Wikidata should be used for automatically generating these
across all Wikipedia language editions, thus 
simplifying maintenance and ensuring consistency.
As of August 2020, the first goal has been reached,
and the second one is getting closer.

The population of the Wikidata KB started out with 
human contributors manually entering statements.
In 2014, Google offered the content of its,
then phased out,
Freebase KB 
\cite{Bollacker:Sigmod2008}
for possible import 
into Wikidata: 3 billion statements about 50 million entities. The Wikidata community refrained from automated import for quality assurance; instead a tool was created that
allowed editors to validate (or discard) individual 
statements before insertion into the KB
\cite{DBLP:conf/www/TanonVSSP16}.
As a result, ca. 17 million statements about 4.5 million entities were added to Wikidata.
This small fraction of the enormous volume of Freebase
underlines the very high quality standards that the
Wikidata community exercises (including the requirement for
reliable references to support statements).
The rapid growth of Wikidata in subsequent years has
been largely based on its human contributors,
but also used bulk imports with humans-in-the-loop for
quality assurance (see below).

Beyond this Wikipedia-centric usage,
it turned out that 
Wikidata
became useful for many other purposes as well:
interlinking and enriching data from public libraries and archives, and managing biomedical and scholarly data (see Section~\ref{subsec:wikidata:examples}).

\subsection{Data Model and KB Life-Cycle}

\begin{figure}
\centering
   \includegraphics[width=0.9\textwidth]{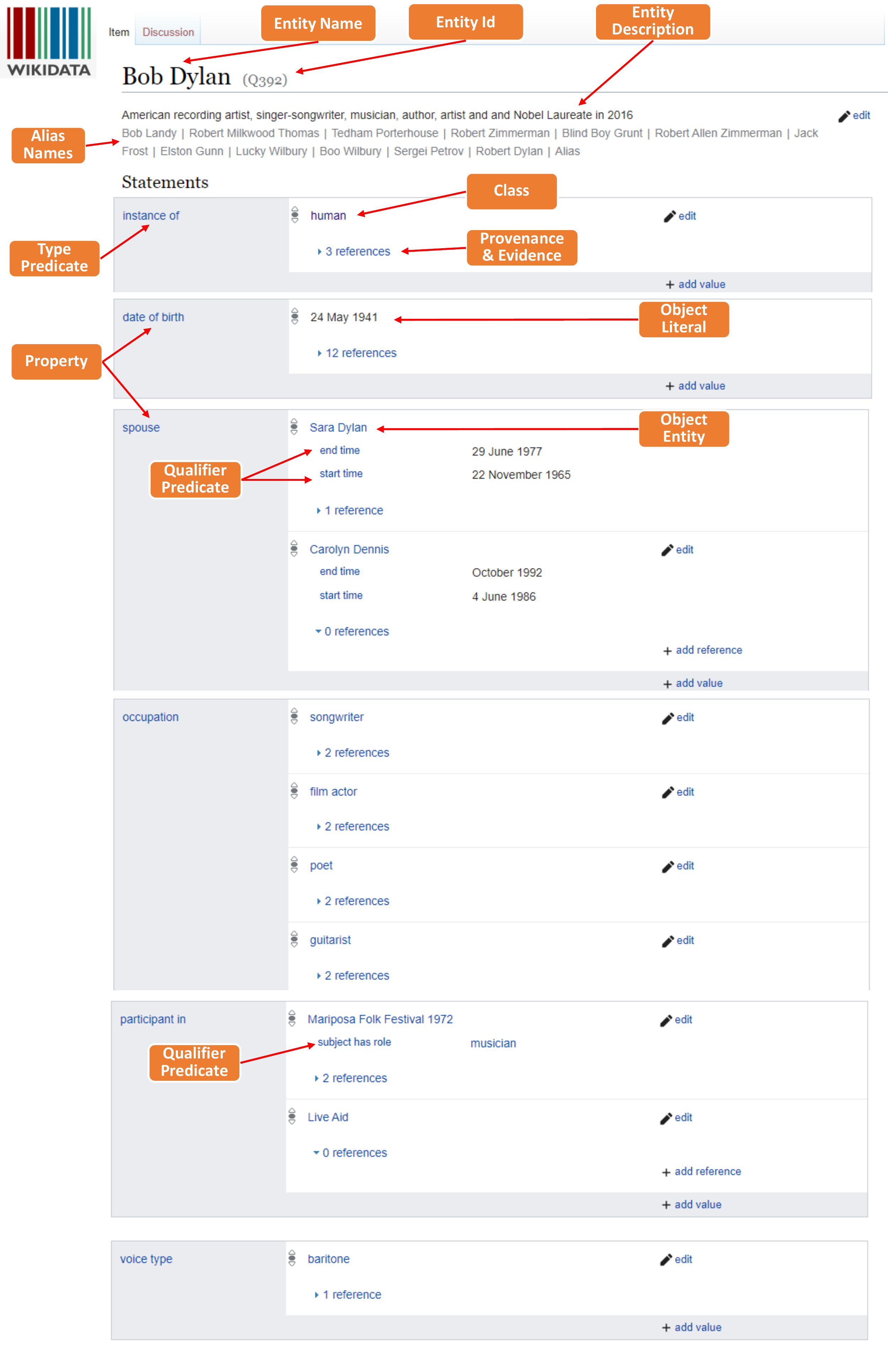}
   \vspace*{-0.3cm}
      \caption{Excerpt of Wikidata page for \textit{Bob Dylan} ({\small\href{https://www.wikidata.org/wiki/Q392}{www.wikidata.org/wiki/Q392}}, August 31, 2020).}
      \label{ch9-fig-wikidata}
\end{figure}

\noindent{\bf Entities (aka. Items) and Statements (aka. Claims):}\\
Wikidata's data model \cite{Vrandecic:CACM2014} revolves around entities, SPO triples, and qualifiers. \textit{Entities}, called {\em items} in Wikidata jargon,
have language-independent Q-code identifiers (e.g.,
Q392 for \textit{Bob Dylan}, or Q214430 for \textit{Like a Rolling Stone}).
Figure \ref{ch9-fig-wikidata} shows an excerpt of the Wikidata page
for Bob Dylan, annotated with some key concepts
of the data model.

{SPO triples},  called 
{\em claims}, are statements about entities. 
Following the RDF model, subjects are entities with Q-codes
and objects can be entities or literals.
The predicates come from a manually specified set of about 7000 properties, identified via language-agnostic P-codes 
(e.g. P569 for date of birth, P26 for spouse, or P577 for publication date). 
This predicate space and the entities alike are
continuously and collaboratively 
expanded by the contributors in the
Wikidata community.

Wikidata circumvents the restrictions of triple-based knowledge representation by reification 
(e.g., Macron's inauguration as French president is itself an entity)
and by \textit{qualifiers} for refining SPO triples
(cf. Section \ref{ch2-subsec:properties}). Qualifiers are a predicates that enrich triples with context, about
sources, dates, reasons, etc.
For example, the \ent{spouse} property comes with
qualifiers for wedding date and divorce date.
Awards, such as Bob Dylan winning the Nobel Prize,
have qualifiers for date, location, field, prize money,
laudation speaker, etc.

\vspace*{0.2cm}
\noindent{\bf Taxonomy and Consistency:}\\
Wikidata items are organized into classes by use of the \textit{instanceOf} property, which in turn are organized using 
\textit{instanceOf} and 
\textit{subClassOf}.
There is no hard distinction, though, between instances and classes; for example, 
\ent{Buddha} ({\small\url{https://www.wikidata.org/wiki/Q7055}}) is both an instance of \ent{religious concept}
and a subclass of \ent{religious ecstasy} and \ent{person}
(as of August 27, 2020), and the latter leads to
super-classes like subject, agent, item, individual.
The complexity, difficult interpretability and
potential inconsistency of the Wikidata class system
was specifically addressed by YAGO 4 (see Section~\ref{sec:yago}).

An important 
principle of the collaborative community
is to allow different perspectives.
Therefore, 
Wikidata refers to statements as {\em claims}.
These do not necessarily capture a single view of the world, with universally agreed-upon facts.
Rather it is possible (and often happens) that
alternative standpoints are inserted as separate statements for the same entity and property.
For example, it is accepted that Jesus has more than one birth date, and the same holds for the death date of the mountaineer
George Mallory (who died somewhere on Mount Everest
with unknown precise date).
In principle, Wikidata would even tolerate entering
alternative death places for Elvis Presley (including perhaps
Mare Elysium on Mars), but the community cross-checks edits and may intervene in such a case.

Wikidata also supports a rich portfolio of \textit{consistency constraints}, including type constraints for properties, 
functional dependencies (called single-value constraints)
and more. However, by the philosophy of capturing potentially conflicting claims, the constraints are not rigorously enforced. Instead, they serve to generate warnings 
when entering new data; so it is up to the contributors
to respect the constraints or not.
Also, users can systematically
look at constraint violations and make corrections as
deemed appropriate.

\vspace*{0.2cm}
\noindent{\bf KB Life-Cycle:}\\
Like Wikipedia, Wikidata follows a collaborative process, under which both data and schema continuously evolve. For inserting entities and statements, an important criterion 
is to include references to reliable sources of evidence.
For Bob Dylan (as of August 2020), for example, these include mostly
digital libraries, archives and LOD sources such
as BBC
({\small\url{http://www.bbc.co.uk/music/sevenages/artists/bob-dylan/}}), 
Musicbrainz ({\small\href{https://musicbrainz.org/artist/72c536dc-7137-4477-a521-567eeb840fa8}{https://musicbrainz.org/artist/72c536dc\dots}}) and also
Wikipedia articles.
For some statements beyond the standard biography and 
creative works, also 
news articles and authoritative websites
such as 
{\small\url{https://www.biography.com/people/bob-dylan-9283052}} are cited as references. An example is Dylan's 
unmarried partnership with Joan Baez, supported by
citing a new article from {\small\url{https://www.dailymail.co.uk/}}.
Schema edits, on the other hand, are typically subject of 
community
discussions. Frequent issues are, for instance, whether a proposed property type will be re-used often enough, whether a a property could also be expressed by class membership, 
and what kinds of constraints should be imposed.

Regardless of its initial intention to support Wikipedia, Wikidata now holds statements about many entities that
are not present in Wikipedia at all.
This is largely data imported from so-called GLAMs (an acronym for galleries, libraries, archives, museums), and government organizations that provide Open Data.
Typically, this involves adding an entity, like a book author or public school, and a set of statements for their identifiers in the original data sources.
The data imports go through a  semi-automated process:
first discussed and approved by the community, 
then carried out by bots. Typical issues for approval are whether the data is of sufficient interest for a KB at all,  how data quality is assured, and how to interlink the new data with the existing content of the KB.
Import procedures are supported by a variety of  tools, such as OpenRefine ({\small\url{https://openrefine.org}}) and Mix'n'match ({\small\url{https://mix-n-match.toolforge.org}}).

\subsection{Applications}
\label{subsec:wikidata:examples}

Beyond Wikipedia, 
Wikidata is utilized in a range of applications.

\vspace*{0.2cm}
\noindent{\bf Galleries, Libraries, Archives and Museums (GLAMs):}\\
These stakeholders are major proponents of open and interlinked data, 
Their entities, like authors, artists and their works,
are increasingly captured in Wikidata, typically with
identifiers that point to the original repositories.
For some entities, these identifiers constitute the majority
of the statements in Wikidata.
GLAMs have high interest in such interlinking,
opening up their contents
for collaborative enrichment and better visibility in search engines. 
An ideal enrichment would follow the ``round-trip''
pattern: Wikidata imports entity identifiers, say about
a lesser known painter from a museum, the Wikidata community augments
this with facts about the painter's biography,
and this added value can be easily combined with the
museum's online contents about the painter's works
\cite{glamreport}.

\vspace*{0.2cm}
\noindent{\bf Scholarly Knowledge:}\\
Knowledge about scientific publications,
authors and their organizations
is another use case that is gaining importance.
Besides existing projects like 
CiteSeerX 
\cite{DBLP:journals/aim/WuWCKCTOJMG15,DBLP:conf/semco/Al-ZaidyG18}, 
SemanticScholar
\cite{ammar2018construction,DBLP:conf/acl/LoWNKW20}, 
AMiner 
\cite{DBLP:conf/kdd/TangZYLZS08,DBLP:journals/dint/WanZZT19},
Open Research Knowledge Graph
\cite{DBLP:conf/jcdl/OelenJSA20,DBLP:conf/ercimdl/BrackHSAE20},
Scholia is a Wikidata-based project
\cite{DBLP:conf/esws/NielsenMW17}.
It aims to provide tools for
the semantic analysis of scholarly topics, author networks, bibliometric measures of impact and more, as open alternative and value-added
extensions to 
prevalent services like
Google Scholar, Microsoft Academic 
or publisher services
\cite{DBLP:conf/www/SinhaSSMEHW15,harzing2016google,DBLP:conf/semweb/Farber19}.

\vspace*{0.2cm}
\noindent{\bf Entity Identification:}\\
Another desirable purpose of KBs is to provide master data for \textit{entity identification} and 
\textit{cross-linkage} between resources.
Wikidata is taking up a central role in the Web of Linked Open Data (LOD, see Section \ref{subsec:webofdata}).
Wikidata identifiers are becoming widespread in 
interlinking datasets and knowledge repositories.
In turn, 
a large fraction of Wikidata statements are 
about external identifiers, such as \textit{TwitterID, VIAF-ID} or \textit{GoogleScholarID}.

\vspace*{0.2cm}
\noindent{\bf Life Science Knowledge:}\\
The life sciences -- biomedicine and health -- 
is a specific domain 
where Wikidata has the potential to become a data hub
\cite{waagmeester2020science}. 
On one hand, the Wikidata KB contains a large amount
of statements about diseases, drugs, proteins etc.
On the other hand, its rich coverage of identifiers
as links to other repositories (see above)
supports combining
from different biomedical sources in a user-friendly
manner (e.g., for data scientists in the health area).

\subsection{Challenges}

Wikidata is today's most prominent endeavor on
collaborative knowledge engineering.
Still, it faces a number of challenges in its future advances.

\vspace*{0.2cm}
\noindent{\bf Quality Assurance and Countering Vandalism:}\\
These are %
never-ending concerns
for collaborative projects.
A good strategy requires balancing
openness of contributions and moderated approvals.
Allowing edits by any contributor
has the risk of introducing errors but also the 
advantage that errors can be quickly caught and corrected by others.
Staggered review and approval, on the other hand, 
can prevent blatant errors, at the expense of
slowing down the KB growth, though.
In addition to vandalism, concerns are also raised over more subtle content distortions with commercial or political interests.

\vspace*{0.2cm}
\noindent{\bf Evolving Scope and Focus:}\\
The scope and focus of Wikidata are repeatedly coming into discussion, especially when new data imports are discussed. In 2019, for instance, debates revolved around the importing of scholarly data (mostly identifiers for authors and publications), which currently makes up 40\% of Wikidata's entities. 
Such imbalances could possibly bias functionality for search and ranking, and puts strain on Wikidata's infrastructure, 
potentially
at the cost of other stakeholders. 
Therefore, decisions on in-scope and out-of-scope topics are 
a recurring concern.

\vspace*{0.2cm}
\noindent{\bf Schema Stability:}\\
Wikidata's 
collaborative processes and continuous evolution 
are better able to keep up with an evolving reality than any expert-level modeling of classes and properties. 
However, this implies that applications relying on Wikidata require continuous monitoring, as changes in property definitions and taxonomic structures can break queries.
Finding the right balance between the community's
grassroots contributions and controlling the
quality and stability of the KB schema will remain
a challenge.

\vspace*{0.2cm}
\noindent{\textbf{Data Duplication:}}\\
Redundancy, and the resulting risk of inconsistency,
are further issues that arise
from the principle of a collaborative community.
At the entity level, this is well under control by
the self-organization among editors which captures
duplicates quickly and resolves them.
However, redundancy is an issue for types and properties.
By independent edits, properties can be
duplicated in forward and backward direction, such as \textit{parent / child}, \textit{has part / is part of}, and \textit{award received / recipient}.
Similarly, existential information is stored in separate properties, such as \textit{child / number of children} and \textit{episode / number of episodes}. These 
make querying more complex, and most critically, 
may easily cause inconsistencies.

\section{Industrial Knowledge Graphs}
\label{ch9-sec:industrialKG}

{\em Knowledge graphs}, or {\em KGs} for short,
is the industry jargon for
knowledge bases -- a widely used but oversimplifying term as KBs comprise much
more than just binary relations. 
KGs started taking an important role in industry in 2012, the year when Google launched knowledge-based search under the slogan
``things, not strings'' \cite{Singhal2012}.
The {Google KG} started from Freebase 
\cite{Bollacker:Sigmod2008},
a KB built by Metaweb, acquired by Google in 2010. 
In the same year, Amazon acquired Evi (formerly True Knowledge) \cite{DBLP:journals/aim/Tunstall-Pedoe10}, whose 
KG laid the foundation for Alexa question answering. 
Knowledge graphs are broadly used in search engines like
Google, Bing and Baidu,
in question answering such as 
Apple Siri, Amazon Alexa and Google Assistant,
and in e-Commerce at Alibaba Taobao, Amazon, eBay, Walmart and others. 
Another early player in industrial KGs has been
Wolfram Alpha \cite{hoy2010wolfphram}, which provided services
to Apple, Amazon, Microsoft, Samsung and others.
Building an authoritative knowledge graph with comprehensive and high-quality data to support the broad range of applications has been a hot topic for industry practice in the past decade \cite{Noy:CACM2019}.

The biggest success for knowledge bases in industry
is mainly for 
{\em popular domains} such as {music, movies, books} and { sports}. Efforts to collect 
{\em long-tail knowledge} started much later (around 2015), but have already been successful in facilitating users to find answers for their hobbies such as yoga or cocktails. 
Gathering and organizing 
{\em retail product knowledge} started 
in 2017,
facing many obstacles but already bearing fruits. We next describe the efforts and progress for each of these three aspects.

\subsection{Curating Knowledge for Popular Domains}

The first pot of gold for knowledge collection in industry is Wikipedia. 
As discussed in 
Chapter \ref{ch2:knowledge-integration}, 
Wikipedia is a
premium source and great starting point with its huge number of entities, informative descriptions
and rich semi-structured contents.
Wikipedia data has played an important role in Google KG, Bing Satori, Amazon Evi, and presumably more.

The knowledge graphs are then extended on a set of popular domains: large domains include {\em Music, Movies, Books, Sports, Geo, etc.}, and medium domains include {\em Organizations, People, Natural World, Cars, etc.} 
Two common features for such domains make them pioneer domains for knowledge collection. 
\squishlist
\item First, there are already rich data sources in (semi-)structured form and of %
high
quality.
For example, IMDB
({\small\href{https://www.imdb.com/interfaces/}{https://www.imdb.com/interfaces/}})
is a well-known authoritative data source for {movies},
and
MusicBrainz
({\small\href{https://musicbrainz.org/doc/MusicBrainz_Database}{https://musicbrainz.org/doc/MusicBrainz\_Database}})
is an authoritative source for {music}
with open data for free download.
Big companies also license data from major data providers, 
and often have their own data sets (for example, Google has rich data on books, locations, etc.). 
\item Second, the complexity of the domain schema is manageable. Continuing with the {movie} domain, the Freebase knowledge graph contained 52 entity types and 155 properties for this domain. The schema (aka. ontology)
to describe these types and properties can be manually 
specified by 
a knowledge engineer
within weeks, especially by leveraging existing data sources.
\squishend

A big challenge in this process is to integrate 
entities and properties from Wikipedia 
with domain-specific data sources. 
This requires alignment between schemas/ontologies
(see Section \ref{ch3-sec:alignment}) and 
entity matching (see Section \ref{sec:entity-matching}).
Schema alignment is carefully curated manually; this is affordable because the size of the schema/ontology for each domain is manageable. On the other hand, manually linking entities that refer to the same real-world person or movie is unrealistic, at the scale of many millions of entities. 
This calls for automatic entity linking (EL) as discussed in
Chapter \ref{ch3-sec-EntityDisambiguation} (see especially Section \ref{sec:entity-matching}). 
To meet the high bar for KB accuracy, the entity linkage needs to have very high precision, obtained by sacrificing recall to an acceptable extent, and by manually checking cases where the EL method has low confidence.

\subsection{Collecting Long-Tail Knowledge}

Long-tail domains are those where the entities are not globally popular; thus oftentimes the number of entities is small: thousands or even hundreds only. Examples of tail domains include gym exercises, yoga poses, cheese varieties, and cocktails. Although a tail domain may not be popular by itself, given the large number of tail domains, they can be collectively impactful, for example, addressing people's hobbies -- an important part of life.

Entity distribution among head, torso, and tail domains observes the power law; that is, a few head domains cover a huge number of entities, whereas a huge number of long-tail domains each covers a fairly small number of entities. The huge number of long-tail domains and the diversity of their attributes make it impossible to collect knowledge 
from a few domain-specific premium sources as we do for head domains. 
Likewise, there is no source that covers 
a significant fraction of different long-tail domains. 
An empirical study with manual curation
of four long-tail domains
(cheese varities, tomatoes, gym exercises, yoga poses)
showed that many entities of interest can be found on
websites, but these entities neither exist in Wikipedia nor in the Freebase KB
\cite{DBLP:journals/pvldb/LiDLL17}.

Millions of tail domains and no single data source to cover all domains call for a different solution for collecting long-tail knowledge. We next describe two different approaches: tooling for curation (Section~\ref{subsec:tooling}) and automatic extraction
from a huge variety of websites
(Sections~\ref{subsec:KV} and \ref{subsec:ceres}).

\subsubsection{Tools for Long-tail Knowledge Curation}
\label{subsec:tooling}

One effective method to extract long-tail knowledge is to provide tools that support knowledge engineers on the manual curation of long-tail domain knowledge. 
The Google Knowledge Graph and 
projects at Amazon that feed into Alexa took this approach.

The process comprises five steps: 
\squishlist
\item[1.] Identify a few data sources for each domain.
\item[2.] Define the schema for the domain according to data in these sources.
\item[3.] Extract instance-level data from the sources 
through annotation tools and hand-crafted patterns.
\item[4.] Link the extracted knowledge 
to existing entities and types in the knowledge base.
\item[5.] Insert all new entities and acquired statements into the knowledge base.
\squishend
Each step involves manual work and human curation. It can easily take a knowledge engineer a few weeks
to curate one long-tail domain. 

There are studies on how to 
accelerate each step in this process and reduce the manual work. The work of \cite{Wang2019Midas} proposed the
{\em MIDAS} method for discovering web sources 
of particular value for a given domain.
The core insight from MIDAS is that 
even if automatic extraction methods may not yield sufficiently high quality overall, they can provide clues about which websites contain a large amount of relevant contents, allow for easy annotation, and are worthwhile for extraction
(see the discussion on source discovery in
Section \ref{ch6:subsubsec:source-discovery}).

A challenge in harvesting these sources, however, is that a 
website often includes facts about multiple groups of entities, having different schemas and requiring different extraction templates (see Section \ref{ch6-sec:properties-from-semistructured}). 
MIDAS discovers groups that contain a sufficient number of  entities and statements that are absent from the KB one wishes to augment, such that the extraction benefits outweigh the efforts. 
MIDAS also helps to identify interesting domains, such as medicinal chemicals or US history events.

Another study~\cite{DBLP:journals/pvldb/LiDLL17} aimed to identify extraction errors 
from particular sources
by verifying the correctness of the collected knowledge. 
This is based on an end-to-end knowledge verification framework, called {\em FACTY}. The method leverages both search-based and extraction-based techniques to find supporting evidence for each triple, and subsequently predicts the correctness of each triple based on the evidence through {\em knowledge fusion} \cite{Dong:KDD2014}
(see Sections \ref{ch8-subsec:qualityprediction} and
\ref{ch8-subsec:datacleaningandknowledgefusion}).
Various types of evidence are considered, including existing knowledge bases, extraction results from websites, search-engine query logs, and results of searching subject and object of SPO triples. 
In the study of \cite{DBLP:journals/pvldb/LiDLL17}, 
this technique achieved high recall:
positive evidence
was found for 60\% of the correct triples on 96\% of the entities in the examined four tail domains. 

Unfortunately, this method also finds positive evidence for many incorrect
statements; so the outlined approach suffered from low precision (fraction of verified triples that are truly correct).
To distinguish correct and wrong triples, knowledge fusion estimates the correctness of each triple, taking into consideration the quality of the evidence sources and the reliability of the techniques for obtaining evidence. With 
this additional quality assurance,
the percentage of correct statements that could be verified dropped from 60\% to 50\%, but the precision increased to 84\% (i.e., among all triples that are verified to be correct, 84\% are indeed correct) \cite{DBLP:journals/pvldb/LiDLL17}.
This is an enormous support and productivity boost
for human curators.

\subsubsection{Knowledge Vault}
\label{subsec:KV}
Tooling can speed up the collection of long-tail knowledge; however, the solution still does not scale to 
hundred thousands of domains.
For high coverage of many domains, it is inevitable
to apply information extraction 
to
millions of diverse websites.
Among such 
endeavors, 
Google's {\em Knowledge Vault (KV)} is a prominent one
(\citet{Dong:KDD2014} \cite{DBLP:journals/pvldb/DongGHHMSZ14}).
Knowledge Vault 
differs
from other 
projects
in two 
important aspects:
\squishlist
\item First, KV extracts knowledge from four types of web sources: 
\squishlist
\item[o] text documents such as news articles, 
\item[o] semi-structured DOM trees, where attribute-value pairs are presented with rich visual and layout features,
\item[o] HTML tables with tabular data 
embedded in web pages, and 
\item[o] human-annotated pages with 
microformats according to ontologies like 
{\em schema.org} \cite{Guha:CACM2016}.
\squishend
\item Second, KV addresses the low quality of the extraction by applying knowledge fusion techniques
(see Sections \ref{ch8-subsec:qualityprediction} and
\ref{ch8-subsec:datacleaningandknowledgefusion}).
\squishend

KV applied 16 types of extractors 
on these four types of web contents. For text and DOM trees, extraction techniques were used as discussed in Sections %
\ref{sec:properties-patterns}
and \ref{ch6-sec:properties-from-semistructured}. 
In particular, KV first identified Freebase entities in a sentence or semi-structured page, and then predicted the relation between a pair of entities. The training data 
was obtained from Freebase, using distant supervision.
To this end, KV utilized the {\em Local Completeness Assumption (LCA)} (see Section \ref{ch8-sec:knowledgebase-completeness}):
when a subject-property combination has at least one object in the KB, then the set of triples with this
SP combination is complete. Conversely, any choice of object O for the same SP pair such that SPO is not in the KB, must be false.
This has been key to generating also negative training
samples, and turned out to be decisive for extraction quality
\cite{Dong:KDD2014}.

For web tables, schema mapping and entity linking 
were employed to map rows and columns to entities and properties, respectively (cf. Section \ref{sec:el-semistructured}). 
For pages with microformat annotations, KV extracted
statements that follows the {\em schema.org} ontology and then transformed them to the Freebase schema.
These mapping were manually specified, as they involved only a few tens of property types.

\noindent{\bf Experimental Findings:}\\
The above methods extracted 2.8 billion SPO triples from over 2 billion web pages. 
Among the four source types, DOM trees for HTML-encoded web pages contributed 75\% of extracted statements. This is not surprising because each semi-structured website normally has one or more back-end databases that yield many instances.
In comparison, web tables contributed the smallest number of extractions, less than 5\%. 
This is partly because web tables are hand-crafted and
thus contain only few rows, in contrast to tables or lists that
are generated from back-end databases.
Another reason is that the extraction techniques faced limitations on aligning column headers with KB properties
(as web tables often have generic and highly ambiguous headers such as 
``Name'' or ``Count'').

Despite the large volume of extractions, extraction quality initially was very low, with precision around 10\%.
KV solved the problem by applying 
{\em knowledge fusion} techniques in an additional curation phase (see Chapter \ref{chapter:KB-curation}, especially Section \ref{ch8-subsec:datacleaningandknowledgefusion}). 
First, it trained a logistic regression model to accept or reject a triple based on from how many web pages it was extracted and by how many extraction models it was returned. Second, KV employed the {\em Path Ranking Algorithm} \cite{lao2011random}
for deduction (see Section \ref{ch8-subsec:rulemining}),
to assess the plausibility of a triple according to paths between the S and O entities. 
Third, KV used a neural network for link prediction, 
based on KG embeddings (see Section \ref{ch8-sec:kgembeddings}).
Note that all three techniques were used conservatively here,
to prune false positives (not for inferring additional triples).

All three approaches achieved over 90\% area-under-the-ROC-curve;
together they reached nearly 95\%.
In the end, from a pool of 1.6 billion candidate triples, 324 million (20\%) were assessed to have confidence 0.7 or higher, and 271 million (17\%) had confidence 0.9 or higher -- high-quality candidates to be considered for KB augmentation.

\vspace*{0.2cm}
\noindent{\bf Lessons Learned:}\\
Despite the promising results, KV did not significantly contribute to the Google Knowledge Graph, mainly because the number of new triples meeting the very high bar for precision was not sufficiently high (the Google KG expected 
99\% precision).
There are several reasons. First, the extractions are restricted to existing entities and properties. Since 
the Google KG already has a high coverage of top entities and
their properties, missing relations between popular entities were not many. Second, the first-stage extraction quality was too low. The knowledge-fusion stage could prune false positives,  but could not add any correct extractions. Third, evaluation in the wild may not be as good as estimated from
the local completeness assumption (LCA, see above).
It is common to observe different quality when the training data and test data have different distributions, and that is more likely to happen when the training data is not randomly sampled and makes certain assumptions.

Nevertheless, the KV data found an interesting application
for estimating the quality of web sources by a
{\em Knowledge-Based Trust} model \cite{DBLP:journals/pvldb/DongGMDHLSZ15}
(see Section \ref{ch8-subsec:datacleaningandknowledgefusion}).
The underlying intuition is that the trustworthiness of a web source  can be measured by the accuracy of the triples provided by the source, which can be approximated 
from the confidence-weighted number of triples extracted from the source. A probabilistic graphical model was devised to
compute the probability of a triple being correct, the probability of an extraction being correct, the precision and recall of an extractor, and the trustworthiness of a data source. 

The method computed trustworthiness scores for 5.6 million websites and led to interesting observations.
First, most of the sources have a score above 0.5,
and some reach 0.8. 
Second, the trustworthiness scores are orthogonal to PageRank
scores, which is based on the popularity of a website rather than its content quality. 
The knowledge-based trust method is the basis for fact verification in knowledge curation tools, described in 
Section \ref{subsec:tooling}.

\subsubsection{Amazon Ceres}
\label{subsec:ceres}

Amazon Ceres 
(\citet{DBLP:journals/pvldb/LockardDSE18} \cite{DBLP:conf/naacl/LockardSD19} \cite{Lockard2020ZeroShotCeres})
is a large-scale effort to collect long-tail knowledge by web extraction. 
Ceres made two design choices that deviate from the
rationale of Knowledge Vault. 
\squishlist
\item First, instead of relying on knowledge fusion to clean up extracted statements, it focuses on improving extraction quality from the beginning. Ceres improved extraction precision from semi-structured sources from 43\% in KV to over 90\%, a quality reasonable for industrial-strength services.
\item Second, Ceres also extracts knowledge about new entities, and even new properties. This way, it extracts head knowledge for tail entities, instead of tail knowledge for head entities, and thus has more room 
to satisfy users' long-tail knowledge needs. 
\squishend
\noindent These differences make Ceres suitable for industry production, used at Amazon to collect long-tail knowledge for Alexa and product knowledge %
for retail.

Ceres focuses on only one type of web content: 
semi-structured web pages. 
Recall that DOM-tree-based contents contributed 75\% of extractions and 94\% of high-confidence extractions in Knowledge Vault; so is a natural top choice among the  different types of web contents. The dense high-quality information in semi-structured sources makes it easier for extraction and entity linking, supporting the two design choices above. 

On the other hand, semi-structured contents pose the challenge that every website follows a different template (or a set of templates). So the model trained on one website normally does not carry over to another website. 
Ceres devised a suite of extraction techniques to address this challenge: Ceres~\cite{DBLP:journals/pvldb/LockardDSE18} for properties specified in the KB schema (referred to as ``Closed IE''), OpenCeres~\cite{DBLP:conf/naacl/LockardSD19} for Open IE
(see Chapter 7), and ZeroShotCeres~\cite{Lockard2020ZeroShotCeres} for  extraction from unseen websites and even unseen subject domains. We next outline the three systems.

\noindent{\bf Ceres for Schema-based ``Closed IE'':}\\
Ceres extracts knowledge according to an existing schema/ontology. Training a different model for a different website would require a large amount of training data, posing a bottleneck. Ceres solves this problem by generating the training data automatically using existing knowledge as seeds. This is the distant supervision technique described in Section \ref{ch6:subsubsec:extraction-from-dom-trees}
and widely used in large-scale knowledge harvesting.
However, compared to Google Knowledge Vault, for example,
Ceres is able to double the extraction precision with two significant enhancements:
\squishlist
\item First, Ceres focuses on entity detail pages
(see Section \ref{ch6:subsubsec:extraction-from-dom-trees}), each of which describes an individual entity.
It employs a two-step annotation algorithm that first identifies the entity that is the primary subject of a page, and then annotates entities on the page that have relations with that entity in the seed knowledge. 
This prevents a large number of spurious outputs.
\item Second, Ceres leverages the common structure among property-object pairs within a web page, and across pages from the same website, to further improve the labeling quality. 
\squishend
\noindent With improved labeling from distant supervision, and thus better training data, Ceres is able to extract knowledge from DOM trees in websites about long-tail domains 
with  precision
above 90\% \cite{DBLP:journals/pvldb/LockardDSE18}.

\vspace*{0.2cm}
\noindent{\bf Open-IE-based Ceres:}\\
{\em OpenCeres} \cite{DBLP:conf/naacl/LockardSD19} and {\em ZeroShotCeres} 
\cite{Lockard2020ZeroShotCeres} address the Open IE problem, with the goal of extracting properties that do not exist in the schema/ontology (see Chapter 7). 
These systems are among the first OpenIE 
approaches specifically geared for semi-structured data, and did so by exploring both structural and layout signals in web pages. 
Between them, OpenCeres trains different models for different websites, and requires the website to contain  seeds from the existing KB. ZeroShotCeres, as the name indicates, can extract statements from a new website in a completely new domain, by training a universal model that applies to all websites. 

OpenCeres explores similarities between different property-object pairs within a page.
The idea is that the format for an unknown property 
(e.g., \ent{coveredBy} for songs) is often similar to that for a known one (e.g., \ent{writtenBy}). 
It applies semi-supervised label propagation 
to generate training data for unseen properties (see Section \ref{ch7-openie-semistructured}). 

ZeroShotCeres \cite{Lockard2020ZeroShotCeres} extends 
this approach by considering similarities between different websites. The underlying intuition is that despite different websites having different templates, there are underlying commonalities in font, style, and layout. 
ZeroShotCeres trains a {\em Graph Neural Network} to encode 
patterns common across different websites. Thus, the
learned model can
generalize to previously unseen templates and even 
new vertical domains.

\subsection{Collection of Retail Product Knowledge}

Recent years have been witnessing KB applications in a new domain: the retail product domain. Similar to generic domains, product knowledge can improve product search, recommendation, and voice shopping. However, this new domain presents  
specific challenges.

\vspace*{0.2cm}
\noindent{\bf Challenges:}\\
First, except for a few categories such as {electronics}, structured data is sparse and noisy across nearly all data sources. This is because the majority of product data resides in catalogs from e-Business websites such as Alibaba, Amazon, Ebay, Walmart, etc., and these big players often rely on data contributed by retailers. 
In contrast to stakeholders for digital products like movies and books, in the retail business, contributing manufacturers and merchants mainly list product features in titles and textual descriptions instead of providing structured data
\cite{DBLP:conf/kdd/ZhengMD018,xu2019scaling}. 
As a result, structured knowledge needs to be mined from textual contents like titles and descriptions. Thousands of product attributes, billions of existing products, and millions of new products emerging on a daily basis require fully automatic and efficient knowledge discovery and update mechanisms.

Second, the domain is 
complex in various ways. The number of product types is towards millions, and there are sophisticated relations between the types like subclasses (e.g., swimsuit vs. athletic swimwear), synonyms (e.g., swimsuit vs. bathing suit), and overlapping types (e.g., fashion swimwear vs. two-piece swimwear). Product attributes vastly differ between types (e.g., compare TVs and dog food), and also evolve over time (e.g., older TVs did not have WiFi connectivity). All of these make it hard to design a comprehensive schema or ontology and keep it up-to-date, thus calling for automatic solutions.
Open catalogs like \url{https://icecat.biz/} do not provide extensive schemas, and face the same issue that 
fine-grained types and properties are merely mentioned
in textual documentation.

Third, the huge variety of different product types makes it even harder to train  models for knowledge acquisition and curation. Product attributes, value vocabularies, text patterns in product titles and descriptions widely differ for different types. Even highly related product types can have rather different attributes.
For example, {\em Coffee} and {\em Tea}, which share the same parent {\em Drink}, describe packaging sizes by different vocabularies and patterns, such as:\\
\hspace*{0.5cm} ``Gound Coffee, 20 Ounce Bag, Rainforest Alliance Certified'' vs. \\
\hspace*{0.5cm} ``Classic Tea Variety Box, 48 Count (Pack of 1)''.\\
\noindent Training a single one-size-fits-all model 
for this huge variety is hopeless.
On the other hand, collecting training data for each of the thousands to millions of product types is prohibitively expensive.

Research on product knowledge graphs is pursued by
major stakeholders including Alibaba Taobao, Amazon, eBay and Walmart \cite{Noy:CACM2019,DBLP:conf/sigmod/LuoLYBCWLYZ20,DBLP:conf/kdd/DongHKLLMXZZSDM20,DBLP:conf/wsdm/XuRKKA20,DBLP:conf/icde/PfadlerZWWHL20}.
Next, we exemplify current approaches to these challenges by discussing methods for building and curating the Amazon Product Graph.

\subsubsection{Amazon Product Graph}

\begin{figure}
\centering
   \includegraphics[width=0.9\textwidth]{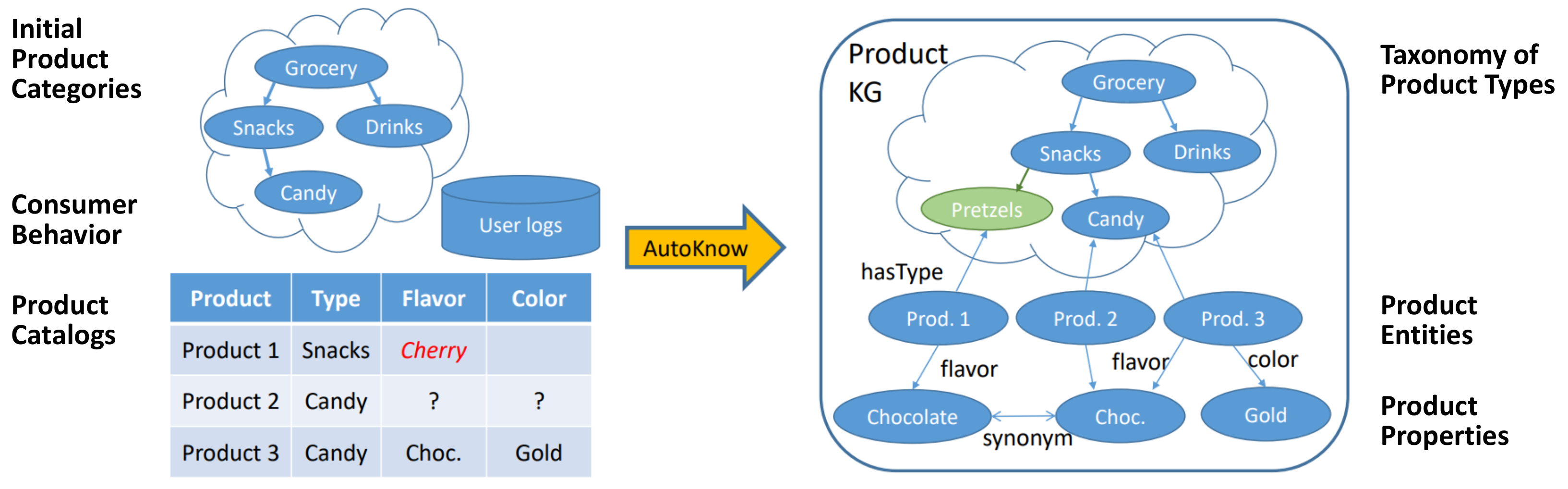}
      \caption{Input and Output of the AutoKnow System (Source: \cite{DBLP:conf/kdd/DongHKLLMXZZSDM20}).}
      \label{ch9-fig:autoknow-overview}
\end{figure}

Amazon is building a Product Graph for retail products. 
Figure \ref{ch9-fig:autoknow-overview} 
illustrates the inputs and outputs for
this endeavor.
The inputs include pre-existing categories,
catalogs and signals from consumers' shopping behaviors,
like queries, clicks, purchases, likes and ratings.
The output is a KB that comprises an enriched and
clean taxonomy and canonicalized entities with
informative properties.
To distinguish the product knowledge graph from a traditional KB (or KG), the output is referred to as a {``broad graph''}. 
This is a bipartite graph, where one side contains nodes representing products and the other side contains nodes representing product properties, such as brands, flavors, and ingredients (lower part of the right side of Figure~\ref{ch9-fig:autoknow-overview}). 

Amazon's knowledge collection system for its product KG
is called 
{\em AutoKnow} 
(\citet{DBLP:conf/kdd/DongHKLLMXZZSDM20}).
The system starts by building a product type {\em taxonomy} and deciding on applicable product properties for each type. 
Subsequently, it infers structured tripes, cleans up noisy values, and identifies synonyms between attribute values
(e.g., ``chocolate'' and ``choc.'' for the \ent{flavor} property).

AutoKnow is designed to operate fully automatic at scale, suited for
a huge number of entities and types in the retail domain. 
For training its machine-learning models, 
it leverages existing catalogs and customer behavior logs.
These are used to generate training samples, 
eliminating the need for manual labeling.
The learned models can generalize to new product types that have never been seen during training (see \cite{DBLP:conf/kdd/DongHKLLMXZZSDM20} for 
technical details).

A few techniques play a key role for scalability and
generality:
\squishlist
\item AutoKnow 
devised a {\em Graph Neural Network} for machine learning,
which lends itself to the structure of the product KG.
\item It takes initial product categories as input signal to train models for building the KG taxonomy. 
\item  It heavily relies on distant supervision and
semi-supervised learning, to minimize the burden of
manual labeling.
\item It jointly infers facts about properties and synonymous expressions for types, properties and values,
to cater for the widely varying vocabularies by
consumers.
\squishend

\subsection{Computational Knowledge: Wolfram Alpha}
\label{subsec:wolfram-alpha}

Wolfram Alpha is a {\em computational knowledge engine} \cite{hoy2010wolfphram,weisstein2014computable} that was
developed in the late 2000s, to support fact-centric question answering (QA).
The Wolfram company is best known for its computer-algebra system
Mathematica. The KB enhances computations with Mathematica by 
encyclopedic facts such as location coordinates and distances.
Conversely, the QA interface, accessible at {\small\url{http://wolframalpha.com}},
makes use of computational inference by functions of the Mathematica library.

The KB that underlies Wolfram Alpha is built by importing and integrating
facts from a variety of structured and well curated sources, such as
CIA World Factbook, US Geological Survey, Dow Jones, feeds on weather data, official statistics, and more.
The specific strength of Wolfram Alpha is the ability to compute derived knowledge.
For example, one can ask about the age of Bob Dylan: the returned answer, as of
January 13, 2021, 
is ``79 years, 7 months and 19 days''. Obviously, this is not stored in the KB,
but derived by a functional program from the KB statement about Dylan's birthdate.
More advanced cases of this kind of computational knowledge are statistical aggregations, comparisons, rankings, trends, etc., mostly regarding geography, climate, economy and finance.
The QA service and other APIs of Wolfram Alpha have been licensed to other
companies, including usage for Apple Siri and Amazon Alexa.

\subsection{Search and Question Answering with Machine Knowledge}

Industrial knowledge bases
at Internet companies
have been applied in four major categories:
{\em search, recommendation, information display}, and {\em question answering / conversations}. 
Search and recommendation are the major ways for users to discover new information, including not only 
web pages, news, images and videos, but also specific entities such as products, businesses, music, movies, house-for-rent, and so on. Information display, often on a detail page or in a knowledge panel, is a way to provide easy-to-understand information about an entity such as a person, song or movie.
In e-commerce, thorough information can help decision making such as whether to purchase a product or rent a vacation house. 
Finally, {question answering (QA) and conversations} play a mixed role, both as information entry points and as detailed information providers. Conversations can be considered as multi-round question answering, allowing both flexibility and broadness of questions and keeping context in consideration to make the answers in each round more appropriate and engaging. 
 
In this section, we focus on search and QA, which are mature use cases in industry to leverage the power of a KB. %
Exploiting knowledge for recommendation and conversation 
are important issues, too, but still open research topics.

\subsubsection{Knowledge-Facilitated Web Search}

\begin{figure}\centering
\includegraphics[width=0.8\textwidth]{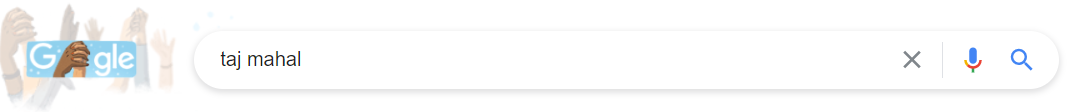}
\includegraphics[width=0.5\textwidth]{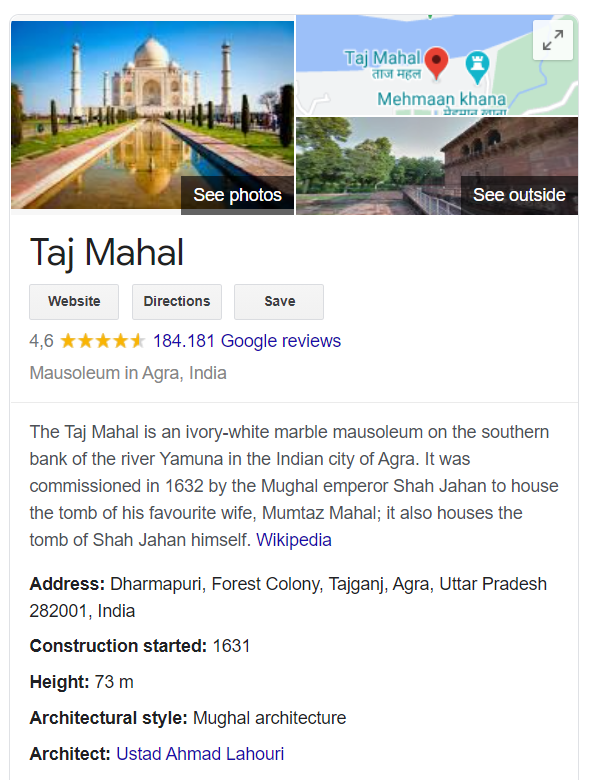}
\caption{Knowledge panel shown for search query ``Taj Mahal'' on Google, as of March 8, 2021.}
\label{fig:taj_panel}
\end{figure}

\begin{figure}\centering
\includegraphics[width=0.9\textwidth]{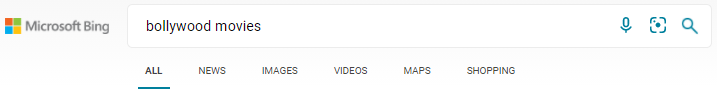}
\includegraphics[width=0.9\textwidth]{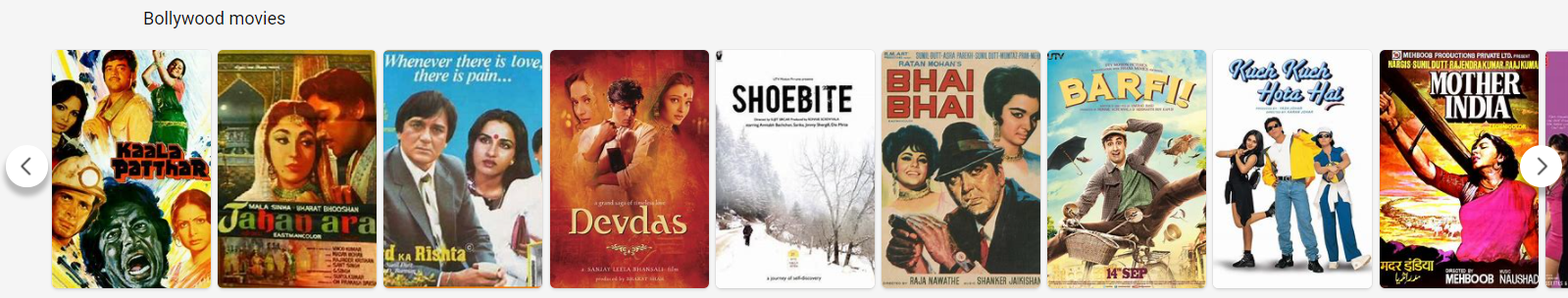}
\caption{Knowledge carousel shown for search query ``Bollywood movies'' 
on Bing, as of March 8, 2021.}
\label{fig:bollywood_carousel}
\end{figure}

The use case that people are most familiar with for KB usage is web search, provided by all major search engines like Google, Bing, Baidu, Yandex and others. 
A huge fraction of queries is about individual entities like celebrities, companies, products, places, events etc.
(see, e.g., {\small\url{https://trends.google.com/trends/yis/2020/US/}} for search-engine statistics).
This dominance of entity queries makes the KB particularly valuable.
For example, when a user searches for ``Taj Mahal'', a {\em knowledge panel} for the Indian palace 
appears on the right side of search results, showing a photo, a brief description, its address, 
architect etc. (see Figure~\ref{fig:taj_panel}).
As another example, when a user searches for
``Bollywood movies", a 
{\em knowledge carousel} will show a list of Bollywood movies ordered by popularity
(see Figure~\ref{fig:bollywood_carousel}).
A modified query "Bollywood movies by year" will trigger the same knowledge carousel but order movies by release time starting from the latest.

Google and Bing adopted knowledge-based search around 2012
\cite{Singhal2012,Qian2013}. 
The value from their KBs is to provide crisp, easy-to-read information such that users do not need to click on the blue links and browse through text documents.
This is especially helpful on small screens such as mobile phones, often with voice input, and also in situations where Internet connectivity is poor or expensive. 
According to the market research company SparkToro~\cite{Fishkin2019}, as of June 2019,
half of Google Search resulted in ``zero-clicks'': 
the users were directly satisfied with seeing a knowledge panel, carousel or other 
KB excerpt,
as opposed to the traditional result page of ten blue links with preview snippets.

Taking knowledge panels as an example, there are several decisions to make by the search engine:
\begin{itemize}
    \item D1: Shall the query trigger the display of a knowledge panel? For example, a knowledge panel is appropriate for the query ``Taj Mahal'', but not for queries in the style of ``democrats vs. republicans'' or ``who will win the upcoming election?''.
    \item D2: If so, which entity is the query asking about? For the ``Taj Mahal'' query, shall we return the famous Indian palace or the American musician with the stage name ``Taj Mahal''? Or should we show both and let the user choose?
    \item D3: Which attributes and relations shall be shown for this entity, and how should they be ordered?
\end{itemize}

We briefly outline factors that search engines 
consider in making these decisions.
For D1, 
search engines invoke a {\em query understanding}
component to identify candidate entities in the KB that best match the query (using NER-style sequence-tagging techniques
as discussed in Chapter \ref{ch3:entities}), and also identify key properties mentioned in the query.
For example, for the query ``taj mahal year'', 
the system could identify \ent{Taj Mahal} (Indian palace)
as an entity and \ent{construction year} as an attribute of interest.
Linking to attributes or relations 
is related to
word sense disambiguation, but is simplified
by the fact that target candidates are canonicalized property types in the KB.
The same holds for linking to classes, such as 
\ent{movies} for a query ``films about AI'',
or even to a refined class \ent{AI movies}.

For decision D2, when there are multiple candidates entities, this is the case for 
entity linking (EL),
as discussed in Chapter \ref{ch3-sec-EntityDisambiguation},
applied to short query strings and based on machine learning techniques. 
In addition, search engines can leverage their huge
collections of query-and-click logs.
Among several candidate entities that match a query, the entity that appears most frequently in top-ranked answers would often be chosen as the one for displaying a knowledge panel. For our example query, the Indian palace would be triggered most of the time, 
as it is way more popular in search than the American musician. 
However, other factors would be taken into account as well, most notably, the geo-location of the user (e.g., American vs. European) and possibly even personalized profiles
(e.g., for product search and entertainment -- once a music context is established, the musician is more likely the user's intent).

Finally, for decision D3, the attributes and 
relations for display are typically chosen according to user query-and-click patterns. For example, if most people who query about palaces or buildings 
search for construction information, 
the relevant attributes, such as 
\ent{architect}, will be 
shown in the
\ent{Taj Mahal} knowledge panel. On the other hand, when users specify a particular property in their queries,
like for ``taj mahal year'' or ``taj mahal fee'', the knowledge panel will first show \ent{construction year} or
\ent{entrance fee}, respectively.

\begin{figure}\centering
\includegraphics[width=0.6\textwidth]{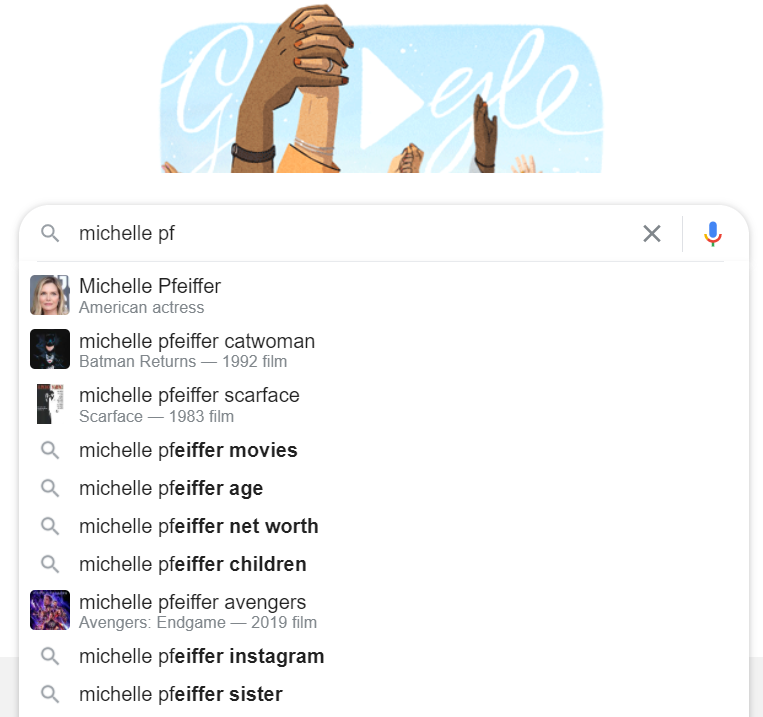}
\caption{Auto-completion suggestions for search query ``Michelle Pf'' 
on Google, as of March 8, 2021.}
\label{fig:michellepf-autocompletion}
\end{figure}

In addition to the case of zero-click search results, the KB also helps to improve standard search results in several ways (standard search is nowadays often referred to as ``organic search'', as opposed to
paid/sponsored search results). 
First, with the entities and properties in the KB, both query understanding and document interpretation are improved, feeding new features to 
the search engine to leverage the query-and-click log 
to train better ranking models (see, e.g., \cite{DBLP:conf/sigir/DietzKM18,Reinanda2020} and references there). 
The techniques are further improved using 
contextual language models such as BERT \cite{DBLP:conf/naacl/DevlinCLT19}. 
Second, KBs help in various {\em search-assistance tasks}, including query auto-completion, spelling correction, query reformulation and query expansion (see, e.g., \cite{Reinanda2020}). 
As an example, when searching for the actor
Michelle Pfeiffer, after typing ``Michelle Pf'' the search engine will consult its KB for 
{\em auto-completion suggestions} such as  ``Michelle Pfeiffer'', 
``Michelle Pfeiffer movies'', ``Michelle Pfeiffer net worth'', 
``Michelle Pfeiffer children'' etc. 
(see Figure \ref{fig:michellepf-autocompletion}).
This exploits both the KB wealth of entities and
its content about relevant property types,
in combination with signals from query-and-click logs
\cite{Marantz2013}.

\subsubsection{Search by Entity Description}

A very large fraction of Internet search queries are lookups of named entities like celebrities, companies, products or events (see, e.g., {\small\url{https://trends.google.com/trends/yis/2020/US/}}).
An interesting variation of this entity search, beyond the bulk of most popular queries, is when the user
describes an entity, or a set of entities, by
its properties rather than by name \cite{DBLP:series/irs/Balog18}.
For example, the query ``Stanford computer science professors'' should return individual people like
Fei-Fei Li, Jeff Ullman, Jennifer Widom etc.
This applies even when there is only one correct answer, for example, for ``Stanford professor author of compiler and database textbooks'' (returning Jeff Ullman).

This case arises especially in {\em product search} when
products do not have an explicit name but are simply
represented by their properties.
An example is ``Seattle's Best Vanilla Flavored Decaf Portside Blend Medium Roast Ground Coffee'', 
which is a concatenation of properties:
\ent{brand} = ``Seattle's Best'',
\ent{flavor} = ``Vanilla'',
\ent{caffein} = ``Decaf'',
\ent{beans} = ``Portside Blend'',
\ent{processing} = ``Medium Roast Ground''.
Note that some relevant properties are still missing
from this description (e.g., no package \ent{weight} given).

Searching products by such, usually more
incomplete, descriptions is an important task
in e-commerce, when the user intent is purchasing.
The search engine attempts to map as many input phrases
as possible to crisp properties in the KB, so as
to narrow down the relevant products and make
the best fitting offer(s) to the user.
Because there is no central subject entity associated
with such queries, it is typically much harder to
find all relevant entities, especially under tight
restrictions for query response time.

\subsubsection{QA Assistants}

Question answering plays an important role for 
(speech-based) personal assistants, such as 
Amazon Alexa, Apple Siri, Google Assistant, Microsoft Cortana, Yandex Alice and others. 
Questions can be explicit such as
``How old is Jeff Bezos?'', or implicit such as
``play Taylor Swift's latest hit''.
They can be generic such as ``What is the weather in Seattle?'', or personal with implicit context such as
``What is the weather?'' where the assistant will tell the weather for the user's current location. 

The underlying idea for QA assistants is similar to that for knowledge-based search. 
The first step is to identify entities, types and properties in the question and link them to KB items to leverage the
KB content. Here, the linking can go beyond standard EL
(in the sense of Chapter \ref{ch3-sec-EntityDisambiguation})
by considering also phrases that denote
types (e.g., ``hit'' mapped to \ent{song})
and attributes or relations
(e.g., ``old'' mapped to \ent{birthdate}).
For the example question about Bezos's age,
the QA engine would first locate the entity
\ent{Jeff Bezos}, then follow the 
\ent{birthdate} relation to the date of birth,
and finally convert the date into the age in years.

Obviously, this question understanding can be quite challenging, and
incomplete linking is a common situation
(e.g., linking only ``Bezos'' to \ent{Jeff Bezos}, 
but missing out on ``old'').
Then, the QA engine needs to cleverly traverse the
KB graph to arrive at answer candidates,
based on supervised machine learning
(see, e.g., \cite{DBLP:conf/emnlp/SunBC19}).
When questions are too complex for the
QA assistant, web-search-based retrieval with
smart extraction of snippets can be a fallback option.

Knowledge-based QA is even more useful but also more challenging when combined with text-based QA over web contents. 
Consider the question 
``Who is the richest tech person?'' (presumably with
Jeff Bezos as top answer).
The KB may have all necessary information for it,
but the QA engine would have to figure out that
it needs to follow properties like
\ent{works for} (company) and \ent{belongs to} 
(high-tech industry), in addition to
identifying the \ent{net worth} attribute and
computing the argmax over a set of candidates.
This sophisticated process may easily fail on the
KB alone, but the web is full of cues that 
Bezos is a tech person.
The easiest way to combine KB-QA and Web-QA is to
retrieve the answer from web pages, and once the answer
is clear, map it to the KB and display the knowledge panel.
The Web-QA itself is usually based on passage retrieval,
which identifies short texts (e.g., a few sentences or a paragraph) that have a high 
density 
of matching the
question words and phrases, or a high matching score
based on neural language models such as BERT.
Subsequently, another learner extracts short text spans
from the best passages as candidate answers,
based on (distantly) supervised training
(see, e.g., \cite{DBLP:conf/acl/ChenFWB17}).
State-of-the-art techniques for this kind of text-oriented
QA and machine reading comprehension are discussed
in the recent survey \cite{DBLP:journals/access/HuangXHWQFZPW20}.

A second and less obvious way of combining 
KB-QA and Web-QA arises when neither the answer from
the KB nor the answer from web contents 
is sufficiently confident to be shown to the user.
Combining both sources, KB and web, could then boost
the joint confidence for displaying an answer.
One can conceive even more integrated ways of
seamlessly computing answers from both KB and web text;
this is a hot research topic 
(see, e.g., \cite{DBLP:conf/emnlp/SunBC19,DBLP:journals/corr/abs-2012-14610} and references there).

\subsection{Outlook}

KB technology has been highly impactful in Internet-centric industry, like search engines, online shopping and
social networks \cite{Noy:CACM2019}.
As of 2021, it remains a hot topic in industry
as the bars for quality and coverage are raised
and use cases are being expanded (see, e.g., 
\cite{DBLP:conf/www/ShinavierBZDGAO19,DBLP:journals/corr/abs-2003-02320}).

The technology has largely been driven by applications
in mainstream Internet business, but there are many
use cases also for other domains such as
health, materials, industrial plants,
financial services, and fashion 
(e.g., \cite{rotmensch2017learning,DBLP:conf/semweb/StrotgenTFMTMA019,DBLP:conf/semweb/HubauerL0H18,DBLP:journals/cacm/QiX18,Kari:DINAcon2019}).
Moreover, as scalable data platforms and machine learning 
are becoming commodities and affordable to small players as well,
KB technology is also advanced by startups and
specialized companies (e.g., \cite{DBLP:conf/sigmod/Zhang0RCN16,mesquita2019knowledgenet}).

\section{Ecosystem of Interlinked Data and Knowledge}
\label{subsec:webofdata}

The \emph{Semantic Web} is 
the
umbrella term for all semantic data and services
that are openly accessible via Web standards,
including several knowledge bases.
The design principles that KBs satisfy to 
be compliant with 
Semantic Web standards are the following:
\begin{itemize}
\item \textbf{Uniform machine-readable data representation:} This principle is implemented by using the RDF/RDFS standards for representing 
knowledge as statements in triple form (see Section~\ref{ch2-sec:knowledgerepresentation}).
\item \textbf{Self-descriptive data:} 
For semantic interpretation, there is no distinction between the instance-level data and the schema. 
The KB can describe its schema, constraints, reasoning rules, and taxonomy/ontology in a single model.
The relevant standards are RDFS, SHACL, and OWL  (see Section~\ref{sec:intensional}). 
\item \textbf{Interlinked data:} A KB can link to other KBs via statements that reference entities in other KBs, or directly at the entity level, by
the \ent{owl:sameAs} property
(cf. Section \ref{sec:entity-matching}),
and can be accessed using standardized protocols.
\end{itemize}
This last point lifts the KB
into an ecosystem of connected KBs and other datasets, which is called the Web of Data, Linked Data, or (if the data is openly available) Linked Open Data (LOD),
or the LOD cloud
(see surveys by \citet{HeathBizer2011}
and \citet{DBLP:books/sp/Hogan20}).
For this to work, the KBs have to provide 
entity identifiers
that are globally unique -- so that the Elvis Presley in DBpedia can be distinguished from the Elvis Presley in YAGO. For this purpose, each KB is identified by a \emph{uniform resource identifier} (URI, a generalized form of URL), such as \emph{https://yago-knowledge.org} for YAGO. Each entity in the KB is identified by a local name in the name space given by that URI, as in \emph{http://yago-knowledge.org/resource/Elvis\_Presley} for Elvis in YAGO. This requirement is already part of the RDF standard. To avoid cumbersome identifiers, RDF allows abbreviating URI prefixes. 
For example, by the abbrevation \emph{yago:} for \emph{http://yago-knowledge.org/resource/}, 
we can simply write \emph{yago:Elvis\_Presley}.

Interestingly, a KB can contain statements about entities from other KBs. For example, DBpedia can assert that its Elvis Presley belongs to the class \emph{singer} that is defined in YAGO, by the statement
\begin{lstlisting}
dbpedia:Elvis_Presley rdf:type yago:Singer
\end{lstlisting}
This has the advantage that a KB can re-use the vocabulary of another KB. 
The prefix ``rdf:'' refers to 
the standard vocabulary of RDF, which can be seen as another KB in the same unified framework.
The same applies to the OWL and SHACL vocabularies. There are also 
vocabularies with extensive modeling of semantic classes,
like \url{schema.org} \cite{Guha:CACM2016}  
(used in Section~\ref{sec:yago}).

The final ingredient for the Web of Data is making these URIs \emph{dereferenceable}: When we access the URI of Elvis in DBpedia, the server of DBpedia replies with a piece of RDF data about Elvis. In this reply, the client can find the URI of a class in YAGO, which it can 
follow
in the same way. By this mechanism, the machine can 
seamlessly
``surf the Semantic Web''
across all its knowledge and data bases, analogous to humans surfing the HTML-based Web. 

The most important links between KBs are \emph{owl:sameAs} links. They are used to assert that an identifier in one KB refers to the same real-world entity denoted by another identifier in another KB:
\begin{lstlisting}
dbpedia:Elvis_Presley owl:sameAs yago:Elvis_Presley
\end{lstlisting}
These links enable applications to complement the data found about an entity in one KB with the data from another KB.
As of 2019, 
more than 1200 data and knowledge resources
are interlinked this way (see \url{https://lod-cloud.net}).

Another 
useful element of the Semantic Web are
annotation languages like RDFa, Microdata, or JSON-LD 
for
augmenting HTML documents with RDF data
\cite{DBLP:conf/semweb/MeuselPB14}. 
Search engines use such annotations to identify, for example, product reviews with their scores, shops with their locations, or movies with their showtimes.
Such micro-data appears in billions of web pages,
with increasing trend
(see, e.g., {\small\url{http://webdatacommons.org/structureddata/}}).
While a large portion of annotations concern 
website meta-data, the 
most widely used class of HTML micro-data is \emph{schema:Product}.

Finally, the Semantic Web has pushed forward the adoption of shared vocabularies. Perhaps the most prominent effect is \emph{schema.org} 
(\citet{Guha:CACM2016}), 
a vocabulary for
thousands of classes, developed by Google, Microsoft, Yahoo  and Yandex and standardized by 
the World Wide Web Consortium.

\clearpage\newpage
\chapter{Wrap-Up}

\section{Take-Home Lessons}

This article 
discussed
concepts and methods for the task of automatically building comprehensive knowledge bases (KBs) of near-human quality.
This entails two strategic objectives:
\squishlist
\item a very high degree of {\em correctness}, with error rates below 5 percent or even under 1 percent, and
\item a very large {\em coverage} of entities, their semantic types and their properties, within the scope and intended usage of the KB. 
\squishend

\vspace*{0.2cm}
\noindent{\bf Low-hanging Fruit First:}
The two goals about correctness and coverage 
are in tension with each other.
The first goal suggests conservative methods, tapping largely into premium sources and using high-precision algorithms such as rules and (learned) patterns.
Premium sources (e.g., Wikipedia, IMDB, Goodreads, MayoClinic etc.) are 
characterized by i) having authoritative
high-quality content, often in
semi-structured form with categories, lists, tables, ii) high coverage
of relevant entities, iii) clean and 
uniform style, structure and layout, and,
therefore, iv) being extraction-friendly.
The second goal calls for aggressive methods that can discover promising sources, identify long-tail entities and extract interesting properties beyond basic facts.
Typically, this involves powerful but riskier methods such as Open IE and deep neural networks

Such aggressive methods can often leverage the outcome of the conservative stage by using high-quality data from an initial KB as {\em seeds for distant supervision}.
This mitigates the typical bottleneck of not having enough training samples.
The bottom line is that KB construction should consider harvesting ``low-hanging fruit'' first and then embark on more challenging methodology as needed.

\vspace*{0.2cm}
\noindent{\bf Core Knowledge on Entities and Types:}
Entities and their semantic types form the backbone of every high-quality KB. Attributes and relations are great assets on top, but they can shine only if the entity-type foundation is proper. 
Search engines and recommender systems often need only entities, class memberships, and entity-entity relatedness. 
This is why this article first 
emphasized the construction of {\em taxonomic knowledge} and the {\em canonicalization of entities} (Chapters 3 through 5).
Based on this sound foundation,
populating the KB with attributes and relations
(Chapters 6 and 7) yields great value
for advanced use cases such as
expert-level QA or 
entity-centric analytics over properties.

\vspace*{0.2cm}
\noindent{\bf Knowledge Engineering:}
KB construction is not a one-time task that can be tackled by a single method in an end-to-end manner.
We need to keep in mind that KBs serve as 
infrastructure assets and must be maintained over long timespans. 
Typically, we start with building an initial KB of limited scope and size, and then gradually grow it over time.
This life-cycle involves tasks like
KB augmentation by adding properties (Chapter 6), 
discovering
new entities, schema expansion (Chapter 7), 
and most importantly, {\em quality assurance} and 
{\em KB curation} (Chapter 8).

For this entire framework, but also for each of its sub-tasks, a substantial amount of engineering is required,
with humans like KB architects and KB curators in the loop.
KB construction relies on clever algorithms and smart machine learning, but orchestrating and steering the complete machinery %
cannot be fully automated.

\vspace*{0.2cm}
\noindent{\bf Diverse Toolbox:}
For many sub-tasks of KB creation and curation, we have presented a variety of alternative methods. 
Some readers may have hoped for a clear recommendation 
of a single best-practice choice, but this is unrealistic.
There are many trade-offs (e.g., precision vs. recall vs. cost) and
dependencies on application requirements.
With varying sweet spots and limitations of different algorithms and learners, 
there is no one-size-fits-all, turn-key method.

Put positively, a wide portfolio of different models, methods and tools for knowledge extraction and knowledge cleaning is a great asset to build on. 
Each sub-task in the KB life-cycle mandates judicious choices from this toolbox, again emphasizing the need for humans in the loop.

\section{Challenges and Opportunities}

In this final section, we sketch some of the open challenges that remain to be addressed towards the next generation of knowledge bases, pointing out opportunities for original and potentially impactful research.

\vspace*{0.2cm}
\noindent{\bf Language Models and Knowledge Bases:}\\
Many methods for knowledge extraction harness different kinds of language models, from lexicons of word senses 
and lexical relations \cite{Fellbaum1998,DBLP:series/synthesis/2016Gurevych}
all the way to contextualized embeddings such as ELMo, BERT and GPT-3 \cite{DBLP:conf/naacl/PetersNIGCLZ18,DBLP:conf/naacl/DevlinCLT19,brown2020language}.
As the latter encode huge text corpora and are trained to predict missing words or phrases, an intriguing idea could be to use {\em neural language models} directly as a proxy KB (see, e.g., \cite{DBLP:conf/emnlp/PetroniRRLBWM19,DBLP:journals/tacl/JiangXAN20}). %
For example, instead of looking up which river flows through Paris, we could ask a neural model to
predict the word or phrase ``[?]'' in
the incomplete sentence\\
\hspace*{1cm}``Paris is located on the banks of the [?]'',\\
Current models correctly return
``Seine'' as the best prediction
(followed by Loire, Danube, Mississippi etc.).
However, when we want to obtain protest songs by Bob Dylan from\\
\hspace*{1cm} ``Dylan wrote protest songs such as [?]'',\\
the top prediction is
``this'', and the predictions for\\
\hspace*{1cm} ``Dylan wrote [?] songs such as Hurricane''\\
are ``popular'', ``other'', ``many'', ``several'',
``three'' etc, but not ``protest'' or 
``political''. These results are 
sort of correct, but completely useless.

Nevertheless, the impressive capabilities of these models for reading comprehension could be leveraged more extensively for KB construction and augmentation.
The interplay of symbolic knowledge and latent understanding of language is an important research avenue \cite{DBLP:journals/corr/abs-2003-08271}.

\vspace*{0.2cm}
\noindent{\bf Credibility and Trust:}\\
We live in times with online information exploding
in volume, velocity and varying shades of veracity. 
Unfortunately, this includes a large fraction of (false) misinformation and (intentionally false) disinformation.
To support people in detecting fake news and assessing the credibility of claims in social media, research on fact checking and propaganda detection has become a major direction
(e.g., \cite{DBLP:conf/emnlp/RashkinCJVC17,DBLP:conf/www/PopatMSW17,DBLP:conf/naacl/ThorneVCM18,DBLP:conf/acl/ZhangIR19,adair2019automated,DBLP:conf/ijcai/MartinoCBYPN20,DBLP:conf/www/JiangBI020}).
When KB construction taps into riskier sources such as polarized news sources or discussion forums, assessing
credibility and trustworthiness becomes crucial.
Conversely, KB statements and reasoning over them
could potentially contribute also to unmasking false claims.

An important use case is the domain of health.
On the one hand, health communities and specific discussion
forums
are sources for learning more about symptoms of diseases,
typical (as opposed to potentially possible) side effects of drugs, the patients' experience with therapies, and
all kinds of cross-talk signals about combinations of
medical treatments.
On the other hand, such social media come with 
a large amount of
noisy and false statements
(see, e.g., \cite{DBLP:conf/kdd/MukherjeeWD14}).
Contemporary examples are 
treatments for
virus-caused epidemic diseases and the pros and cons of vaccinations.
Even scientific news
are often under debate
(e.g., \cite{DBLP:conf/www/Smeros0A19,wadden2020fact}).
Tackling these concerns about veracity is an inherent part
of knowledge acquistion for the health domain.

\vspace*{0.2cm}
\noindent{\bf Commonsense and Socio-Cultural Knowledge:}\\
Commonsense knowledge (CSK) is the AI term for non-encyclopedic world knowledge
that virtually all humans agree on. 
This comprises:
\squishlist
\item Notable {\em properties of everyday objects}, such as:
mountains are high and steep, they are taller than 
man-made buildings,
they may be snow-covered or
rocky (but they are never fast or funny);
cups are usually round and are used to hold liquids such as coffee, tea etc.
\item  {\em Behavioral patterns} and {\em causality}, such as:
children live with their parents, pregnancy leads to birth, and so on.
\item {\em Human activities} and their typical settings, such as:
concerts involve musicians, instruments and an audience;
rock concerts involve amplifiers and usually take place in
large concert halls or open air festivals (rather than cozy bars or dinner places).
\squishend

This kind of CSK about general concepts and activities (as opposed to
individual named entities) is obvious for humans: every young adult or even child
knows this. However, CSK is surprisingly difficult to acquire by machines.
There is ample AI work on knowledge representation to this end \cite{davis2014representations}, introducing,
for example, epistemic logics with modalities like \ent{always}, \ent{sometimes}, \ent{typically}
or \ent{never}, but there is not much work on building large-scale CSK collections.
Some works consider concept hierarchies like WordNet-style hypernymy as CSK,
but this can equally be seen as part of encyclopedic KBs and has been intensively researched
(see Chapters 3 and 4).

CSK may not be needed for today's mega-applications like search engines,
but it could become a crucial building block for next-generation AI.
An envisioned use case is to equip conversational assistants, like chatbots,
with commonsense background knowledge.
This would enable them to understand their human dialog partners better
(e.g., humor) and behave more robustly in their generated utterances
(e.g., avoiding absurdities and offensive statements).
Interpreting visual contents, especially videos with speech, would be another
potential use case of high importance
\cite{DBLP:conf/cvpr/ZellersBFC19}.

A variation of CSK that matters for human-computer interaction is {\em socio-cultural knowledge}:
behavioral patterns that do not necessarily hold universally, but are widely agreed upon
within a large socio-cultural group.
For example, people in the western world greet each other by handshakes, but this
is very uncommon in 
large parts of Asia.
Similarly, there are preferred styles of eating meals, with different utensils,
different ways of sharing, etc.
These are not just specific to geo-regions, but need to consider social and cultural
backgrounds; for example, children and juveniles do not greet each other by handshakes
even in the western world.

Various projects have applied knowledge extraction methods, like those in Chapters 6 and 7,
to gather CSK statements from online contents
(e.g., 
\cite{Tandon2014,DBLP:conf/cikm/TandonMDW15,DBLP:journals/tacl/ZhangRDD17,DBLP:journals/tacl/MishraTC17,DBLP:conf/aaai/SapBABLRRSC19,DBLP:conf/cikm/RomeroRPPSW19,chalier2020joint,TuanPhongNguyen:WWW2021}),
tapping into non-standard sources like book texts, online forums, frequent queries and 
image descriptions.
A major difficulty to tackle is that mundane CSK is rarely expressed explicitly \cite{schubert2002can}
(hence the need to tap non-standard sources), and when it is, this is often
in a very biased, atypical or misleading manner \cite{gordon2013reporting}.
For example, web-frequency signals would strongly suggest that most
\ent{programmers} are 
\ent{lonely} and \ent{socially awkward} (which is disproven by the readers of this article, hopefully).
The most successful CSK projects so far seem to be the conservative ones that relied
on human inputs by crowdsourcing, like ConceptNet \cite{singh2002open,Speer:LREC2012,DBLP:conf/aaai/SpeerCH17},
or knowledge engineers, like Cyc \cite{Lenat:CACM1995,DBLP:conf/aaai/MatuszekWKCSSL05}, but these are fairly limited in scope and scale.
There are great research opportunities to re-think these prior approaches and devise
new ones to advance
CSK acquisition.

\vspace*{0.2cm}
\noindent{\bf Personal and Subjective Knowledge:}\\
Another step away from encyclopedic knowledge of universal interest is to capture knowledge about individual users, like their habits, tastes and preferences. This would be based on the digital footprint that a user leaves by her online behavior, including clicks, likes, purchases etc.
This kind of knowledge  for {\em user
profiling} is collected by major platforms for shopping, entertainment, social media, etc.,
and leveraged to personalize recommendations and services.

However, there would be big merit in
providing a user with her {\em personal KB} also
on the user's desktop and/or smartphone.
This KB could serve as an ``augmented memory'' that assists the user by smart search over 
her personal digital history.
For example, the user could easily retrieve restaurants where the user dined during her vacation in Italy
and could recollect which places and dishes she liked best.
Research on this notion of a personal KB dates
back to the theme of ``Personal Information Management'' (e.g., \cite{DBLP:conf/sigir/DumaisCCJSR03,DBLP:conf/cidr/DongH05,DBLP:conf/vldb/DittrichSKB05})
and is being revived now
(e.g., \cite{DBLP:conf/www/MontoyaTASS18,DBLP:conf/ictir/BalogK19}).

In addition to such services for end-users, there is an even more compelling reason for
personal KBs under user control, namely,
privacy and, more specifically, the 
need for {\em inverse privacy} \cite{DBLP:journals/cacm/GurevichW16}:
knowing what others (the big providers) 
know about you.
Although modern legislation such as the EU-wide GDPR
(General Data Protection Regulation)
mandate that users have the right to
inquire provider-side profiling and
have unwanted information removed,
the implementation is all but easy.
Users leave their data in a wide variety of
web and cloud services over a very long timeframe, and most will simply forget which
traces they have left where and when.
A digital assistant could intercept the user's
online access and record personal information locally, this way maintaining a longitudinal
personal KB, under the control of the user herself (see, e.g., \cite{DBLP:conf/cikm/KalokyriBM18}).
A practically viable solution still faces research challenges, though: 
what to capture and how to represent it,
how to contextualize statements for
user interpretability, 
how to empower end-users to manage the
life-cycle of their personal KBs, and more.

While personal knowledge is still factual
but refers only to one individual, 
there are also use cases for capturing
{\em subjective knowledge},
regarding beliefs and argumentations.
For example, it could be of interest
to have statements of the form:
\begin{lstlisting}
Fabian believes id1: (Elvis livesOn Mars)
Gerhard believes id2: (Dylan deserves PeaceNobelPrize)
Luna supports Gerhard on id2
Simon opposes Gerhard on id2
\end{lstlisting}
\noindent These kinds of statements, with
second-order predicates about attribution and stance, are essential to capture people's 
positions and potential biases in discussions and
controversies. Obviously, the stakeholders of
interest could be important people such
as Emmanuel Macron or Angela Merkel rather
than the above insignificant ones.
Use cases include credibility analysis
(see point {\em Credibility and Trust} above), argumentation mining,
analyzing debates, political opinion mining,
uncovering propaganda and other manipulation,
reasoning over legal texts, and more
(see, e.g., \cite{DBLP:conf/wsdm/AwadallahRW12,DBLP:journals/coling/StabG17,DBLP:journals/debu/Halevy19,DBLP:conf/naacl/ChenK0CR19,DBLP:journals/fdata/Olteanu00K19,DBLP:conf/ecir/Suchanek20,Bhutani:AKBC2020,DBLP:conf/aaai/KhatibHWJB020}).
How to automatically capture this kind of subjective knowledge for systematic 
organization in a KB is a widely open
research area.

\vspace*{0.2cm}
\noindent{\bf Entities with Quantities:}\\
KBs should also support knowledge workers like (data) journalists, (business and media) analysts, health experts, and more.
Such advanced users 
go beyond
finding entities or looking up their properties,
and often desire to filter, compare, group, aggregate and rank entities based on
{\em quantities}: financial, physical, technological, medical and other measures, such as 
annual revenue or estimated worth, distance or speed,
energy consumption or CO2 footprint,
blood lab values or drug dosages.
Examples of quantity-centric information needs are:
\squishlist
\item Which runners have 
ten or more marathons under 
2:10:00 hours?
\item How do the sales/downloads, earnings and wealth of male and female singers compare?
\item How do Japanese electric cars compare to US-made models, in terms of energy efficiency, carbon footprint and cost/km?
\squishend
These kinds of analyses would be straightforward, using SQL or SPARQL queries and data-science tools, if the underlying data were stored in a single database or knowledge base.
Unfortunately, this is not the case. KBs are notoriously sparse regarding quantities; for example, Wikidata contains several thousand marathon runners but knows their best times 
only for a few tens (not to speak of all their races).
Instead of a KB, we could turn to domain-specific databases on the web, but finding the right sources in an ``ocean of data''
and assessing their quality, freshness and completeness 
is itself a big challenge.

Extracting statements about quantities, 
from text or tables,
poses major issues
\cite{DBLP:conf/kdd/SarawagiC14,DBLP:conf/cikm/IbrahimRW16,DBLP:conf/semweb/HoIPBW19,DBLP:journals/debu/Weikum20,VinhThinhHo:WWW2021}: 
\squishlist
\item[i)] detecting and normalizing 
quantities
that appear with varying values (e.g., estimated or stale), with different scales (e.g., with modifiers ``thousand'', 
``K'' or ``Mio'') and units
(e.g., MPG-e (miles per gallon equivalent) vs. kWh/100km),
\item[ii)] inferring to which entity and measure a quantity mention refers, and
\item[iii)] contextualizing entity-quantity pairs
with enough data for proper interpretation in downstream analytics -- all this with very high quality and coverage.
\squishend

\vspace*{0.2cm}
\noindent{\bf Knowledge Base Coverage:}\\
Commonsense and quantity knowledge are two dimensions on which today's KBs fall short. 
In general, notwithstanding good progress over
the last decade, coverage is a major concern even for very large KBs \cite{DBLP:conf/cikm/RazniewskiD20}.
Typically, they cover basic facts about people, places and products very well,
but often miss out on more sophisticated, highly notable points such as:
\squishlist
\item Johnny Cash performed a free concert in a prison (in the 1960s!).
\item The Shoshone woman Sacagawea carried her newborn child when serving as guide and interpreter for the Lewis and Clark expedition.
\item Frida Kahlo, the surrealistic Mexican painter, suffered her whole life from injuries in a bus accident.
\item Cixin Liu's book Three Body features locations like Tsinghua University and Alpha Centauri.
\squishend
These facts are prominently mentioned in the respective Wikipedia articles and would
be remembered as salient points by most readers, yet they are completely absent in today's KBs.
Despite advances on Open IE (Chapter 7) to discover 
new predicates,
coverage is a major pain point.
This calls for re-thinking the notion of knowledge saliency
and the approaches to capture ``unknown unknowns''.

\vspace*{0.2cm}
\noindent{\bf Human in the Loop:}\\
Throughout this article, we have emphasized the goal of automating KB construction and curation as much as possible. However, human inputs do play a role and their extent can vary widely. 
The most obvious step is manually checking the validity of KB statements for quality assurance in industrial-strength KBs, which can adopt a variety of strategies (see, e.g., \cite{DBLP:conf/wsdm/TanAIG14}).
Going beyond this stage,
seed-based learning benefits from hand-crafted seeds 
and from feedback on learned patterns and rules (see, e.g.,  Section~\ref{ch3-subsect-patterns}
and Section~\ref{sec:nell}). This human intervention serves to flag incorrect or misleading signals for knowledge extraction as early as possible, and helps steering the KB construction process towards high quality.

It is also conceivable to rely entirely on humans for
KB construction. This can either be via human experts (as in the early days of Cyc~\cite{Lenat:CACM1995}) or crowdsourcing workers (as for large parts of
ConceptNet~\cite{conceptnet17} and ImageNet~\cite{Deng:CVPR2009}).
Wikidata builds on community inputs \cite{Vrandecic:CACM2014}, where 
many 
contributors 
have programming skills
and provide bulk imports from high-quality data sources.
Yet another variant of the human-in-the-loop theme is {\em games with a purpose} (such as
ESP~\cite{DBLP:conf/chi/AhnD04} or Verbosity~\cite{von2006verbosity}).

The natural conclusion is to carefully combine human input with large-scale automation. How do this in the best way for low cost and high quality, is another research challenge for next-generation knowledge bases (see, e.g., \cite{DBLP:journals/cacm/DoanRH11,DBLP:journals/cacm/DoanKCGPCMC20}).

\clearpage\newpage

\chapter*{Acknowledgements}

Many thanks to Mike Casey and other staff at
NOW Publishers.
We are most grateful for the thoughtful and
extremely helpful comments by Soumen Chakrabarti,
AnHai Doan and two other (anonymous) reviewers.
We also appreciate the sustained encouragement
and support by our editors Surajit Chaudhuri,
Joe Hellerstein and Ihab Ilyas.
It is a great pleasure and honor to have such
wonderful colleagues in our research community.


\backmatter  


\clearpage\newpage
{
\setlength{\leftmargin}{0pt}
\begin{flushleft}
\printbibliography
\end{flushleft}
}

\end{document}